\documentclass{article}
\usepackage[top=1.5cm, bottom=1.5cm, left=1.5cm, right=1.5cm]{geometry}
\usepackage{authblk}

\usepackage{amsfonts}
\usepackage{amssymb}
\usepackage{bm}
\usepackage{url}
\usepackage{amsmath}
\usepackage{graphicx}
\usepackage{subfigure}
\usepackage[ruled, vlined]{algorithm2e}
\usepackage{mathtools}
\usepackage{threeparttable}
\usepackage{multirow}
\usepackage{morefloats}

\usepackage{adjustbox}
\usepackage{dsfont}

\usepackage[titletoc,title]{appendix}

\newcommand{\commentout}[1]{}
\newcommand\eval[1]{\begin{array}[t]{@{}c@{\,}|@{\,}}%
		\raisebox{0pt}[0.85\height][1.33\depth]{$ \displaystyle#1 $}\end{array}}

\title{AdjointBackMapV2: Precise Reconstruction of Arbitrary CNN Unit's Activation via Adjoint Operators}
\author[1]{Qing Wan\thanks{frankqingwan@gmail.com}}
\author[2]{Siu Wun Cheung\thanks{cheung26@llnl.gov}}
\author[3]{Yoonsuck Choe\thanks{choe@tamu.edu}}
\affil[1]{School of Computer Science and Technology, Zhejiang Gongshang University \protect \\Zhejiang Province 310018, China}
\affil[2]{Center for Applied Scientific Computing, Lawrence Livermore National Laboratory \protect \\Livermore, CA 94550, USA}
\affil[3]{Department of Computer Science \& Engineering, Texas A\&M University \protect \\College Station, TX, USA, 77843}

\begin{document}
	
	\maketitle
	
	\begin{abstract}
		Adjoint operators have been found to be effective in the exploration of CNN's inner workings \cite{wan2022adjointbackmap}. However, the previous no-bias assumption restricted its generalization. We overcome the restriction via embedding input images into an extended normed space that includes bias in all CNN layers as part of the extended space and propose an adjoint-operator-based algorithm that maps high-level weights back to the extended input space for reconstructing an effective hypersurface. Such hypersurface can be computed for an arbitrary unit in the CNN, and we prove that this reconstructed hypersurface, when multiplied by the original input (through an inner product), will precisely replicate the output value of each unit. We show experimental results based on the CIFAR-10 and CIFAR-100 data sets where the proposed approach achieves near $0$ activation value reconstruction error.
	\end{abstract}
	
	\section{Introduction}
	\label{Introduction}
	Convolutional Neural Network (CNN) has achieved superior performance in Computer Vision (CV). Its innovation comes from the building block of learning a large number of kernels in parallel automatically, greatly facilitated by the availability of GPUs \cite{nickolls2008scalable, chetlur2014cudnn}. Also, architectural innovations \cite{krizhevsky2012imagenet, simonyan2014very, he2016deep, zoph2018learning} significantly enhanced their generalization capability.
	
	Despite many successes in computer vision, CNN's inner workings remain to be explained. Usually, convolutional or pooling layers heap up inside a CNN, and nonlinear activations and shortcuts are laced among them for connections. This sophistication increases CNN's opacity and difficulty in the exploration of its internal functional principles. As AI-powered systems have been deployed massively in industrial applications, the demand for transparency and explainability has also grown rapidly. Recent techniques have aimed to achieve transparency/explainability in CNNs. In general, these techniques can be grouped into three approaches: (1) inverse of feature maps \cite{zeiler2014visualizing, springenberg2014striving, samek2016interpreting, dosovitskiy2016inverting}, (2) perturbation \cite{baehrens2010explain, simonyan2013deep, ribeiro2016should, koh2017understanding, kim2018interpretability}, and (3) activation maps \cite{zhou2016learning, selvaraju2017grad}. These methods estimate the features that contribute most to CNN's decision. However, inverse requires bijection, which does not hold for CNNs. Perturbation gauges important features locally, which might be a small part of the prediction factors. Besides, once the input changes (the vector drifts), the conclusion might vary simultaneously. Activation maps heat objects up from the input using a resized high-level weighted activation map (heatmap) to be overlaid on the original image. For example, suppose the high-level heatmap (shape: $8 \times 8$) will be scaled up to fit the $32 \times 32$ input image for highlighting the features of interest of the CNN. The potential assumption is that each of the $4 \times 4$ area on the input corresponds to a single pixel in the heatmap. As we know, the effective receptive field of a convolutional layer's unit will dilate as the layer goes higher up. Thus, the effective receptive field of the heatmap's pixel may exceed a $4 \times 4$ area in the input image. Restricting the heatmap unit to represent only a $4 \times 4$ area would violate this scale.
	
	Despite these efforts, a fundamental question is still open: Given a CNN unit in a convolved feature map or a class output value from the fully connected (FC) layer inside a CNN, can the full computational effect of its weight (convolutional kernel or wegith vector in the FC layer) be represented in the input space? This paper investigates this seemingly intractable question. As our previous work \cite{wan2022adjointbackmap} has revealed the effectiveness of Adjoint operators \cite{banach1987theory} in modeling and analyzing CNN models without bias units inside, we are motivated to convert a trained CNN to an equivalent computational topology (similar to Fig.\ref{fig:Model:cnn_with_bias_and_its_equivalence}), which overcomes the bias-free restriction in \cite{wan2022adjointbackmap}. We project any high-level weight vector back to an extended normed space, taking bias as a part of the input, to reconstruct an effective hypersurface that determines the output value of a unit from an out-channel (out-ch) feature map or the predicted output value of the fully connected (FC) layer via an AdjointBackMapV2 (Algorithm.\ref{algorithm:RM4_to_RM0}, upgraded from our earlier work \cite{wan2022adjointbackmap}). We prove that high precision from \cite{wan2022adjointbackmap} is preserved over any reconstructed effective hypersurface for CNNs using bias or batch normalization (BN) as long as the necessary conditions in Section \ref{Model:Algorithm} are satisfied. Our experiments on both CIFAR-10/CIFAR-100 datasets \cite{krizhevsky2009learning} show that the proposed approach achieves near $0$ reconstruction error.
	
	\paragraph{Paper Organization}
	The rest of the paper is organized as follows: Section \ref{Related_Works} will summarize recent progress towards unveiling CNN's inner workings. Section \ref{Model} will introduce our theory and discuss how the limitations in our previous algorithm  \cite{wan2022adjointbackmap} can be addressed. Section \ref{Experiments} will experiment with three prevalent CNN architectures to verify our theory.
	
	\section{Related works}
	\label{Related_Works}
	CNN's internal working principles remain unclear despite many efforts to achieve explainability. As found in \cite{ghorbani2019interpretation, heo2019fooling, dombrowski2019explanations}, the various methods for interpretability/explainability are fragile, where adversarial examples specifically targeted to fool the interpretation mechanism itself could be generated. To overcome this issue, a principled approach based on mathematical theory is needed. Early attempts \cite{wiatowski2017mathematical, qiu2018dcfnet} approached CNNs using Wavelet Theory to find correlations between CNN and filter banks for a better interpretation. A recent study \cite{jacot2018neural} proposed the Neural Tangent Kernel (NTK) method for approximating a neural network trained through gradient descent with kernel regression. However, vulnerabilities are inevitable with these methods. Wavelet theory considers a convolutional layer as an LTI (Linear Time-Invariant) system, which is not appropriate due to the layer's bias and the nonlinear activation function. The foundation of NTK relies on an approximation between the weight update policy and the following differential equation:
	\begin{equation} \label{eq:NTK_approx}
		\begin{split}
			\mathbf{w}(t + 1) = \mathbf{w}(t) - \frac{\partial{L(\mathbf{x}, \mathbf{y}, \mathbf{w}(t))}}{\partial{\mathbf{w}}},\\
			\frac{d\mathbf{w}(t)}{dt} = \lim_{\epsilon \to 0} \frac{\mathbf{w}(t + \epsilon) - \mathbf{w}(t)}{\epsilon}
			= -\frac{\partial{L(\mathbf{x}, \mathbf{y}, \mathbf{w}(t))}}{\partial{\mathbf{w}}},
		\end{split}
	\end{equation}
	where $\mathbf{w}(t)$ is the weights at time $t$, and $L(\mathbf{x}, \mathbf{y}, \mathbf{w}(t))$ is the loss function with inputs $\mathbf{x}$ and labels $\mathbf{y}$. This approximation commits relative errors. Whether these errors are negligible remains questionable due to high dimensional weight matrices. Besides, the NTK regime suffers from theoretical challenges \cite{hayou2019mean}. Our own theory in this paper circumvents these weaknesses, as Adjoint operator is defined via the dual form \cite{banach1987theory} that entails an equality relationship. This equality guarantees that the effective hypersurfaces recovered with our method, when multiplied by dot-product with the input (a concatenation of the input image and the bias vectors from the entire CNN), will precisely replicate a convolved feature map's activation value or the FC layer's output value in a CNN (see Section \ref{Model} for details).
	
	\section{Model}
	\label{Model}
	\subsection{Notation}
	\label{Model:Notation}
	Mathematical notations are summarized below.
	\begin{enumerate}
		\item $\circledast$ denotes convolution;
		\item $\mathbf{\theta}_{\mathcal{X}}$ denotes the origin of the vector space $\mathcal{X}$;
		\item $\bigtimes$ denotes a Cartesian product;
        \item $f \circ g$ is the composition of two functions $f(\cdot)$ and $g(\cdot)$;
		\item $[\mathbf{x}_{n}; \mathbf{x}_{b}]$ denotes a concatenation of two tensors, $\mathbf{x}_{n}$ and $\mathbf{x}_{b}$;
		\item $\mathbf{x}[\mathbf{b}_{i}]$ represents an operation that retrieves the corresponding part of $\mathbf{b}_{i}$ from $\mathbf{x}$ ($\mathbf{x}$ is a sequential concatenation of tensors $\{\mathbf{b}_{i}\}$);
		\item $\langle \mathbf{x} \mid \mathbf{y} \rangle$ denotes the inner product of $\mathbf{x}, \mathbf{y} \in \mathcal{X}$;
		\item $\mathcal{X^*}$ denotes the algebraic dual of $\mathcal{X}$, i.e., the space of all linear functionals on $\mathcal{X}$;
		\item $\langle \mathbf{x}, \mathbf{x}^{*} \rangle$ denotes the value of a linear functional $\mathbf{x}^{*} \in \mathcal{X^*}$ at $\mathbf{x} \in \mathcal{X}$;
		\item $B(\mathcal{X}, \mathcal{Y})$ denotes the space of all bounded linear operators from $\mathcal{X}$ to $\mathcal{Y}$.
	\end{enumerate}

	\subsection{Theory}
	\label{Model:Theory}
\begin{figure}[!h] 
		\centering
		\begin{tabular}{@{}c@{}c@{}c@{}}
			\includegraphics[width=0.3\textwidth]{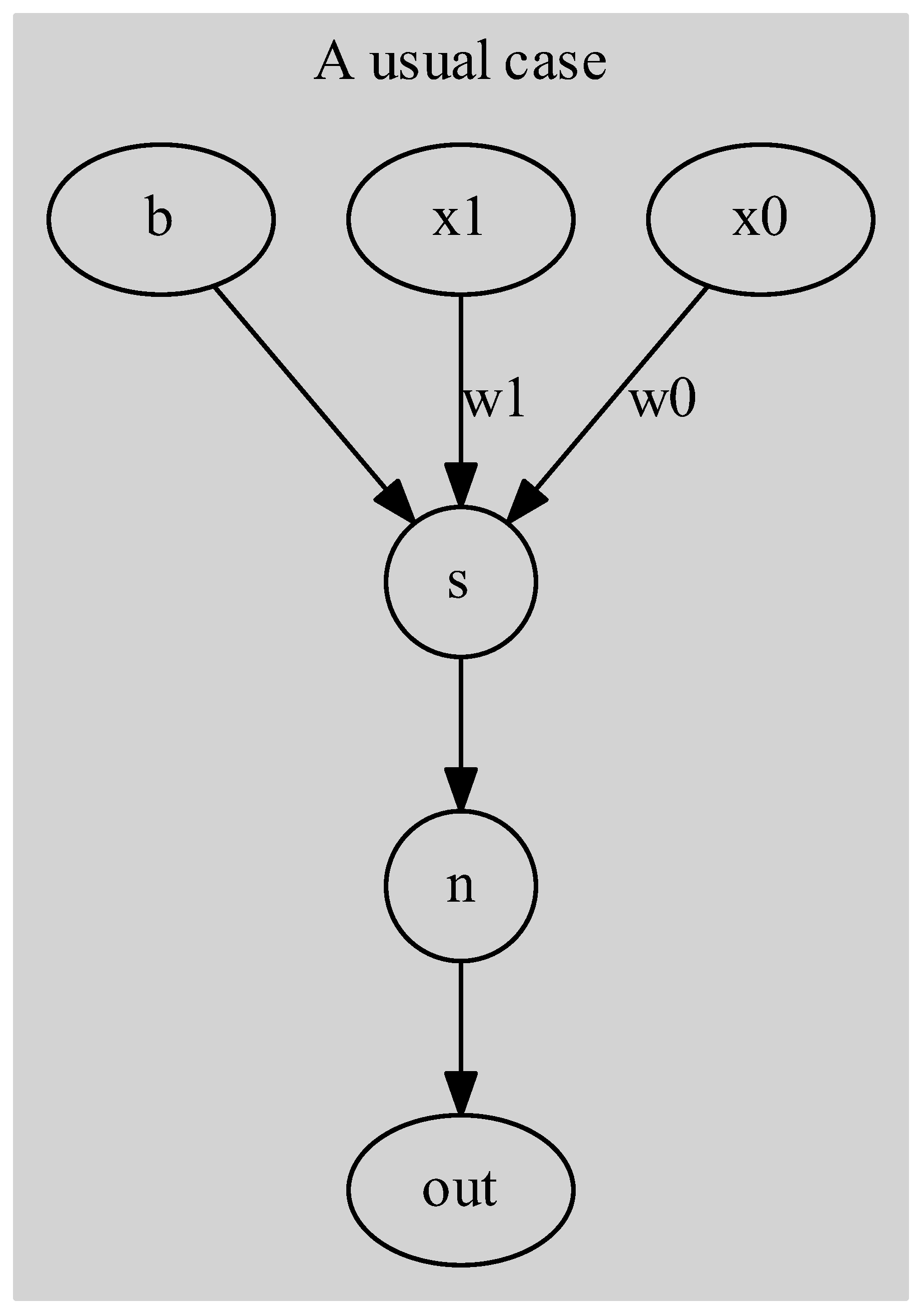} &
			\includegraphics[width=0.3\textwidth]{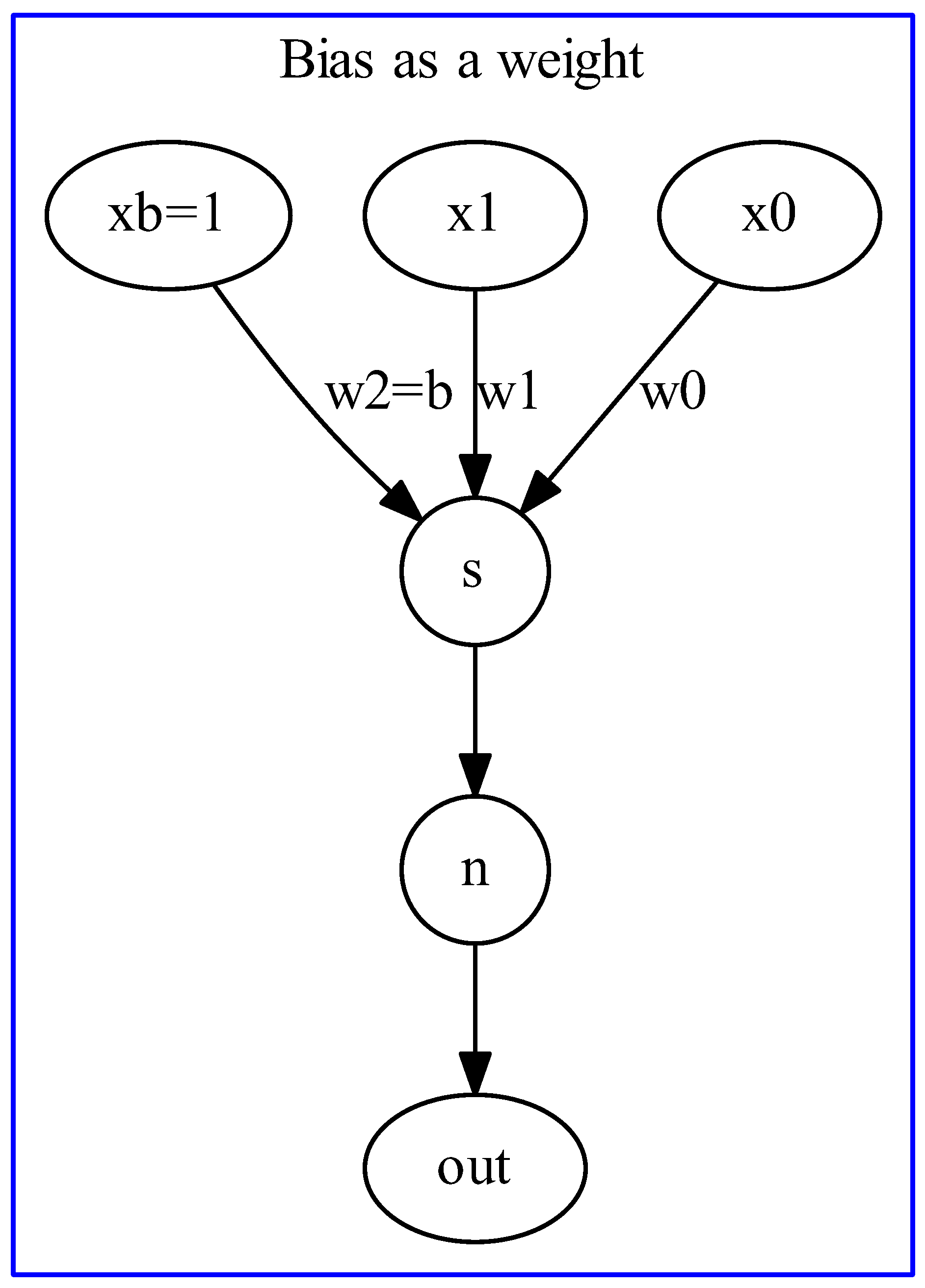} &
			\includegraphics[width=0.3\textwidth]{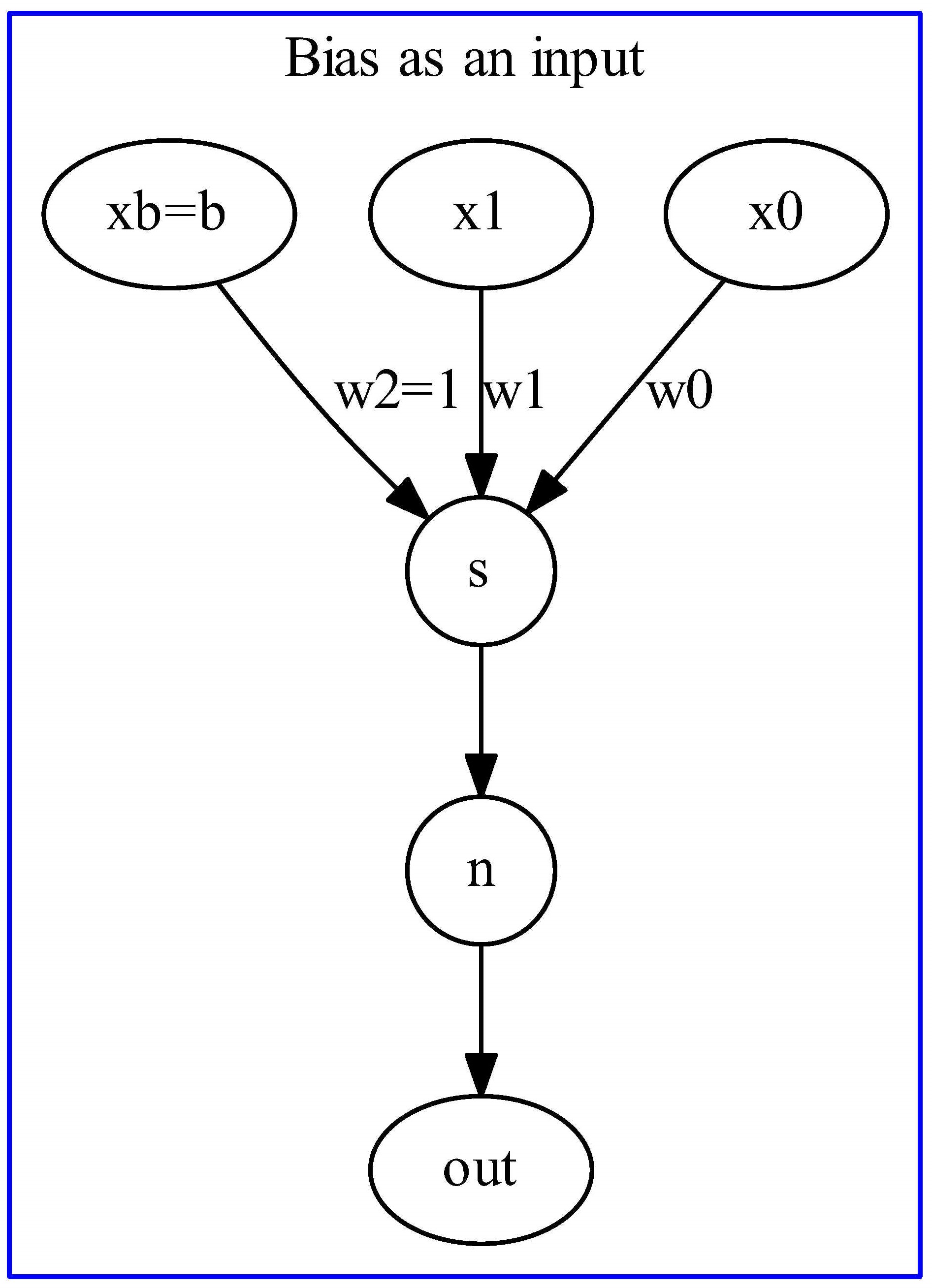} \\
		\end{tabular}
		\caption{Two equivalent models of a trained artificial neuron. {\bf Left}: A conventional artificial neuron with two inputs $x_0, x_1$, whose weight is $w_0, w_1$, respectively. Node $b$ denotes a bias. Node $s$ denotes a summation unit, and node $n$ denotes an activation unit. {\bf Middle}: An equivalent model to the left. We treat the bias $b$ as the multiplication of the third input, $x_b$, and a weight $w_2=b$. {\bf Right}: Another equivalent model to the left. We still treat bias $b$ as the third input but multiplying a weight $w_2=1$. We use the right one since it has fewer computations ($w_2=1$ can be ignored). Note that this assumes that the weights and the biases are already learned and fixed.}
		\label{fig:Model:single_neuron_equivalence}
	\end{figure}

	Applying adjoint operators to a general CNN is intractable since a bias vector usually exists right after convolution. Specifically, the challenge comes from the $\mathbf{F}(\mathbf{\theta}_{\mathcal{X}})$ (the full computation leading up to the receptive field in the in-channel feature map of a convolutional layer or the in-channel feature vector in the FC layer, prior to multiplying with the kernel or the weight vector: Eq.(2) in \cite{wan2022adjointbackmap}; also see Fig.\ref{fig:Model:adjoint_ops_blueprint}) not being equal to $\mathbf{\theta}_{\mathcal{Y}}$. However, this challenge can be addressed if we treat the bias weight value itself as another component of the input that multiplies with a fixed unit weight (Fig.\ref{fig:Model:single_neuron_equivalence}).
	
	\begin{figure}[!h]  
		\centering
		\begin{tabular}{@{}c@{}}
			\includegraphics[width=1\textwidth]{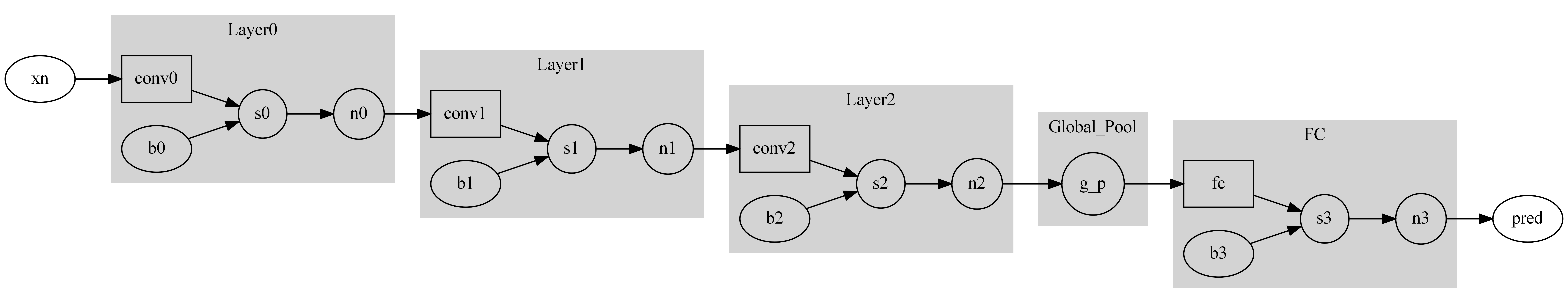} \\
			\includegraphics[width=1\textwidth]{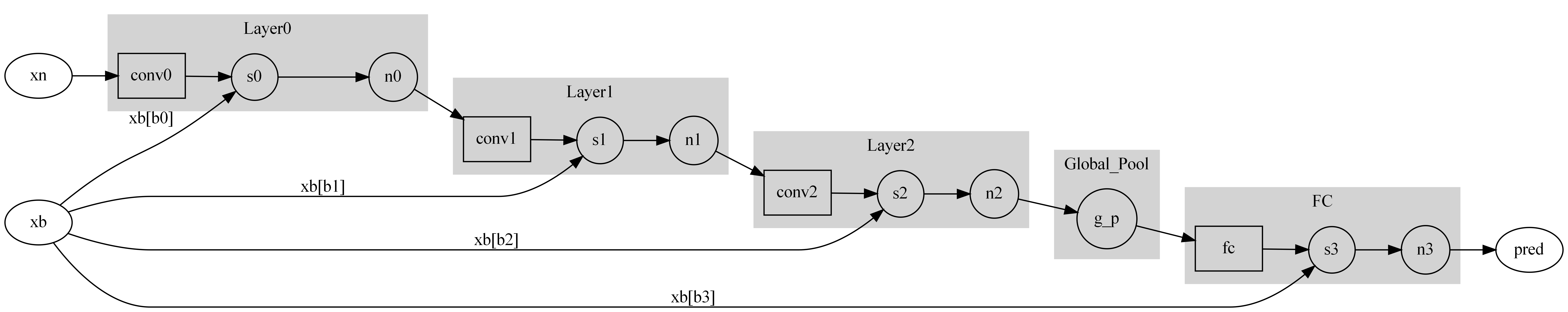} \\
		\end{tabular}
		\caption{An equivalent model of a trained 4-layer CNN. {\bf Top}: A CNN consists of three conv layers, one global pooling layer, and one FC layer. Once training was finished, all parameters would be fixed. {\bf Bottom}: We convert the trained CNN to an equivalent model. In this case, bias vectors from all layers are sequentially concatenated as a big vector, $\mathbf{x}_{b} = [\mathbf{b}_0; \mathbf{b}_1; \mathbf{b}_2; \mathbf{b}_3]$. That vector will be fed as an additional input tensor. Each layer picks its own bias from the $\mathbf{x}_b$ to compute. $\mathbf{x}_b[\mathbf{b}_i]$ denotes that the corresponding part for $\mathbf{b}_i$ will be recovered from the tensor $\mathbf{x}_b$. Note: We can do this since $\mathbf{x}_{b}$ is fixed after the CNN is trained, regardless of the input $\mathbf{x}_{n}$ that is presented during inference.}
		\label{fig:Model:cnn_with_bias_and_its_equivalence}
	\end{figure}
	
	Let $\mathcal{X}_{I}, \mathcal{X}_{b}$ be two subspaces containing the images and the bias vectors, respectively, and embedded in a big normed space $\mathcal{X}$, $\mathcal{X} = \mathcal{X}_{I} \bigtimes \mathcal{X}_{b}$. We consider an element-wise inner product (which induces the norm on $\mathcal{X}$) defined as,
	\begin{equation} \label{eq:inner_product}
		\begin{split}
			\langle [\mathbf{x}_{H \times W \times C}; \mathbf{b}_{L \times 1}] \mid [\mathbf{y}_{H \times W \times C}; \mathbf{d}_{L \times 1}] \rangle
			= \sum_{i=0}^{H-1} \sum_{j=0}^{W-1} \sum_{k=0}^{C-1} x_{i, j, k} y_{i, j, k} + \sum_{l=0}^{L-1}b_{l} d_{l} \\
			= \langle \mathbf{x}_{H \times W \times C} \mid \mathbf{y}_{H \times W \times C} \rangle + \langle \mathbf{b}_{L \times 1} \mid \mathbf{d}_{L \times 1} \rangle,
		\end{split}
	\end{equation}
	where $H, W, C$ are the height, the width, and the number of channels of an image; the first ``$=$'' is by definition, and we use the second ''$=$'' for tensor implementation.
	
	Suppose $\mathbf{F}$ denotes a CNN's forward computation path from an input to an $r_1 \times r_2$ receptive field in an in-channel (in-ch) feature map (represented by a normed space $\mathcal{Y}$); Bias vectors from all CNN's layers are sequentially concatenated to a big one, $\mathbf{x}_{b}$ (Fig.\ref{fig:Model:cnn_with_bias_and_its_equivalence}). $\mathbf{F}$ takes the combination of an image $\mathbf{x}_{n} \in \mathcal{X}_{I}$ and the bias vector $\mathbf{x}_{b} \in \mathcal{X}_{b}$ as the input. Then, a kernel $\mathbf{w}_{r_1 \times r_2}$ convolving on the receptive field, $\mathbf{F}([\mathbf{x}_{n}; \mathbf{x}_{b}])$, before adding the bias vector, can be described in a dual form \cite{banach1987theory, luenberger1997optimization},
	\begin{equation} \label{eq:approx}
		\begin{split}
			c([\mathbf{x}_{n}; \mathbf{x}_{b}]) = \mathbf{F}([\mathbf{x}_{n}; \mathbf{x}_{b}]) \circledast \mathbf{w}_{r_1 \times r_2} \\
			= \langle \mathbf{F}([\mathbf{x}_{n}; \mathbf{x}_{b}]), \mathbf{w}_{r_1 \times r_2} \rangle
			= \langle \mathbf{J}_{\mathbf{F}}(\mathbf{z}([\mathbf{x}_{n}; \mathbf{x}_{b}]))[\mathbf{x}_{n}; \mathbf{x}_{b}], \mathbf{w}_{r_1 \times r_2} \rangle,
		\end{split}
	\end{equation}
	where $c:\mathcal{X} \to \mathbb{R}$ is the activation value of a unit in the convolved feature map (before the bias compensation); $\mathbf{J}_{F}: \mathcal{X} \to B(\mathcal{X}, \mathcal{Y})$ is a Jacobian operator; The third ``='' holds when the CNN is activated with ReLU or Leaky ReLU, and $\exists k \in \mathbb{R}^{+}, \mathbf{z}([\mathbf{x}_{n}; \mathbf{x}_{b}]) = k[\mathbf{x}_{n}; \mathbf{x}_{b}]$ (proved in Section \ref{Appendix:Proof_of_eq_approx}). In fact, Eq.\eqref{eq:approx} here rewrites Eq.(3) in \cite{wan2022adjointbackmap} and inherits their intrinsic idea that $\mathbf{w}_{r_1 \times r_2}$ ($\in \mathcal{Y}^{*}$) is a local hyperplane on $\mathcal{Y}$. Also, we can remove the bias-free restriction by slightly extending the input space to take the bias vectors as part of the input (Fig.\ref{fig:Model:cnn_with_bias_and_its_equivalence}).
	
	For a fixed image $\mathbf{x}_{0}$ and the bias vector $\mathbf{x}_{b}$ extracted from a trained CNN, the Adjoint operator $\mathbf{J}^{*}_{\mathbf{F}}(\mathbf{z}([\mathbf{x}_{0}; \mathbf{x}_{b}]))$ will lead to the following equation:
	\begin{equation} \label{eq:adjoint}
		\begin{split}
			c([\mathbf{x}_{0}; \mathbf{x}_{b}])
			= \langle \mathbf{J}_{\mathbf{F}}(\mathbf{z}([\mathbf{x}_{0}; \mathbf{x}_{b}])) [\mathbf{x}_{0}; \mathbf{x}_{b}], \mathbf{w}_{r_1 \times r_2} \rangle
			= \langle [\mathbf{x}_{0}; \mathbf{x}_{b}] \mid \mathbf{J}^{*}_{\mathbf{F}}(\mathbf{z}([\mathbf{x}_{0}; \mathbf{x}_{b}])) \mathbf{w}_{r_1 \times r_2} \rangle \\
			= \langle [\mathbf{x}_{0}; \mathbf{x}_{b}] \mid \mathbf{J}_{\mathbf{F}}^{\intercal}(\mathbf{z}([\mathbf{x}_{0}; \mathbf{x}_{b}])) \mathbf{w}_{r_1 \times r_2} \rangle \\
			= \langle \mathbf{x}_{0} \mid (\mathbf{J}_{\mathbf{F}}^{\intercal}(\mathbf{z}([\mathbf{x}_{0}; \mathbf{x}_{b}]))[\mathbf{x}_{0}]) \mathbf{w}_{r_1 \times r_2} \rangle
			+ \langle \mathbf{x}_{b} \mid (\mathbf{J}_{\mathbf{F}}^{\intercal}(\mathbf{z}([\mathbf{x}_{0}; \mathbf{x}_{b}]))[\mathbf{x}_{b}]) \mathbf{w}_{r_1 \times r_2} \rangle
		\end{split}
	\end{equation}
	where $\mathbf{J}_{\mathbf{F}}^{\intercal}(\mathbf{z}([\mathbf{x}_{0}; \mathbf{x}_{b}]))$ is the real instance of the Adjoint operator $\mathbf{J}^{*}_{\mathbf{F}}(\mathbf{z}([\mathbf{x}_{0}; \mathbf{x}_{b}]))$; $\mathbf{J}_{\mathbf{F}}^{\intercal}(\mathbf{z}([\mathbf{x}_{0}; \mathbf{x}_{b}]))[\mathbf{x}_{0}], \mathbf{J}_{\mathbf{F}}^{\intercal}(\mathbf{z}([\mathbf{x}_{0}; \mathbf{x}_{b}]))[\mathbf{x}_{b}]$ intends to divide the adjoint operator into two parts (two operators) in response to $\mathbf{x}_{0}, \mathbf{x}_{b}$; The second ``$=$'' and the last ``$=$'' are possible due to the Riesz Representation theorem \cite{luenberger1997optimization} and the distributive property of a linear operator, respectively. Riesz Representation also contributes to the unification of $\mathcal{X}$ and $\mathcal{X}^{*}$ or $\mathcal{Y}$ and $\mathcal{Y}^{*}$. Therefore, we have $\mathbf{w}_{r_1 \times r_2} \in \mathcal{Y}$ and $\mathbf{J}^{*}_{\mathbf{F}}(\mathbf{z}([\mathbf{x}_{0}; \mathbf{x}_{b}])) \in B(\mathcal{Y}, \mathcal{X})$. The two operators (Eq.\eqref{eq:adjoint_to_2ops}) will map a convolution kernel from a convolutional layer or a weight vector from the FC layer back to the image subspace $\mathcal{X}_{I}$ and the bias subspace $\mathcal{X}_{b}$, respectively.
	\begin{equation} \label{eq:adjoint_to_2ops}
		\begin{split}
			\mathbf{J}^{*}_{\mathbf{F}}(\mathbf{z}([\mathbf{x}_{0}; \mathbf{x}_{b}]))[\mathbf{x}_{0}]\mathbf{w}_{r_1 \times r_2} \in \mathcal{X}_{I}, \\
			\mathbf{J}^{*}_{\mathbf{F}}(\mathbf{z}([\mathbf{x}_{0}; \mathbf{x}_{b}]))[\mathbf{x}_{b}]\mathbf{w}_{r_1 \times r_2} \in \mathcal{X}_{b}.
		\end{split}
	\end{equation}
    The two will jointly reconstruct an effective hypersurface representing all decision hyperplanes forward from the input to a unit of the out-ch feature map or an output activation value of the FC layer. We name this method AdjointBackMapV2. This method has the following properties:
	\begin{enumerate}
		\item The mapping $\mathbf{J}_{\mathbf{F}}(\mathbf{z}([\mathbf{x}; \mathbf{x}_{b}]))$ (i.e., $(\mathbf{J}_{\mathbf{F}}) \circ (\mathbf{z}): \mathcal{X} \to B(\mathcal{X}, \mathcal{Y})$) is not linear since $\exists \mathbf{x}_{1}, \mathbf{x}_{2} \in \mathcal{X}_{I}$, such that,
		$\mathbf{J}_{\mathbf{F}}(\mathbf{z}(\alpha[\mathbf{x}_1, \mathbf{x}_{b}] + \beta[\mathbf{x}_2, \mathbf{x}_{b}]))
		\ne \alpha \mathbf{J}_{\mathbf{F}}(\mathbf{z}([\mathbf{x}_1, \mathbf{x}_{b}])) + \beta \mathbf{J}_{\mathbf{F}}(\mathbf{z}([\mathbf{x}_2, \mathbf{x}_{b}]))$ 
		for two scalars $\alpha$ and $\beta$;
		
		\item Eq.\eqref{eq:adjoint_to_2ops} suggests a single effective hyperplane will have two components instead of two effective hyperplanes;
		
		\item A connection to the theory of \cite{wan2022adjointbackmap} comes from Eq.\eqref{eq:adjoint} where if all bias were zeroed out from the CNN's layers, the big norm space $\mathcal{X}$ would shrink to the $\mathcal{X}_{I}$ which is the same input space as in \cite{wan2022adjointbackmap}.
	\end{enumerate}

	\subsection{Algorithm}
	\label{Model:Algorithm}
\begin{figure}[!h]  
		\centering
		\begin{tabular}{@{}c@{}c@{}}
			\includegraphics[width=0.45\textwidth]{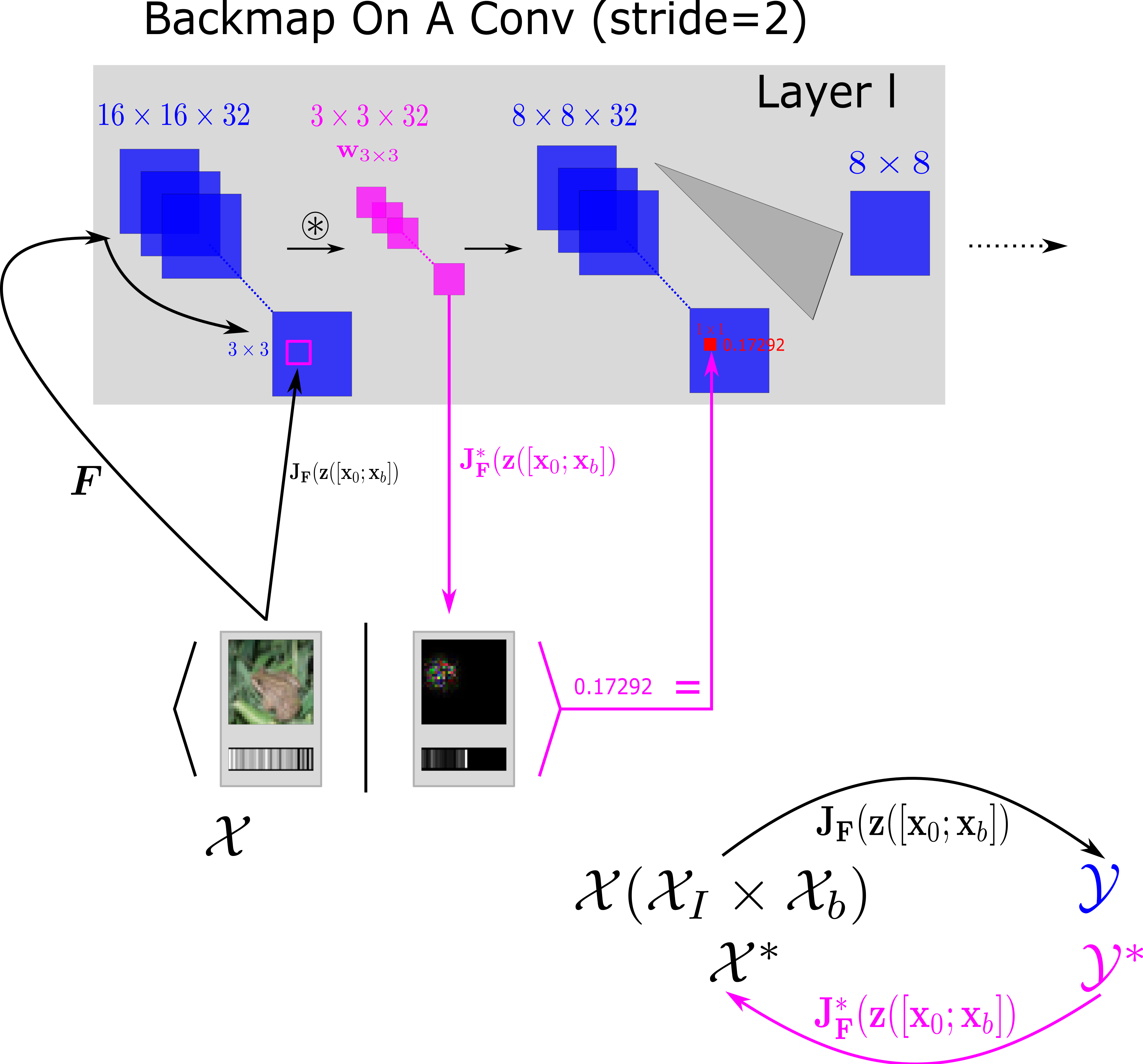} \quad &
			\includegraphics[width=0.45\textwidth]{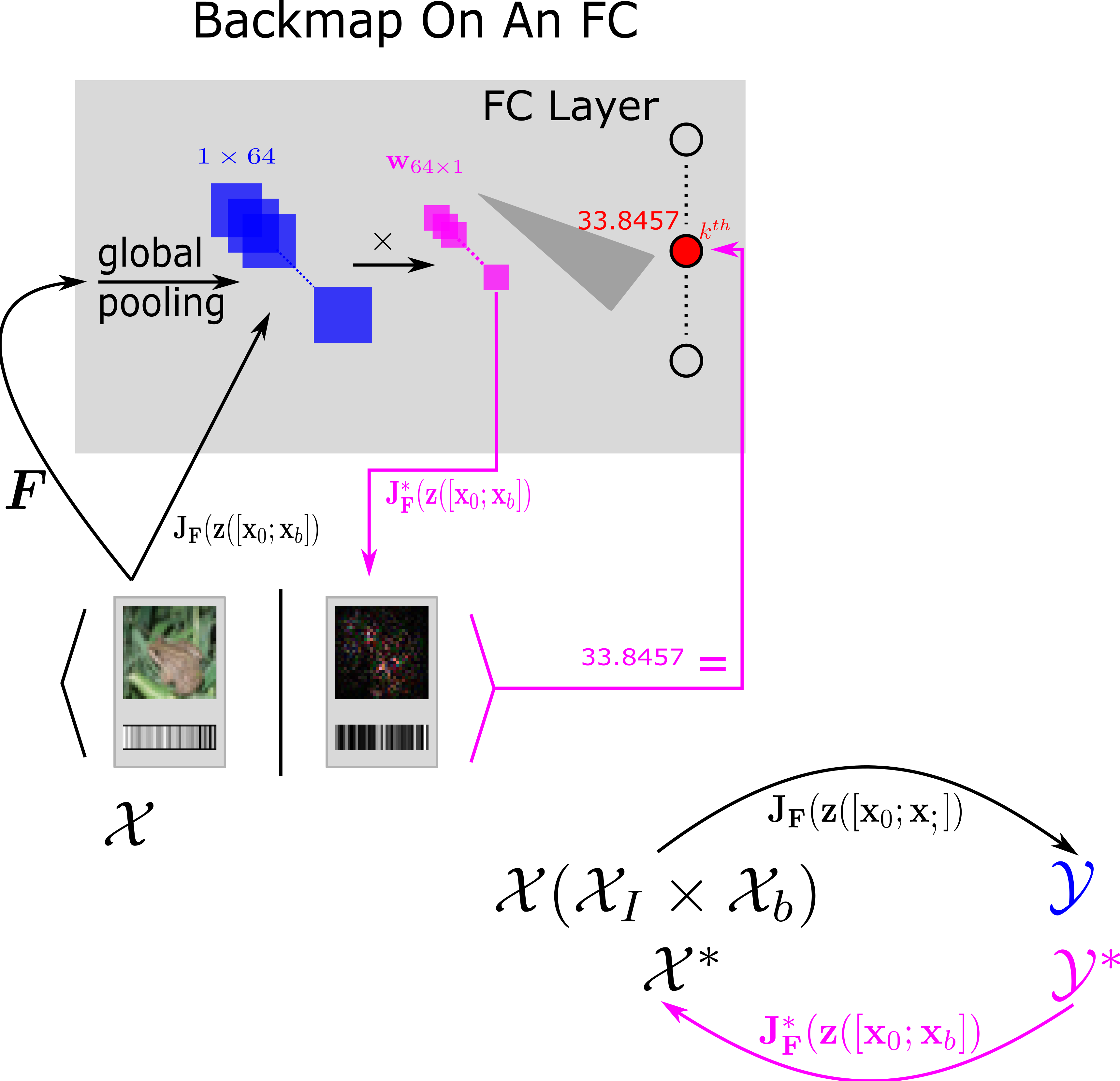} \\
			(a) & (b) \\
		\end{tabular}
		\caption{Principles of AdjointBackMapV2. Our reconstruction method applies to units in (a) Conv layers, and (b) FC layers. We color elements from the same normed space identically. The normed $\mathcal{X}$ is an input space that involves two embedded subspaces $\mathcal{X}_{I}$ and $\mathcal{X}_{b}$. Considering a fixed image $\mathbf{x}_{0}$ and the trained CNN (a fixed $\mathbf{x}_{b}$ from the bias values, shown below the input image above as a rectangle), $\mathbf{F}$ depicts the CNN's full forward computation from the input end to an $r_1 \times r_2$ receptive field in an in-ch feature map or the global pooling layer; Its adjoint operator, $\mathbf{J}^{*}_{\mathbf{F}}(\mathbf{z}([\mathbf{x}_{0}; \mathbf{x}_{b}]))$, projects the corresponding kernel or weights back to the input space $\mathcal{X}$; Riesz Representation unites $\mathcal{X}, \mathcal{X}^{*}$ or $\mathcal{Y}, \mathcal{Y}^{*}$, together. The symbol $\langle \cdot \mid \cdot \rangle$ refers to a dot-product defined in Eq.\eqref{eq:inner_product}; A symbol ``$=$'' means a dot-product between an input and the effective hyperplane is equal to the unit's activation value of a convolved feature map or an FC layer's predicted class as long as the CNN is activated with ReLU or Leaky ReLU. See Section \ref{Model:Theory} for a step-by-step walkthrough.}
		\label{fig:Model:adjoint_ops_blueprint}
	\end{figure}

	\paragraph{Target layers}
	Generally, AdjointBackMapV2 is designed to work on two types of layers inside a CNN. Fig.\ref{fig:Model:adjoint_ops_blueprint} illustrates our principles in detail.
	\begin{enumerate}
		\item Any kernel of a convolutional layer (except the kernels from the first layer, which have already been elements in $\mathcal{X}^{*}$) can be projected back to the joint space $\mathcal{X}$ concatenating the input image and the bias vectors. This back-mapped pattern fully determines the unit's activation value of an out-ch feature map, given an input image;
		\item Any weight vector of the FC layer can be projected back to the joint space $\mathcal{X}$ as well. This back-mapped pattern fully determines the activation value of a predicted output value, given an input image.
	\end{enumerate}
	
	\paragraph{Premise}
	Our algorithm's necessary condition is the third equality (``$=$'') between the left and right hand sides in Eq.\eqref{eq:approx}. Alternatively, any CNN unit holding eq.\eqref{eq:proof_of_eq_approx:piecewise_act} (Appendix) should satisfy our requirements. The proof in Section \ref{Appendix:Proof_of_eq_approx} reveals that any piecewise linear operations attached to a CNN's layers will not affect the equality. Thus, usual architectural techniques like shortcuts/concatenations/multiple receptive-field kernels' sizes are within the scope of our analysis. With this, typical CNNs activated with ReLUs or Leaky ReLUs can be analyzed with our reconstruction method. In general, we insist that the numerical precision should be considered when studying a CNN's inner workings, while conventional trials of shaping a kernel as a filter do not achieve this. Note that our method might not be appropriate for investigating a network activated with functions whose derivatives are not piecewise constants (such as tanh).

	\paragraph{Incorporating batch normalization}
	Usually, bias serves a CNN in two operational modes:
	\begin{enumerate}
		\item Acting as trainable parameters, attached after convolutions.
		\item Acting as moving parameters that channel-wise average batch values (BN, batch normalization \cite{ioffe2017batch}).
	\end{enumerate}
	 The first one has a similar network topology as Fig.\ref{fig:Model:cnn_with_bias_and_its_equivalence}, which is trivial for our theory to accommodate. The batch normalization case is more complex than the first. We will discuss it below. 
	
\begin{figure}[!h]
		\centering
		\begin{tabular}{@{}c@{}c@{}}
			\includegraphics[width=0.5\textwidth]{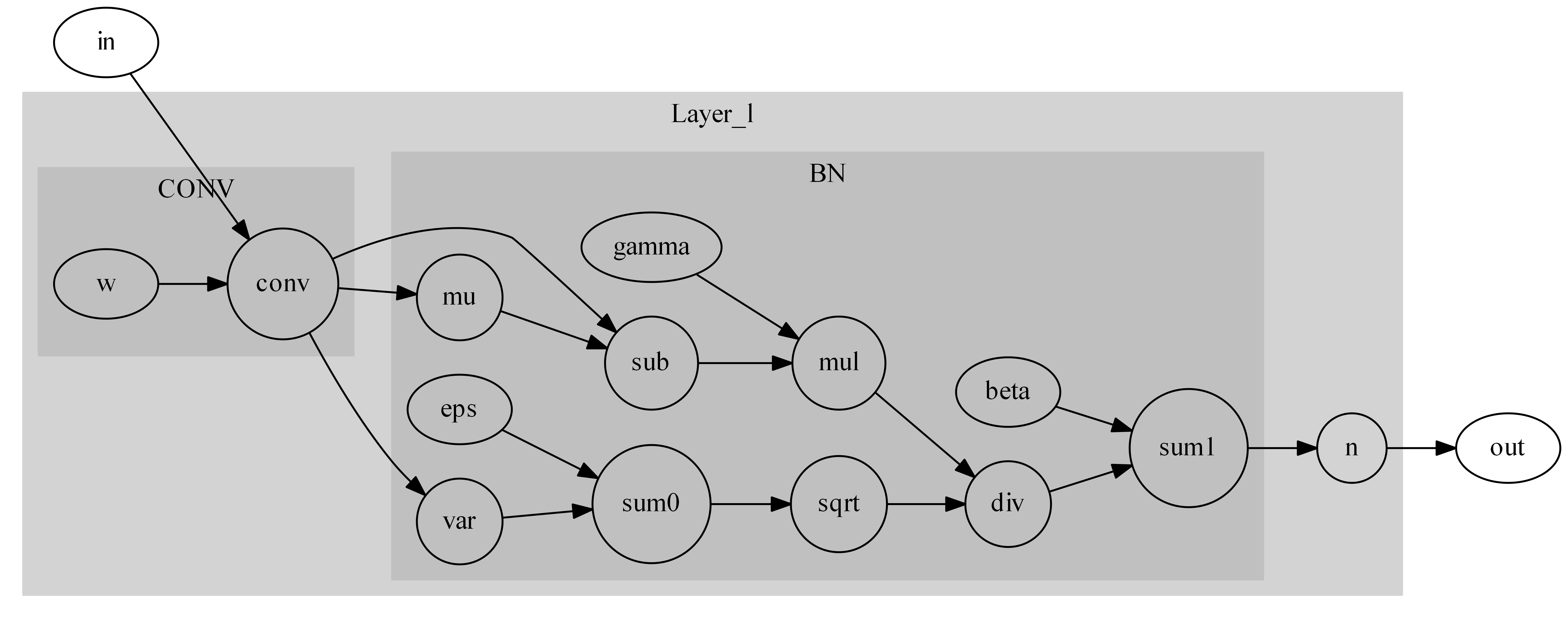} &
			\includegraphics[width=0.5\textwidth]{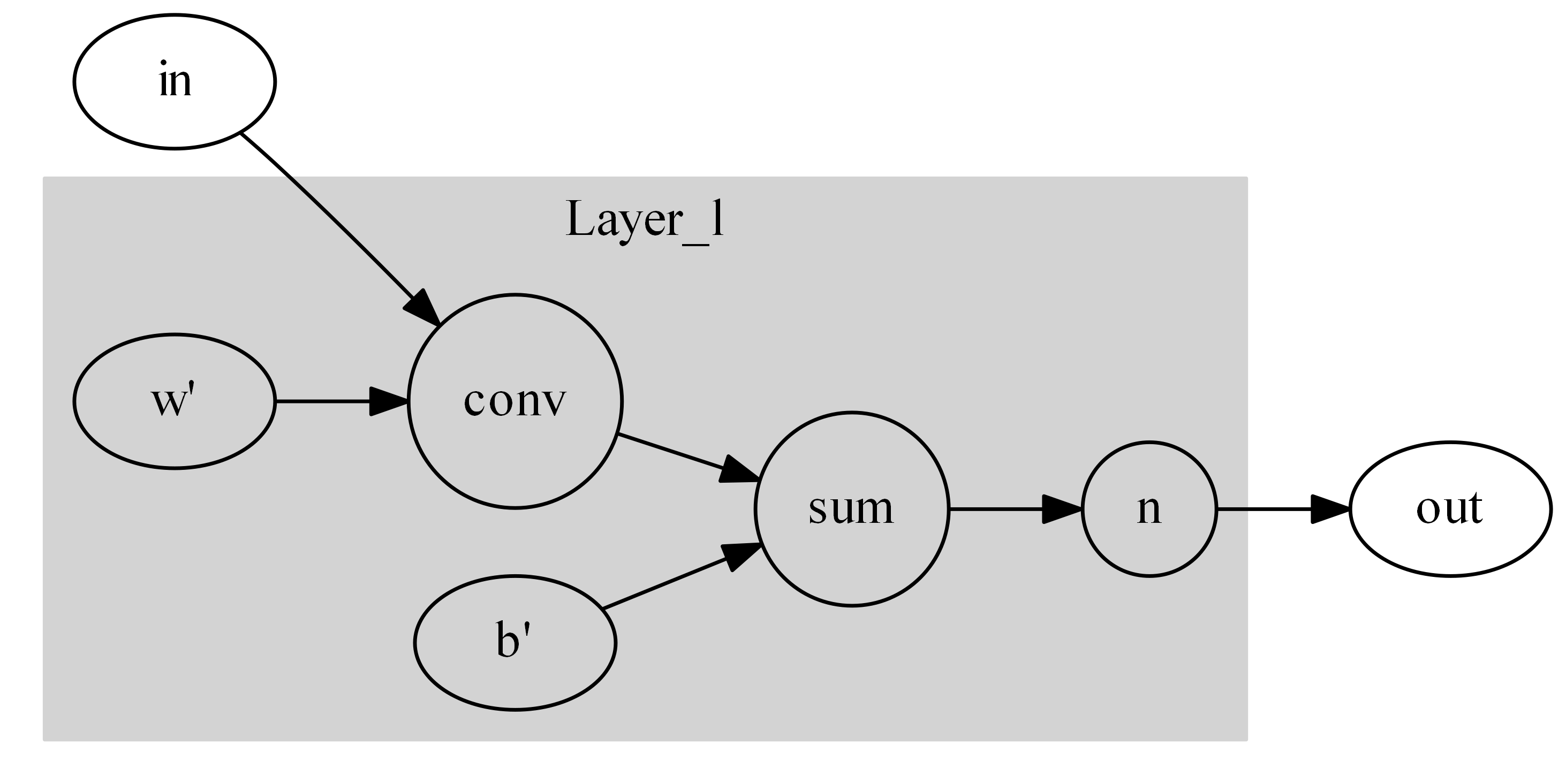} \\
			(a) & (b) \\
		\end{tabular}
		\caption{Batch normalization (BN) and its reduction. (a) A standard BN's implementation flow according to Eq.\eqref{eq:BN}; (b) Reduce the BN's complexity as a computational node similar to Fig.\ref{fig:Model:cnn_with_bias_and_its_equivalence} via Eq.\eqref{eq:BN_reduced}; After that, our theory can work on a CNN using BNs. Besides, this reduction lowers the CNN's computations and DRAM consumption.}
		\label{fig:Model:BN_reduced}
	\end{figure}

	In the ${l}^{th}$ layer, in-ch feature maps $\mathbf{x}_{l-1}$ will convolve with the layer's kernels $\mathbf{w}_{l}$. The moving mean vector, $\mathbf{\mu}$, and moving variance vector, $\mathbf{\sigma}^{2}$, will normalize the convolved results before activating the layer's unit. We summarize the BN as:
	\begin{equation} \label{eq:BN}
		\mathbf{BN}(\mathbf{x}_{l-1}) = \mathbf{\gamma} \times \frac{(\mathbf{x}_{l-1} \circledast \mathbf{w}_{l} - \mathbf{\mu})}{\sqrt{\mathbf{\sigma}^{2} + \epsilon}} + \mathbf{\beta},
	\end{equation}
	where $\mathbf{\gamma}, \mathbf{\beta}$ are two learnable parameters that weight the feature maps channel-wise; a preset $\epsilon$ prevents any divide by zero exceptions (TensorFlow \cite{abadi2016tensorflow} sets $\epsilon = 0.001$). We can reduce its complexity using Eq.\eqref{eq:BN_reduced} (Fig.\ref{fig:Model:BN_reduced}),
	\begin{equation} \label{eq:BN_reduced}
		\begin{split}
			\mathbf{BN}([\mathbf{x}_{l-1}; \mathbf{b}^{'}]) = \mathbf{x}_{l-1} \circledast \mathbf{w}_{l}^{'} + \mathbf{b}^{'} \\
			\mathbf{w}_{l}^{'} = \frac{(\mathbf{\gamma} \times \mathbf{w}_{l})}{\sqrt{\mathbf{\sigma}^{2} + \epsilon}}, \;
			\mathbf{b}{'} = (\mathbf{\beta} - \frac{\mathbf{\gamma} \times \mathbf{\mu}}{\sqrt{\mathbf{\sigma}^{2} + \epsilon}}).
		\end{split}
	\end{equation}
	(Note again that this is possible since we are working with a fully trained CNN where all the weights and the BN parameters are learned and fixed.)
	This suggests that a CNN trained to use BNs is equivalent to a similar network graph illustrated in Fig.\ref{fig:Model:cnn_with_bias_and_its_equivalence} when the reduced $\mathbf{w}_{l}^{'}$ and $\mathbf{b}^{'}$ are used in place of the trained CNN. Therefore, applying our theory to a CNN with BNs is as trivial as the first operational mode. Besides, Eq.\eqref{eq:BN_reduced} significantly reduces both computations and DRAM consumption since fewer multiplications and parameters are required when compared to Eq.\eqref{eq:BN}. We will elaborate more on our experiments below.

\begin{figure}[!h]
		\centering
		\begin{tabular}{@{}c@{}c@{}}
			\includegraphics[height=0.4\textwidth]{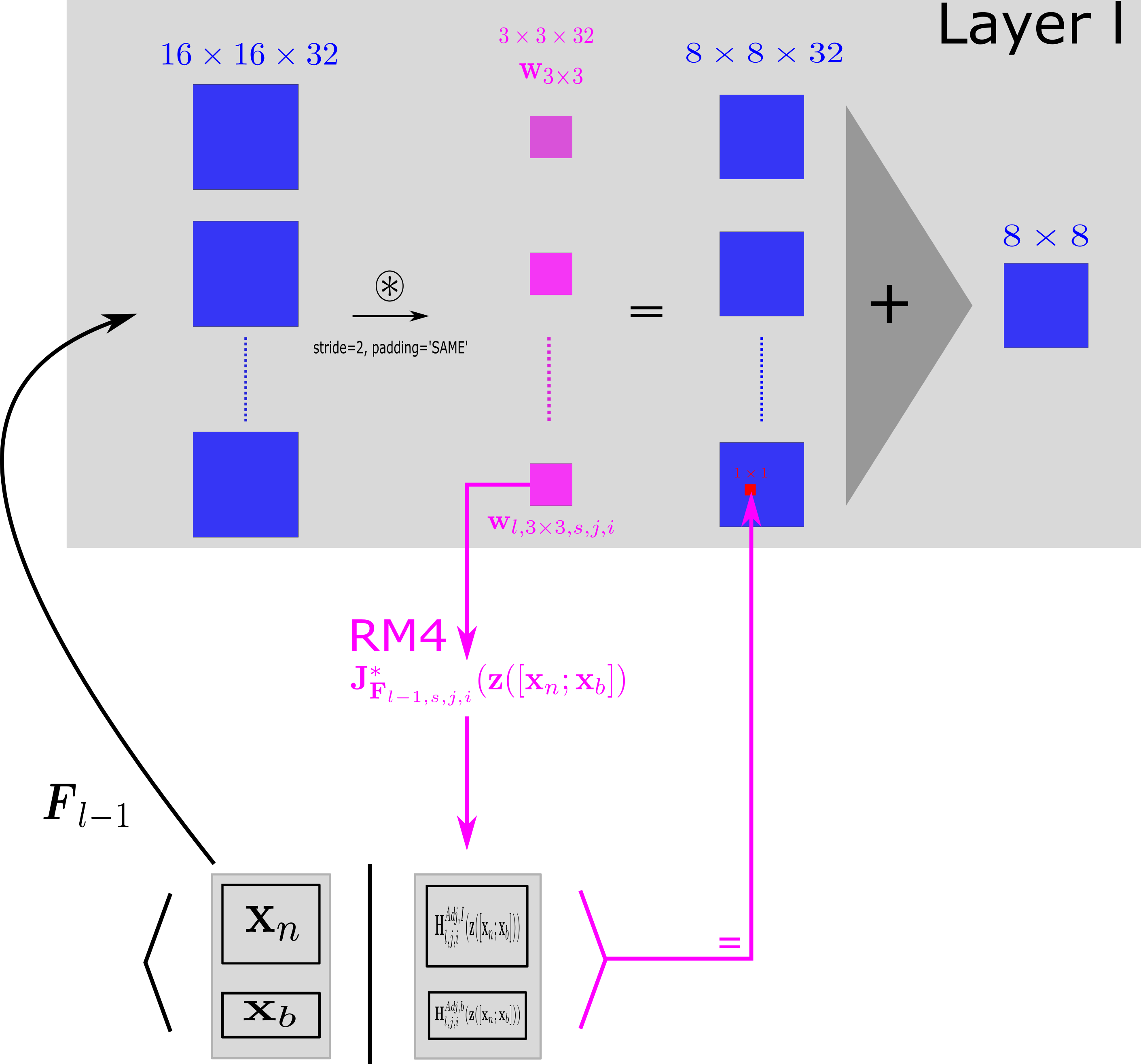} \quad &
			\includegraphics[height=0.4\textwidth]{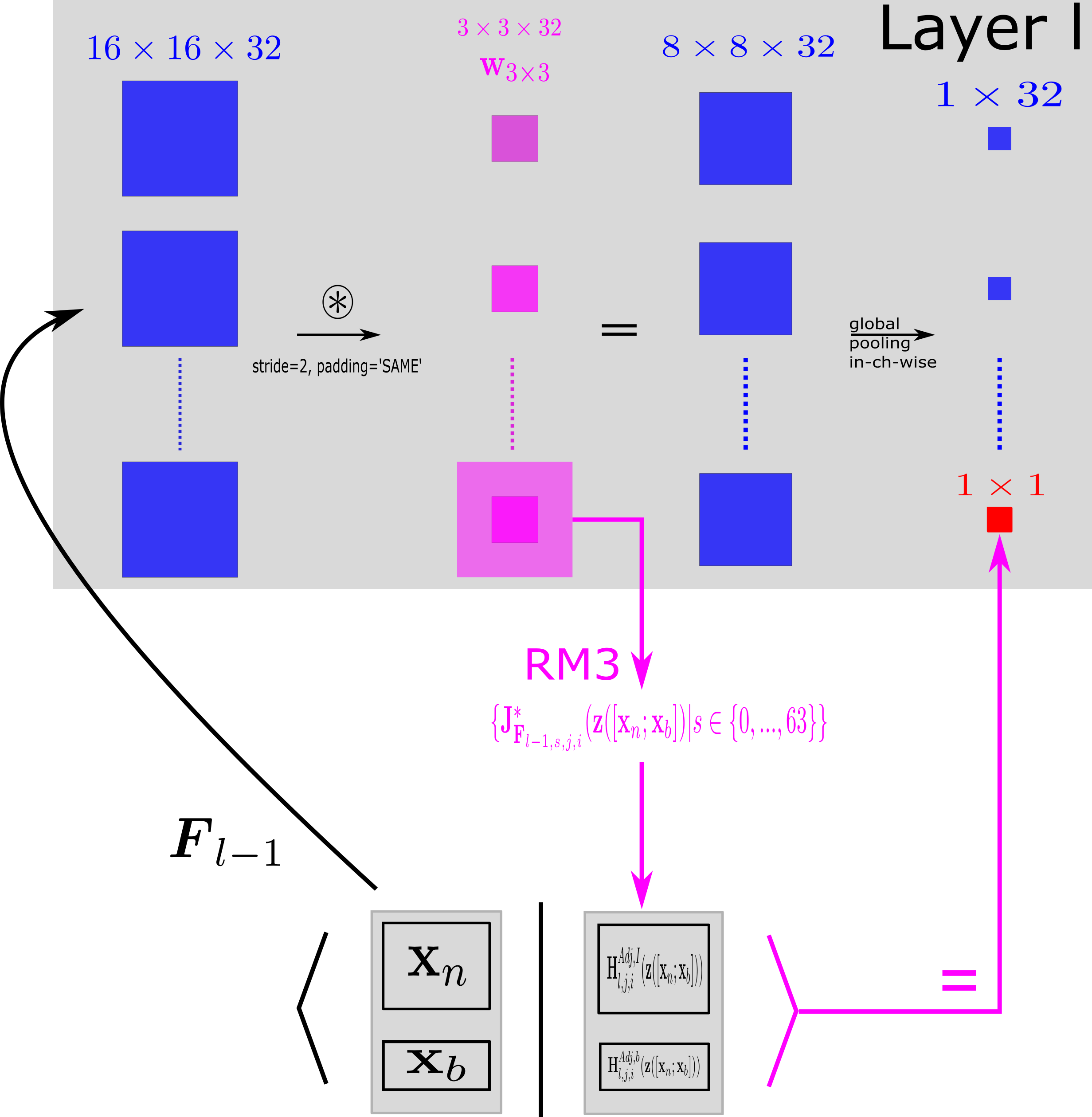} \\
			(a) $RM4$ & (b) $RM3$ \\
			\includegraphics[height=0.4\textwidth]{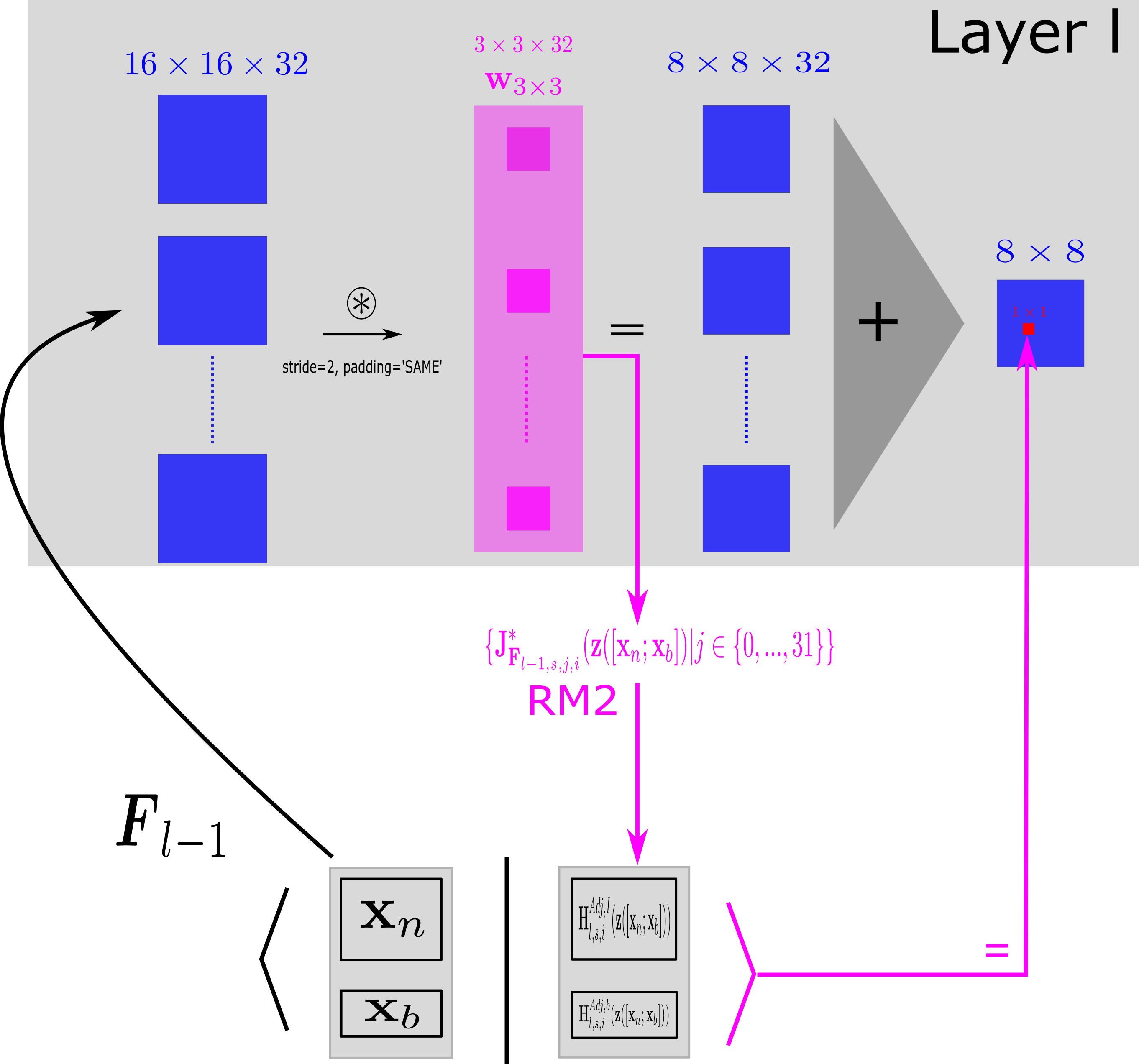} \quad &
			\includegraphics[height=0.4\textwidth]{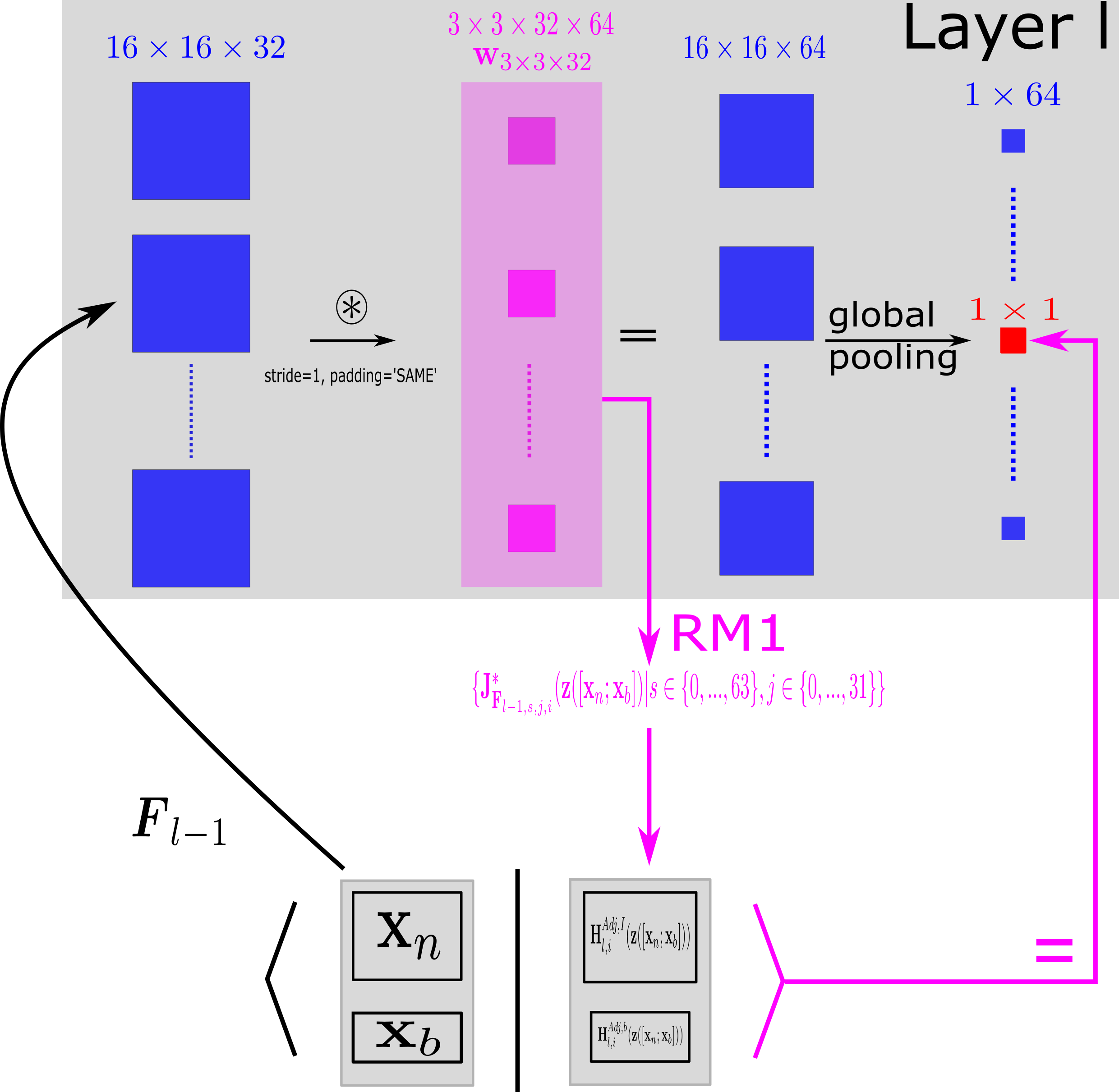} \\
			(c) $RM2$ & (d) $RM1$ \\
		\end{tabular}
		\caption{Two factors (in-ch merge \& global pooling (g\_p)) determine four RMs involved in the Conv layers. (a) $RM4$: Without either in-ch merge or g\_p; (b) $RM3$: Without in-ch merge and with g\_p; (c) $RM2$: With in-ch merge but without g\_p; (d) $RM1$: With both in-ch merge and g\_p. The usage of colors is similar to Fig.\ref{fig:Model:adjoint_ops_blueprint}. An oversized pink mask in either (b) or (d) shows an effective hypersurface reconstructed in response to a stride-wise summation of the convolved feature map. The symbol $\langle \cdot \mid \cdot \rangle$ refers to the inner product defined in Eq.\eqref{eq:inner_product}.}
		\label{fig:Model:two_factors}
	\end{figure}

	\paragraph{Five Reconstruction Modes (RMs)}
	Our AdjointBackMapV2 upgrades the five reconstruction modes (RMs) inherited from \cite{wan2022adjointbackmap} for reconstructing an effective hypersurface, depending on the location of the unit in the CNN, and on the convolution operation's variant: $RM0$ will project from the FC layer, and the remaining four ($RM4 \sim RM1$) will project from a convolutional layer. Generally, two factors will distinguish one RM from others, for $RM4 \sim RM1$:
	\begin{enumerate}
		\item With or without in-ch merge during convolution;
		\item With or without global pooling.
	\end{enumerate}
	All RMs and their algebraic relationships between the forward and backward computation paths are illustrated in Fig.\ref{fig:Model:two_factors}. Note: Although the factors for classifying this section's RMs are similar to those mentioned in \cite{wan2022adjointbackmap}, their principles are significantly different from \cite{wan2022adjointbackmap}'s due to bias being considered.
	
	 To help keep track of the steps in the following, we elaborate on an example using concrete image and kernel sizes. Suppose a CNN activated with Leaky ReLUs has been trained on the CIFAR-10. It takes a $32 \times 32 \times 3$ (height $\times$ width $\times$ channels) RGB image $\mathbf{x}_{n}$ to predict 10 classes. Its ${l}^{th}$ conv layer has kernels $\mathbf{w}_{l, 3 \times 3 \times 32 \times 64}$ (h $\times$ w $\times$ in-chs $\times$ out-chs) convolving on $16 \times 16 \times 32$ in-ch feature maps with the training stride $= 2$ and padding = ``SAME'' \cite{abadi2016tensorflow}. The implementation of 2-D convolution \cite{abadi2016tensorflow} states that a kernel $\mathbf{w}_{3 \times 3}$ only convolves on its corresponding $16 \times 16$ in-ch feature map through $8 \times 8$ moves (a stride move $s$ ranges from $0$ to $63$); Then, all $32$ convolved feature maps are added together in-channel-wise (in-ch summation) as an $8 \times 8$ out-ch feature map that will flow to fuse with bias. Thus, the $l^{th}$ layer is supposed to produce an $8 \times 8 \times 64$ out-ch feature maps. Besides, it may have a global pooling \cite{lin2013network} (g\_p) applied right after the activated feature maps for the FC layer. 
	
	$\mathbf{RM4}$ (Fig.\ref{fig:Model:two_factors}(a)): Neither merging nor global pooling is performed on either in-ch kernels or training strides. In this case, a kernel will be individually mapped via $\mathbf{H}^{Adj}_{l, s, j, i}$, composed of $\mathbf{H}^{Adj, I}_{l, s, j, i}$ and $\mathbf{H}^{Adj, b}_{l, s, j, i}$, to space $\mathcal{X}$.
	\begin{equation} \label{eq:RM4}
		\begin{split}
			\mathbf{H}^{Adj, I}_{l, s, j, i}(\mathbf{z}([\mathbf{x}_{n}; \mathbf{x}_{b}]))
			= (\mathbf{J}^{*}_{\mathbf{F}_{l-1, s, j, i}}(\mathbf{z}([\mathbf{x}_{n}; \mathbf{x}_{b}]))[\mathbf{x}_{n}]) \mathbf{w}_{l, 3 \times 3, s, j, i}, \\
			\mathbf{H}^{Adj, b}_{l, s, j, i}(\mathbf{z}([\mathbf{x}_{n}; \mathbf{x}_{b}]))
			= (\mathbf{J}^{*}_{\mathbf{F}_{l-1, s, j, i}}(\mathbf{z}([\mathbf{x}_{n}; \mathbf{x}_{b}]))[\mathbf{x}_{b}]) \mathbf{w}_{l, 3 \times 3, s, j, i} \\
			s \in \{ 0, 1, ..., 63 \}, j \in \{ 0, 1, .., 31 \}, i \in \{ 0, 1, ..., 63 \},
		\end{split}
	\end{equation}
	where $\mathbf{F}_{l-1, s, j, i}$ denotes the forward path from the input end to an in-ch $3 \times 3$ receptive field that will be convolved to the $(j, i)$ out-ch unit at the stride location $s$. $\mathbf{H}^{Adj}_{l, s, j, i}$ reflects how the local kernel weights on the combined image and bias input when the stride moves. $\langle [\mathbf{x}_{n}; \mathbf{x}_{b}] \mid [\mathbf{H}^{Adj, I}_{l, s, j, i}(\mathbf{z}([\mathbf{x}_{n}; \mathbf{x}_{b}])); \mathbf{H}^{Adj, b}_{l, s, j, i}(\mathbf{z}([\mathbf{x}_{n}; \mathbf{x}_{b}])) \rangle$ should be equal to the unit's activation value before the in-ch summation. Fig.\ref{fig:Model:two_factors}(a) shows the details.
	
	$\mathbf{RM3}$ (Fig.\ref{fig:Model:two_factors}(b)): No merging is taken on any in-ch kernel's convolution. However, global pooling will be applied for mapping, i.e., the effective hypersurfaces from an individual kernel sum together pixel-wise to reconstruct an effective hypersurface, $\mathbf{H}^{Adj}_{l, j, i}$. That effective hypersurface describes the local kernel's weighting in the input space, considering all stride moves merged. In other words, it reveals how a feature sum could be generated from the space $\mathcal{X}$'s perspective, when an in-channel convolved feature map is pooling globally, i.e.,
	\begin{equation} \label{eq:derive_RM3}
		\begin{split}
			\sum_{s=0}^{63} \mathbf{F}_{l-1, s, j, i}([\mathbf{x}_{n}, \mathbf{x}_{b}]) \circledast \mathbf{w}_{l, 3 \times 3, s, j, i}
			= \langle [\mathbf{x}_{n}; \mathbf{x}_{b}], \mathbf{H}^{Adj}_{l, j, i}(\mathbf{z}([\mathbf{x}_{n}; \mathbf{x}_{b}])) \rangle \\
			= \langle \mathbf{x}_{n}, \sum_{s=0}^{63}(\mathbf{J}^{*}_{\mathbf{F}_{l-1, s, j, i}}(\mathbf{z}([\mathbf{x}_{n}; \mathbf{x}_{b}]))[\mathbf{x}_{n}]) \mathbf{w}_{l, 3 \times 3, s, j, i} \rangle \\
			+ \langle \mathbf{x}_{b}, \sum_{s=0}^{63}(\mathbf{J}^{*}_{\mathbf{F}_{l-1, s, j, i}}(\mathbf{z}([\mathbf{x}_{n}; \mathbf{x}_{b}]))[\mathbf{x}_{b}]) \mathbf{w}_{l, 3 \times 3, s, j, i} \rangle,
		\end{split}
	\end{equation}
	where the distributive law and linearity in dual space support the last ``$=$''. Thus, $\mathbf{H}^{Adj}_{l, j, i}$ is composed of two operators: $\mathbf{H}^{Adj, I}_{l, j, i}$ and $\mathbf{H}^{Adj, b}_{l, j, i}$,
	\begin{equation} \label{eq:RM3}
		\begin{split}
			\mathbf{H}^{Adj, I}_{l, j, i}(\mathbf{z}([\mathbf{x}_{n}; \mathbf{x}_{b}]))
			= \sum_{s=0}^{63}\mathbf{H}^{Adj, I}_{l, s, j, i}(\mathbf{z}([\mathbf{x}_{n}; \mathbf{x}_{b}])), \\
			\mathbf{H}^{Adj, b}_{l, j, i}(\mathbf{z}([\mathbf{x}_{n}; \mathbf{x}_{b}]))
			= \sum_{s=0}^{63}\mathbf{H}^{Adj, b}_{l, s, j, i}(\mathbf{z}([\mathbf{x}_{n}; \mathbf{x}_{b}])). \\
		\end{split}
	\end{equation}
	The relationship to $RM4$ is obvious in Eq.\eqref{eq:RM3} as well (i.e., summation of Eq.\eqref{eq:RM4} over $s$ gives Eq.\eqref{eq:RM3}).
	
	$\mathbf{RM2}$ (Fig.\ref{fig:Model:two_factors}(c)): We do not apply the global pooling. Instead, the kernels' back maps will merge in-channel-wise to reconstruct an effective hypersurface $\mathbf{H}^{Adj}_{l, s, i}$ that determines the unit's activation value of an out-ch feature map. Its two operators, $\mathbf{H}^{Adj, I}_{l, s, i}$ and $\mathbf{H}^{Adj, b}_{l, s, i}$, then become:
	\begin{equation} \label{eq:RM2}
		\begin{split}
			\mathbf{H}^{Adj, I}_{l, s, i}(\mathbf{z}([\mathbf{x}_{n}; \mathbf{x}_{b}]))
			= \sum_{j=0}^{31}\mathbf{H}^{Adj, I}_{l, s, j, i}(\mathbf{z}([\mathbf{x}_{n}; \mathbf{x}_{b}])), \\
			\mathbf{H}^{Adj, b}_{l, s, i}(\mathbf{z}([\mathbf{x}_{n}; \mathbf{x}_{b}]))
			= \sum_{j=0}^{31}\mathbf{H}^{Adj, b}_{l, s, j, i}(\mathbf{z}([\mathbf{x}_{n}; \mathbf{x}_{b}])). \\
		\end{split}
	\end{equation} 
	Eq.\eqref{eq:RM2} also relates $RM2$ to $RM4$ (i.e., summation of Eq.\eqref{eq:RM4} over $j$ gives Eq.\eqref{eq:RM2}).
	
	$\mathbf{RM1}$ (Fig.\ref{fig:Model:two_factors}(d)): We map with both merging and global pooling considered. Then, a reconstruction will be conducted using $\mathbf{H}^{Adj}_{l, i}$ that determines the activation value of the feature maps for the global-pooling layer (Fig.\ref{fig:Model:two_factors}(b)). Its two parts, $\mathbf{H}^{Adj, I}_{l, i}$ and $\mathbf{H}^{Adj, b}_{l, i}$, and their relationships to $RM3$, $RM2$ are summarized below.
	\begin{equation} \label{eq:RM1}
		\begin{split}
			\mathbf{H}^{Adj, I}_{l, i}(\mathbf{z}([\mathbf{x}_{n}; \mathbf{x}_{b}]))
			= \sum_{j=0}^{31} \mathbf{H}^{Adj, I}_{l, j, i}(\mathbf{z}([\mathbf{x}_{n}; \mathbf{x}_{b}]))
			= \sum_{s=0}^{63} \mathbf{H}^{Adj, I}_{l, s, i}(\mathbf{z}([\mathbf{x}_{n}; \mathbf{x}_{b}])) \\
			\mathbf{H}^{Adj, b}_{l, i}(\mathbf{z}([\mathbf{x}_{n}; \mathbf{x}_{b}]))
			= \sum_{j=0}^{31} \mathbf{H}^{Adj, b}_{l, j, i}(\mathbf{z}([\mathbf{x}_{n}; \mathbf{x}_{b}]))
			= \sum_{s=0}^{63} \mathbf{H}^{Adj, b}_{l, s, i}(\mathbf{z}([\mathbf{x}_{n}; \mathbf{x}_{b}])). \\
		\end{split}
	\end{equation}
	
	$\mathbf{RM0}$ (Fig.\ref{fig:Model:adjoint_ops_blueprint}(b)): Mapping an FC weight vector $\mathbf{w}_{k}$ is independent of the factors that govern the convolution operation. An effective hypersurface $\mathbf{H}^{Adj}_{k}(\mathbf{z}([\mathbf{x}_{n}; \mathbf{x}_{b}]))$ reconstructed in this way represents the whole decision process towards a predicted value for the $k^{th}$ class. $\mathbf{H}^{Adj}_{k}$ consists of two operators $\mathbf{H}^{Adj, I}_{k}$ and $\mathbf{H}^{Adj, b}_{k}$ ($k \in \{ 0, 1, .., 9 \}$) as well.
	\begin{equation} \label{eq:RM0}
		\begin{split}
			\mathbf{H}^{Adj, I}_{k}(\mathbf{z}([\mathbf{x}_{n}; \mathbf{x}_{b}]))
			= (\mathbf{J}^{*}_{\mathbf{F}_{k}}(\mathbf{z}([\mathbf{x}_{n}; \mathbf{x}_{b}]))[\mathbf{x}_{n}]) \mathbf{w}_{k}, \\
			\mathbf{H}^{Adj, b}_{k}(\mathbf{z}([\mathbf{x}_{n}; \mathbf{x}_{b}]))
			= (\mathbf{J}^{*}_{\mathbf{F}_{k}}(\mathbf{z}([\mathbf{x}_{n}; \mathbf{x}_{b}]))[\mathbf{x}_{b}]) \mathbf{w}_{k}.
		\end{split}
	\end{equation}
	
	\paragraph{Implementation}
	Computing a Jacobian is expensive, so we use convolution to accelerate our effective hypersurface reconstruction. Eq.\eqref{eq:RM4} $\sim$ \eqref{eq:RM0} are optimized and compiled in Algorithm.\ref{algorithm:RM4_to_RM0}. Padding should be identical to training. Tensorflow \cite{abadi2016tensorflow} functions: matmul, unstack, stack, expanddim, conv2d, sum, are used in the algorithm. The transpose in Eq.\eqref{eq:adjoint} is achieved via an 'axis' option of the conv2d function. From the algorithm, we can see how the different reconstruction modes (RMs) are related.
 
    \scalebox{0.9}{
    \begin{minipage}{1.0\linewidth}
        \begin{algorithm}[H]
    		\DontPrintSemicolon
    		\caption{AdjointBackMapV2 with five modes ($RM_4$ to $RM_0$)}
    		\label{algorithm:RM4_to_RM0}
    		\KwIn{1. $\mathbf{x}$: input image (shape: $d_1 = H \times W \times C$); 2. $\mathbf{b}$: bias vector concatenated from a trained model (shape: $d_2$); 3. $\mathbf{z}$: the function for Eq.\eqref{eq:approx}; 4. $\mathbf{F}_{l}$: CNN's forward mapping leading to in-channels of the $l^{th}$ layer; 5. $\mathbf{w}_l$: the weights at the $l^{th}$ layer; 6. $s$: the convolution stride; 7. $M$: RM selection.}
    		\KwOut{Two parts of an effective hypersurface: $\mathbf{H}^{Adj, I}(\mathbf{z}([\mathbf{x}; \mathbf{b}])), \mathbf{H}^{Adj, b}(\mathbf{z}([\mathbf{x}; \mathbf{b}]))$}
    		\SetKwProg{Fn}{Function}{:}{}
    		\Fn{AdjointBackMapV2($\mathbf{x}, \mathbf{b}, \mathbf{z}, \mathbf{F}, \mathbf{w}_l, s, L$)}
    		{
    			$\mathbf{z}_{0} = \mathbf{z}([\mathbf{x}; \mathbf{b}])$ \\
    			\If(\tcp*[f]{1. FC layer ($RM_0$)}){$M$ is `$RM_0$'}
    			{
    				$\mathbf{J}_{\mathbf{F}, \mathbf{x}, d_1 \times c_{g\_p, in}}, \mathbf{J}_{\mathbf{F}, \mathbf{b}, d_2 \times c_{g\_p, in}} = \frac{\partial \mathbf{F}_{c_{g\_p, in}}}{\partial \mathbf{x}}, \frac{\partial \mathbf{F}_{c_{g\_p, in}}}{\partial \mathbf{b}}$ \\
    				
    				\Return matmul($\mathbf{J}_{\mathbf{F}, \mathbf{x}, d_1 \times c_{g\_p, in}}(\mathbf{z_{0}})$, $\mathbf{w}_{fc, c_{g\_p, in} \times c_{labels}}$, axis=`$c_{g\_p, in}$'), matmul($\mathbf{J}_{\mathbf{F}, \mathbf{b}, d_2 \times c_{g\_p, in}}(\mathbf{z_{0}})$, $\mathbf{w}_{fc, c_{g\_p, in} \times c_{labels}}$, axis=`$c_{g\_p, in}$') \\
    			}
    			$\mathbf{J}_{\mathbf{F}, \mathbf{x}, d_1 \times h_l \times w_l \times c_{l, in}}, \mathbf{J}_{\mathbf{F}, \mathbf{b}, d_2 \times h_l \times w_l \times c_{l, in}} = \frac{\partial \mathbf{F}_{h_l \times w_l \times c_{l, in}}}{\partial \mathbf{x}}, \frac{\partial \mathbf{F}_{h_l \times w_l \times c_{l, in}}}{\partial \mathbf{b}}$ \\
    			\Switch{$M$}
    			{
    				\Case(\tcp*[f]{2. Without in-ch merge}){`$RM_4$' or `$RM_3$'}
    				{
    					$\mathbf{w}_{r_1 \times r_2 \times c_{l, out}} =$ unstack($\mathbf{w}_{l, r_1 \times r_2 \times c_{l, in} \times c_{l, out}}$, axis=`$c_{l, in}$') \\
    					$\mathbf{J}_{\mathbf{F}, \mathbf{x}, d_1 \times h_l \times w_l} = $ unstack($\mathbf{J}_{\mathbf{F}, \mathbf{x}, d_1 \times h_l \times w_l \times c_{l, in}}$, axis=`$c_{l, in}$') \\
    					$\mathbf{J}_{\mathbf{F}, \mathbf{b}, d_2 \times h_l \times w_l} = $ unstack($\mathbf{J}_{\mathbf{F}, \mathbf{b}, d_2 \times h_l \times w_l \times c_{l, in}}$, axis=`$c_{l, in}$') \\
    					$j = 0$, Empty container $R_x$, $R_b$ \\
    					\While{$j < c_{l, in}$}
    					{
    						$\mathbf{J}_{\mathbf{F}, \mathbf{x}}=$ expanddim($\mathbf{J}_{\mathbf{F}, \mathbf{x}, d_1 \times h_l \times w_l}[j]$, axis=`$c_{l, in}$') \\
    						$\mathbf{J}_{\mathbf{F}, \mathbf{b}}=$ expanddim($\mathbf{J}_{\mathbf{F}, \mathbf{b}, d_2 \times h_l \times w_l}[j]$, axis=`$c_{l, in}$') \\
    						$\mathbf{w} = $ expanddim($\mathbf{w}_{r_1 \times r_2 \times c_{l, out}}[j]$, axis=`$c_{l, in}$') \\
    						$R_x$.append(conv2d($\mathbf{J}_{\mathbf{F}, \mathbf{x}}(\mathbf{z}_{0})$, $\mathbf{w}$, stride=$s$, axis=`$(h_l, w_l, c_{l, in}, c_{l, out})$')) \\
    						$R_b$.append(conv2d($\mathbf{J}_{\mathbf{F}, \mathbf{b}}(\mathbf{z}_{0})$, $\mathbf{w}$, stride=$s$, axis=`$(h_l, w_l, c_{l, in}, c_{l, out})$')) \\
    						$j = j + 1$ \\
    					}
    					$\mathbf{H}_{I}, \mathbf{H}_{b} = $ stack($R_x$, axis=`$c_{l, in}$'), stack($R_b$, axis=`$c_{l, in}$') \\
    					\If(\tcp*[f]{2.1 without g\_p ($RM_4$)}){$M$ is `$RM_4$'}
    					{
    						\Return $\mathbf{H}_{I}$, $\mathbf{H}_{b}$
    					}\ElseIf(\tcp*[f]{2.2 with g\_p ($RM_3$)}){$M$ is `$RM_3$'}
    					{
    						\Return sum($\mathbf{H}_{I}$, axis=`$(h_{l, o}, w_{l, o})$'), sum($\mathbf{H}_{b}$, axis=`$(h_{l, o}, w_{l, o})$')\\
    					}
    				}
    				\Case(\tcp*[f]{3. With in-ch merge}){`$RM_2$' or `$RM_1$'}
    				{
    					$\mathbf{H}_{I} = $ conv2d($\mathbf{J}_{\mathbf{F}, \mathbf{x}, d_1 \times h_l \times w_l \times c_{l, in}}(\mathbf{z}_{0})$, $\mathbf{w}_{l, r_1 \times r_2 \times c_{l, in} \times c_{l, out}}$, stride=$s$, axis=`$(h_l, w_l, c_{l, in}, c_{l, out})$') \\
    					$\mathbf{H}_{b} = $ conv2d($\mathbf{J}_{\mathbf{F}, \mathbf{b}, d_2 \times h_l \times w_l \times c_{l, in}}(\mathbf{z}_{0})$, $\mathbf{w}_{l, r_1 \times r_2 \times c_{l, in} \times c_{l, out}}$, stride=$s$, axis=`$(h_l, w_l, c_{l, in}, c_{l, out})$') \\
    					\If(\tcp*[f]{3.1 without g\_p ($RM_2$)}){$M$ = `$RM_2$'}
    					{
    						\Return $\mathbf{H}_{I}$, $\mathbf{H}_{b}$ \\
    					}\ElseIf(\tcp*[f]{3.2 with g\_p ($RM_1$)}){$M$ = `$RM_1$'}
    					{
    						\Return sum($\mathbf{H}_{I}$, axis=`$(h_{l, o}, w_{l, o})$'), sum($\mathbf{H}_{b}$, axis=`$(h_{l, o}, w_{l, o})$') \\
    					}
    				}
    			}
    			\Return NULL
    		}
        \end{algorithm}
    \end{minipage}%
    }

	\section{Experiments and results}
	\label{Experiments}
	This section will further verify our theory experimentally.
	
	As we mentioned in Section \ref{Model:Algorithm}, there are two operational modes of CNN's bias: (1) Conventional parameters trained for channels' compensation; (2) Auxiliary parameters trained for Batch Normalization. We select three models that involve these two operational modes. Also, these three models include both assumed activations: ReLU and Leaky ReLU. Generally, we will verify the nontrivial Eq.\eqref{eq:approx} on every layer of a CNN (except the first layer, the reason for which has been discussed in Section \ref{Model:Algorithm}).
	
	\subsection{Pre-trained CNNs and their conversions}
	We elaborate the training/validation/test settings on all models. We also discuss how we convert a model to an equivalent version using Eq.\eqref{eq:BN_reduced}.
	
	\paragraph{Dataset and augmentation} 
	We used the CIFAR-10 and CIFAR-100 data (\cite{krizhevsky2009learning}; RGB images of $10$/$100$ classes; Resolution: $32 \times 32$; Pixel value range: $[0, 1]$; Each one includes two sets: $50k$ training, $10k$ test). In each dataset, training was conducted on $45k$ of the $50k$ set (randomly selected), and validation is conducted on the remaining $5k$; The test is performed on the original $10k$ test set. Data augmentation methods are employed in training. An input image sequentially goes through the random left or right flipping, the random adjustments of saturation (within $[0, 2.0]$)/contrast (within $[0.4, 1.6]$)/brightness (within $0.5$), and random croppings to $32 \times 32 \times 3$ after resizing to $36 \times 36 \times 3$.

	\paragraph{Models and their conversions}
	We use three standard models: VGG7 \cite{simonyan2014very} with 7 ReLUs, ResNet20 (\cite{he2016deep}, code: \cite{ResNetTF2018}) with 20 Leaky ReLUs, and ResNet20-Fixup (the rightmost one in Figure 1 in \cite{zhang2019fixup}. The input image requires normalization on its RGB channels using the mean $[0.4914, 0.4822, 0.4465]$ and variance $[0.2023, 0.1994, 0.2010]$, code: \cite{FixupResNetPyTorch2019}) with the Fixup initialization and 20 ReLUs. The first two use the second bias operational mode (batch normalization), and the last uses the first bias operational mode (conventional bias mode). Their parameters are listed in tables \ref{table:params:vgg7}, \ref{table:params:resnet20}, and \ref{table:params:resnet20_fixup}, respectively. VGG7 and ResNet20 are trained with BN layers first. After training, we extract $\mathbf{w}, \mathbf{\gamma}, \mathbf{\beta}, \mathbf{\mu}, \mathbf{\sigma}^{2}$ (Eq.\eqref{eq:BN}) to compute and collect the corresponding $\mathbf{w}^{'}$ and $\mathbf{b}^{'}$ with Eq.\eqref{eq:BN_reduced}, layer by layer. Then, we rebuild every architectural layer with its $\mathbf{w}^{'}$ and $\mathbf{b}^{'}$ to construct an equivalent model (Fig.\ref{fig:Model:BN_reduced}). Similarly, ResNet20-Fixup is trained first; We extract all kernels and biases after training. We rebuilt every layer (we merge any convolutional layer having a multiplier, check the third column of table \ref{table:params:resnet20_fixup}) to construct an equivalent model. We verify that every rebuilt model achieves identical accuracy as its pre-trained version. We employ the rebuilt ones for our experiments.

	\paragraph{Loss function, accuracy, training, validation, test}
	We used a summation of the kernels' regularization on the $L_1$ norm and the cross-entropy on a prediction's softmax as the loss function. We used Top-1 accuracy. Training, validation, and test were conducted on an RTX3090 GPU. All three CNN models were trained with Gradient Descent optimizers (GD) on a batch size of $100$. Tables \ref{table:train_val_test_details:cifar10} and \ref{table:train_val_test_details:cifar100} showed additional details. We trained on the $45k$ samples every epoch and validated the trained model on the $5k$ samples every two epochs. The trained model would be saved if a higher validation accuracy was reached. We tested with the $10k$ test samples.
	
	\begin{table}[!h]
		\centering
		\begin{adjustbox}{width=0.9\textwidth}
			\begin{tabular}{|c|c|c|c|c|c|}
				\hline
				Models/lr intervals & $1^{st}$ & $2^{nd}$ & $3^{rd}$ & Total & Test Acc\\
				\hline
				VGG7 & $0.01, [0, 299]$ & $0.002, [300, 399]$ & $0.0005, [400, 500]$ & $501$ & 89.5\% \\
				\hline
				ResNet20 & $0.01, [0, 99]$ & $0.001, [100, 149]$ & $0.0002, [150, 200]$ & $201$ & 91.6\% \\
				\hline
				ResNet20-Fixup & $0.002, [0, 79]$ & $0.001, [80, 119]$ & $0.0005, [120, 150]$ & $151$ & 91.2\% \\
				\hline
			\end{tabular}
		\end{adjustbox}
		\caption{Details for training VGG7/ResNet20/ResNet20-Fixup on CIFAR-10. ``$0.01, [0, 299]$'' denotes the learning rate (lr) $0.01$ during the training interval, maintained during the $0^{th}$ to the $299^{th}$ training epoch. Note that the classification accuracies are modest at best, but since our thrust is primarily theoretical, we did not attempt to further fine-tune the default hyperparameters in the available open-source implementations that we used.}
		\label{table:train_val_test_details:cifar10}
	\end{table}
	
	\begin{table}[!h]
		\centering
		\begin{adjustbox}{width=0.9\textwidth}
			\begin{tabular}{|c|c|c|c|c|c|}
				\hline
				Models/lr intervals & $1^{st}$ & $2^{nd}$ & $3^{rd}$ & Total & Test Acc\\
				\hline
				VGG7 & $0.01, [0, 499]$ & $0.002, [500, 599]$ & $0.0005, [600, 700]$ & $701$ & 63.2\% \\
				\hline
				ResNet20 & $0.01, [0, 99]$ & $0.001, [100, 149]$ & $0.0002, [150, 200]$ & $201$ & 68.8\% \\
				\hline
				ResNet20-Fixup & $0.002, [0, 39]$ & $0.001, [40, 79]$ & $0.0005, [80, 120]$ & $121$ & 64.0\% \\
				\hline
			\end{tabular}
		\end{adjustbox}
		\caption{Details for training VGG7/ResNet20/ResNet20-Fixup on CIFAR-100. Check Table \ref{table:train_val_test_details:cifar10} for notations.}
		\label{table:train_val_test_details:cifar100}
	\end{table}

	\subsection{Verify Eq.\eqref{eq:approx} on CNNs}
	As we mentioned, we will experimentally verify that $\mathbf{z}([\mathbf{x}_{n}; \mathbf{x}_{b}]) = \frac{[\mathbf{x}_{n}; \mathbf{x}_{b}]}{8}$ achieves the equality in Eq.\eqref{eq:approx}.
	
	We verify it on two types of layers: every Conv layer, and the FC layer. In other words, we verify that the activation value of an out-ch feature map's unit or the FC layer's output unit, computed on a CNN directly, should be equal to the dot-product between the input, $[\mathbf{x}_{n}; \mathbf{x}_{b}]$, and the reconstructed effective hyperplane $\mathbf{H}^{Adj}(\frac{[\mathbf{x}_{n}; \mathbf{x}_{b}]}{8})$. The verification can be described with two equations, Eq.\eqref{eq:verify_conv} (Conv) and Eq.\eqref{eq:verify_FC} (FC). Principles of the verification are illustrated using a symbol ``$=$'' in these previous figures: Fig.\ref{fig:Model:two_factors}(c) (Conv), and Fig.\ref{fig:Model:adjoint_ops_blueprint}(b) (FC).
	\begin{equation} \label{eq:verify_conv}
		\begin{split}
			c_{l, s, i} = \underbrace{\mathbf{F}_{l-1, s, i}([\mathbf{x}_{n}; \mathbf{x}_{b}]) \circledast \mathbf{w}_{l, s, i}}_{\mbox{\tiny Original CNN activation (Conv)}}
			= \hat{c}_{l, s, i} \\
			= \underbrace{\langle \mathbf{x}_{n} \mid \mathbf{H}^{Adj, I}_{l, s, i}(\frac{[\mathbf{x}_{n}; \mathbf{x}_{b}]}{8}) \rangle + \langle \mathbf{x}_{b} \mid \mathbf{H}^{Adj, b}_{l, s, i}(\frac{[\mathbf{x}_{n}; \mathbf{x}_{b}]}{8}) \rangle}_{\mbox{\scriptsize Reconstructed Conv unit activation using Adjoint}},
		\end{split}
	\end{equation}
	
	\begin{equation} \label{eq:verify_FC}
		\begin{split}
			c_{fc, k} = \underbrace{\mathbf{F}_{fc, k}([\mathbf{x}_{n}; \mathbf{x}_{b}])}_{\mbox{\tiny Original CNN activation (FC)}}
			= \hat{c}_{fc, k} \\
			= \underbrace{\langle \mathbf{x}_{n} \mid \mathbf{H}^{Adj, I}_{k}(\frac{[\mathbf{x}_{n}; \mathbf{x}_{b}]}{8}) \rangle + \langle \mathbf{x}_{b} \mid \mathbf{H}^{Adj, b}_{k}(\frac{[\mathbf{x}_{n}; \mathbf{x}_{b}]}{8}) \rangle}_{\mbox{\scriptsize Reconstructed FC unit activation using Adjoint}},
		\end{split}
	\end{equation}
	where $c_{l, s, i}$ and $c_{fc, k}$ are true values (corresponding to the left hand side of the third ``$=$'' in Eq.\eqref{eq:approx}) while $\hat{c}_{l, s, i}$ and $\hat{c}_{fc, k}$ are approximated ones (corresponding to the right hand side of the third ``$=$'' in Eq.\eqref{eq:approx}); $c_{l, s, i}$ labels a pixel at the stride move $s$, in the $i^{th}$ out-ch feature map of the $l^{th}$ conv layer (i.e., a unit of $\mathbf{c}_{l}$); $c_{fc, k}$ labels the $k^{th}$ entry in the $\mathbf{c}_{fc}$, from the FC layer, for activating a specific class. In detail, for an image $\mathbf{x}_{n}$, a dot-product should be verified to replicate a unit's activation value from the conv or the FC layer, and this verification should go through the units of all layers except for the first one (which has been explained why in Section \ref{Model:Algorithm}); Also, $RM2$ results are reused in $RM4, RM3, RM1$ via simple summation (Eq.\eqref{eq:RM4} $\sim$ \ref{eq:RM1}), which implies verifying $RM2$ is equivalent to verifying $RM4 \sim RM1$. Since TensorFlow uses the FP32 datatype (single precision) to compute a decimal, rounding errors are inevitable and result in a fractional mismatch between $\mathbf{c}_{l}$ and $\hat{\mathbf{c}}_{l}$ due to different computation paths. We measure relative errors $\mathbf{\epsilon}$ for evaluating approximations, i.e.,
	\begin{equation} \label{eq:relative_err}
		\mathbf{\epsilon}_{l} = \frac{\hat{\mathbf{c}}_{l} - \mathbf{c}_{l}}{\mathbf{c}_{l, \ne 0}},
	\end{equation}
	where $l$ is either the conv layer index or the FC layer; $\mathbf{c}_{l, \ne 0}$ substitutes all zeros inside $\mathbf{c}_{l}$ with the smallest positive float of FP32 to avoid any divide by zero exception.

	\subsection{Results} 
\begin{figure}[!h]
		\centering
		\begin{tabular}{@{}c@{}c@{}c@{}c@{}c@{}c@{}}
			\includegraphics[height=0.12\textwidth]{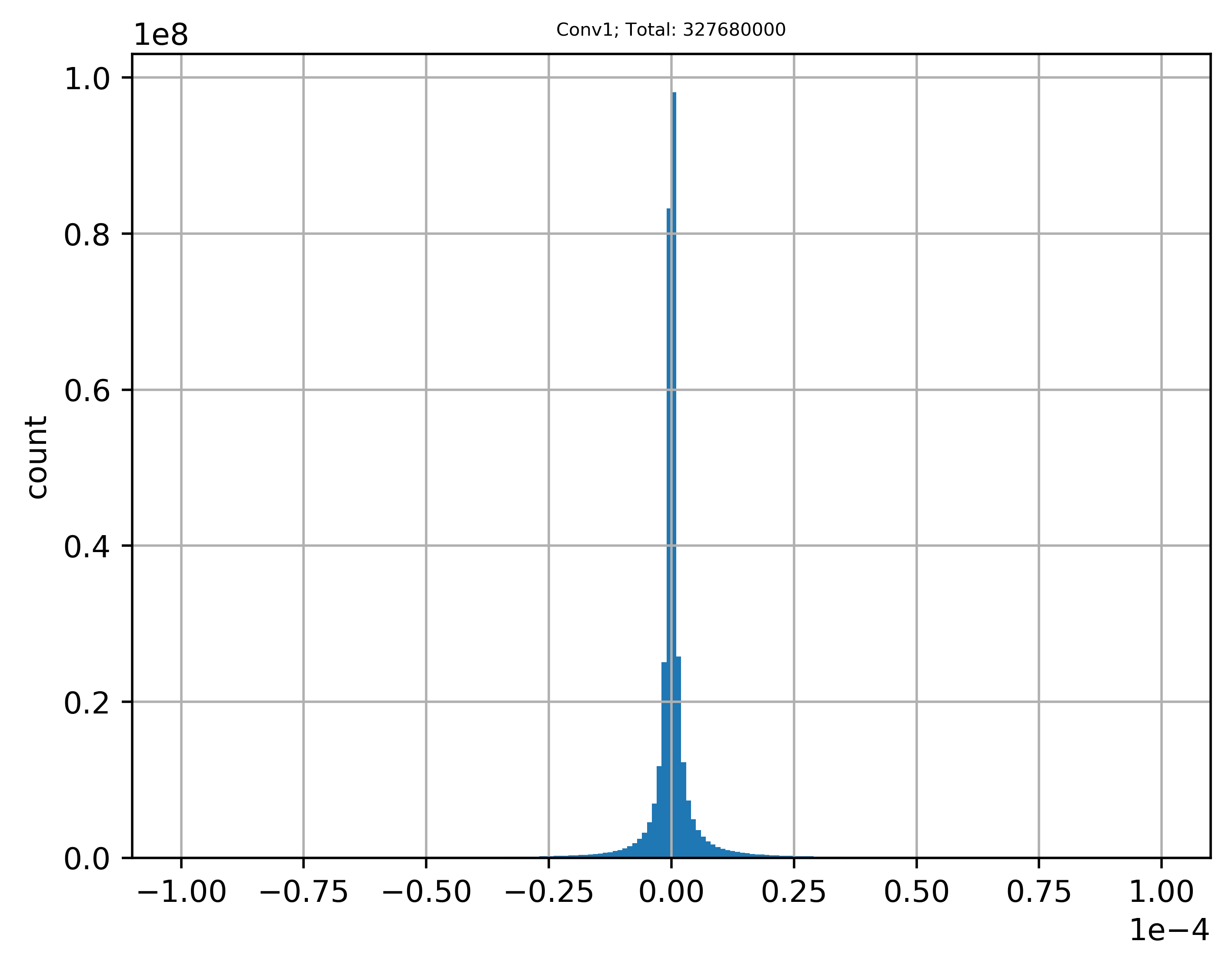} & 
			\includegraphics[height=0.12\textwidth]{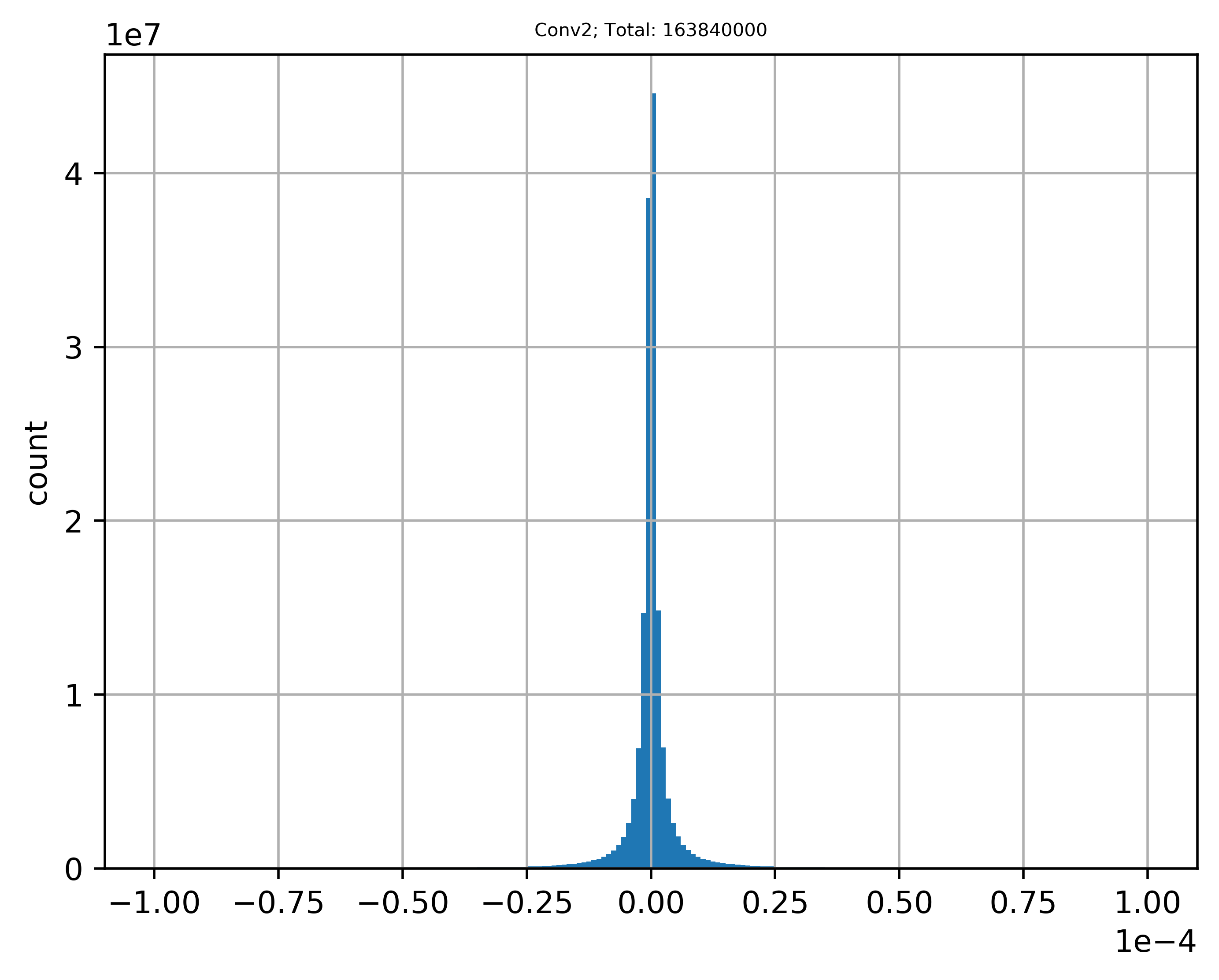} &
			\includegraphics[height=0.12\textwidth]{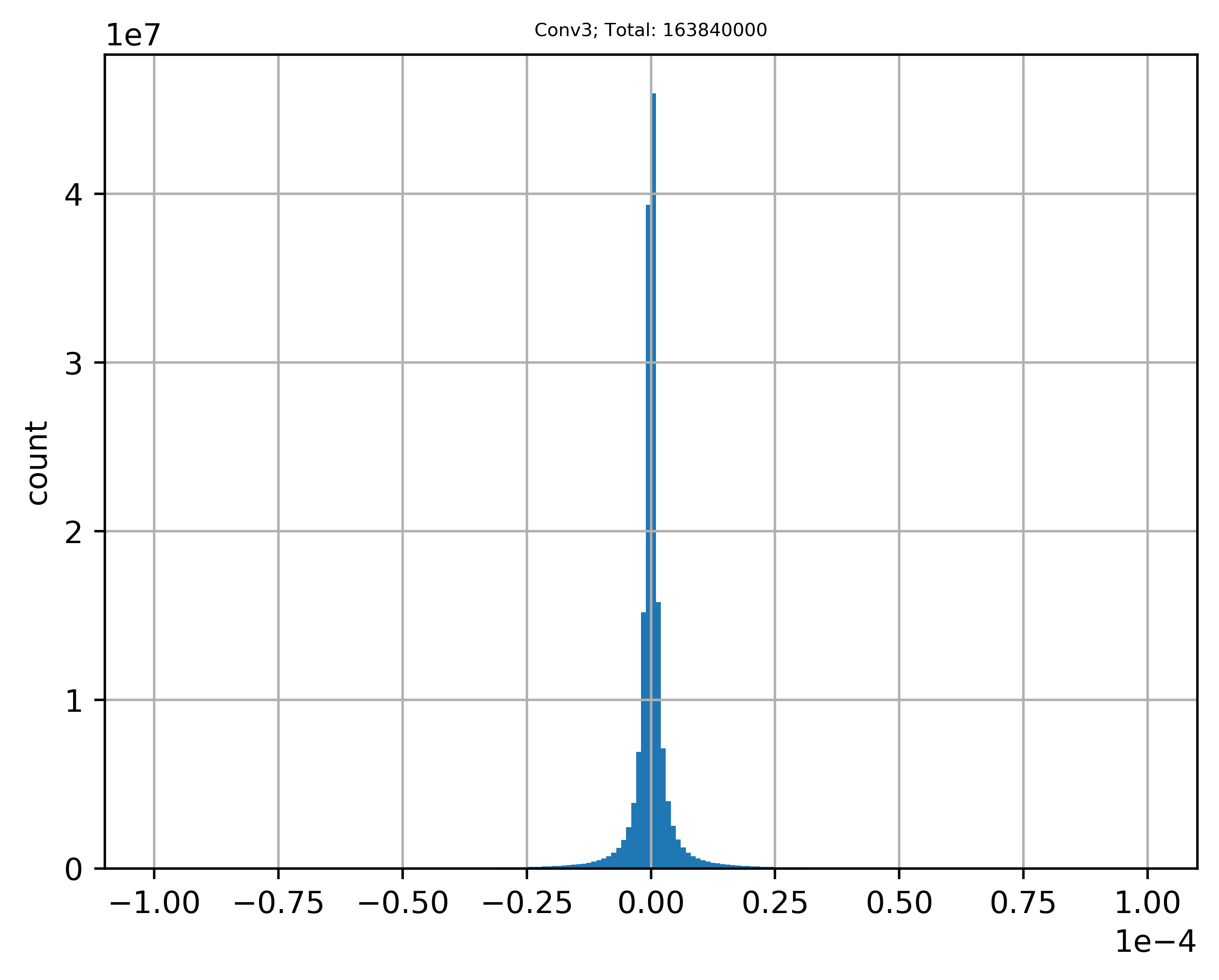} &
			\includegraphics[height=0.12\textwidth]{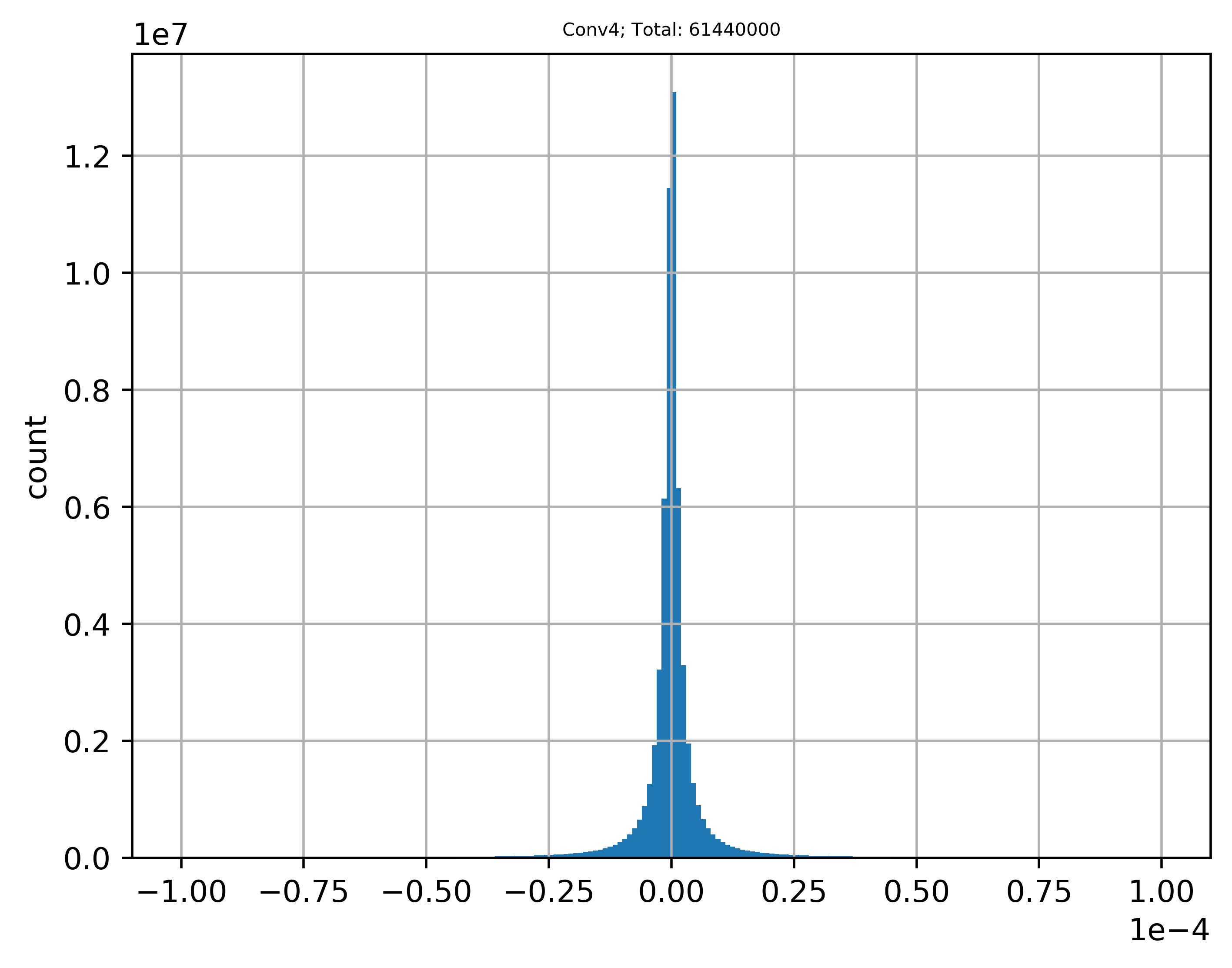} &
			\includegraphics[height=0.12\textwidth]{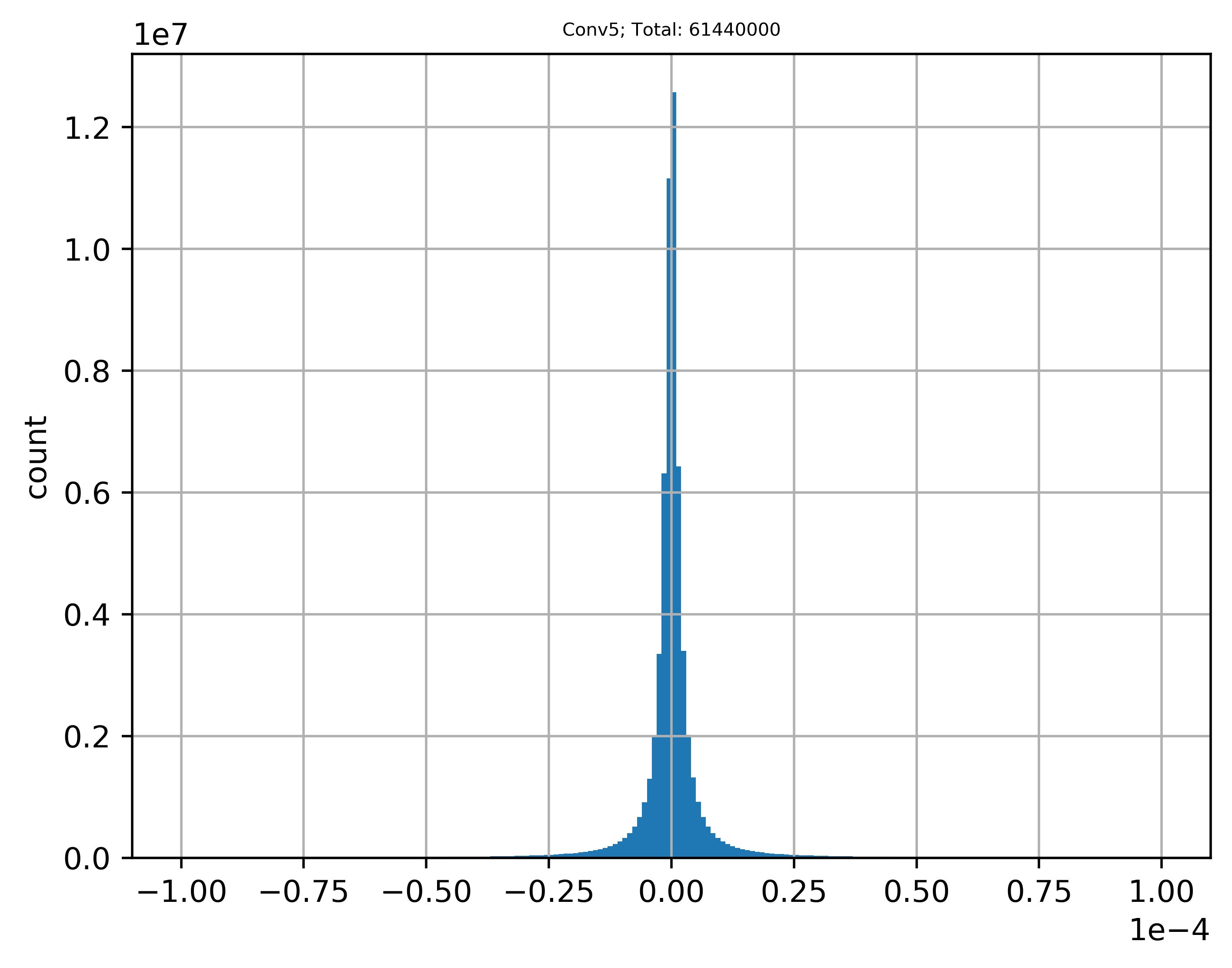} &
			\includegraphics[height=0.12\textwidth]{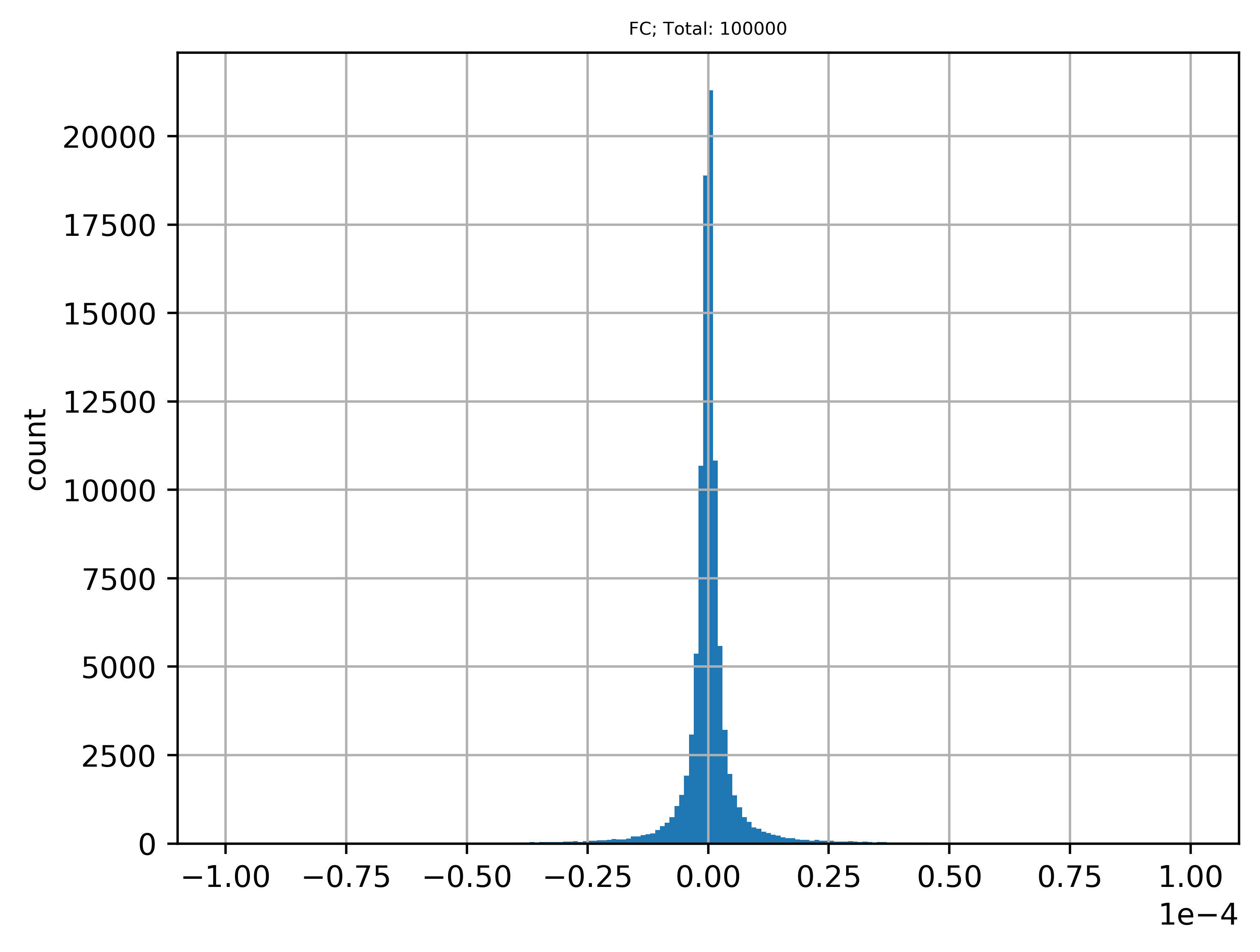} \\
			($a_1$) & ($a_2$) & ($a_3$) & ($a_4$) & ($a_5$) & ($a_6$) \\
		\end{tabular}

		\caption{VGG7 (CIFAR-10). Histograms of relative errors (Eq.\eqref{eq:relative_err}) between units directly computed by CNN ($\mathbf{c}_{l}$) and their values reconstructed by our method ($\hat{\mathbf{c}}_{l}$) on CIFAR-10 using VGG7. The x-axis and y-axis indicate the error and frequency ($10k$ test samples), respectively. $a_1$$\sim$$a_5$ is collected from Conv1$\sim$5 layer, and $a_6$ is collected from the FC layer. In terms of the $4^{th}$ column of table.\ref{table:params:vgg7}, $a_1$ to $a_6$ should have relative errors from $327.68m(=32,768 \times 10k), 163.84m(=16,384 \times 10k), 163.84m, 61.44m(= 6,144 \times 10k), 61.44m$, and $100k$, units, respectively. Percentages of $\mathbf{\epsilon}_{l} \le 1\%$ for all subplots are listed in Table \ref{table:percentage_of_errs:cifar10}. Note that the $x$-axis ranges from $-0.0001$ to $0.0001$, which is an extremely small range. The same is true for all the following figures.}
		\label{fig:Statistics:cifar10:vgg7}
	\end{figure}

\begin{figure}[!h]
		\centering
		\begin{tabular}{@{}c@{}c@{}c@{}c@{}c@{}c@{}c@{}}
			\includegraphics[height=0.1\textwidth]{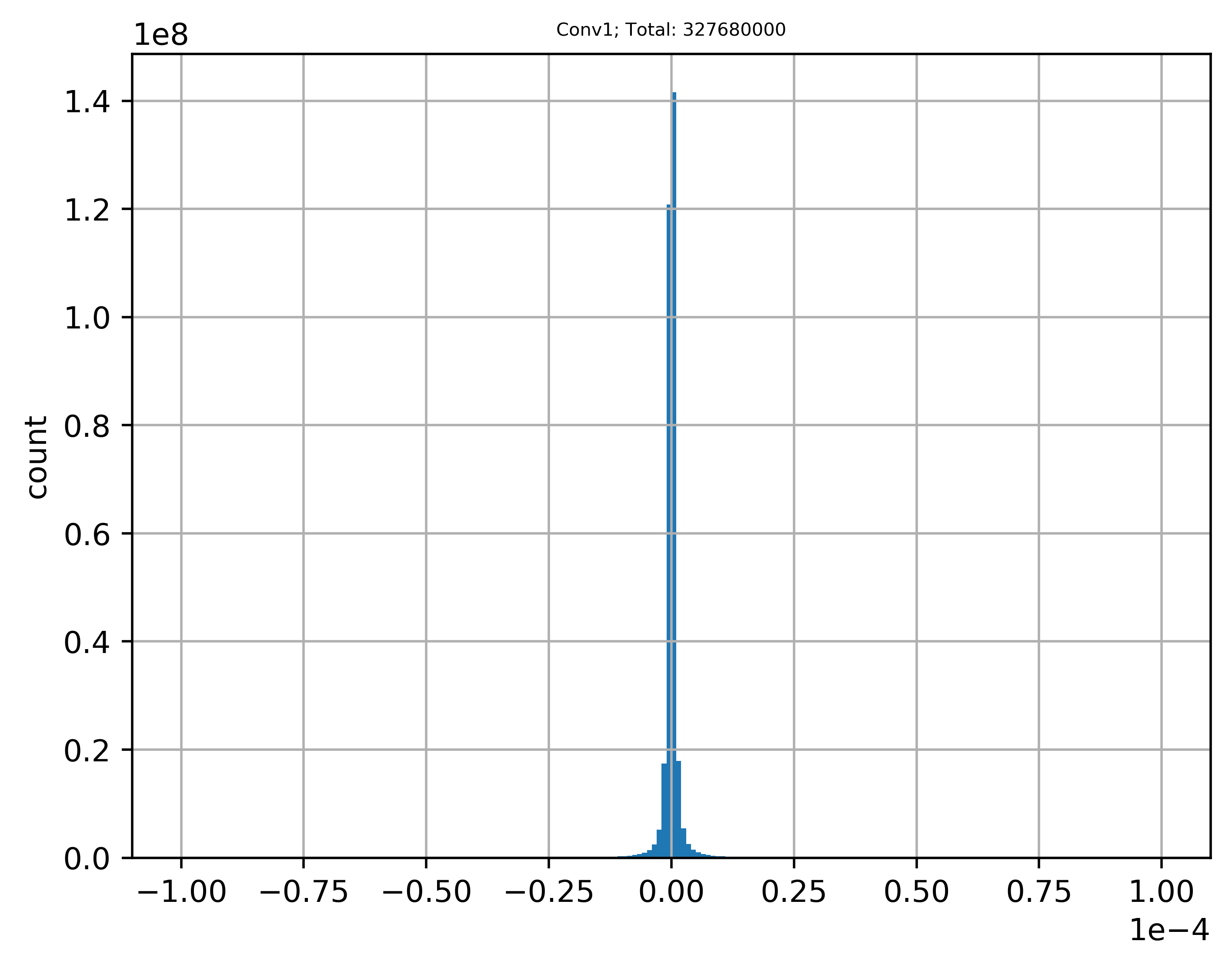} & 
			\includegraphics[height=0.1\textwidth]{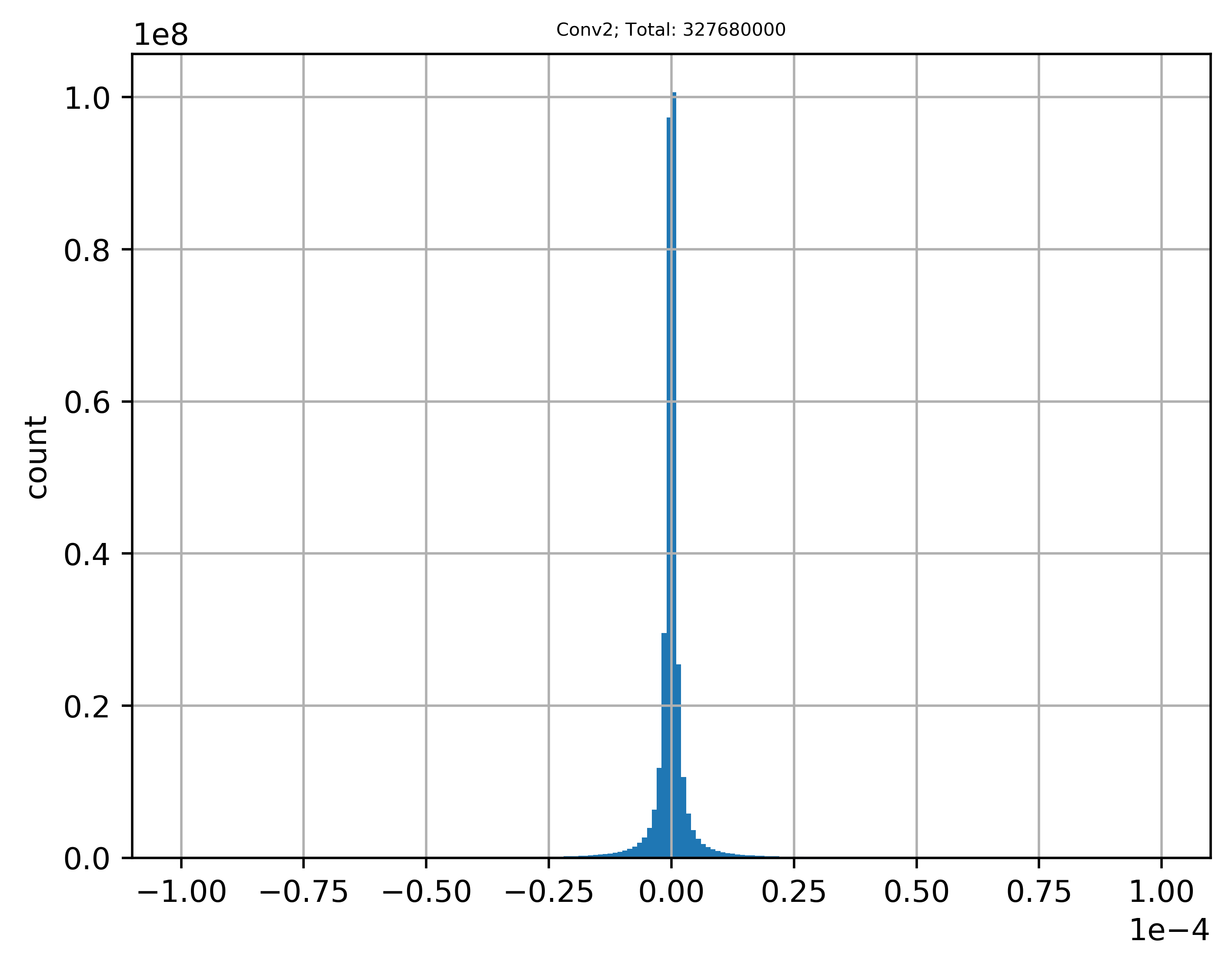} &
			\includegraphics[height=0.1\textwidth]{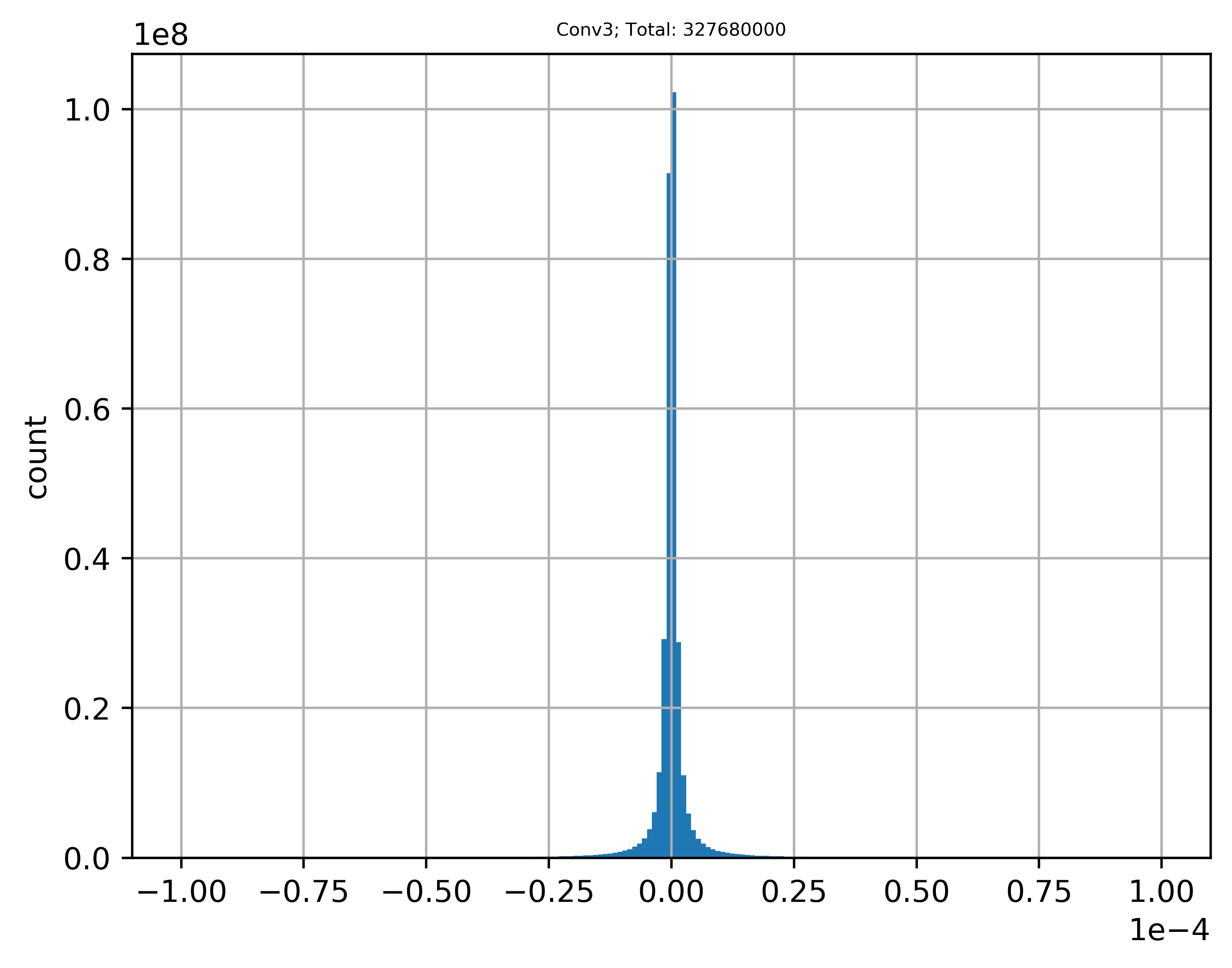} &
			\includegraphics[height=0.1\textwidth]{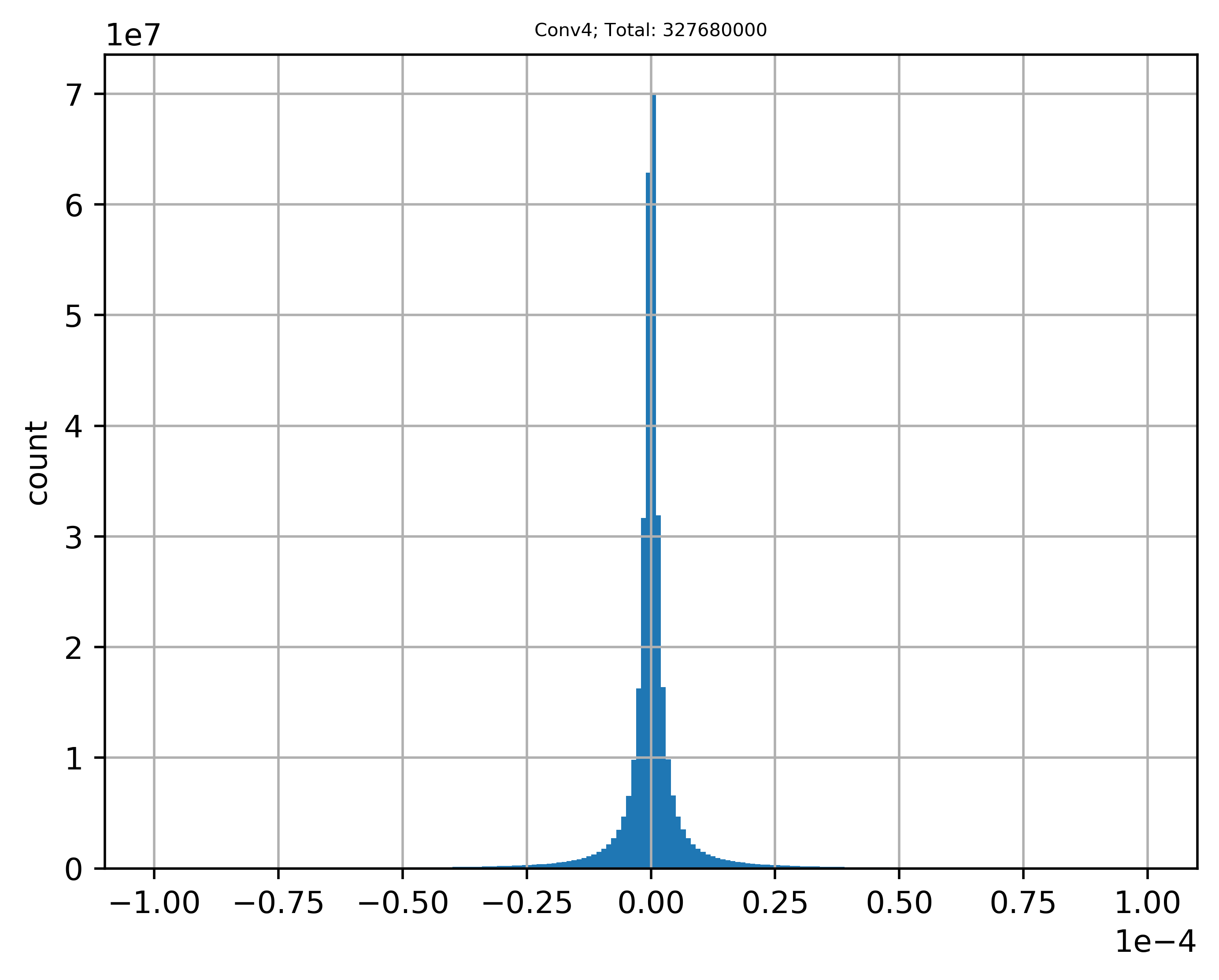} &
			\includegraphics[height=0.1\textwidth]{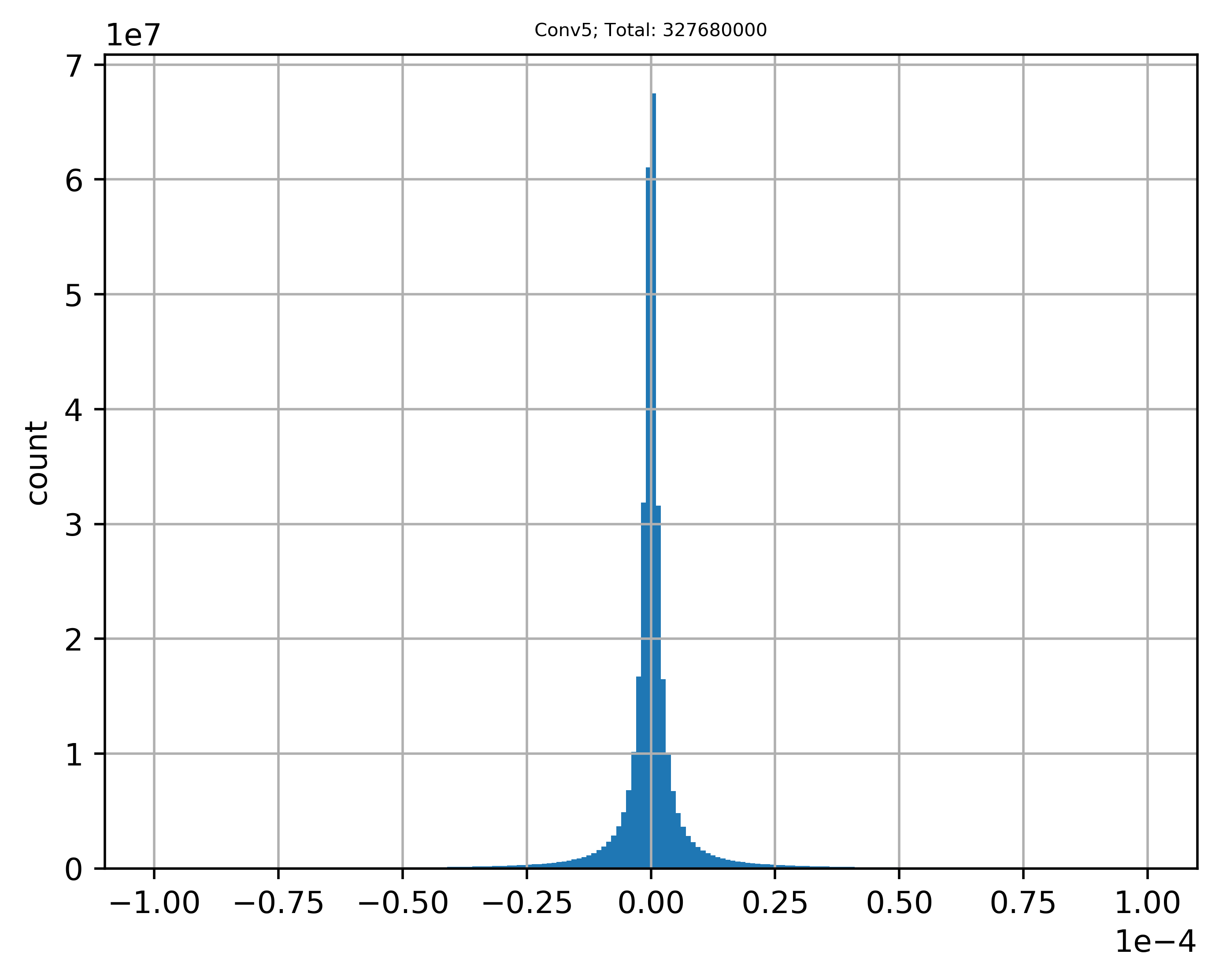} &
			\includegraphics[height=0.1\textwidth]{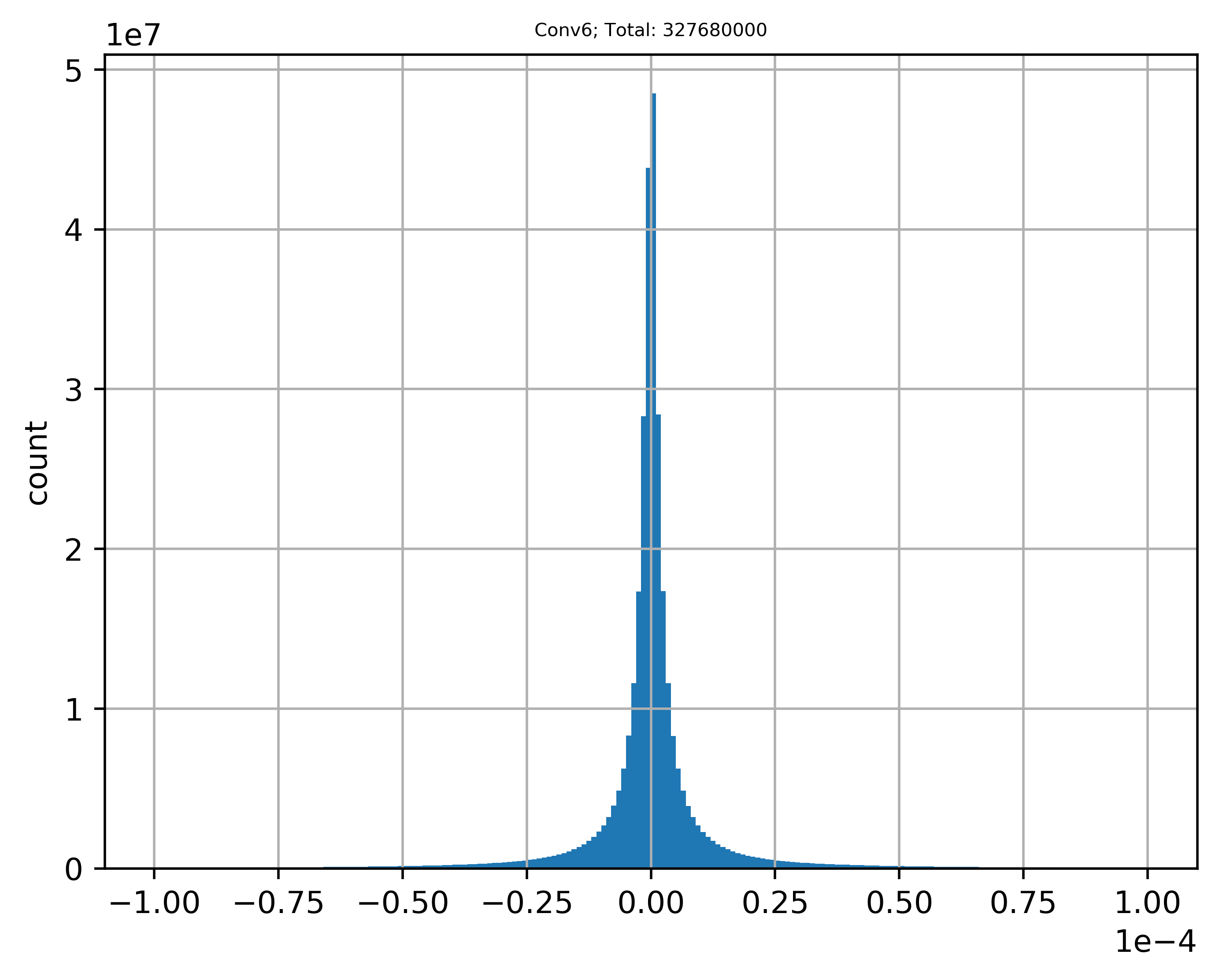} &
			\includegraphics[height=0.1\textwidth]{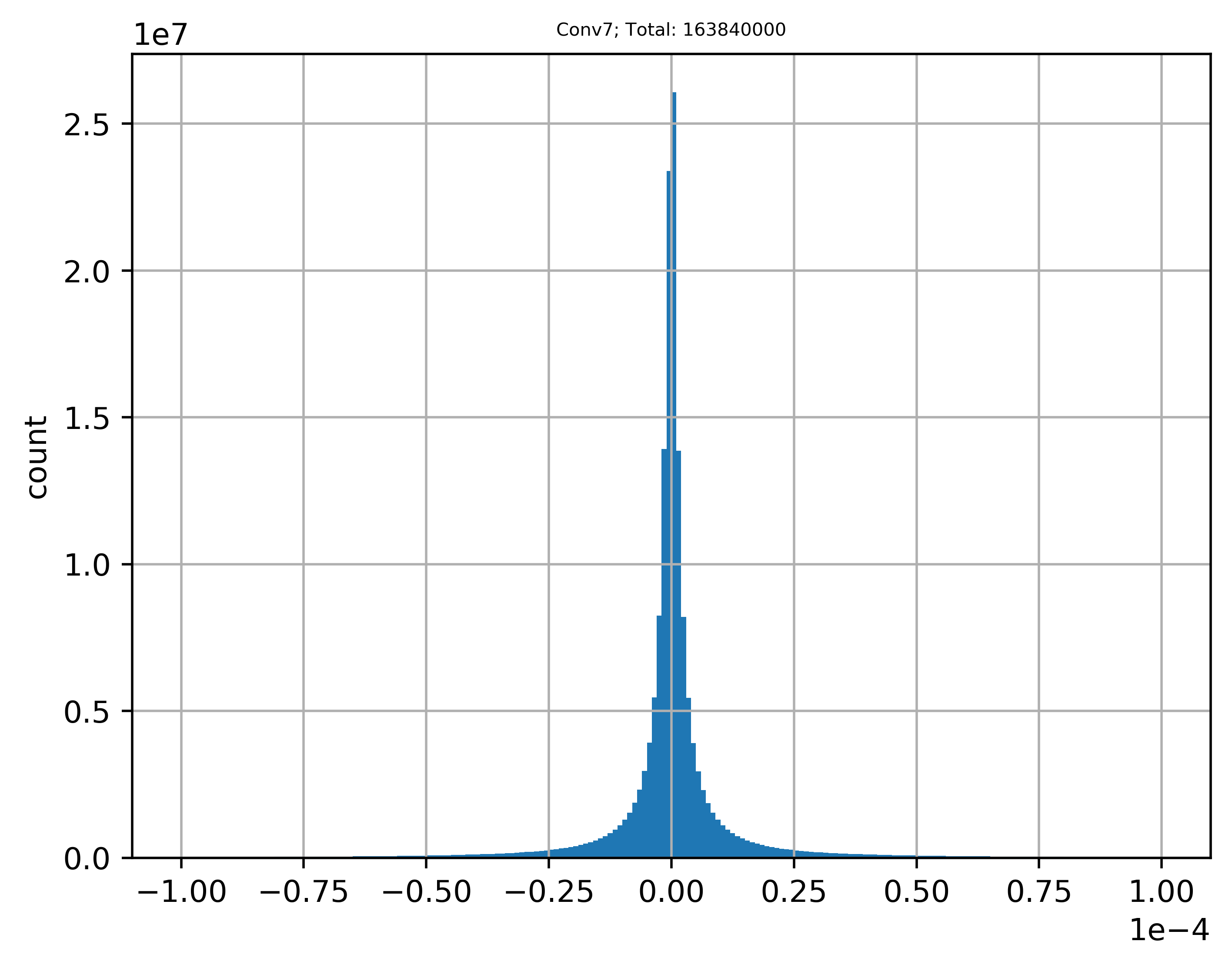} \\
			($a_1$) & ($a_2$) & ($a_3$) & ($a_4$) & ($a_5$) & ($a_6$) & ($a_7$) \\
			\includegraphics[height=0.1\textwidth]{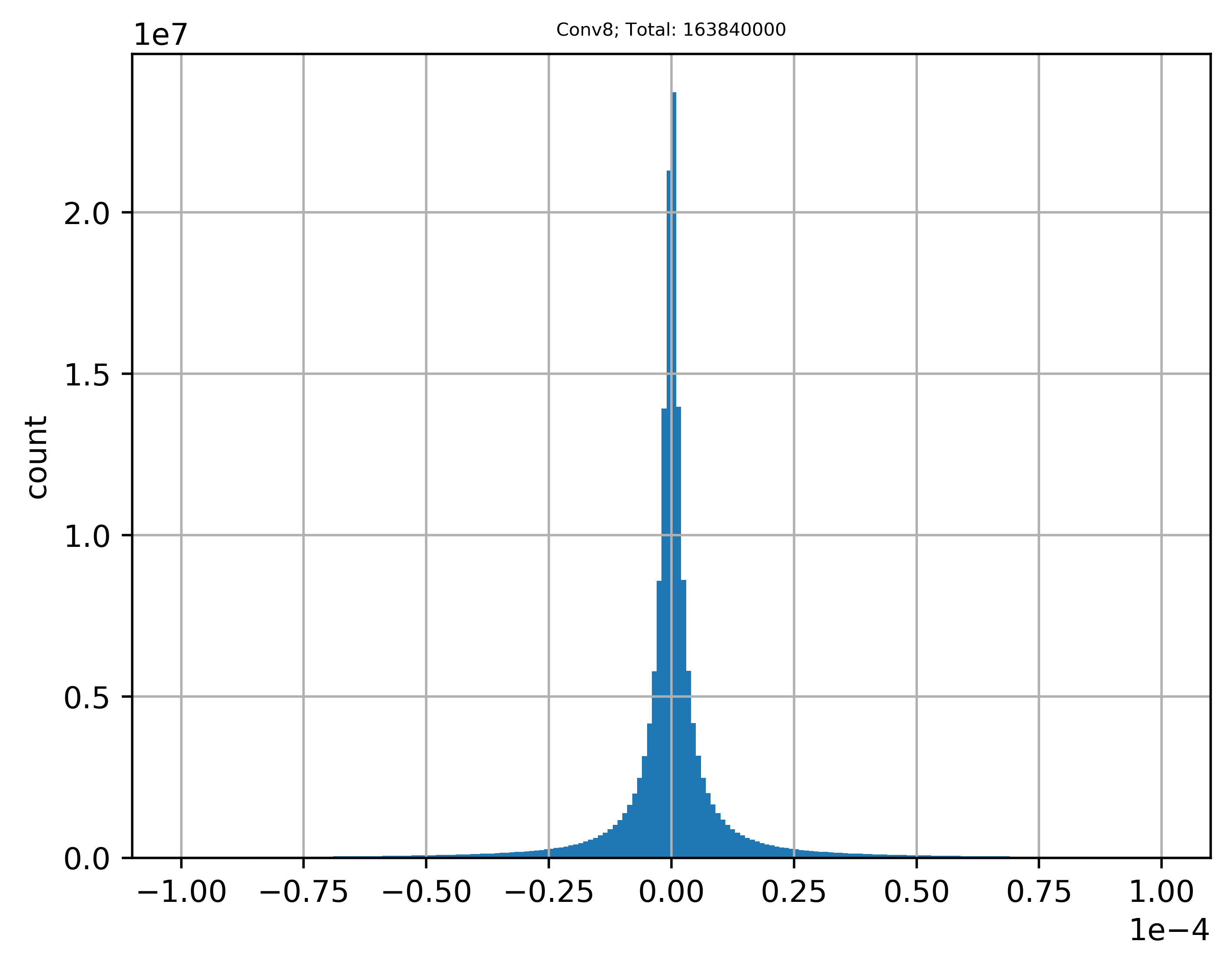} &
			\includegraphics[height=0.1\textwidth]{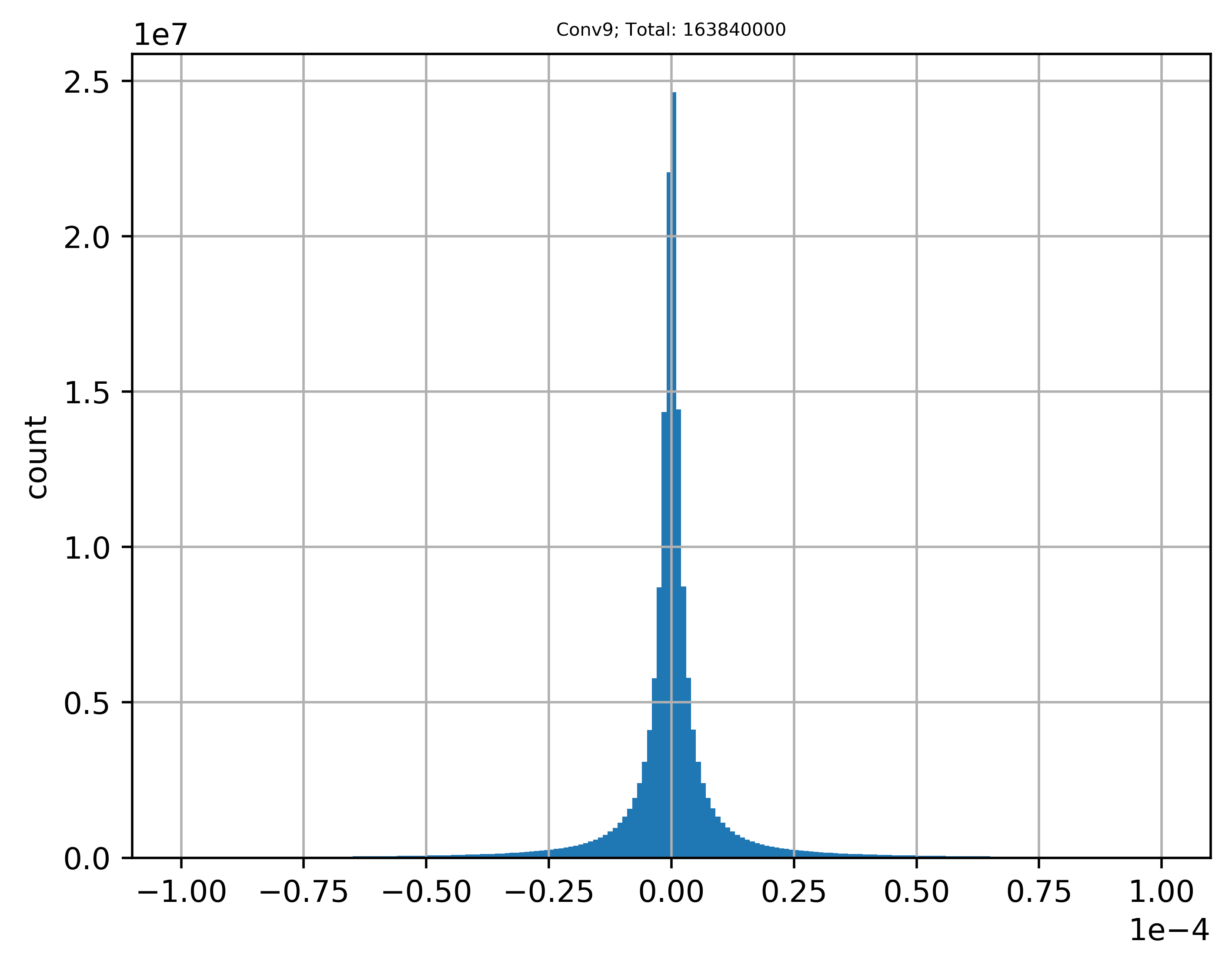} &
			\includegraphics[height=0.1\textwidth]{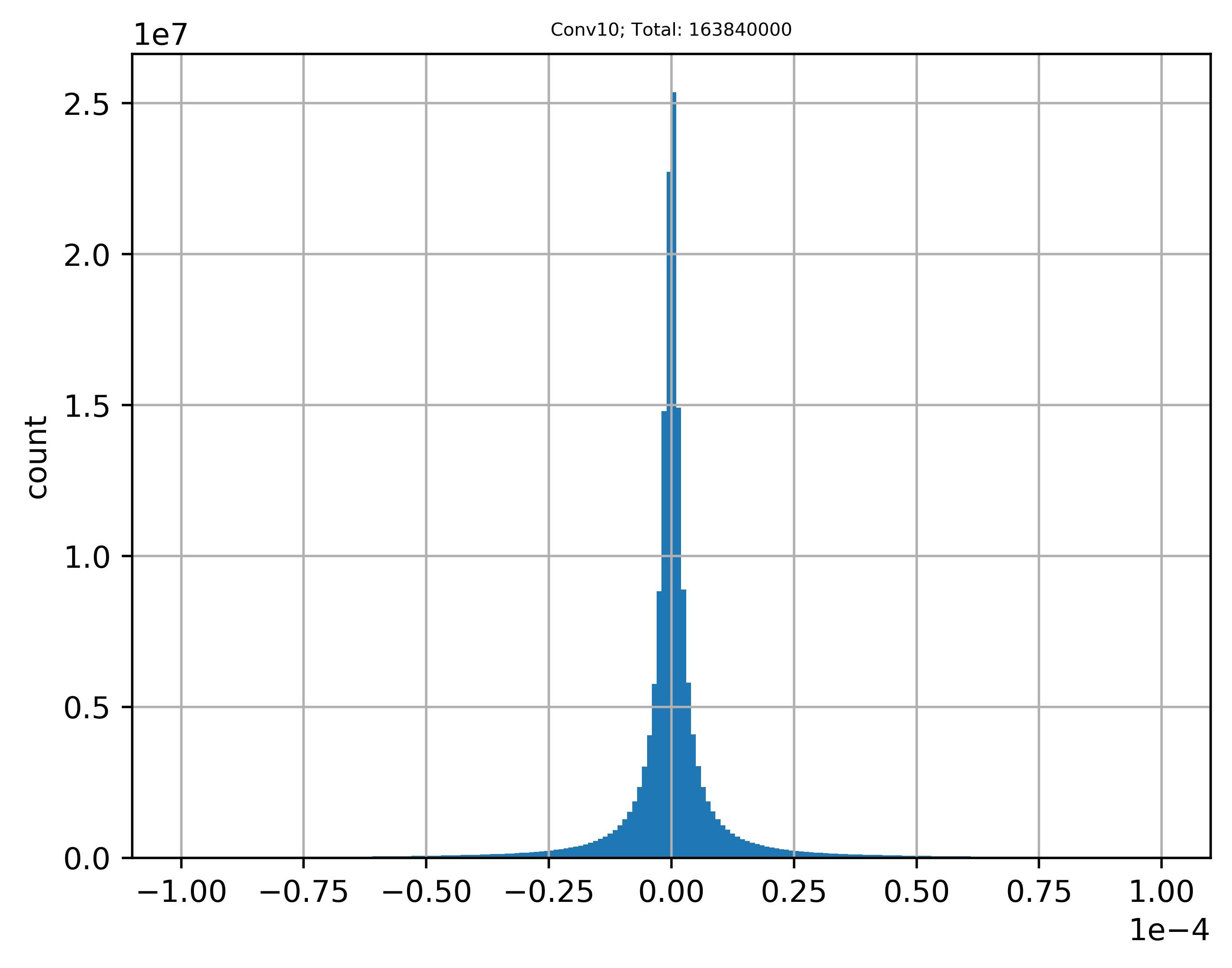} &
			\includegraphics[height=0.1\textwidth]{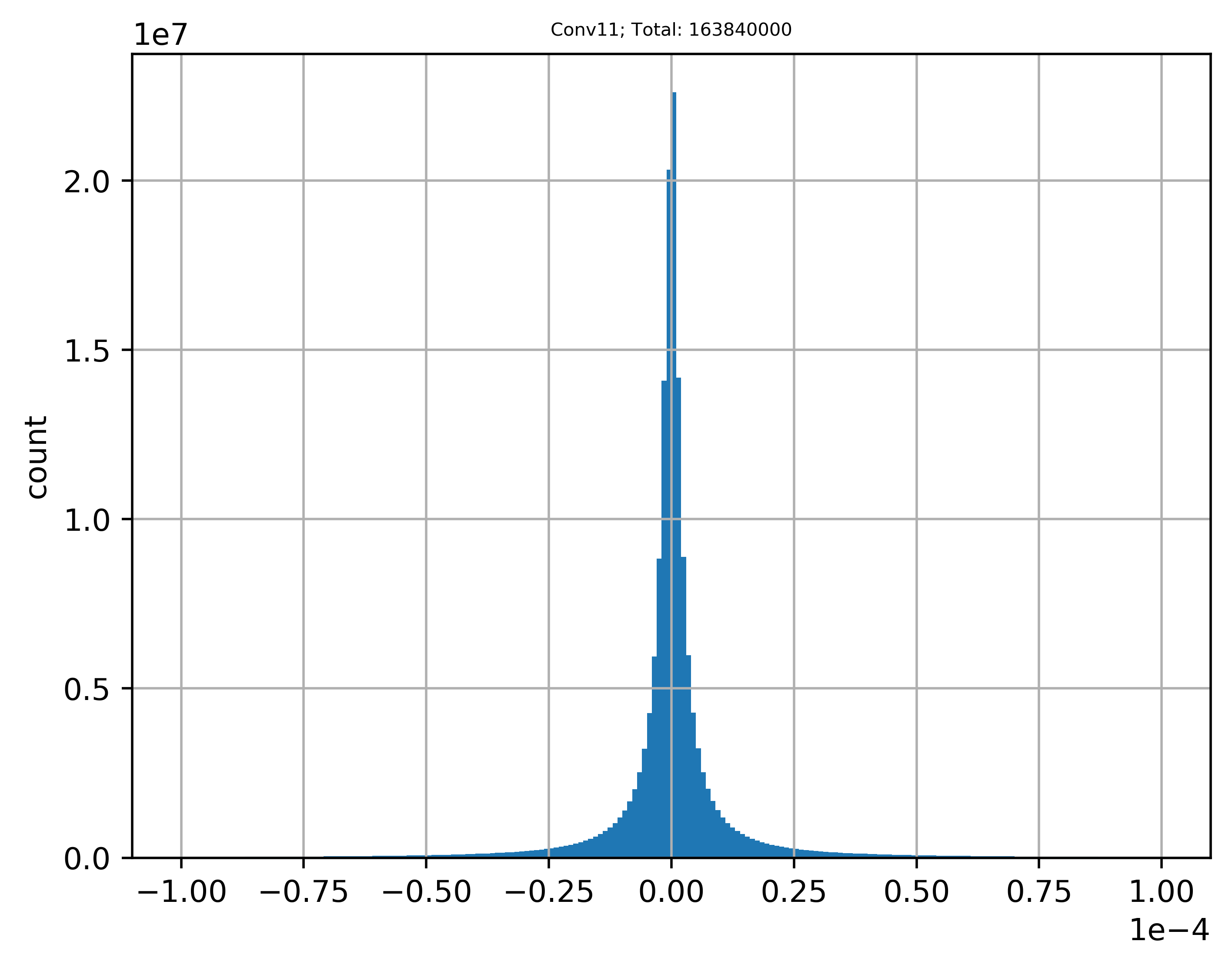} &
			\includegraphics[height=0.1\textwidth]{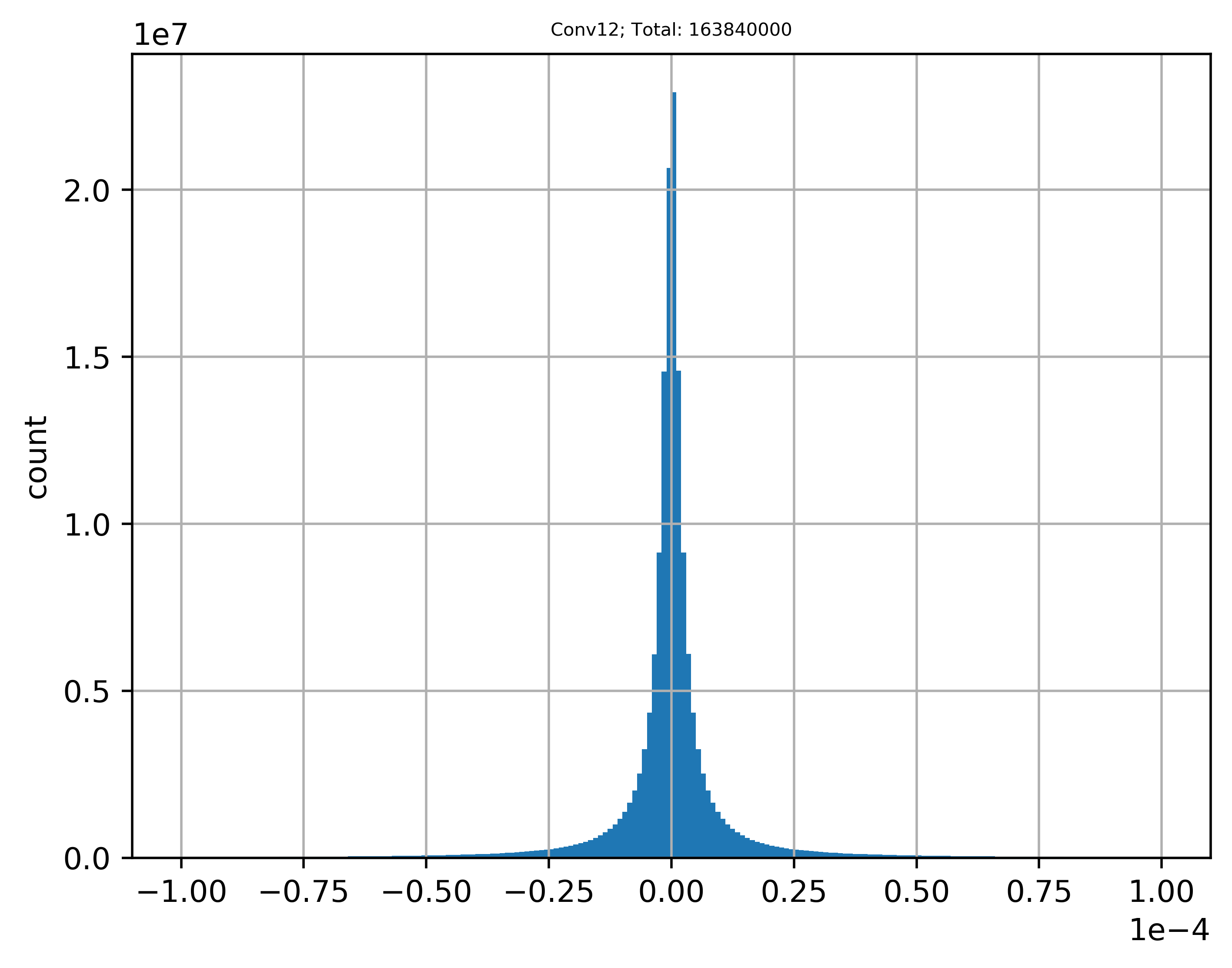} &
			\includegraphics[height=0.1\textwidth]{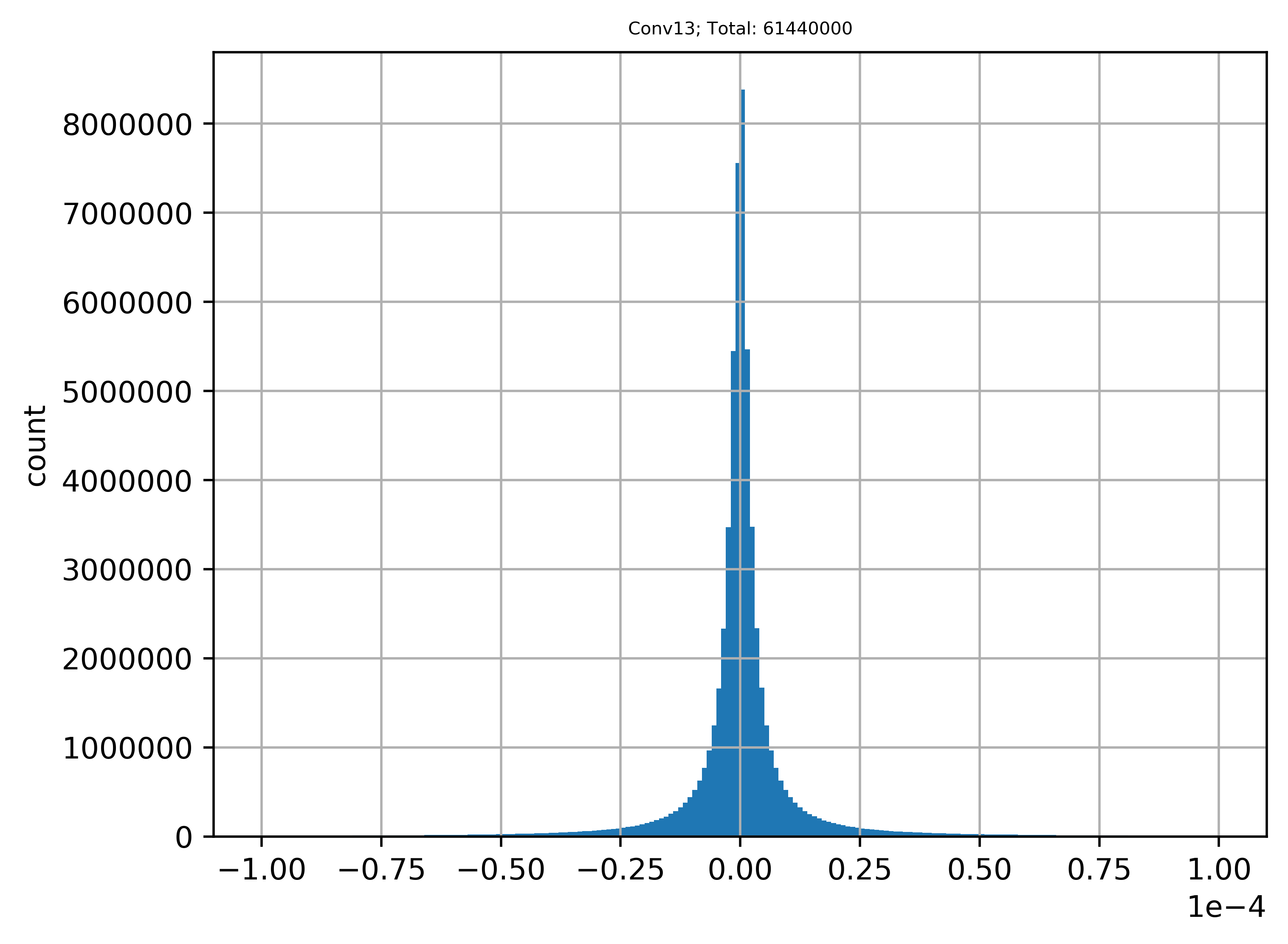} &
			\includegraphics[height=0.1\textwidth]{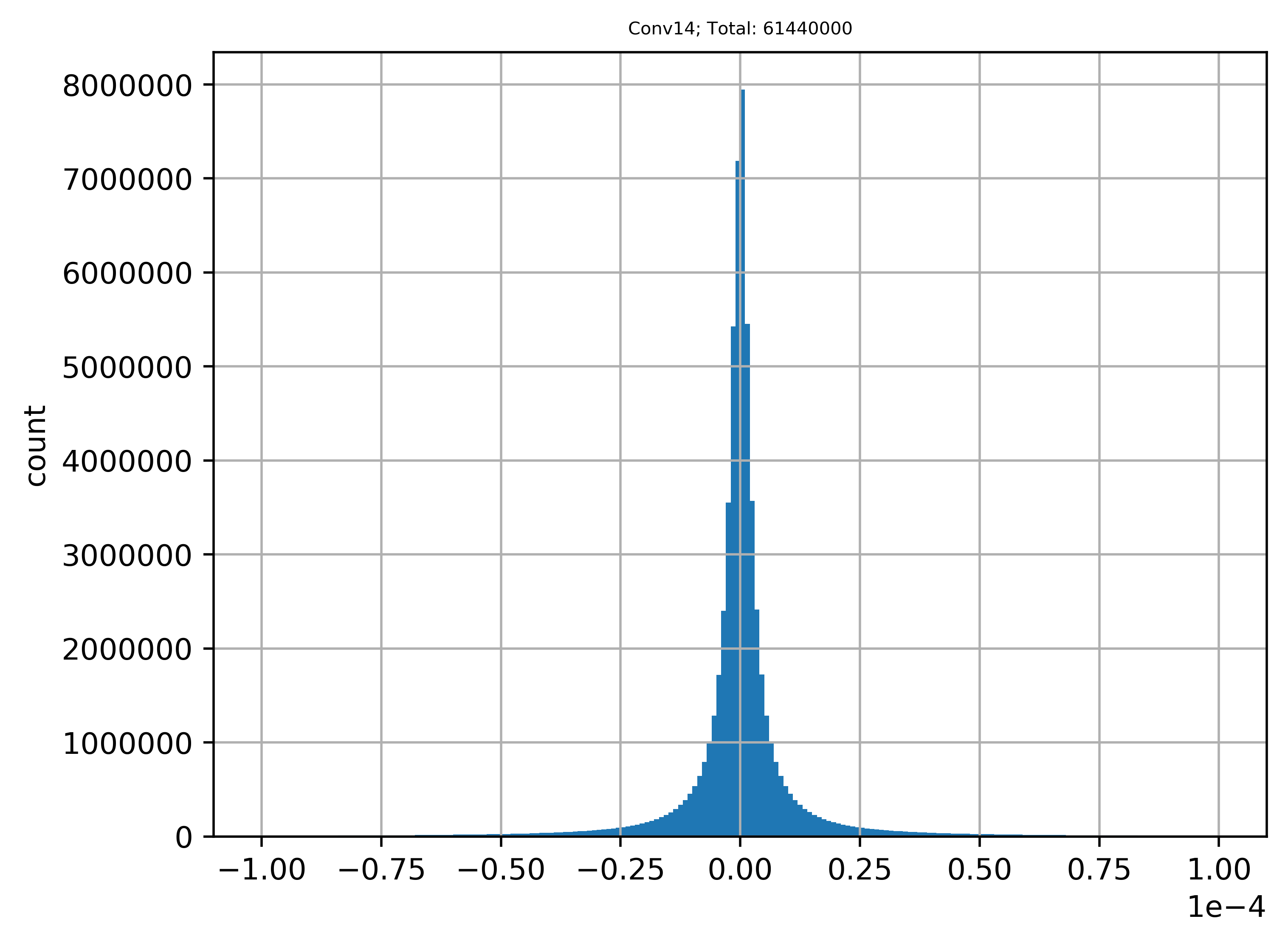} \\
			 ($a_8$) & ($a_9$) & ($a_{10}$) & ($a_{11}$) & ($a_{12}$) & ($a_{13}$) & ($a_{14}$) \\
			\includegraphics[height=0.1\textwidth]{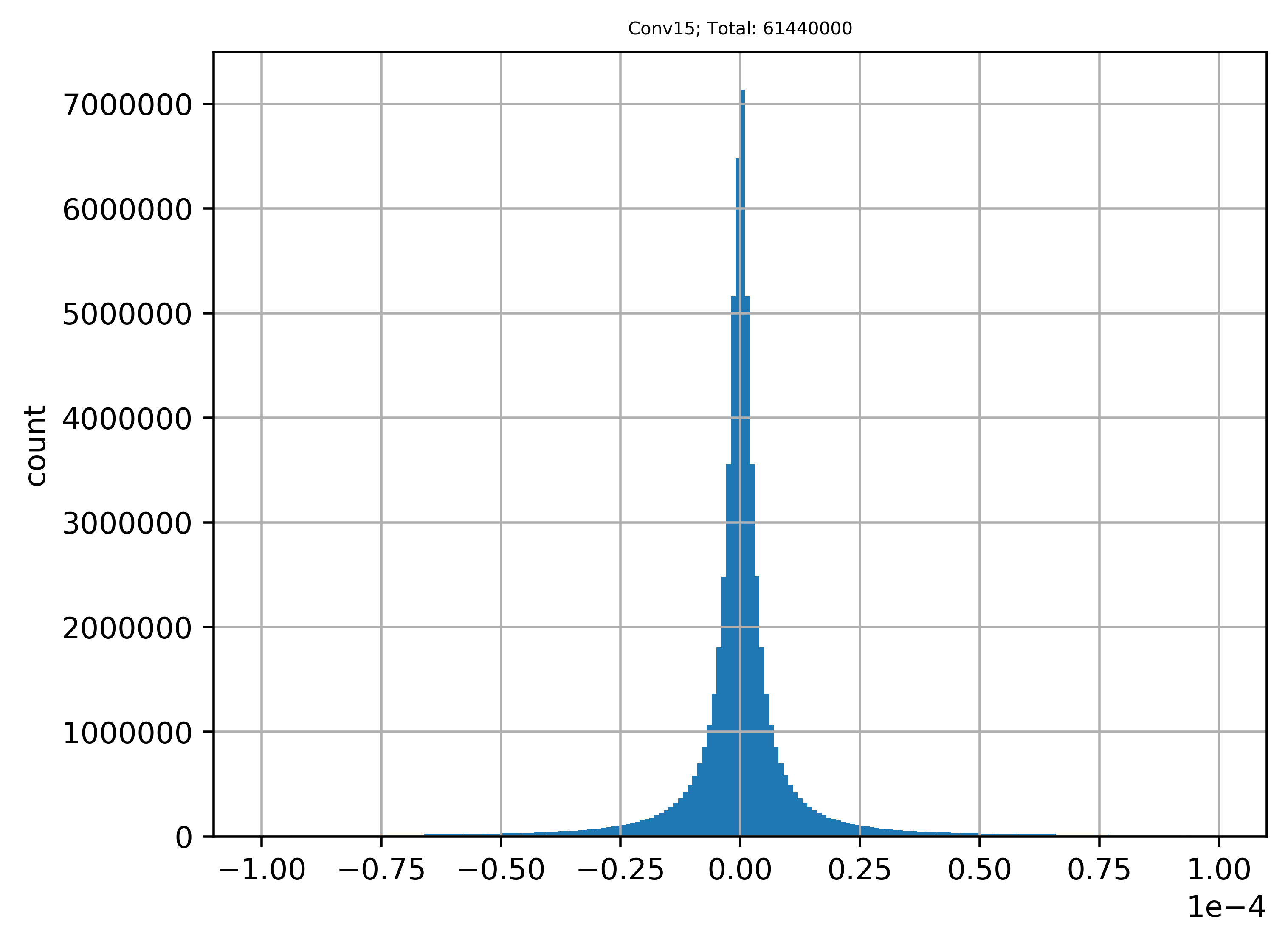} &
			\includegraphics[height=0.1\textwidth]{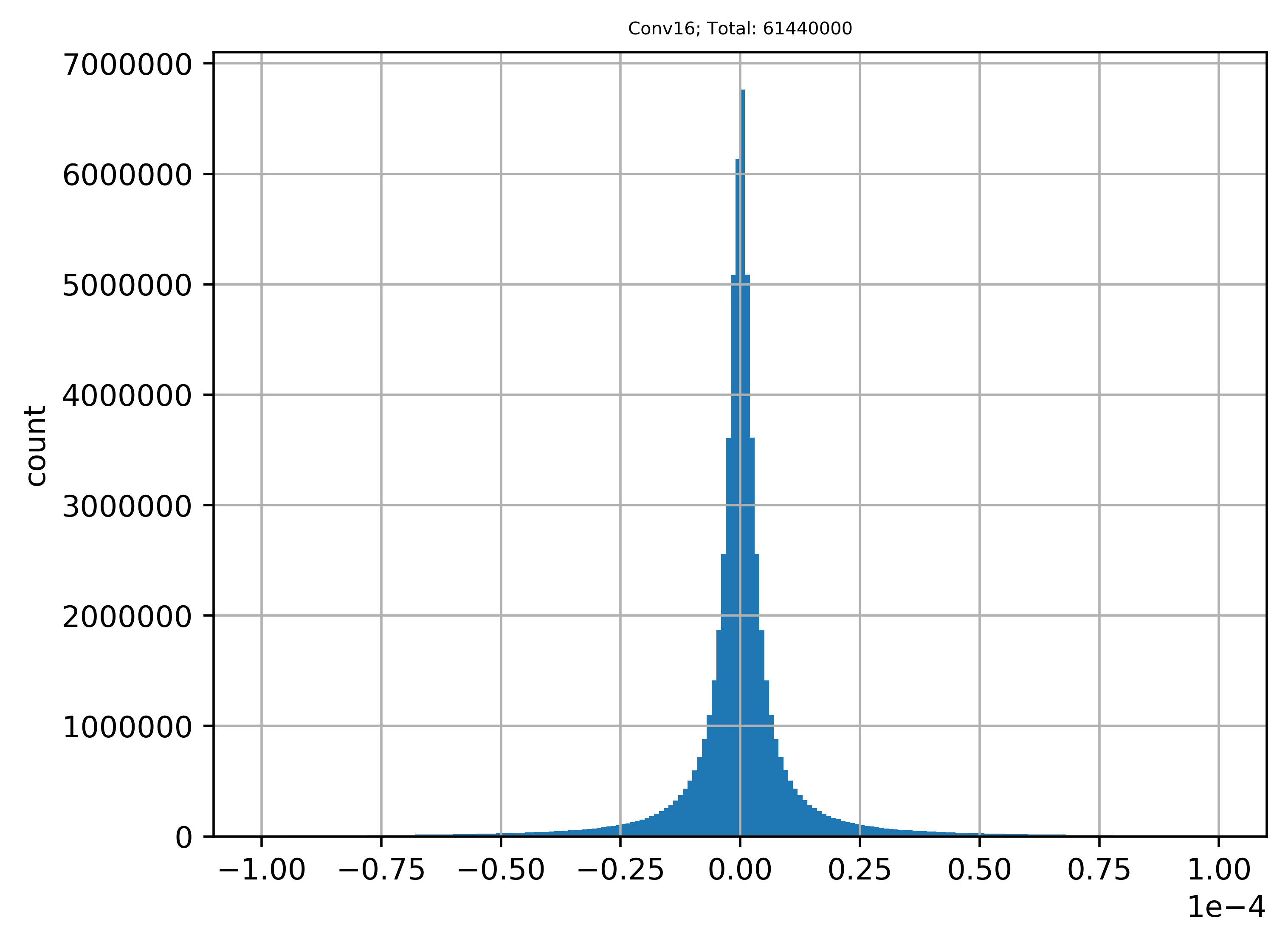} &
			\includegraphics[height=0.1\textwidth]{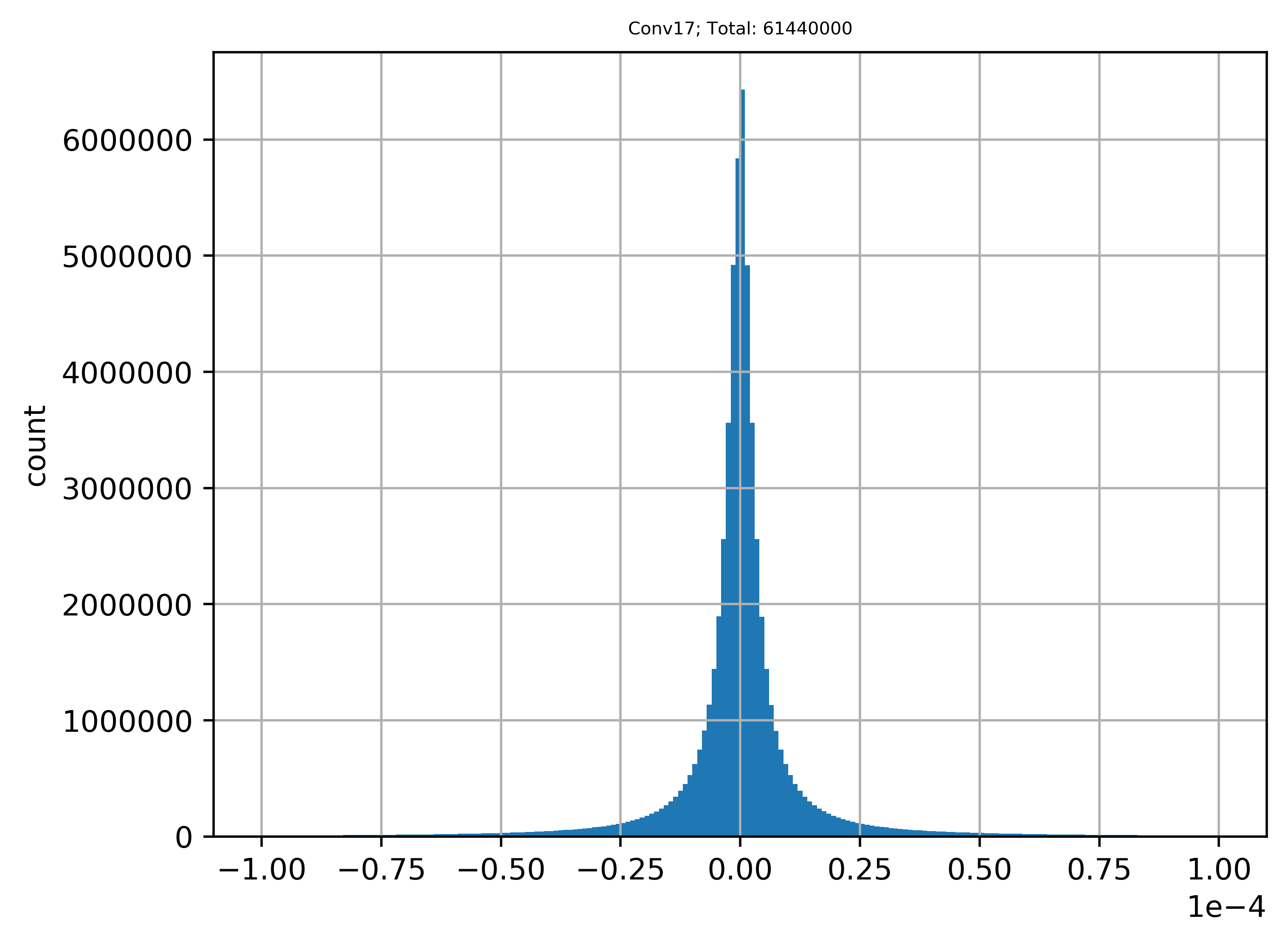} &
			\includegraphics[height=0.1\textwidth]{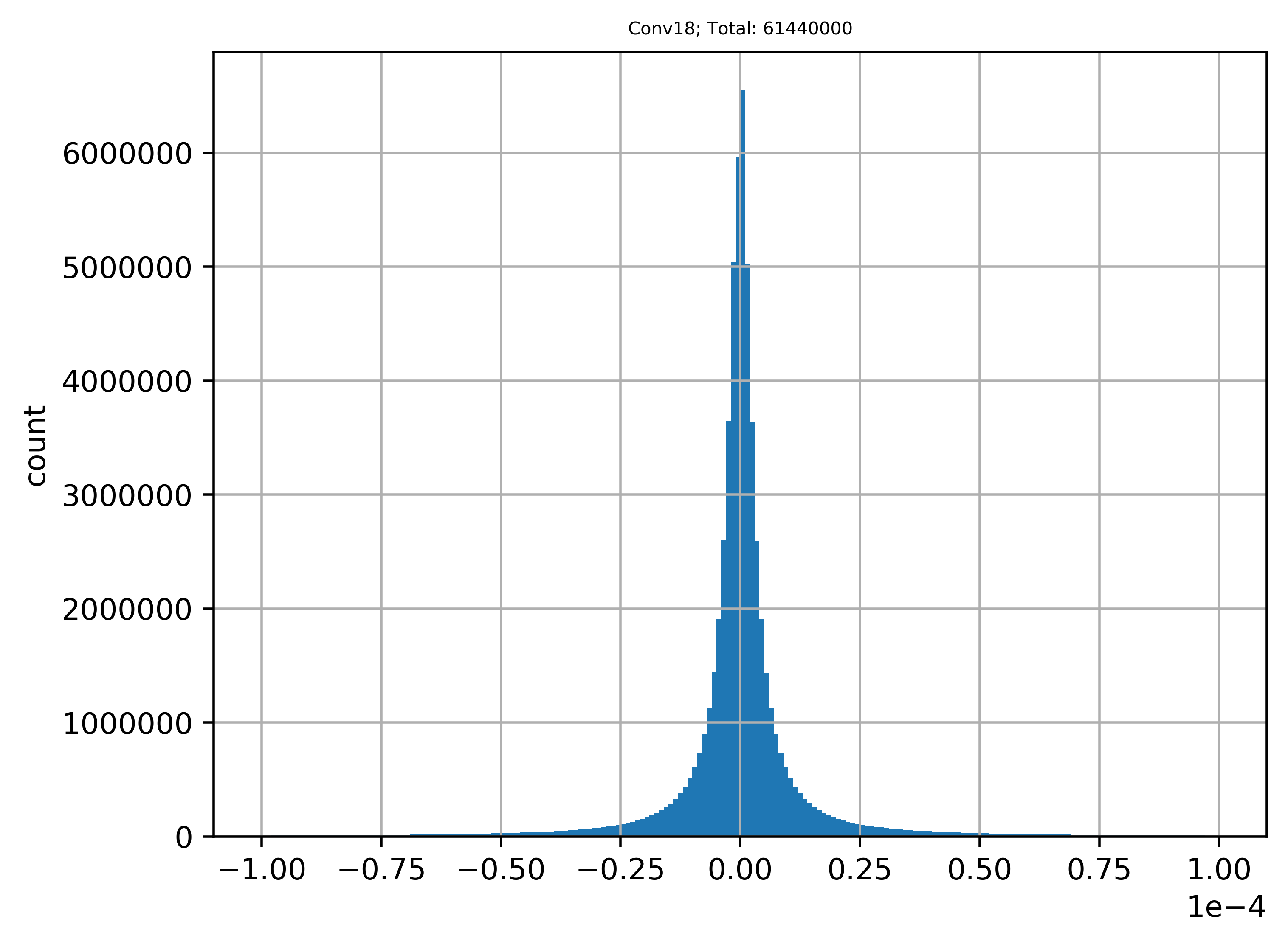} &
			\includegraphics[height=0.1\textwidth]{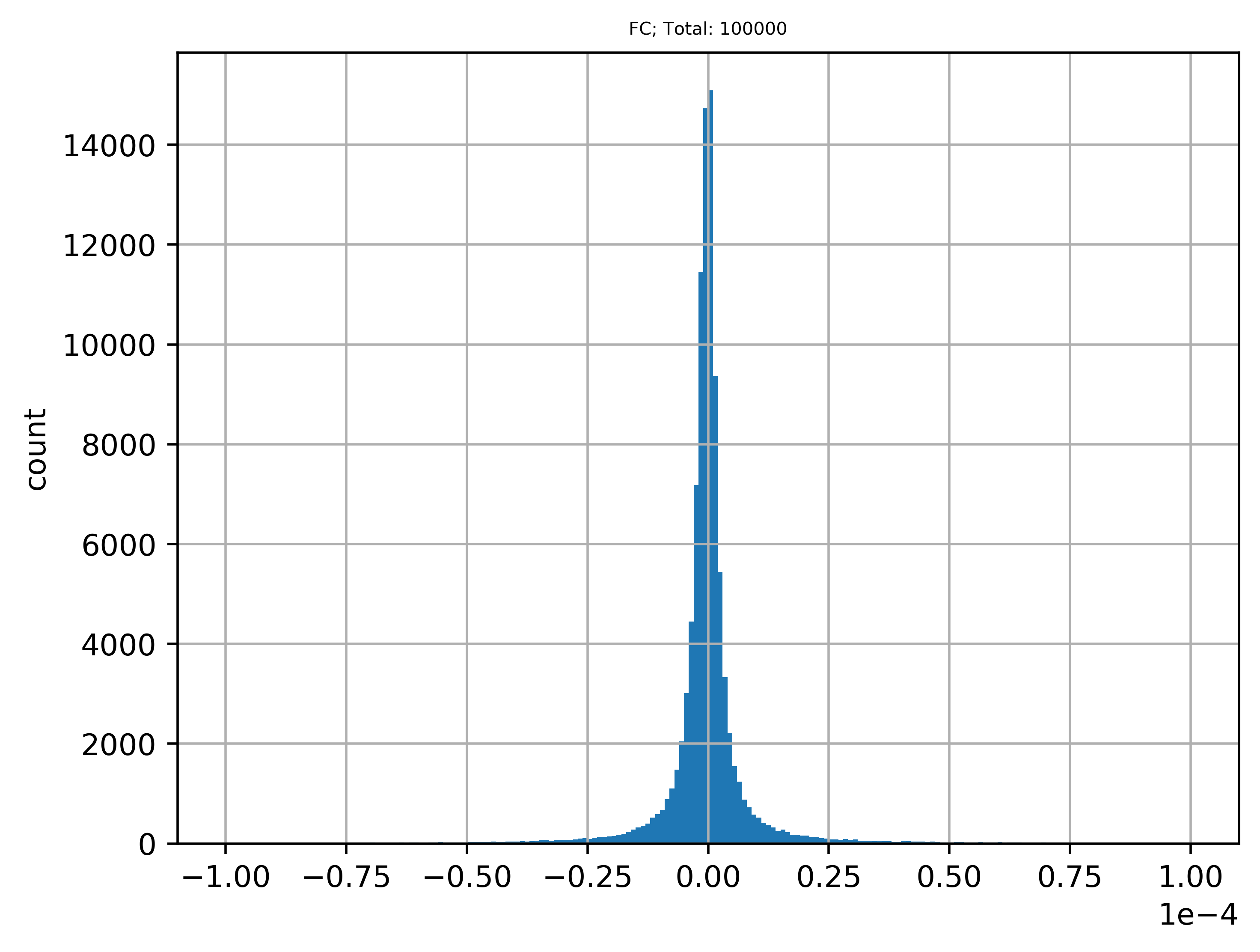} & & \\
			 ($a_{15}$) & ($a_{16}$) & ($a_{17}$) & ($a_{18}$) & ($a_{19}$) & & \\
		\end{tabular}
		\caption{ResNet20 (CIFAR-10). Histograms of relative errors (Eq.\eqref{eq:relative_err}) between units directly computed by CNN ($\mathbf{c}_{l}$) and their values reconstructed by our method ($\hat{\mathbf{c}}_{l}$) on CIFAR-10 using ResNet20. Similar to Fig.\ref{fig:Statistics:cifar10:vgg7}, $a_1$$\sim$$a_{18}$ is collected from Conv1$\sim$18 layer, and $a_{19}$ is collected from the FC layer. In terms of the $5^{th}$ column of table.\ref{table:params:resnet20}, $a_1$ to $a_{19}$ should have relative errors from $327.68m$($a_1$$\sim$$a_6$), $163.84m$($a_7$$\sim$$a_{12}$), $61.44m$($a_{13}$$\sim$$a_{18}$), and $100k$($a_{19}$), units, respectively. Percentages of $\mathbf{\epsilon}_{l} \le 1\%$ for all subplots are listed in Table \ref{table:percentage_of_errs:cifar10}.}
		\label{fig:Statistics:cifar10:resnet20}
	\end{figure}

\begin{figure}[!h]
		\centering
		\begin{tabular}{@{}c@{}c@{}c@{}c@{}c@{}c@{}c@{}}
			\includegraphics[height=0.1\textwidth]{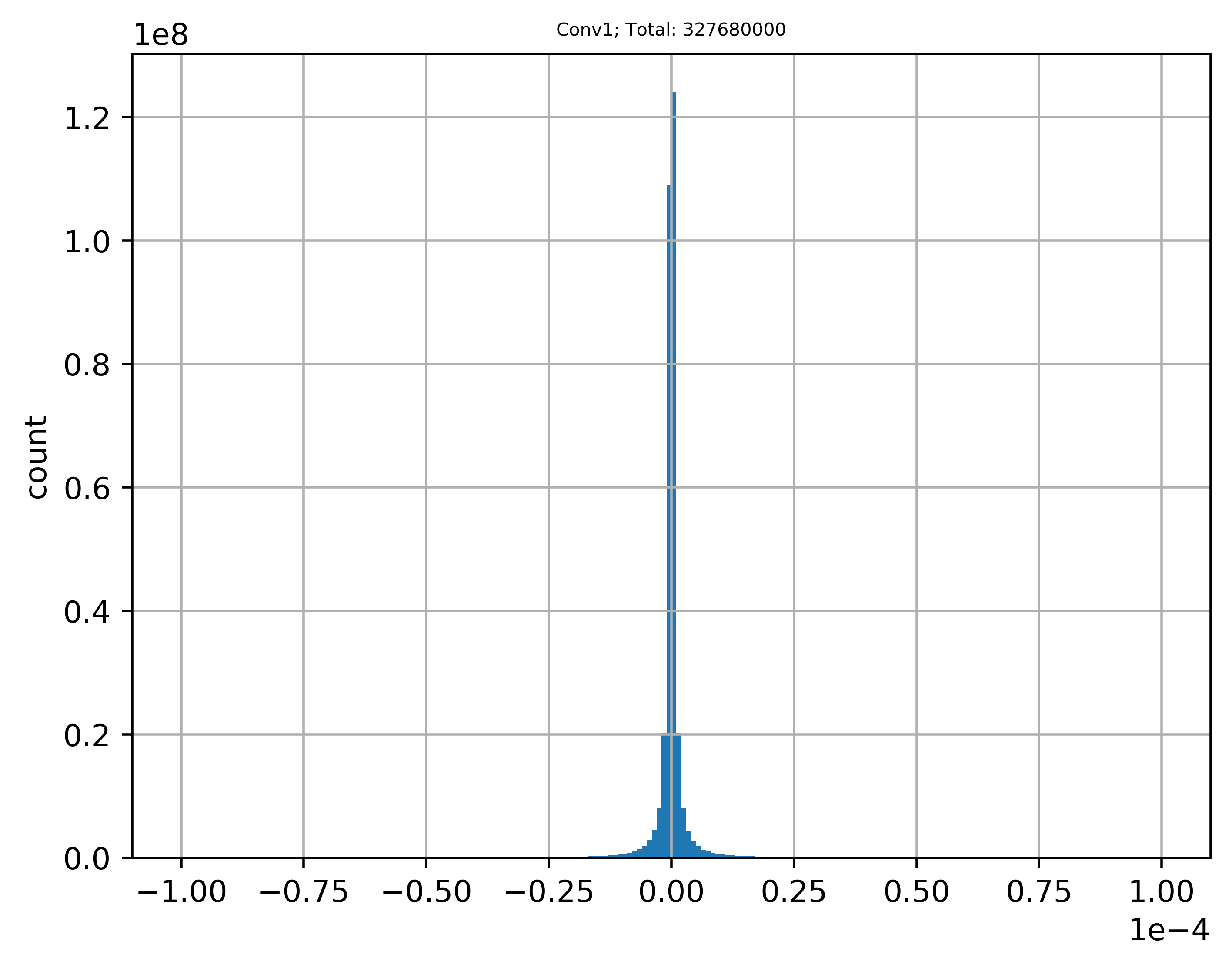} & 
			\includegraphics[height=0.1\textwidth]{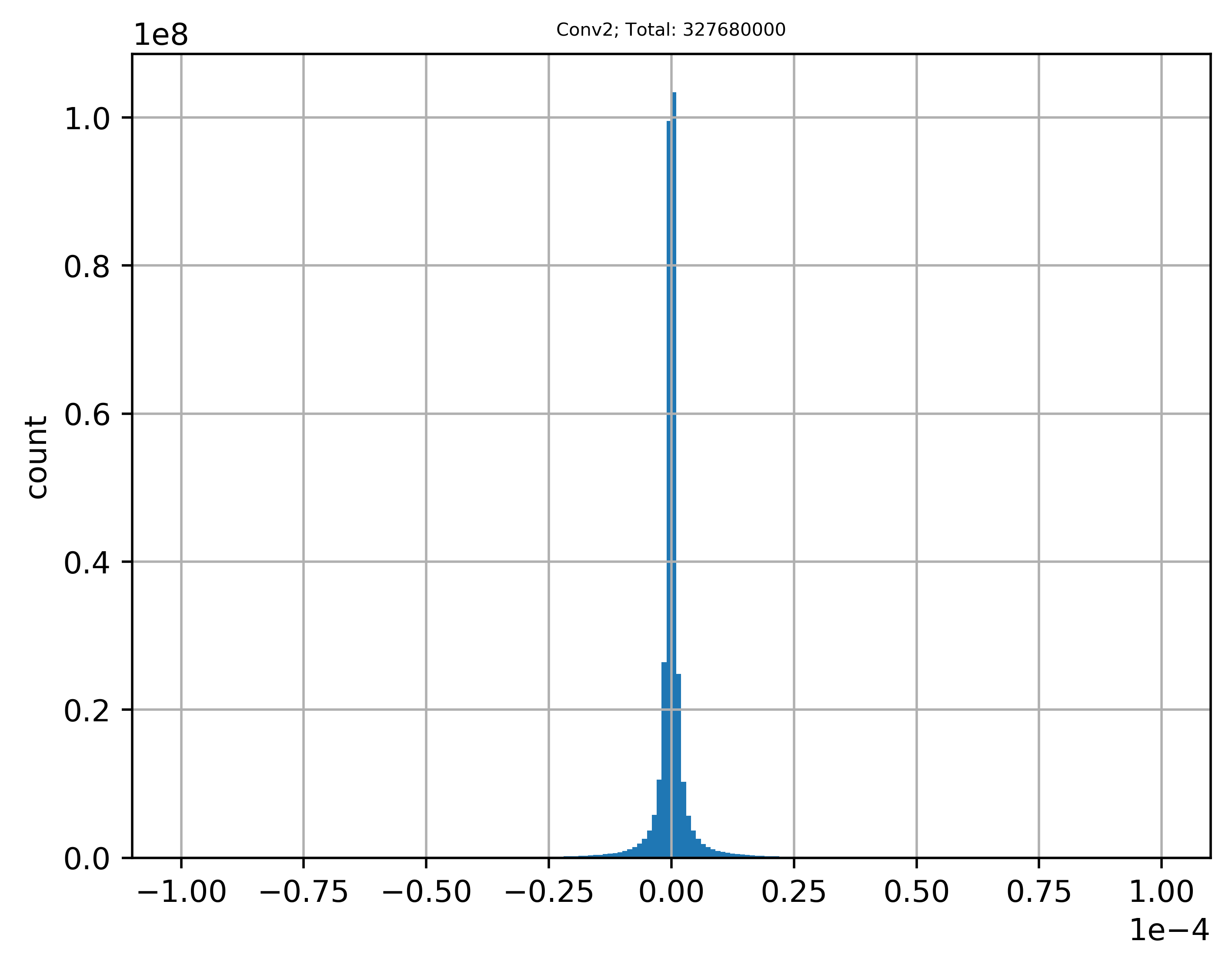} &
			\includegraphics[height=0.1\textwidth]{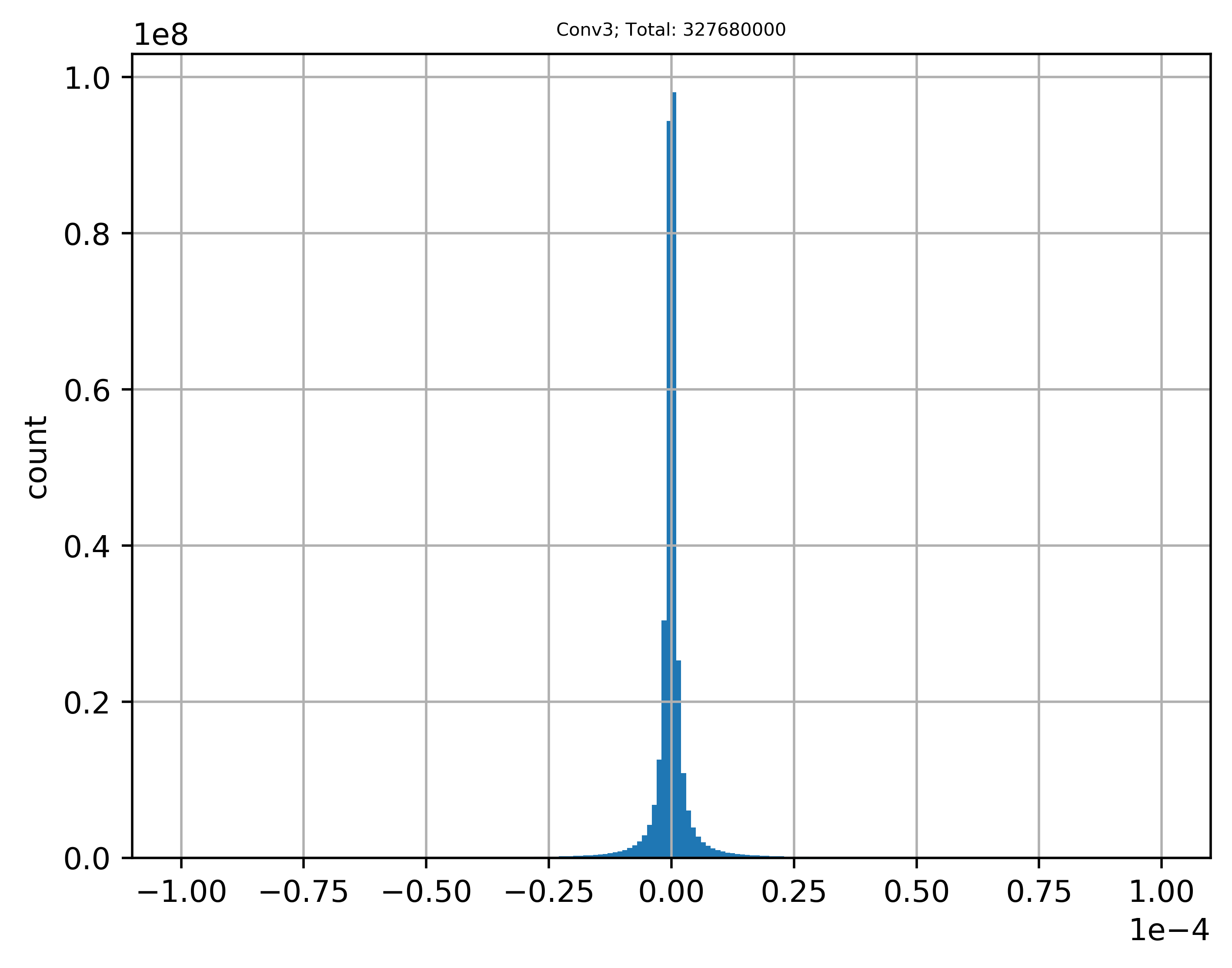} &
			\includegraphics[height=0.1\textwidth]{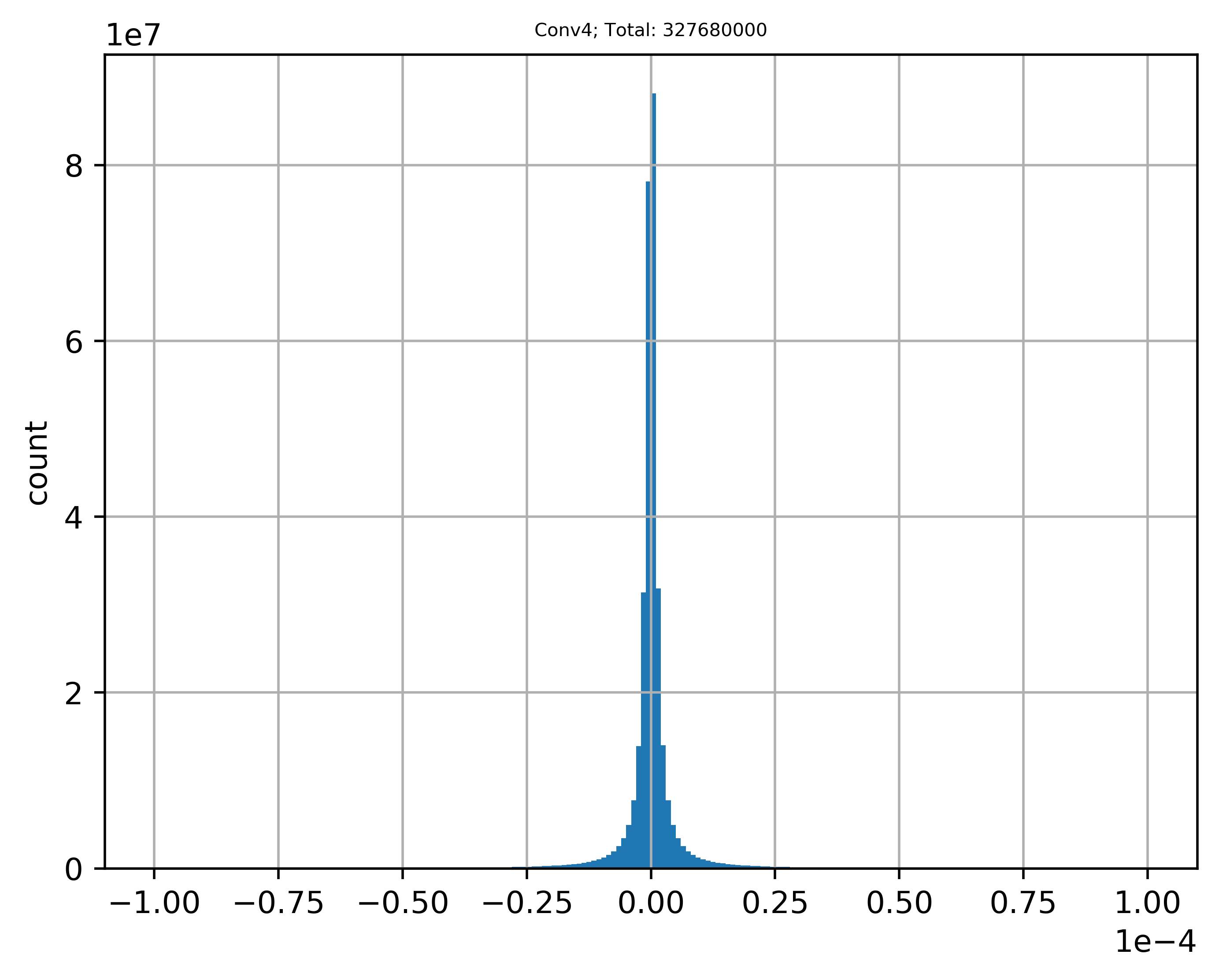} &
			\includegraphics[height=0.1\textwidth]{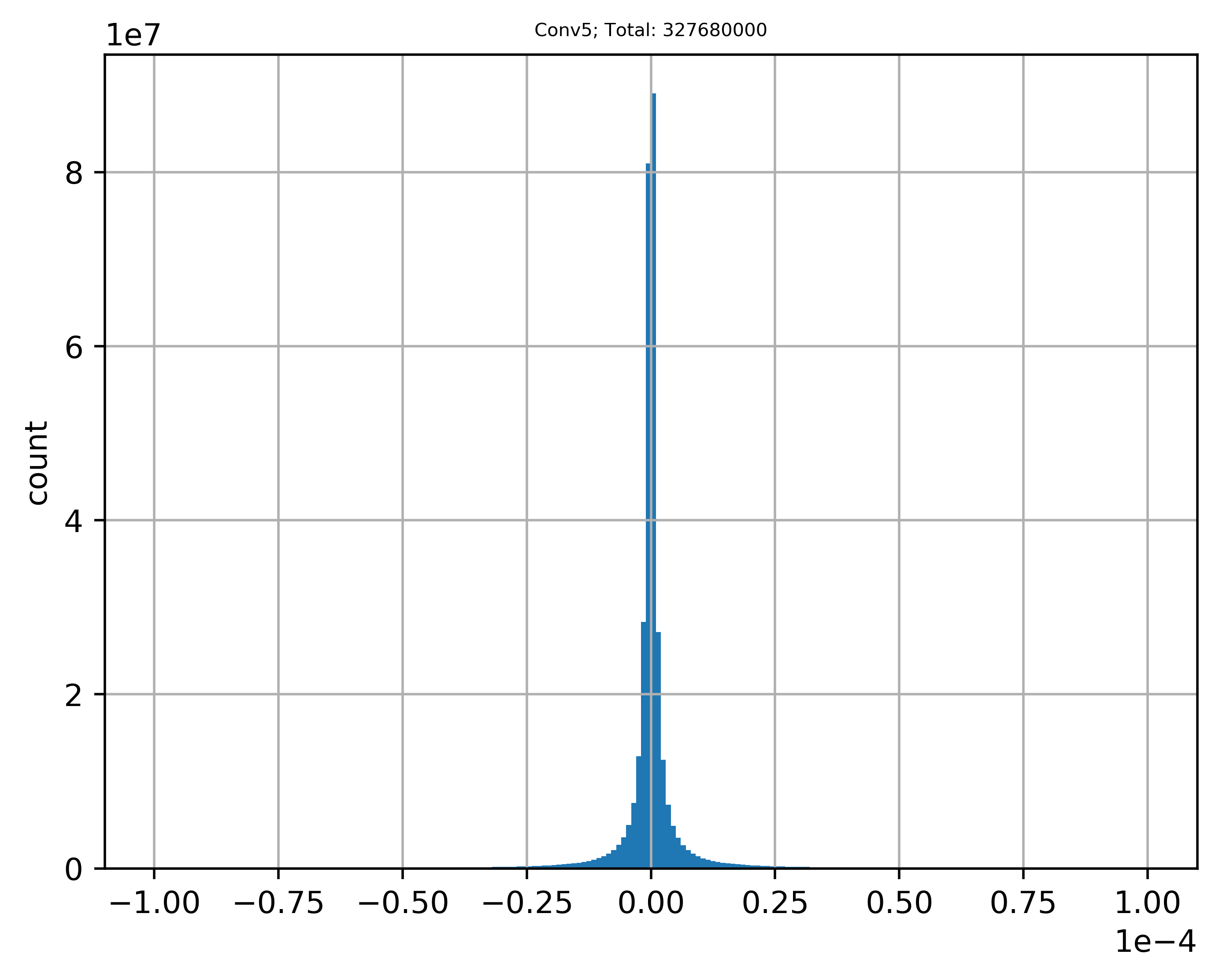} &
			\includegraphics[height=0.1\textwidth]{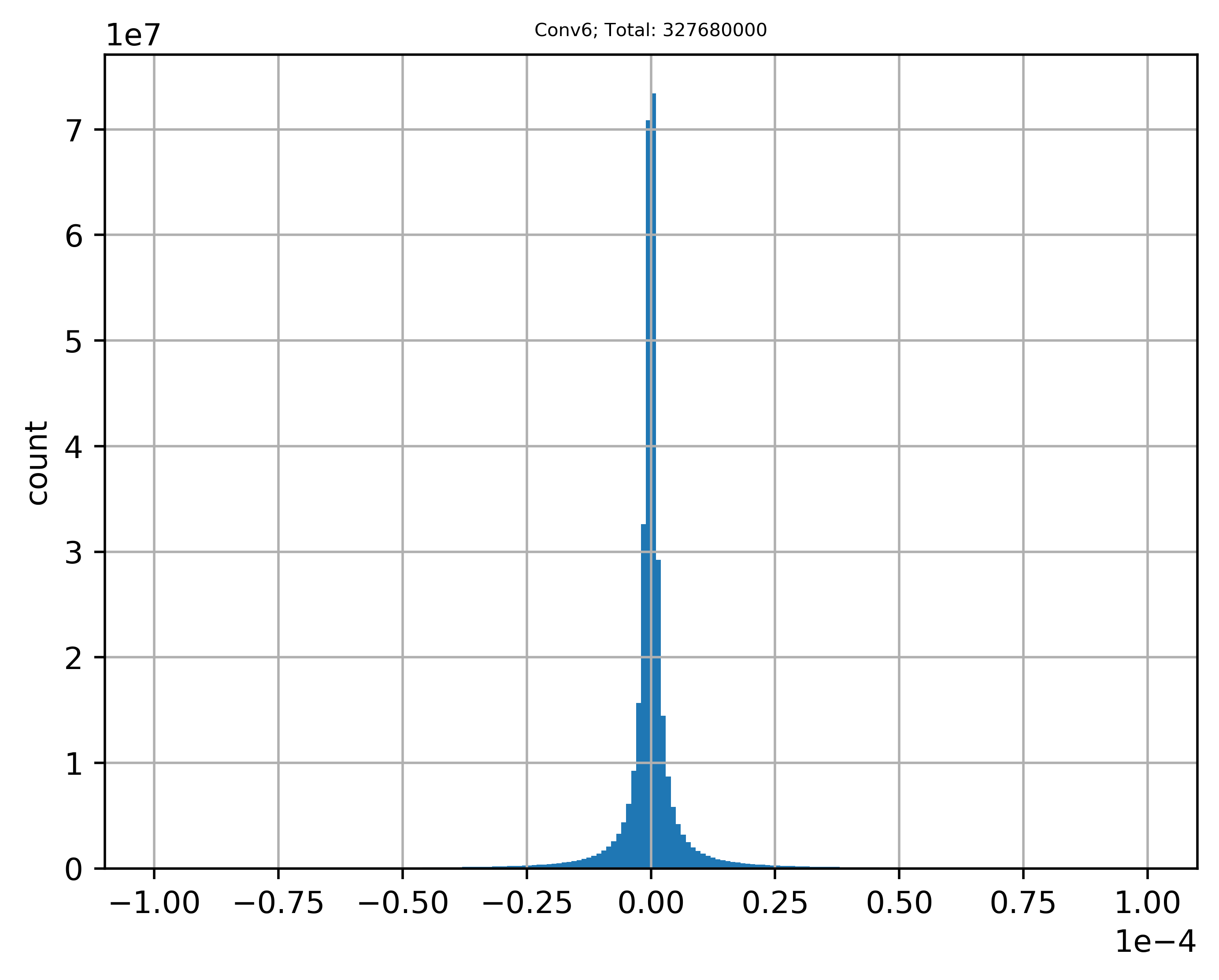} &
			\includegraphics[height=0.1\textwidth]{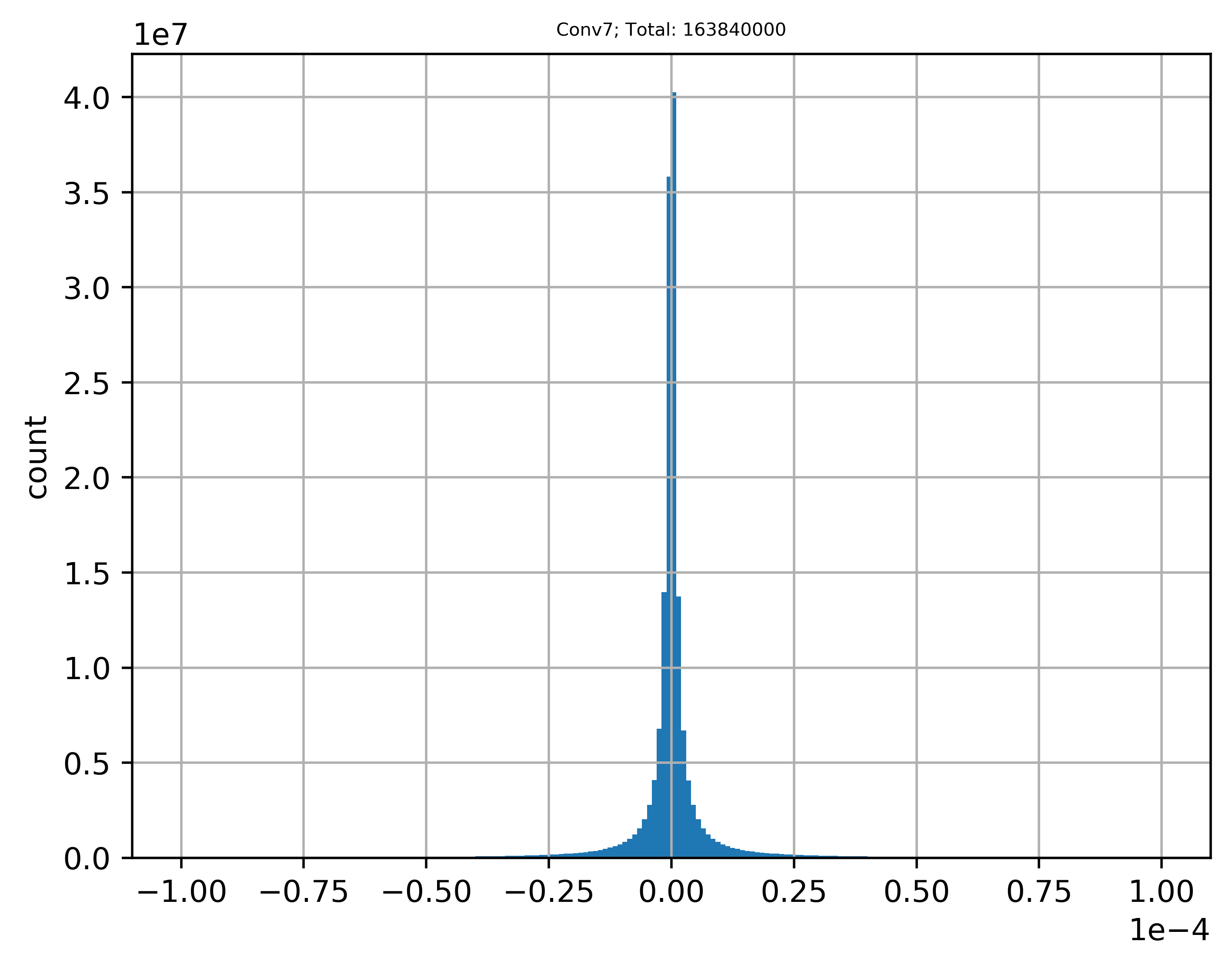} \\
			($a_1$) & ($a_2$) & ($a_3$) & ($a_4$) & ($a_5$) & ($a_6$) & ($a_7$) \\
			\includegraphics[height=0.1\textwidth]{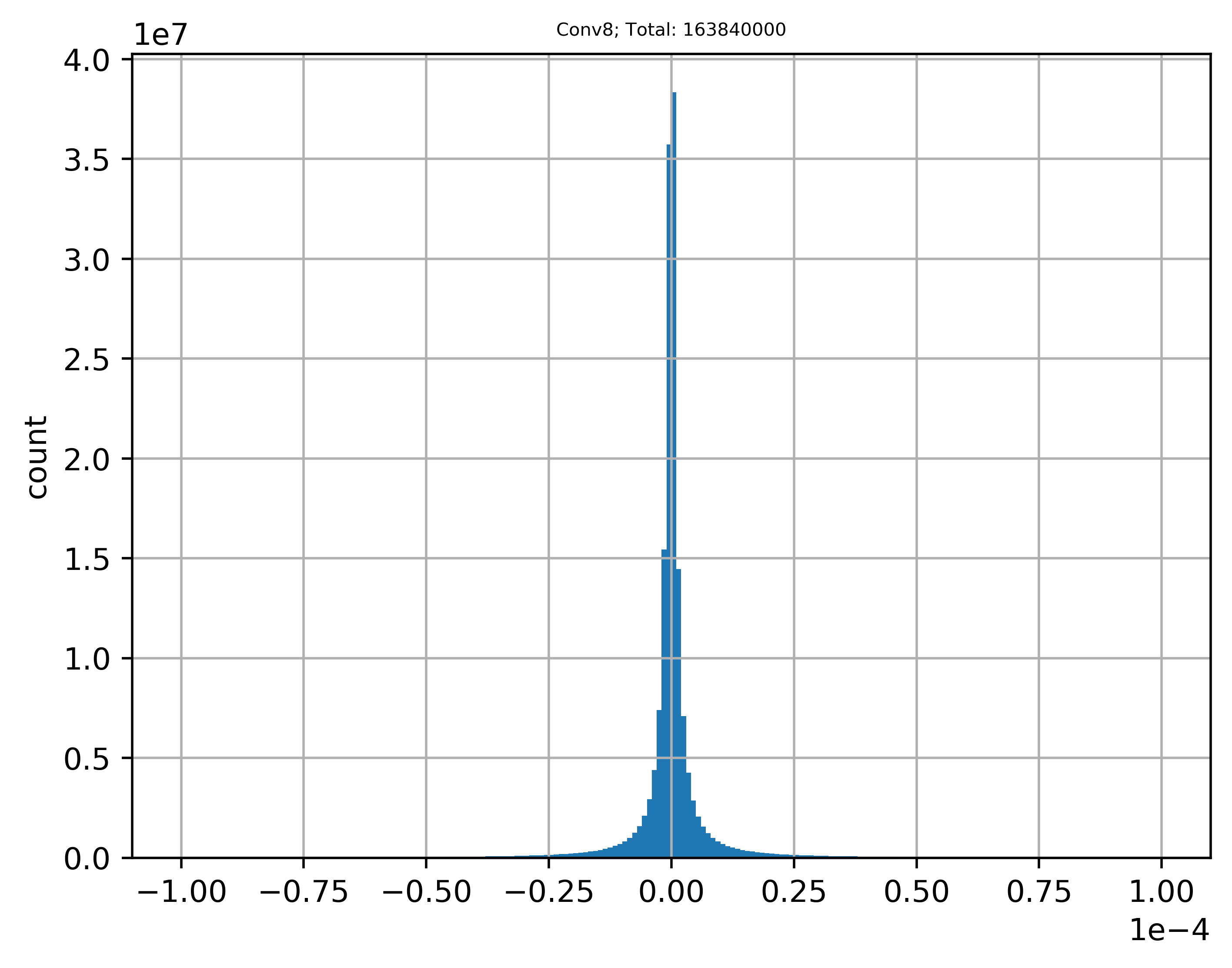} &
			\includegraphics[height=0.1\textwidth]{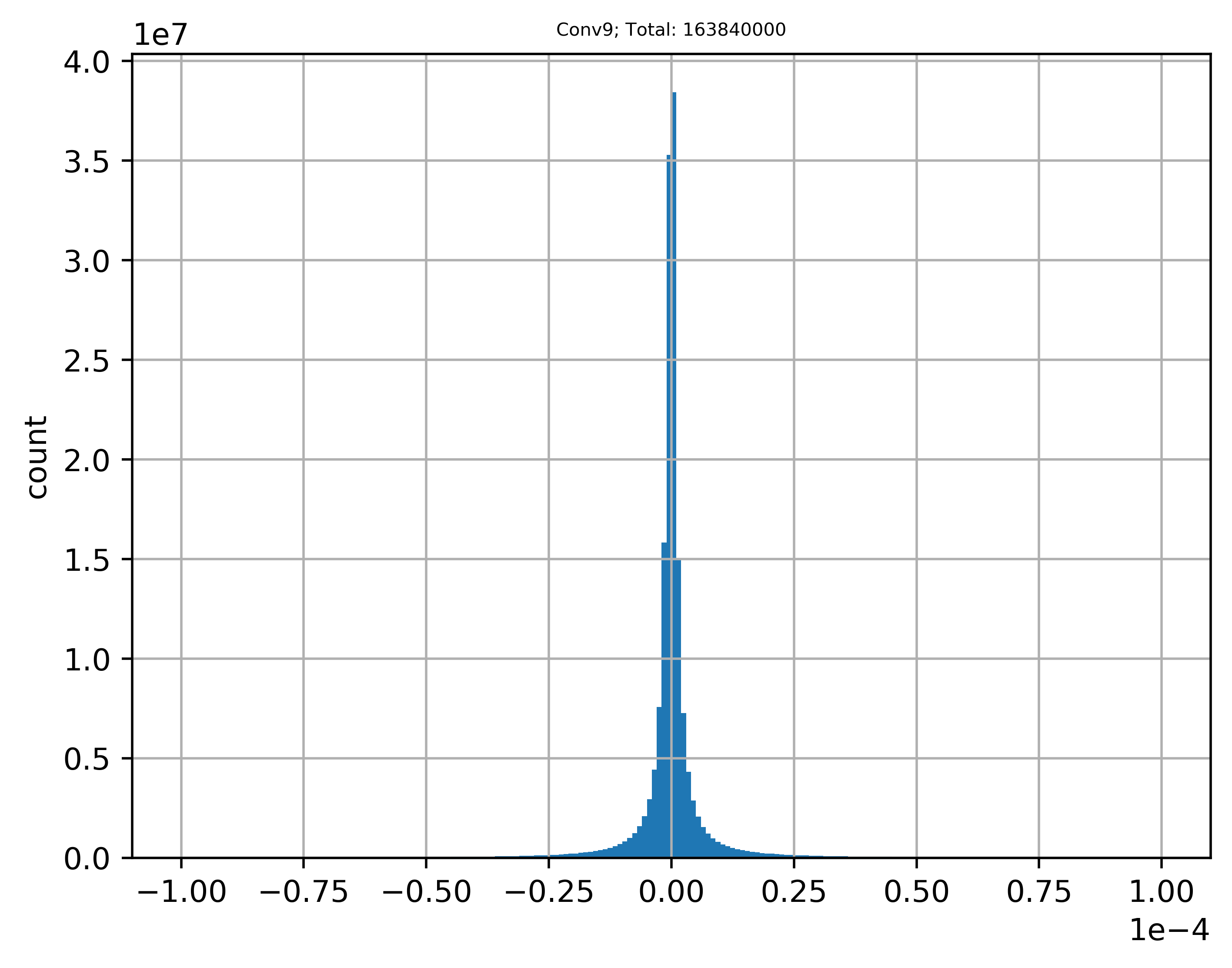} &
			\includegraphics[height=0.1\textwidth]{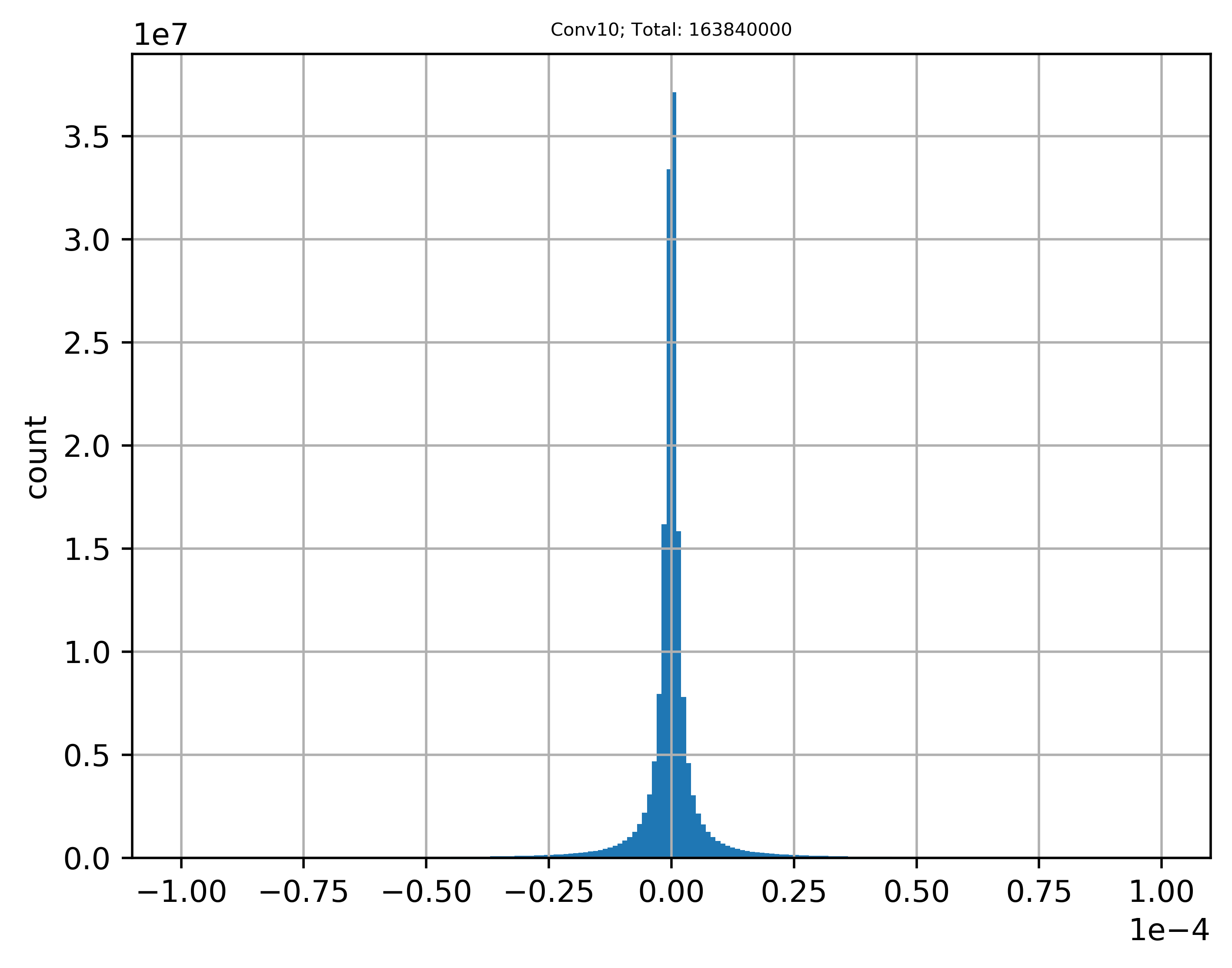} &
			\includegraphics[height=0.1\textwidth]{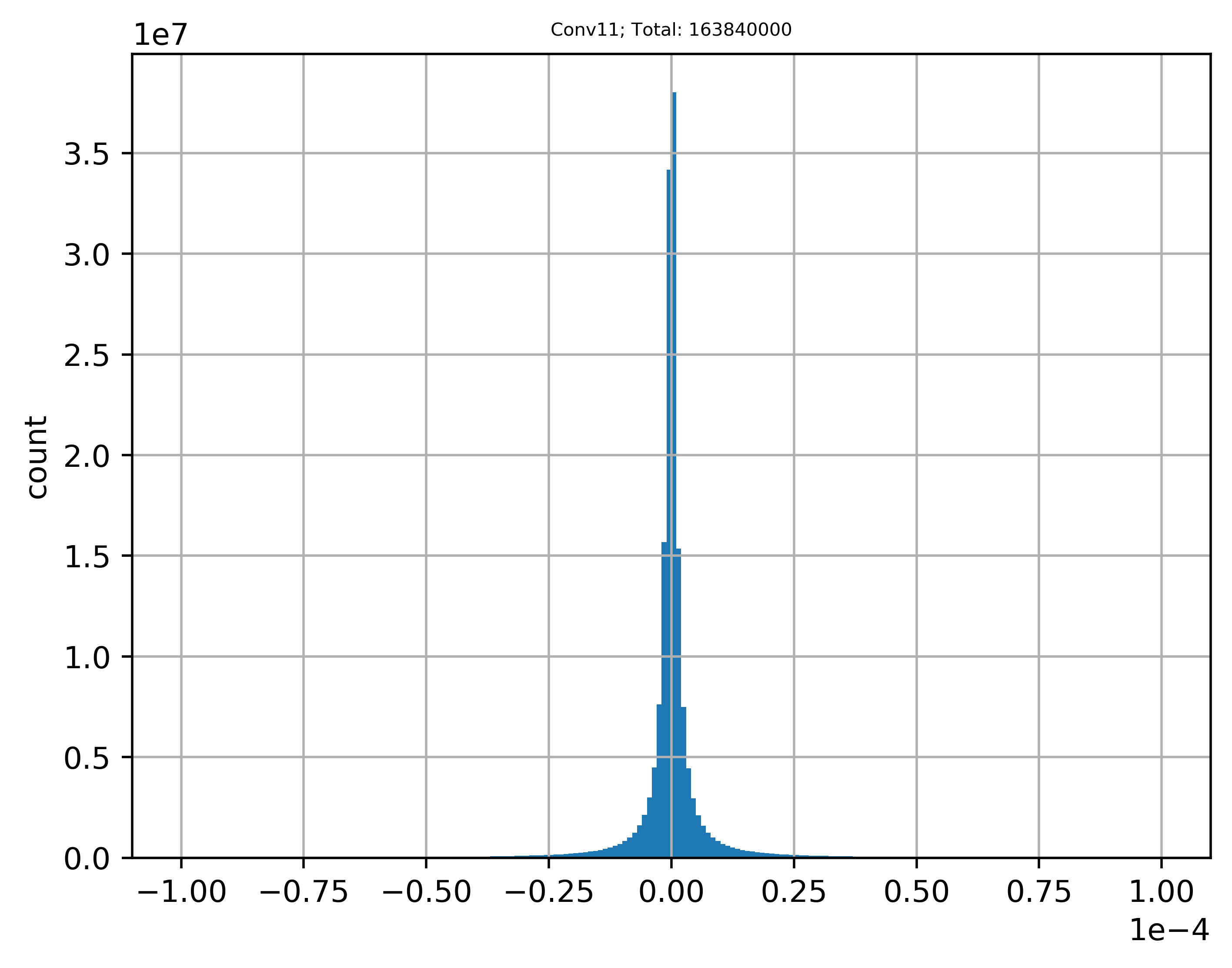} &
			\includegraphics[height=0.1\textwidth]{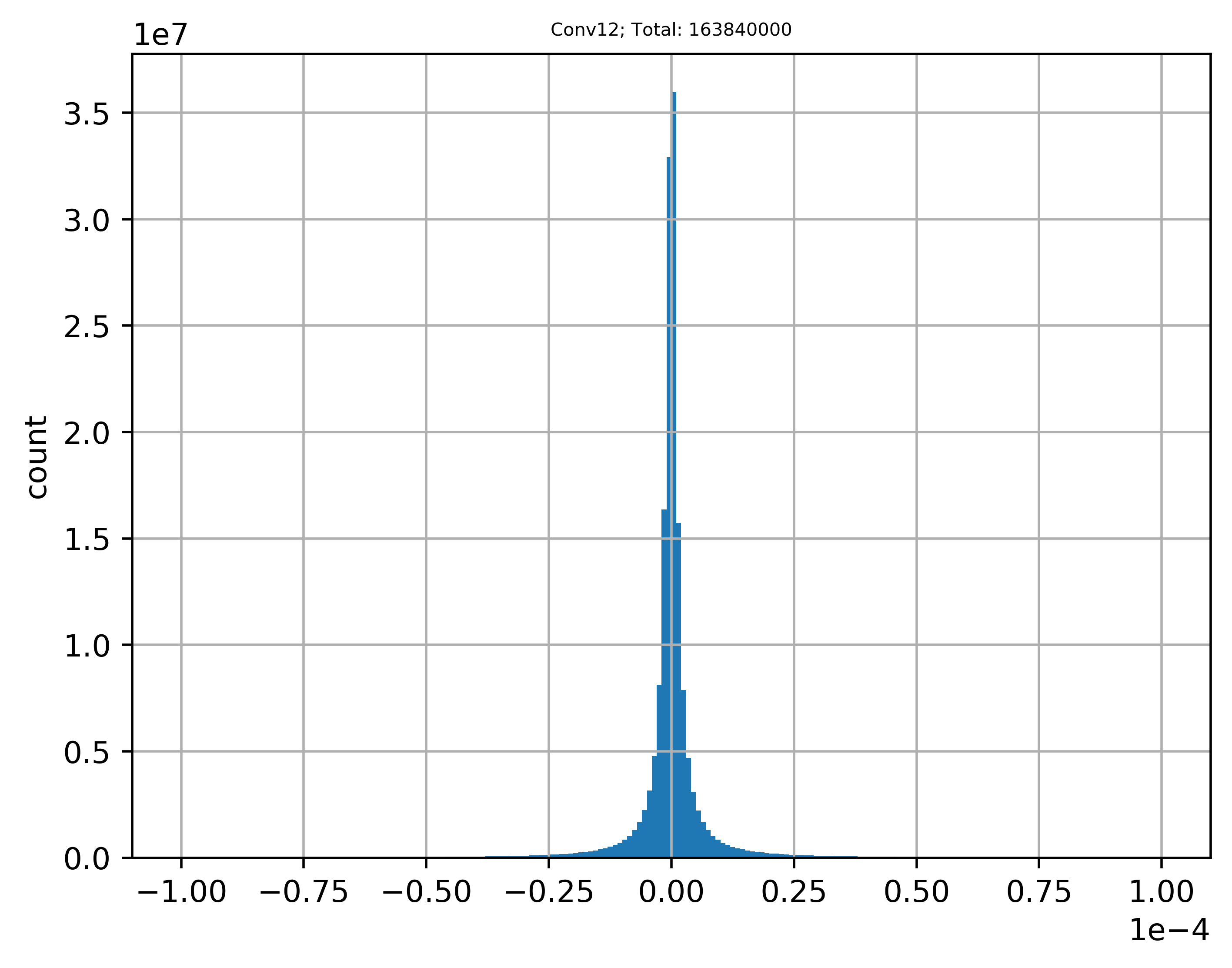} &
			\includegraphics[height=0.1\textwidth]{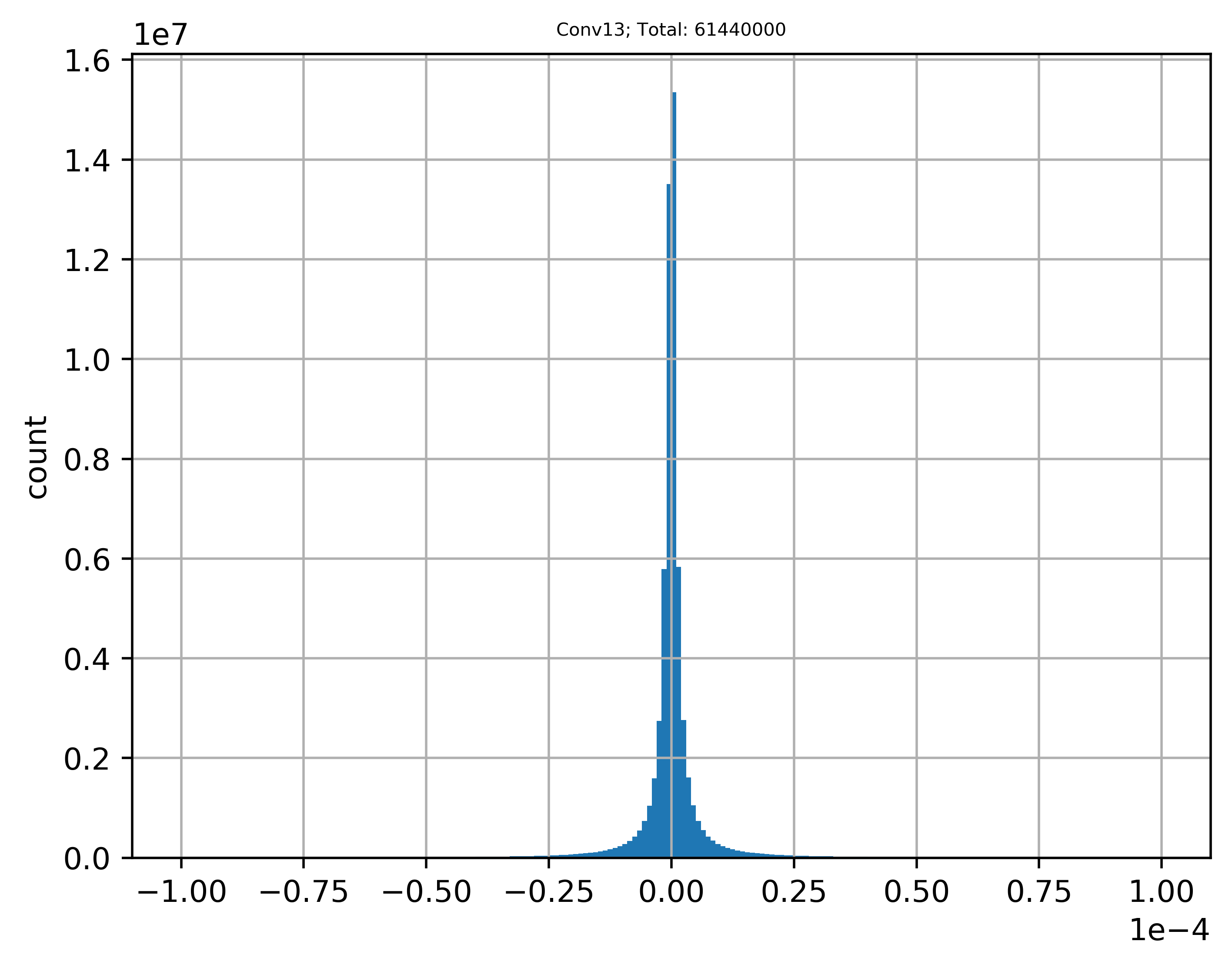} &
			\includegraphics[height=0.1\textwidth]{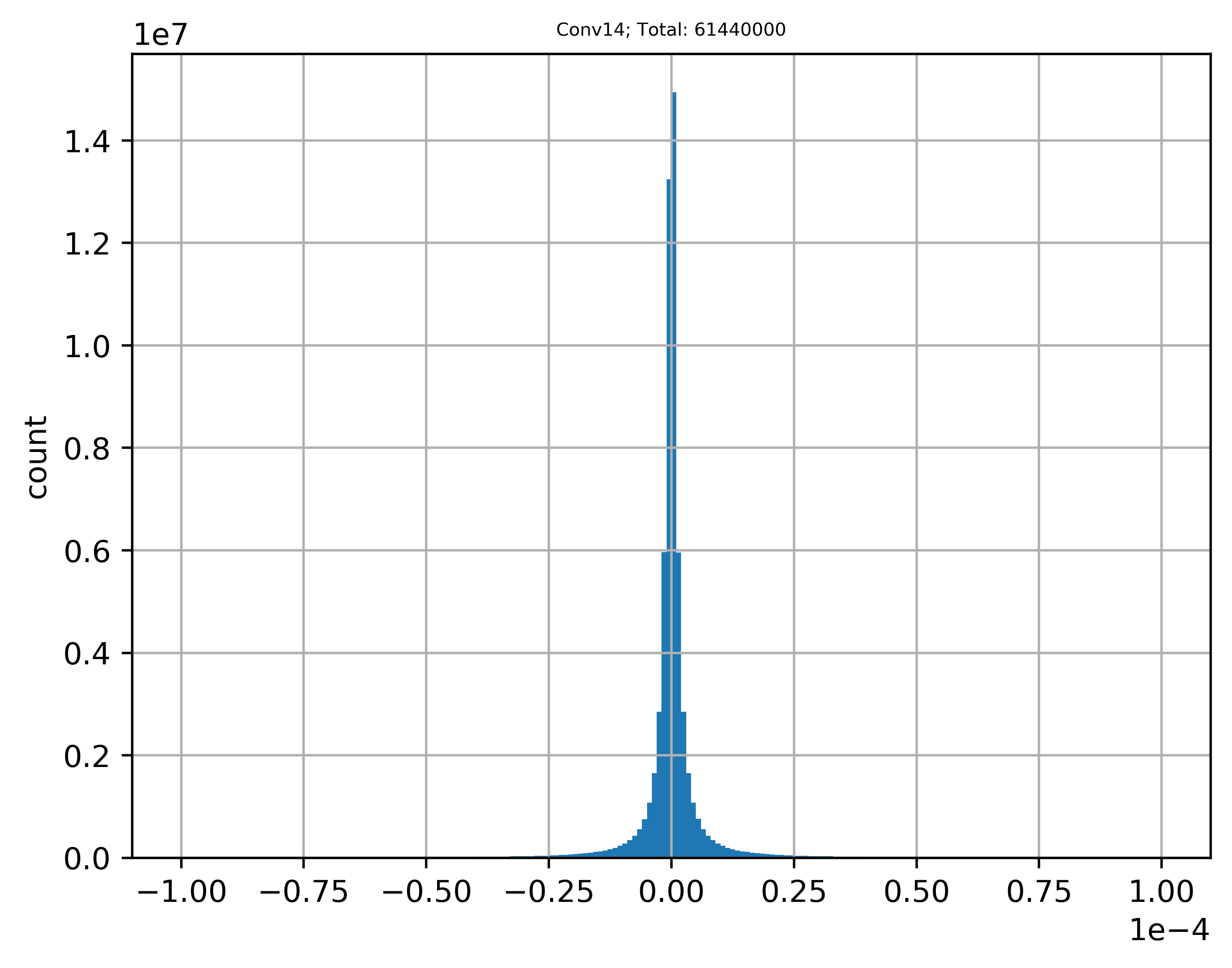} \\
			($a_8$) & ($a_9$) & ($a_{10}$) & ($a_{11}$) & ($a_{12}$) & ($a_{13}$) & ($a_{14}$) \\
			\includegraphics[height=0.1\textwidth]{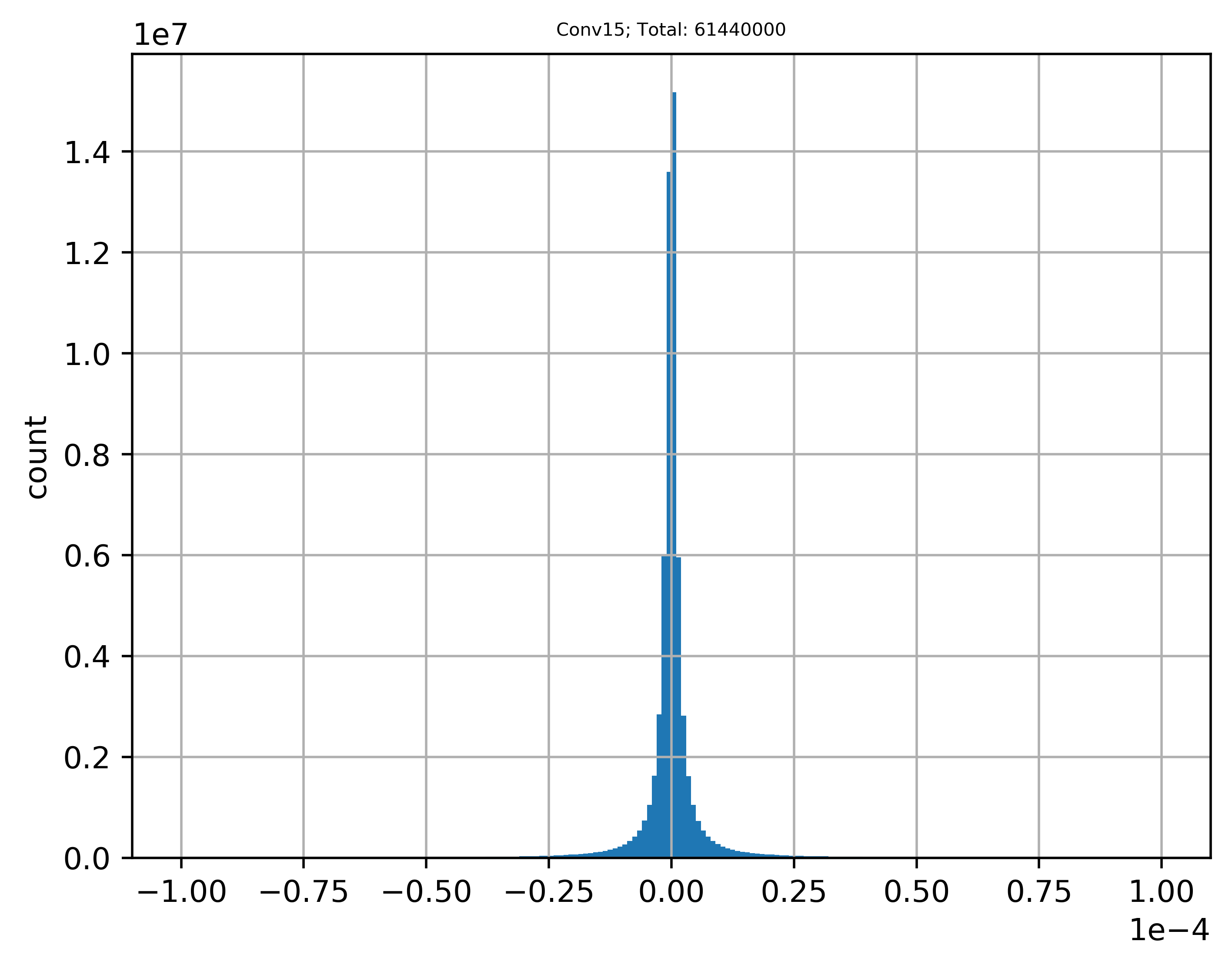} &
			\includegraphics[height=0.1\textwidth]{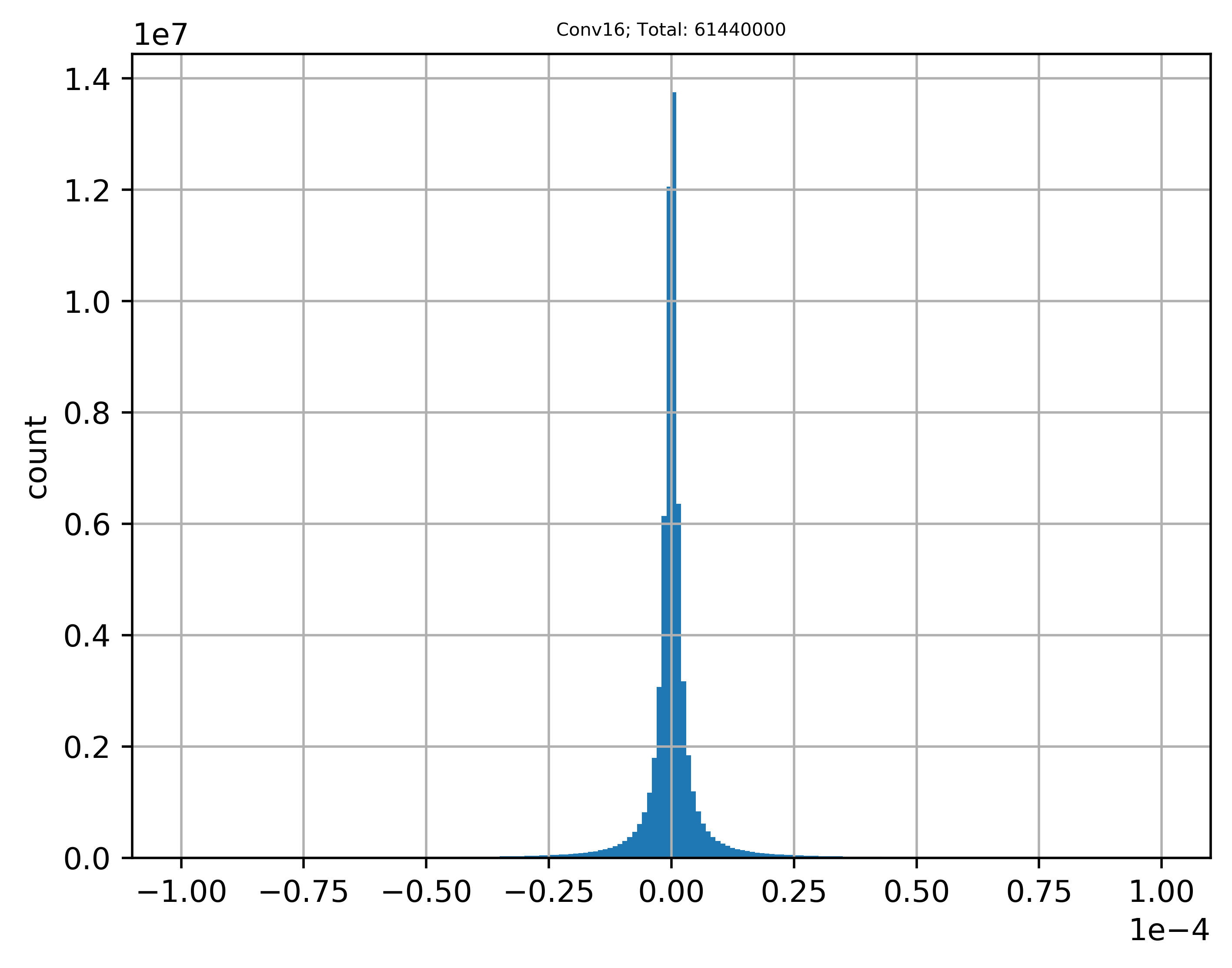} &
			\includegraphics[height=0.1\textwidth]{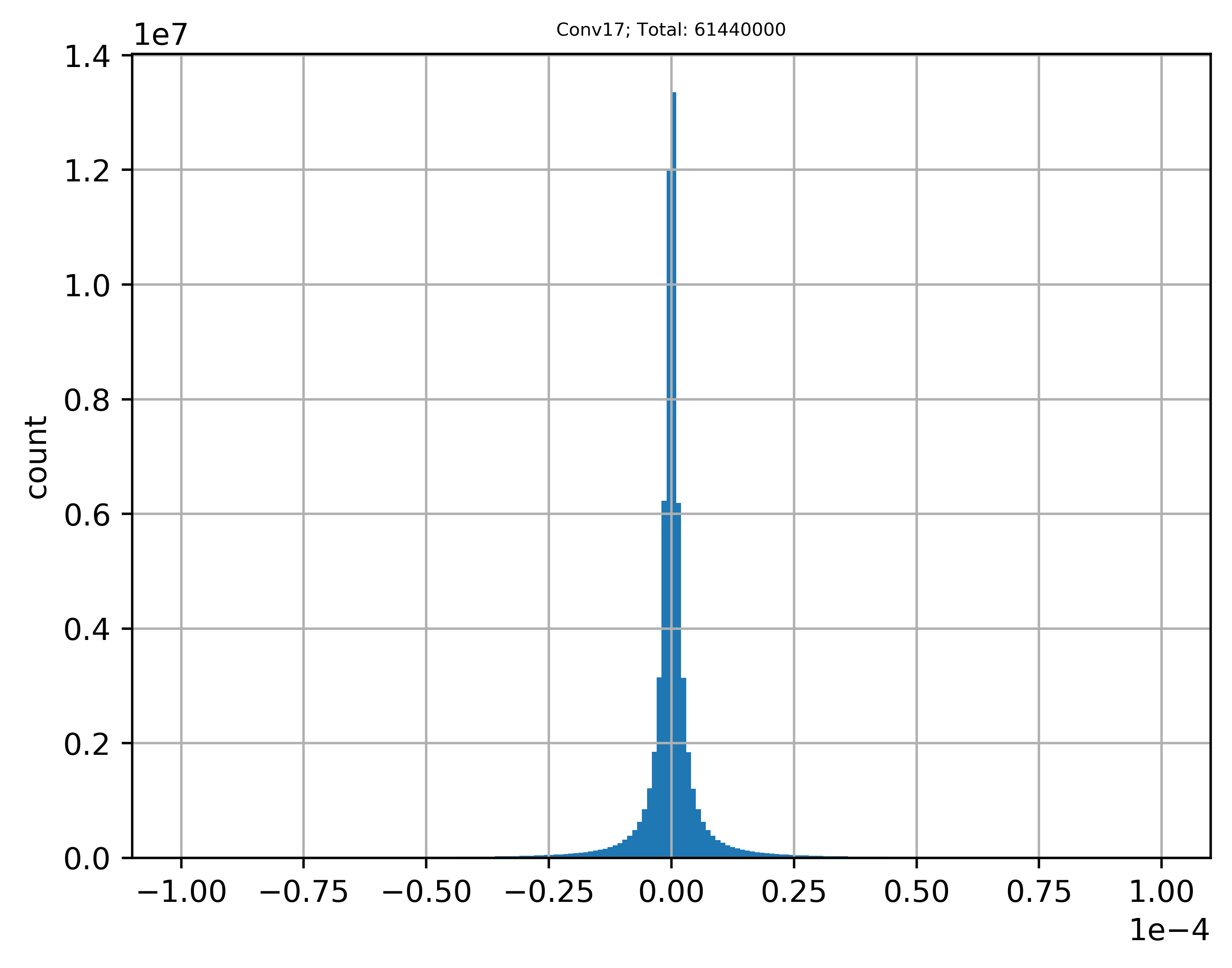} &
			\includegraphics[height=0.1\textwidth]{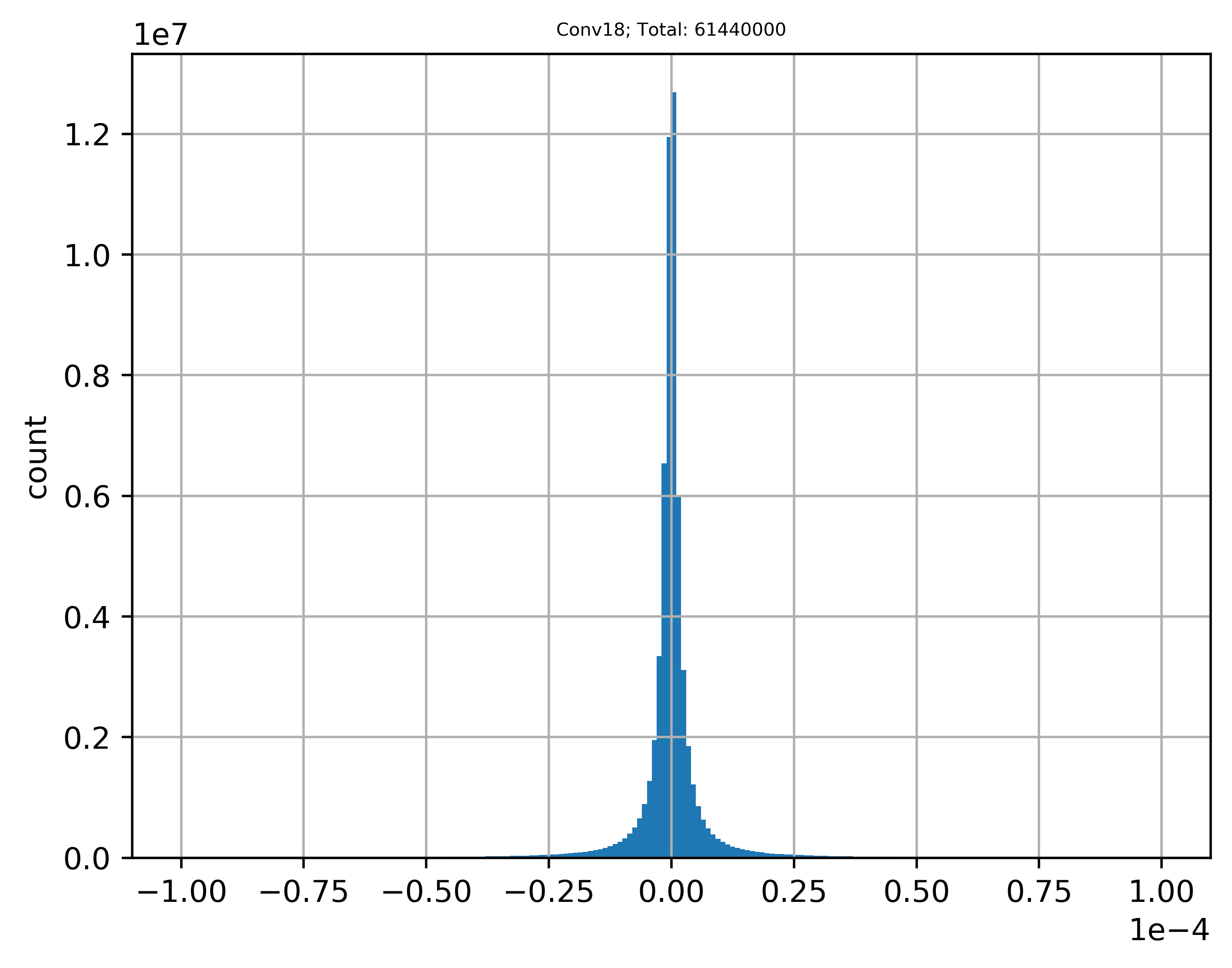} &
			\includegraphics[height=0.1\textwidth]{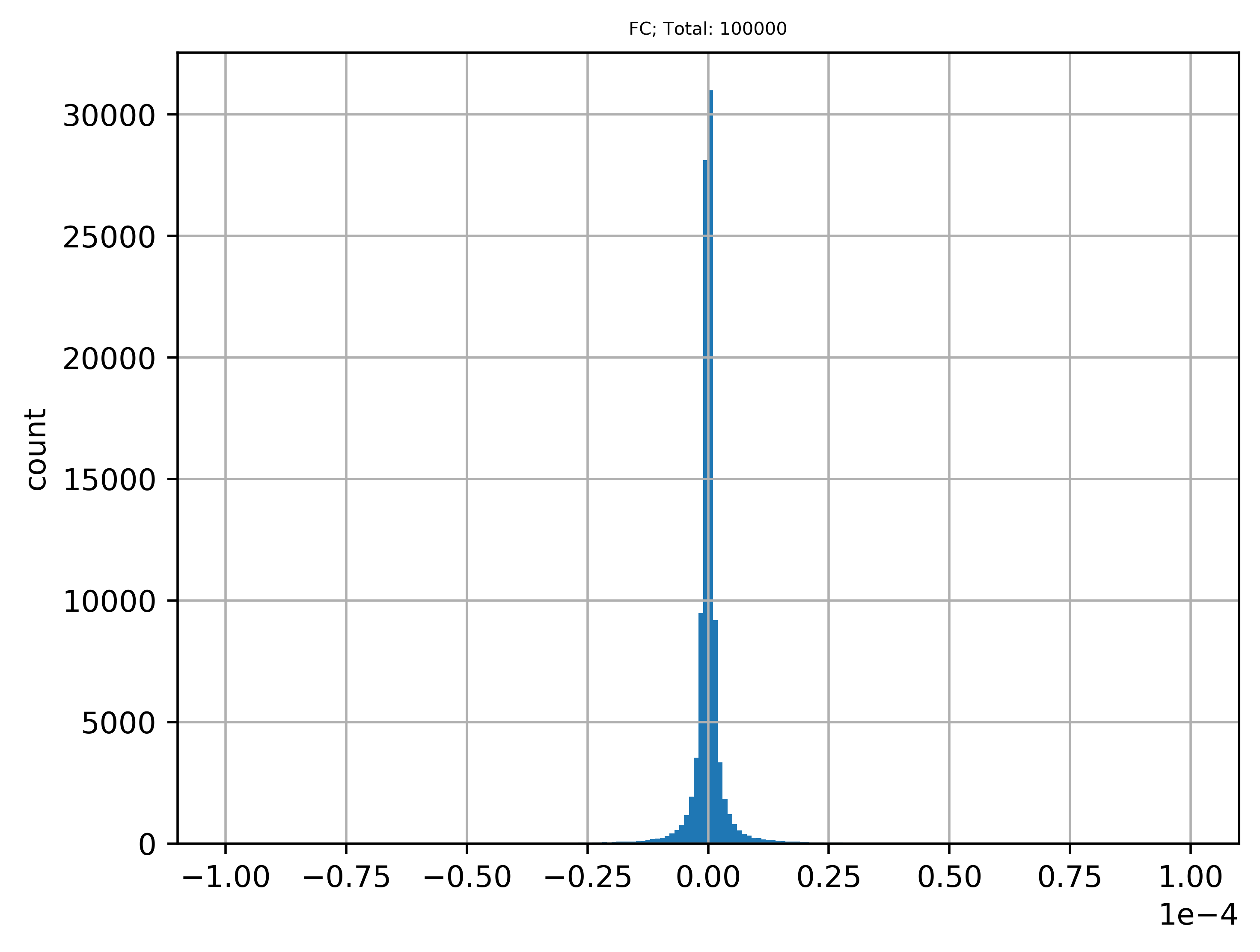} & & \\
			($a_{15}$) & ($a_{16}$) & ($a_{17}$) & ($a_{18}$) & ($a_{19}$) & & \\
		\end{tabular}
		\caption{ResNet20-Fixup (CIFAR-10). Histograms of relative errors (Eq.\eqref{eq:relative_err}) between units directly computed by CNN ($\mathbf{c}_{l}$) and their values reconstructed by our method ($\hat{\mathbf{c}}_{l}$) on CIFAR-10 using ResNet20-Fixup. Details are similar to Fig.\ref{fig:Statistics:cifar10:resnet20}. Percentages of $\mathbf{\epsilon}_{l} \le 1\%$ for all subplots are listed in Table \ref{table:percentage_of_errs:cifar10}.}
		\label{fig:Statistics:cifar10:resnet20_fixup}
	\end{figure}

\begin{figure}[!h]
		\centering
		\begin{tabular}{@{}c@{}c@{}c@{}c@{}c@{}c@{}}
			\includegraphics[height=0.12\textwidth]{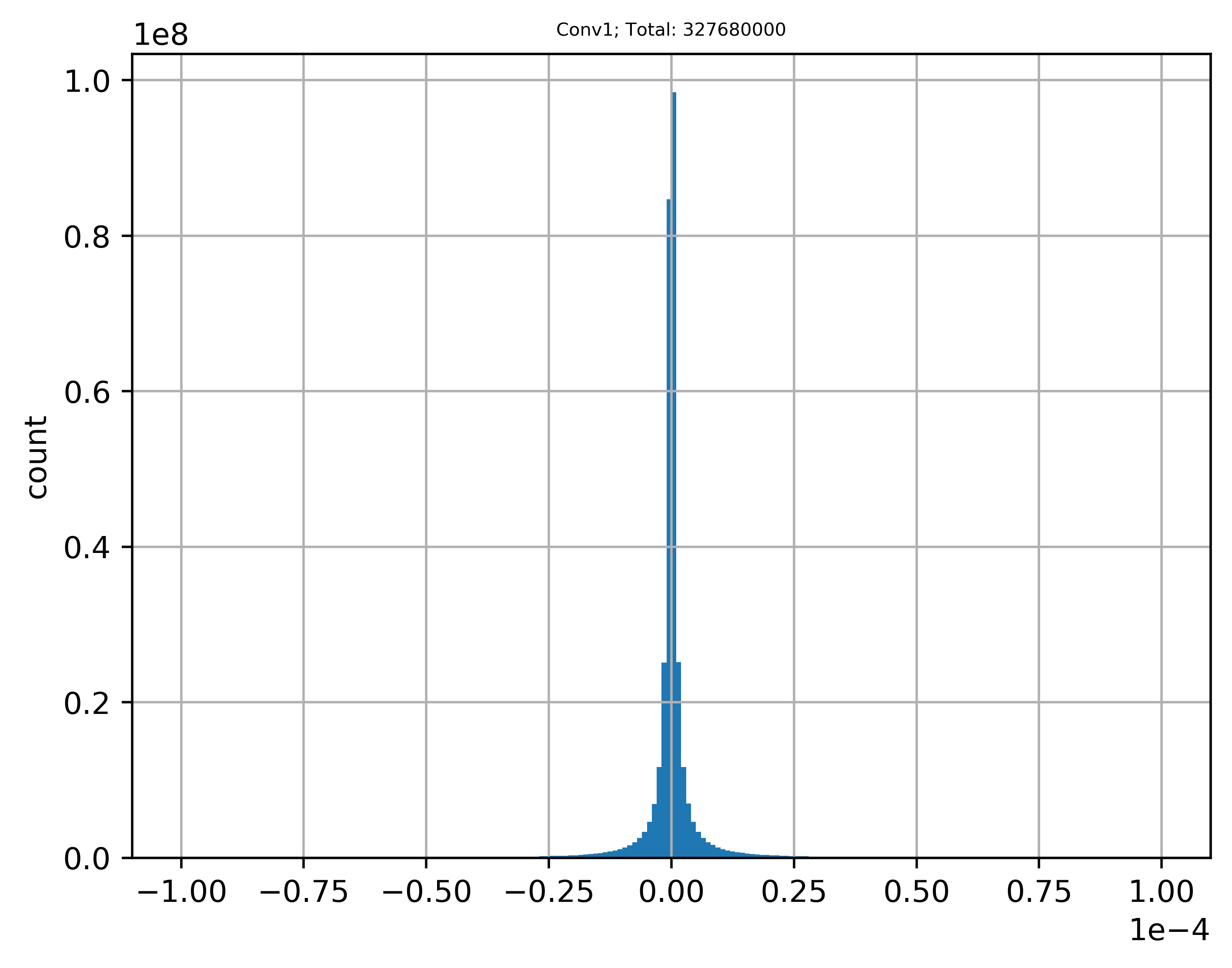} & 
			\includegraphics[height=0.12\textwidth]{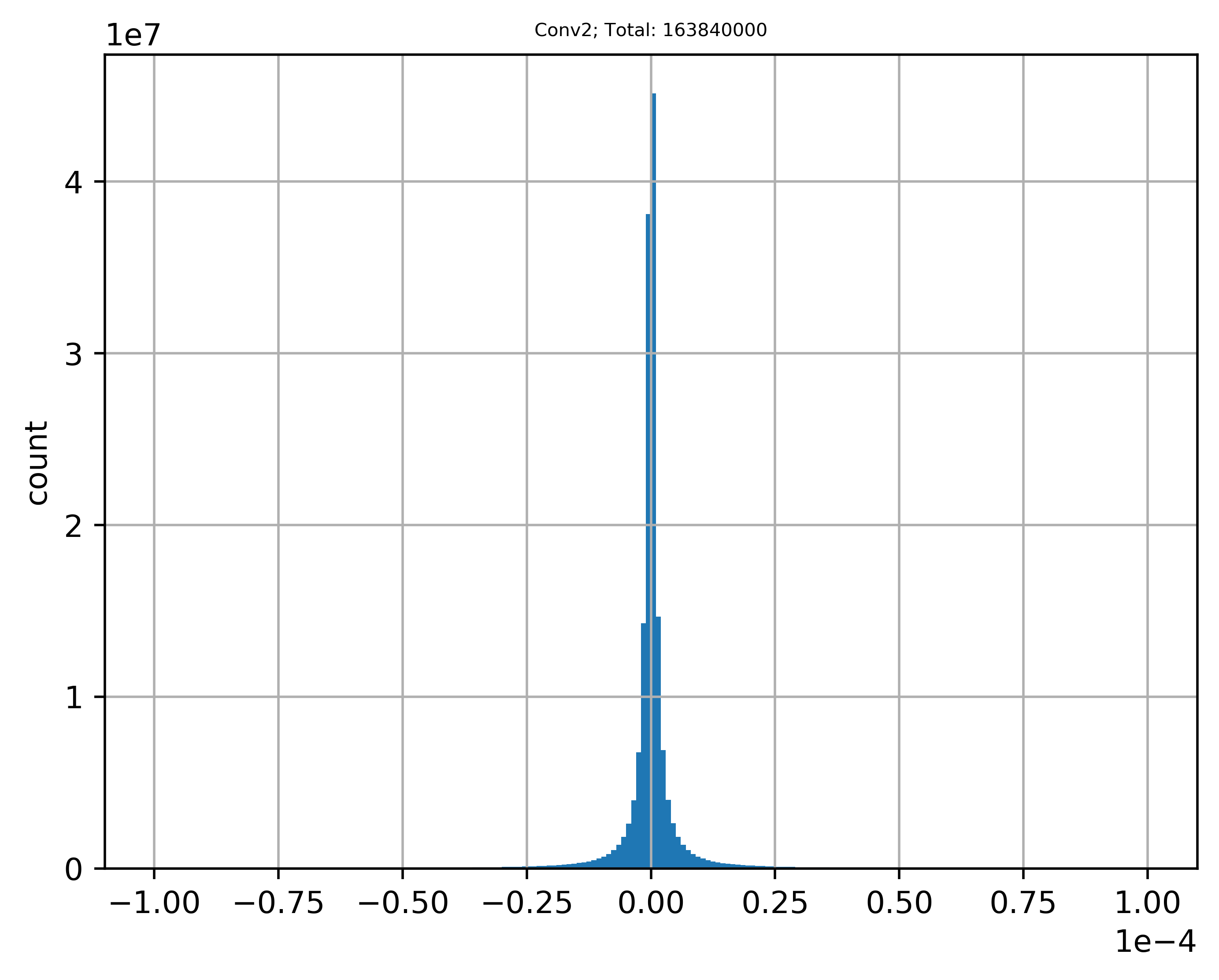} &
			\includegraphics[height=0.12\textwidth]{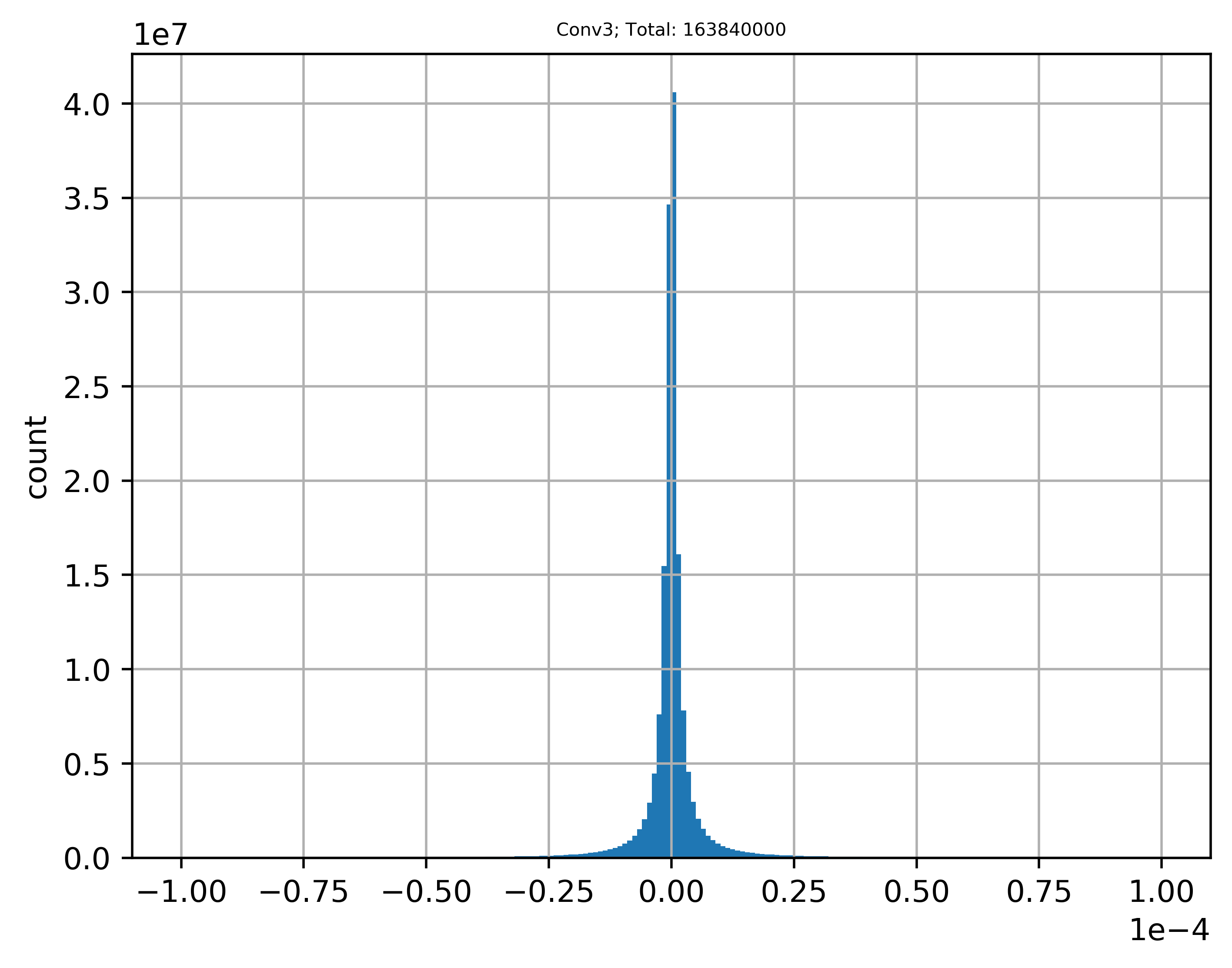} &
			\includegraphics[height=0.12\textwidth]{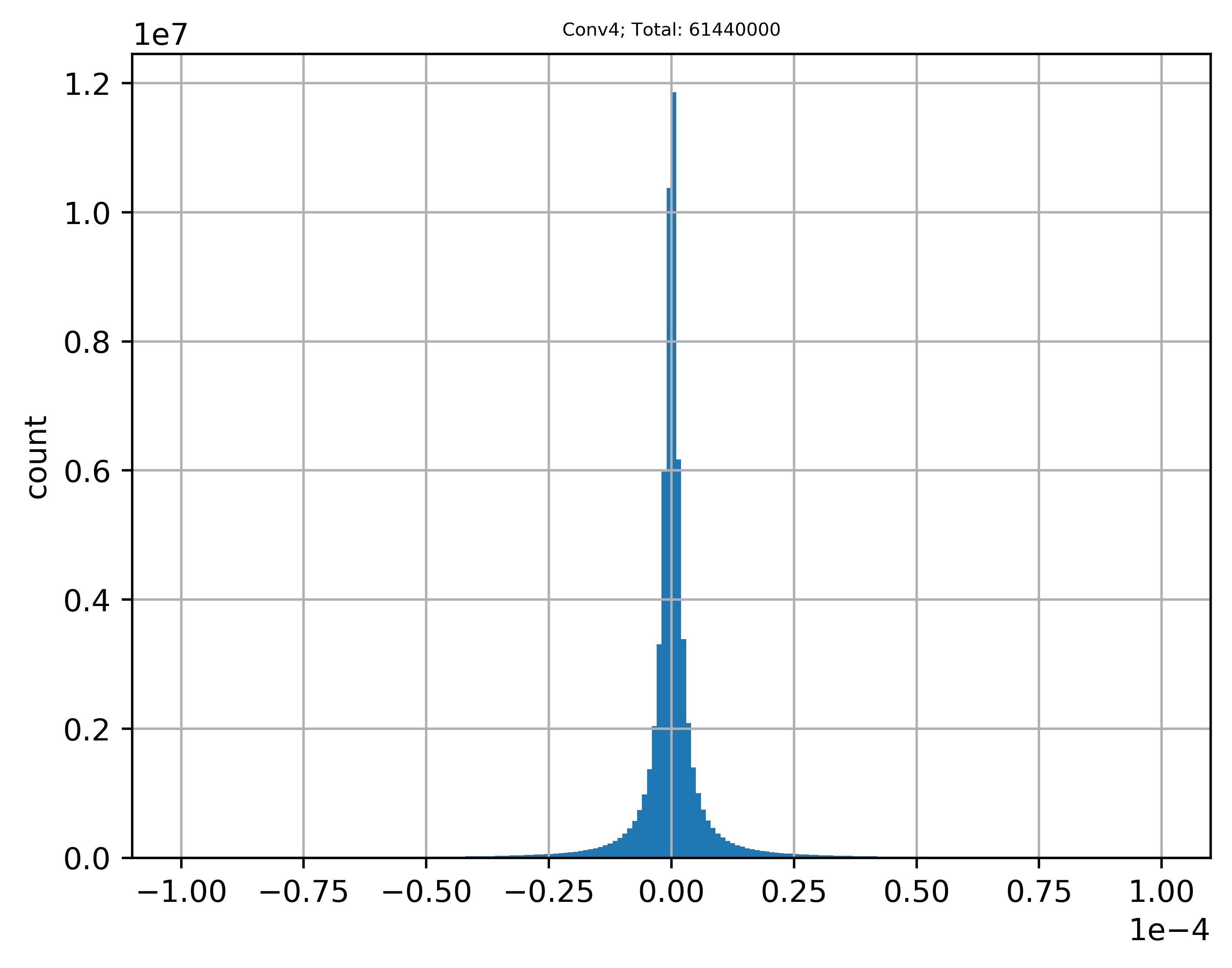} &
			\includegraphics[height=0.12\textwidth]{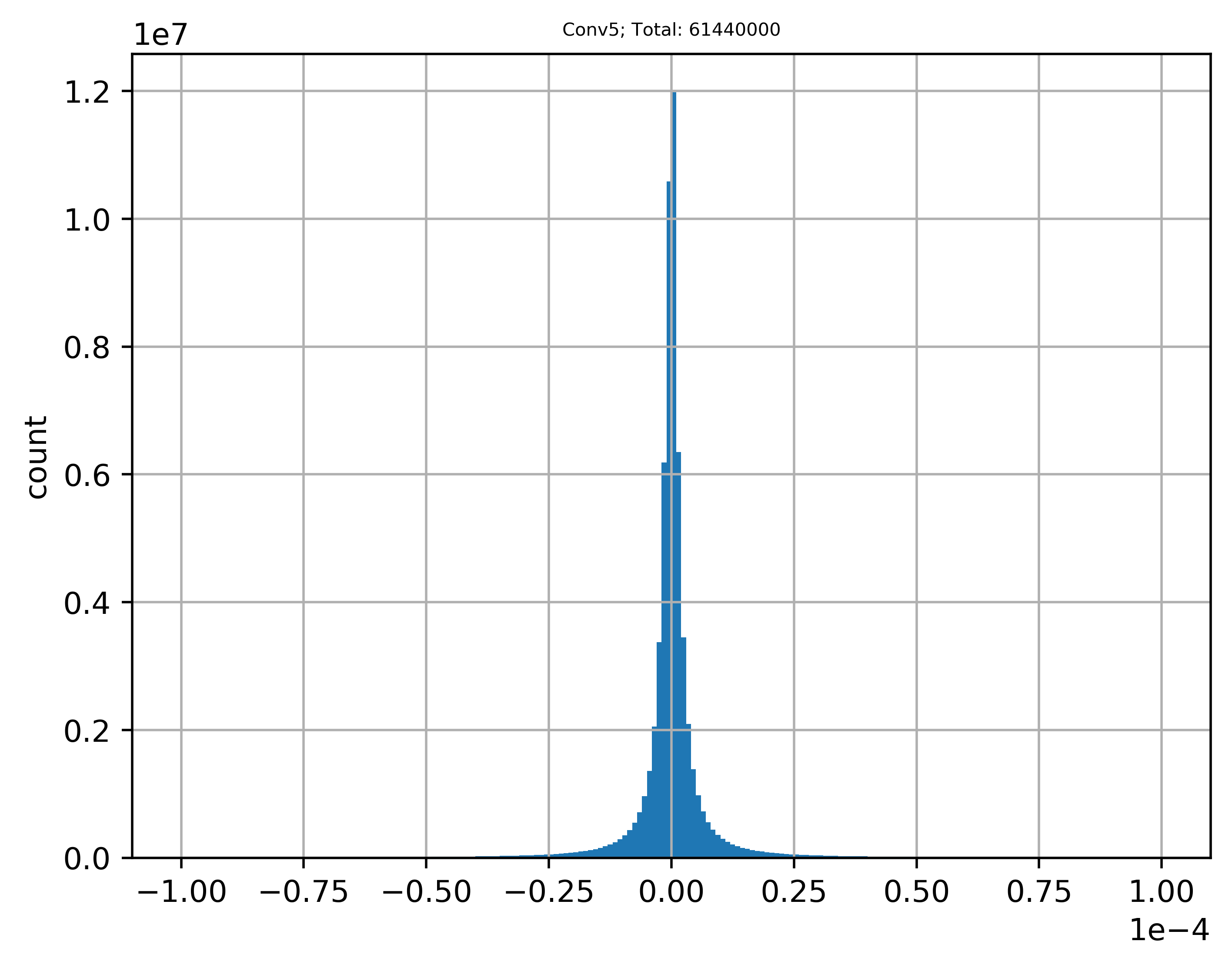} &
			\includegraphics[height=0.12\textwidth]{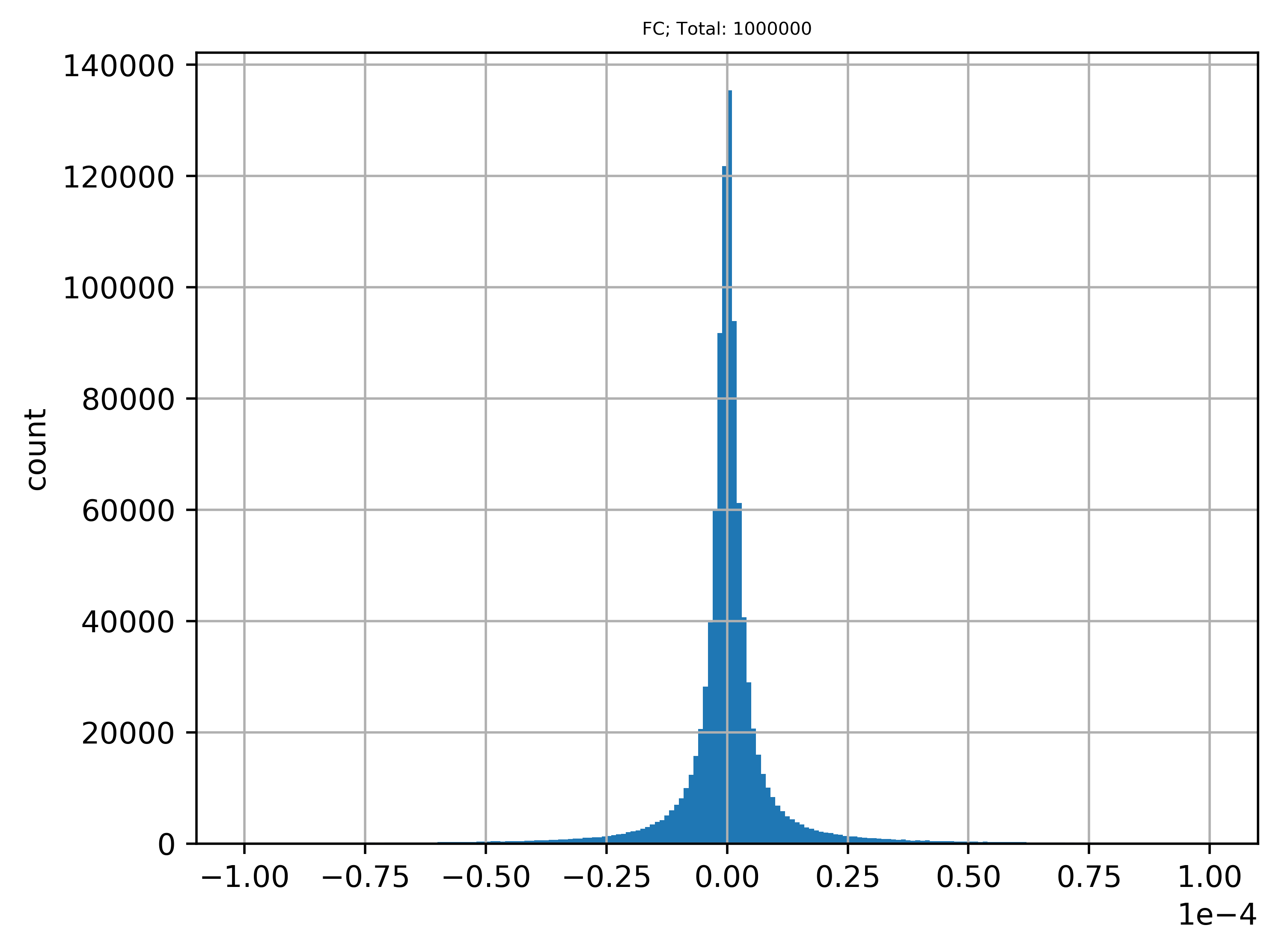} \\
			($a_1$) & ($a_2$) & ($a_3$) & ($a_4$) & ($a_5$) & ($a_6$) \\
		\end{tabular}
		\caption{VGG7 (CIFAR-100). Histograms of relative errors (Eq.\eqref{eq:relative_err}) between units directly computed by CNN ($\mathbf{c}_{l}$) and their values reconstructed by our method ($\hat{\mathbf{c}}_{l}$) on CIFAR-100 using VGG7. Details are similar to Fig.\ref{fig:Statistics:cifar10:vgg7} except that the quantity of relative errors in $a_6$ is $1m$. Percentages of $\mathbf{\epsilon}_{l} \le 1\%$ for all subplots are listed in Table \ref{table:percentage_of_errs:cifar100}.}
		\label{fig:Statistics:cifar100:vgg7}
	\end{figure}

\begin{figure}[!h]
		\centering
		\begin{tabular}{@{}c@{}c@{}c@{}c@{}c@{}c@{}c@{}}
			\includegraphics[height=0.1\textwidth]{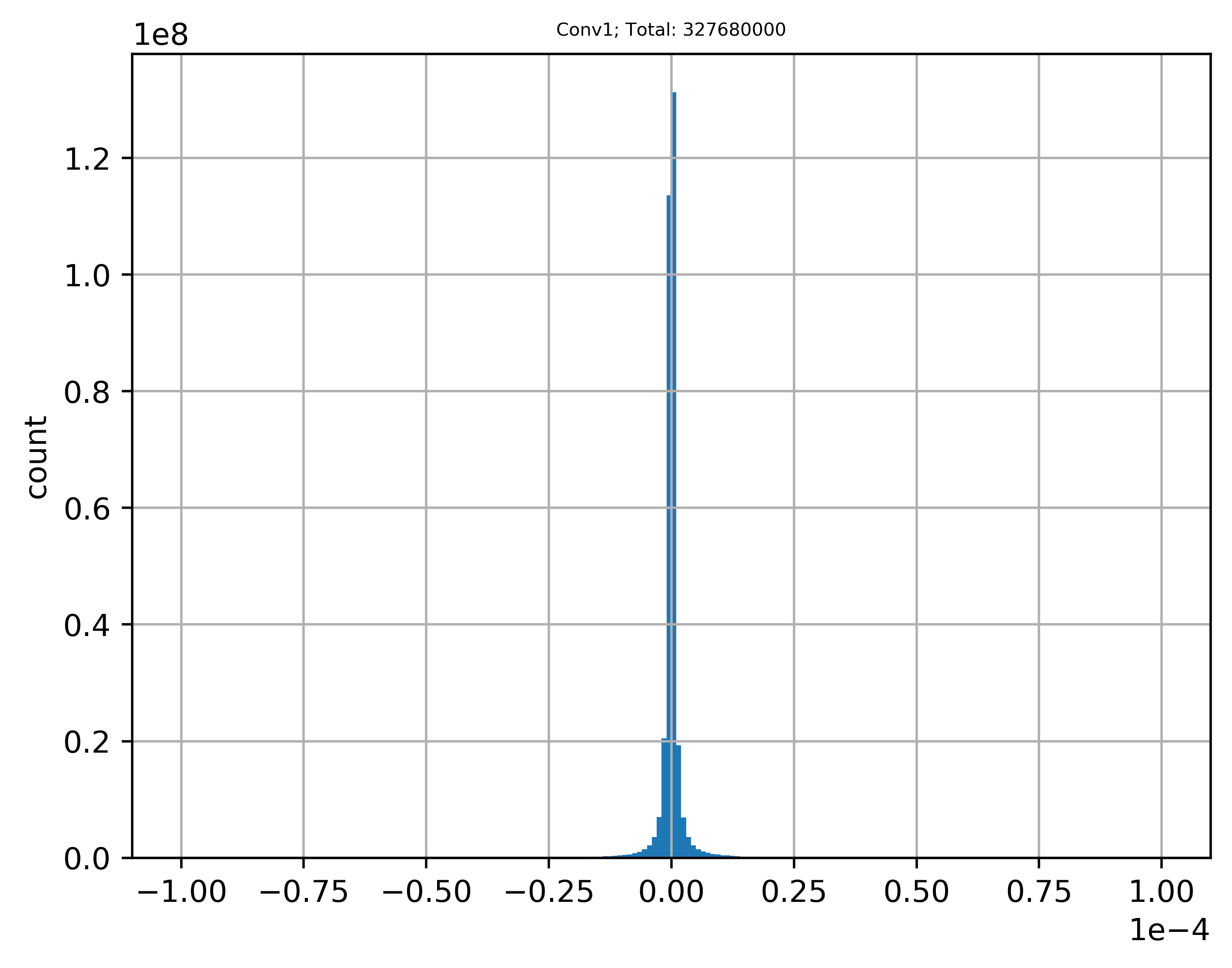} & 
			\includegraphics[height=0.1\textwidth]{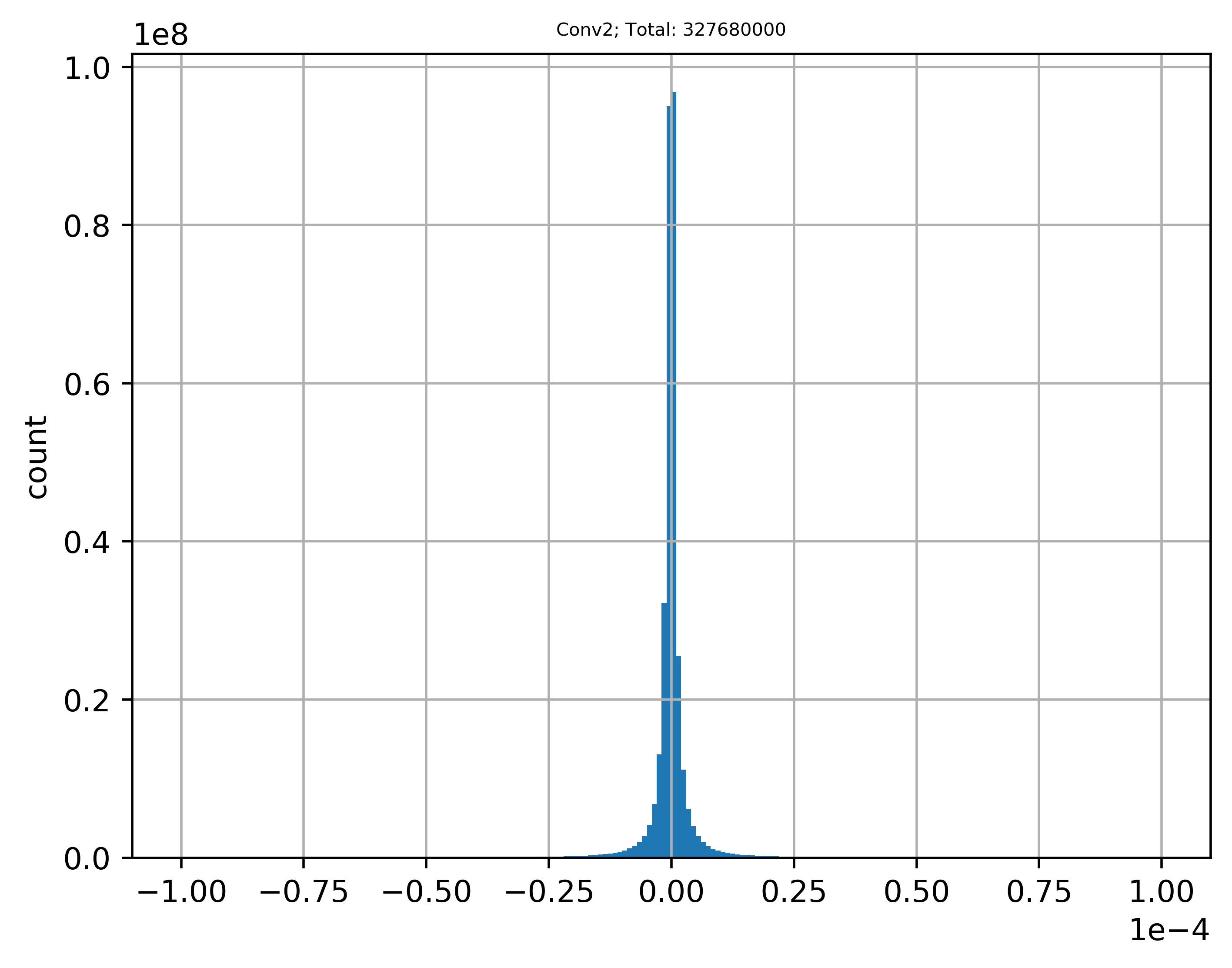} &
			\includegraphics[height=0.1\textwidth]{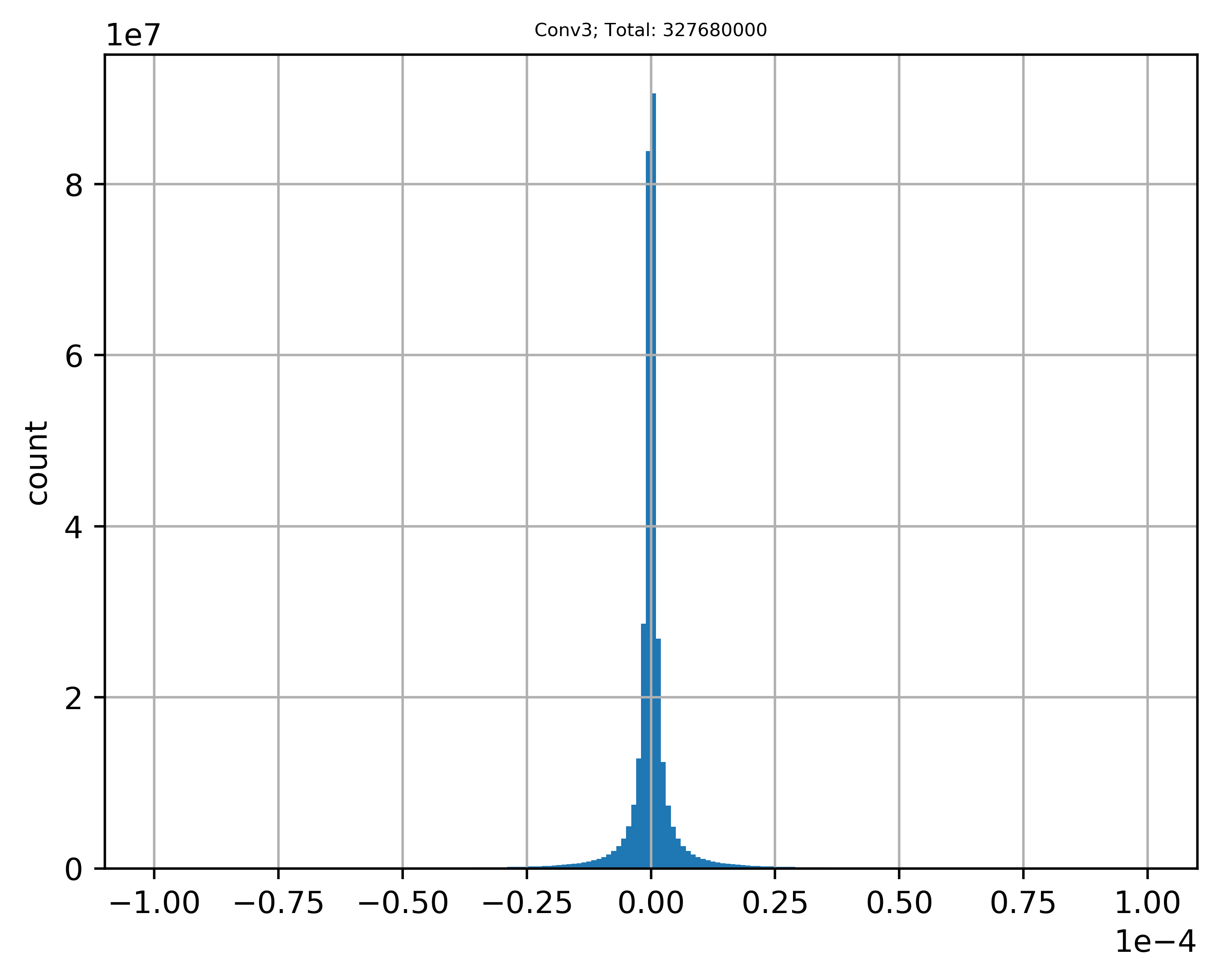} &
			\includegraphics[height=0.1\textwidth]{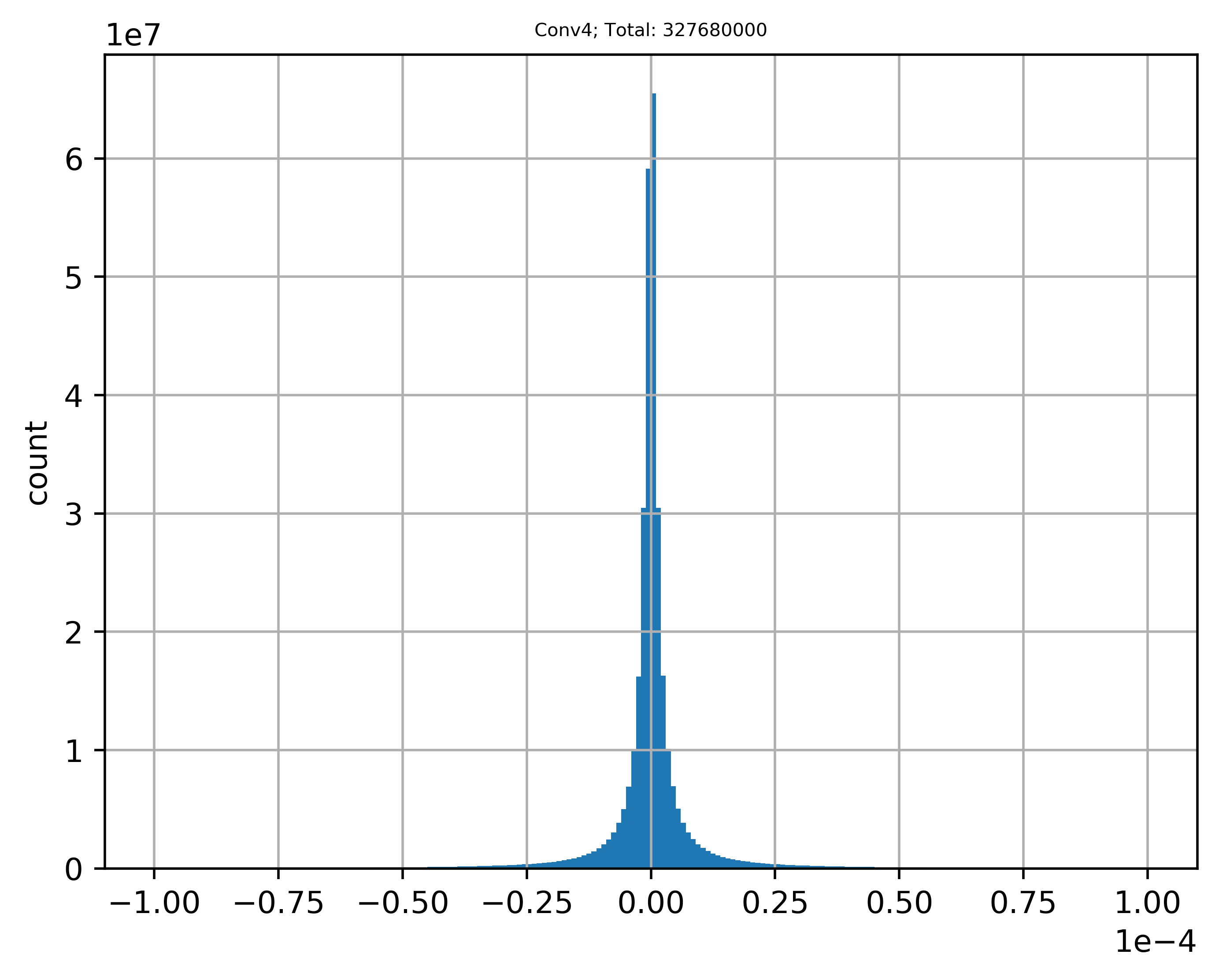} &
			\includegraphics[height=0.1\textwidth]{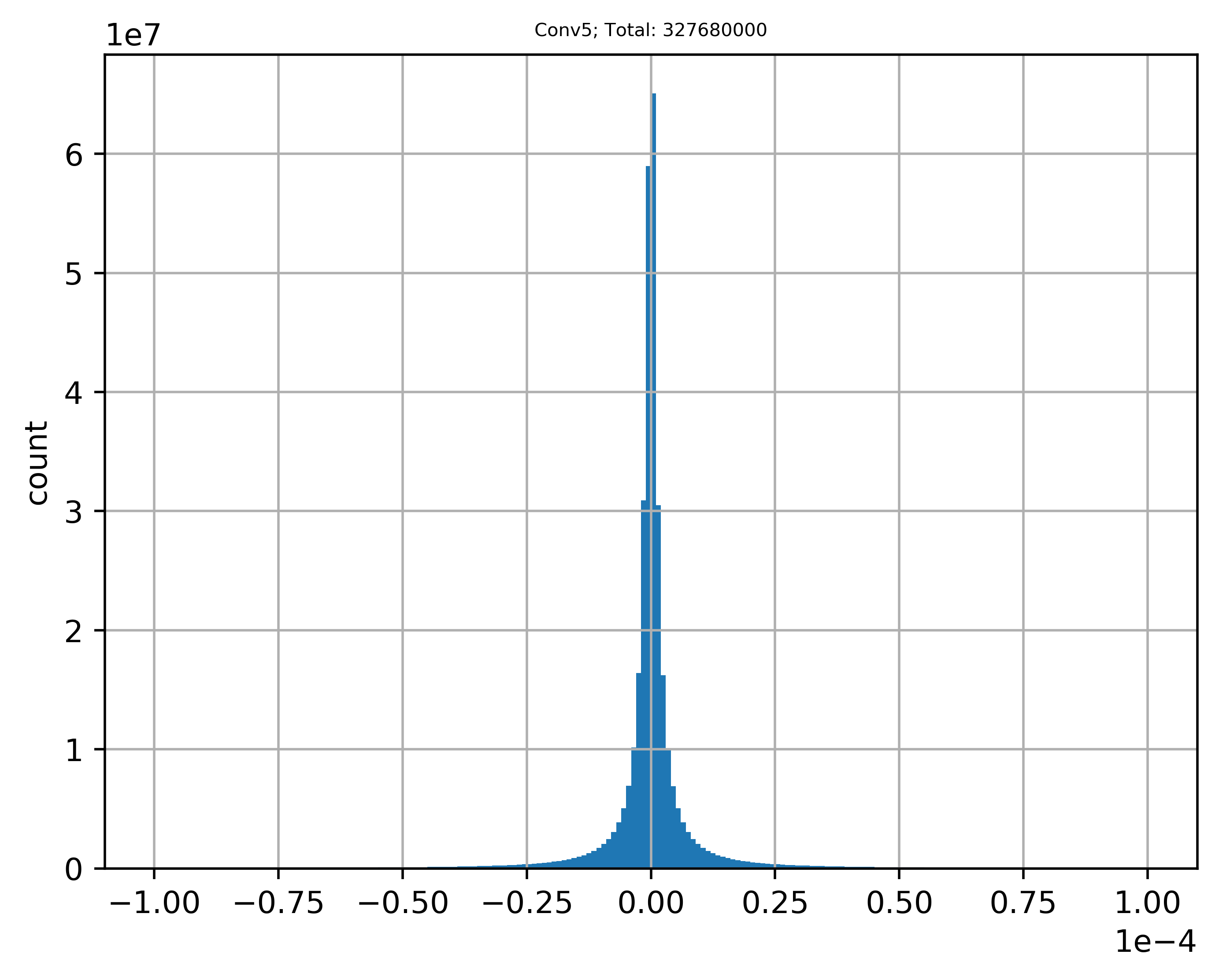} &
			\includegraphics[height=0.1\textwidth]{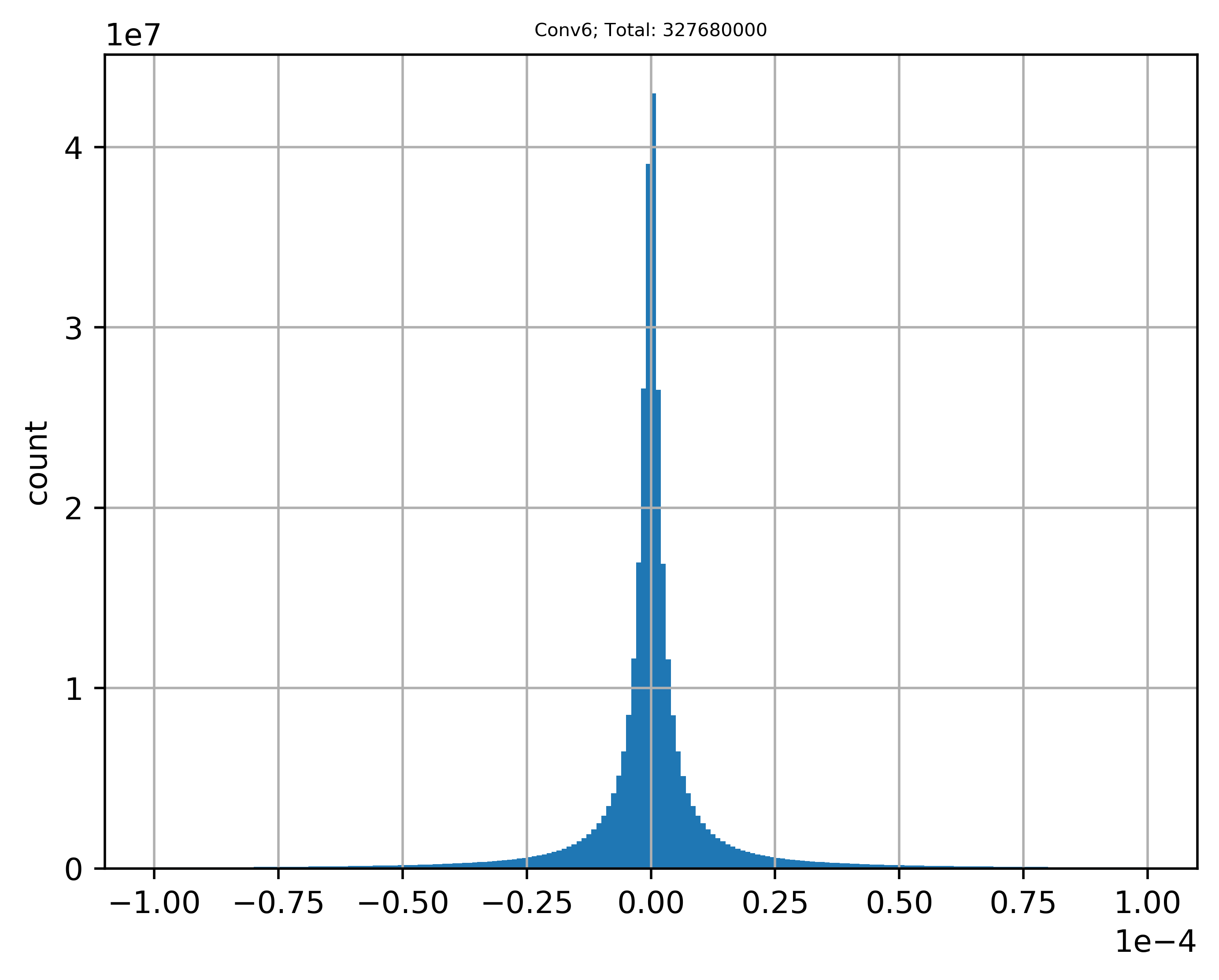} &
			\includegraphics[height=0.1\textwidth]{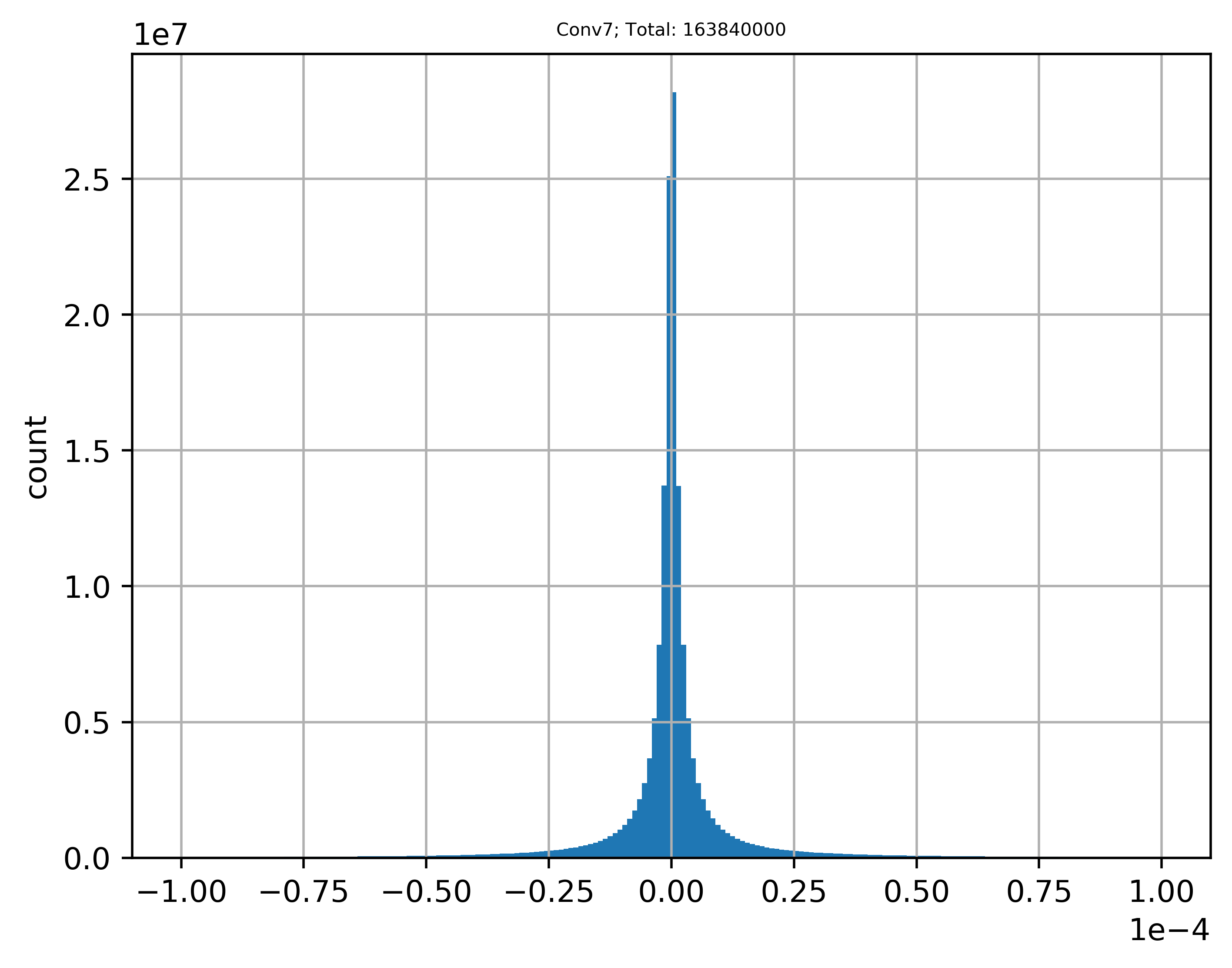} \\
			($a_1$) & ($a_2$) & ($a_3$) & ($a_4$) & ($a_5$) & ($a_6$) & ($a_7$) \\
			\includegraphics[height=0.1\textwidth]{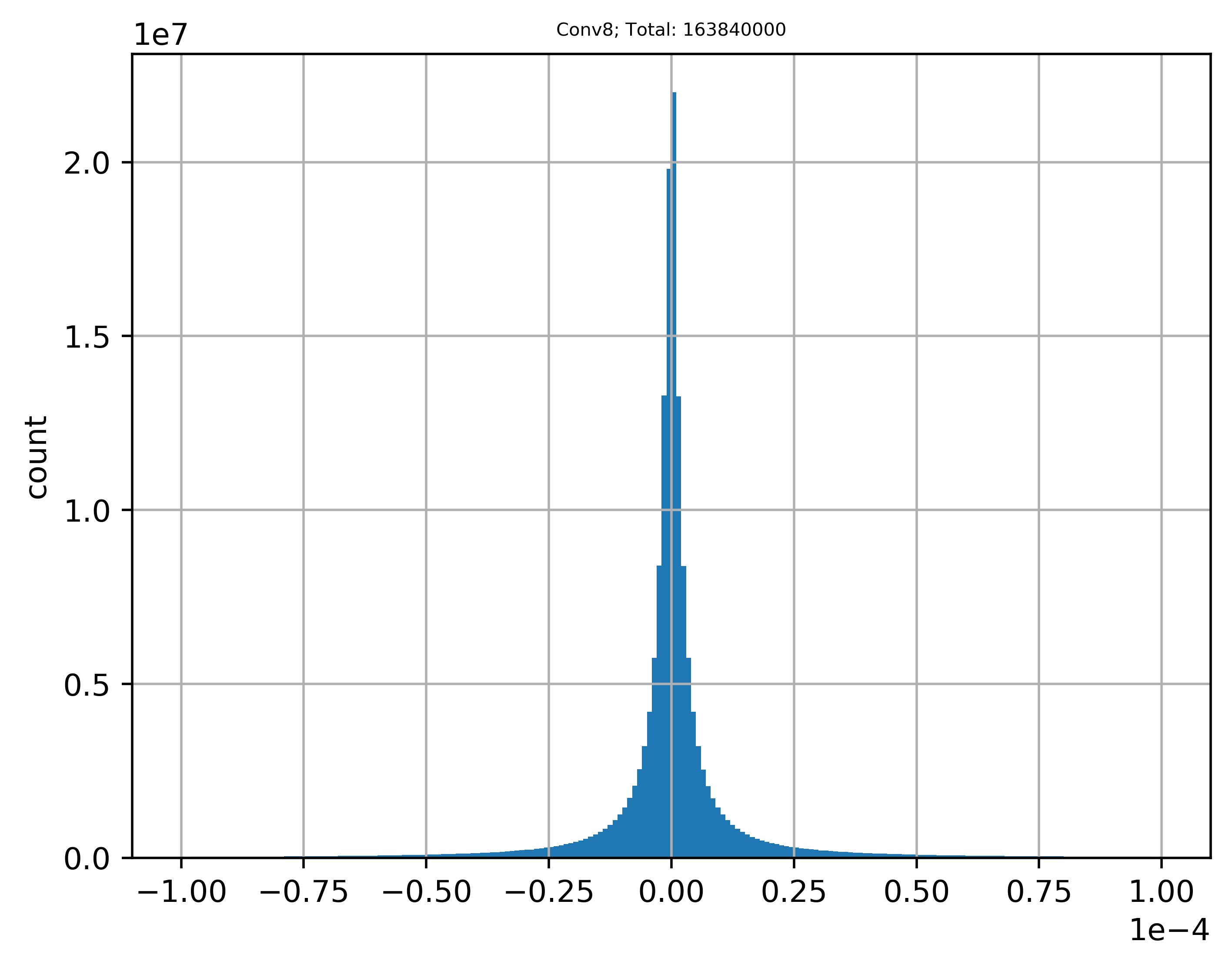} &
			\includegraphics[height=0.1\textwidth]{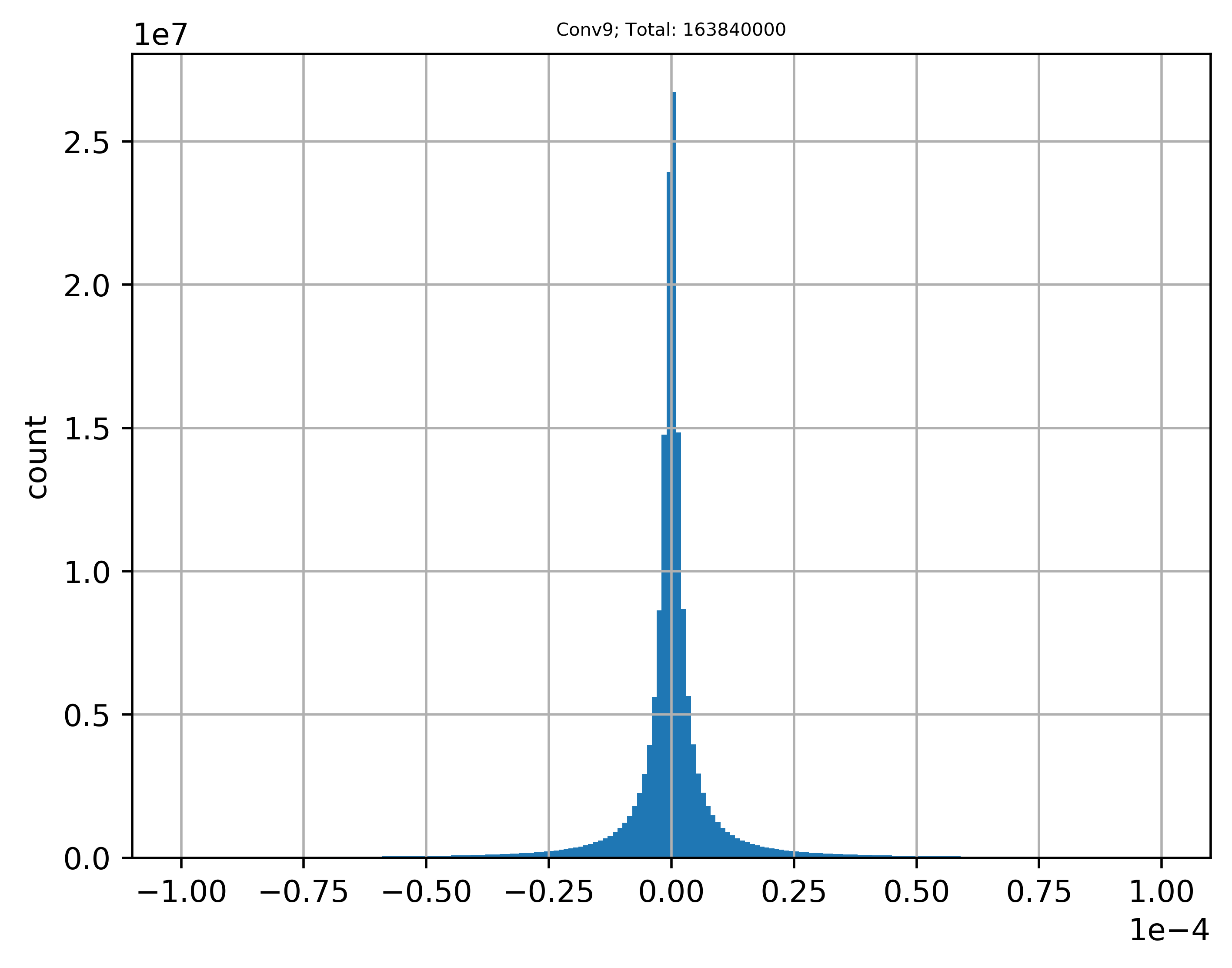} &
			\includegraphics[height=0.1\textwidth]{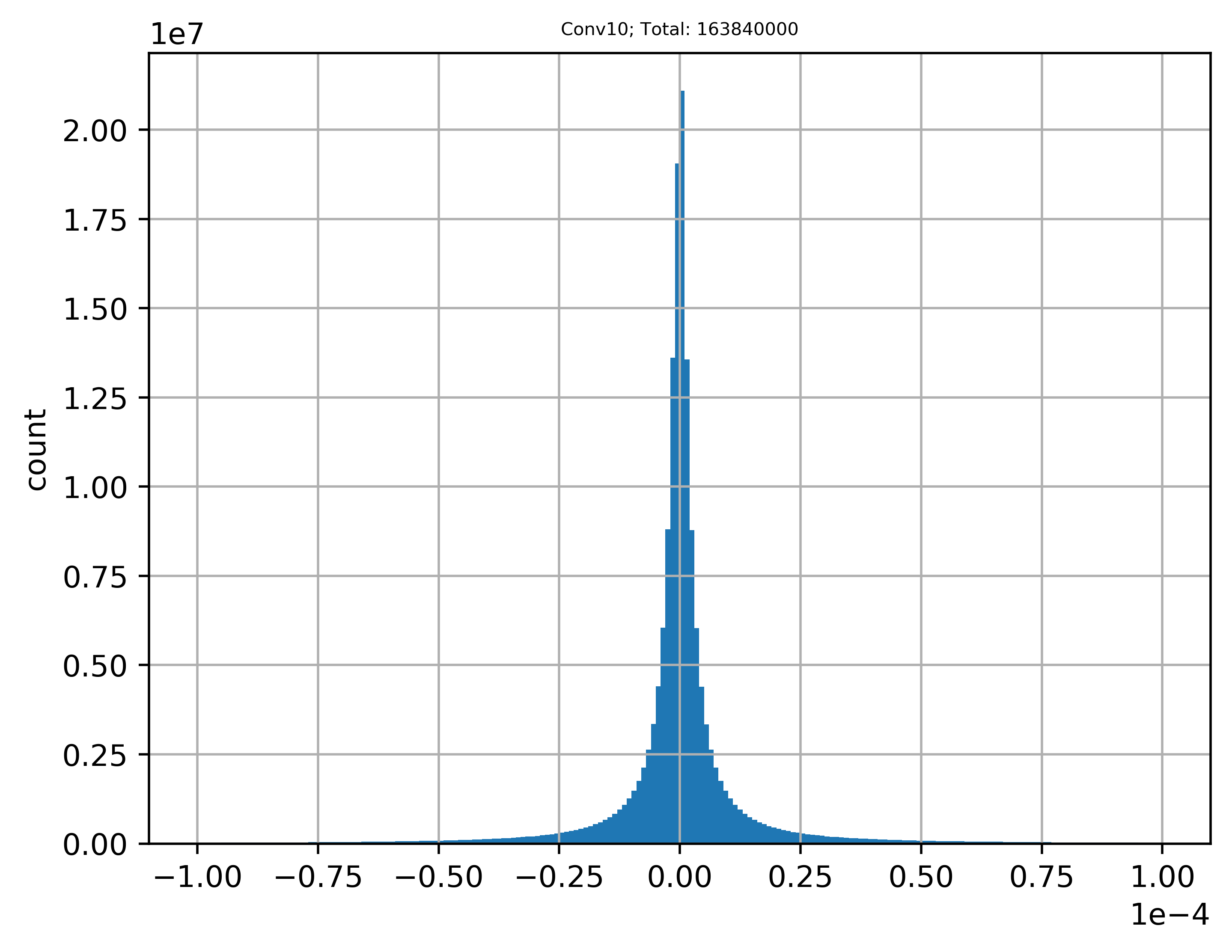} &
			\includegraphics[height=0.1\textwidth]{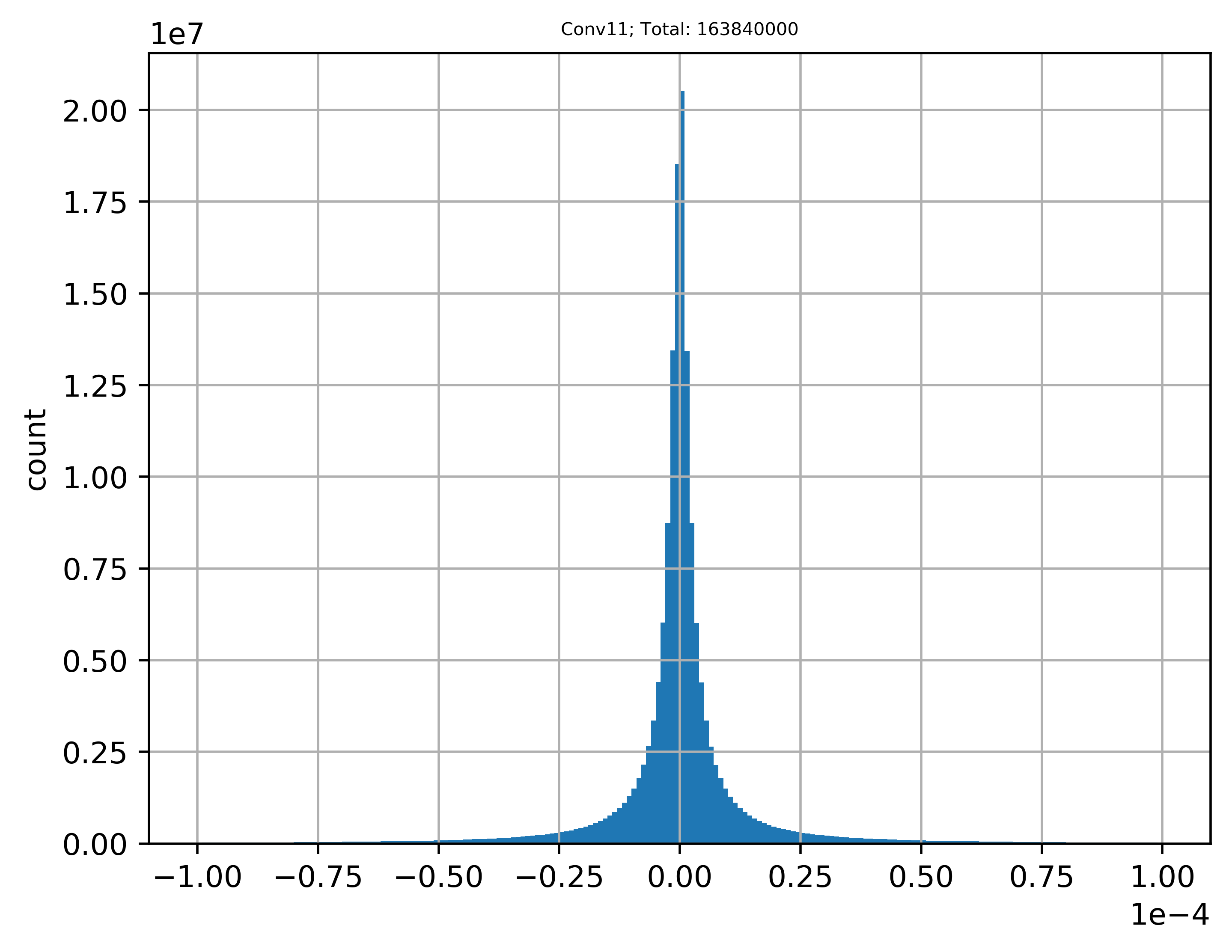} &
			\includegraphics[height=0.1\textwidth]{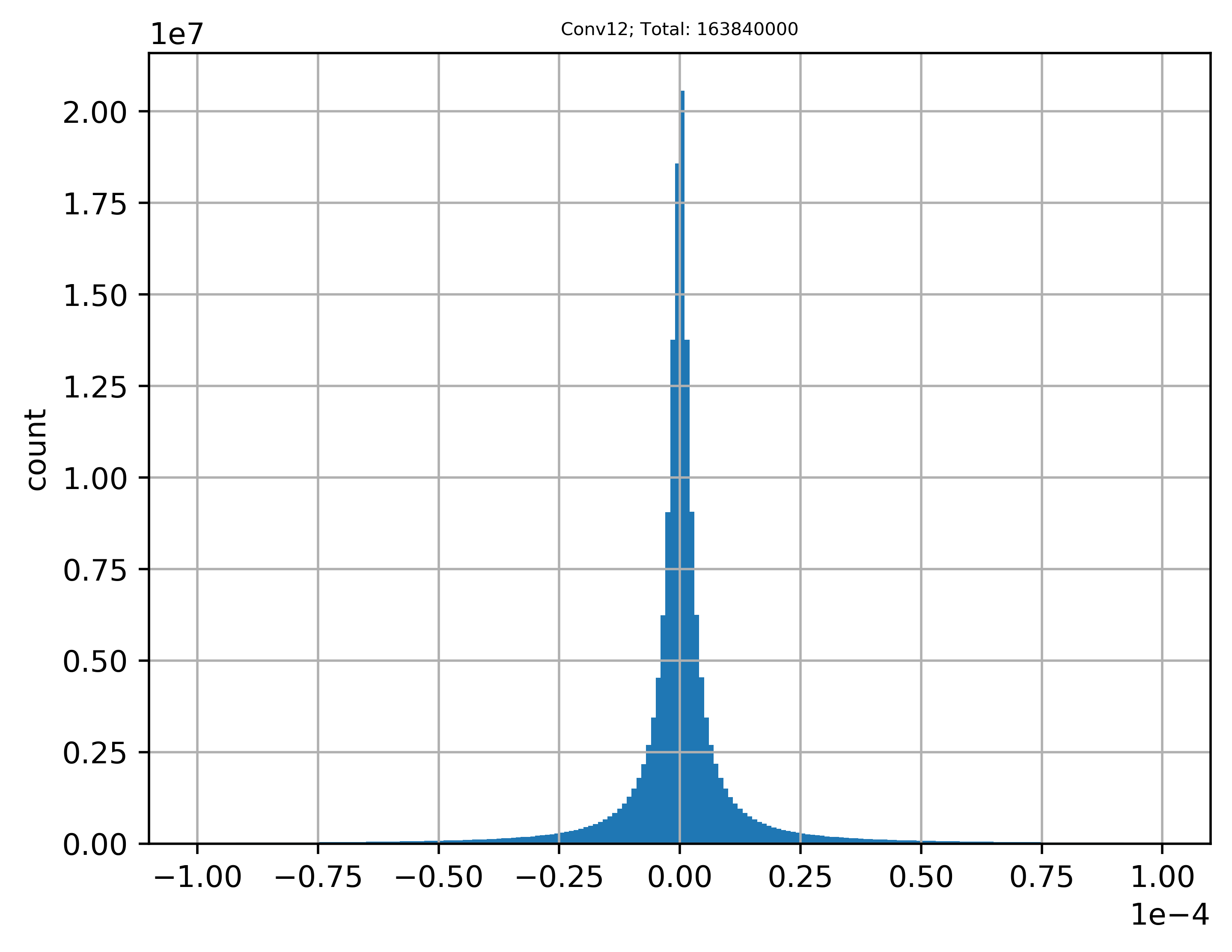} &
			\includegraphics[height=0.1\textwidth]{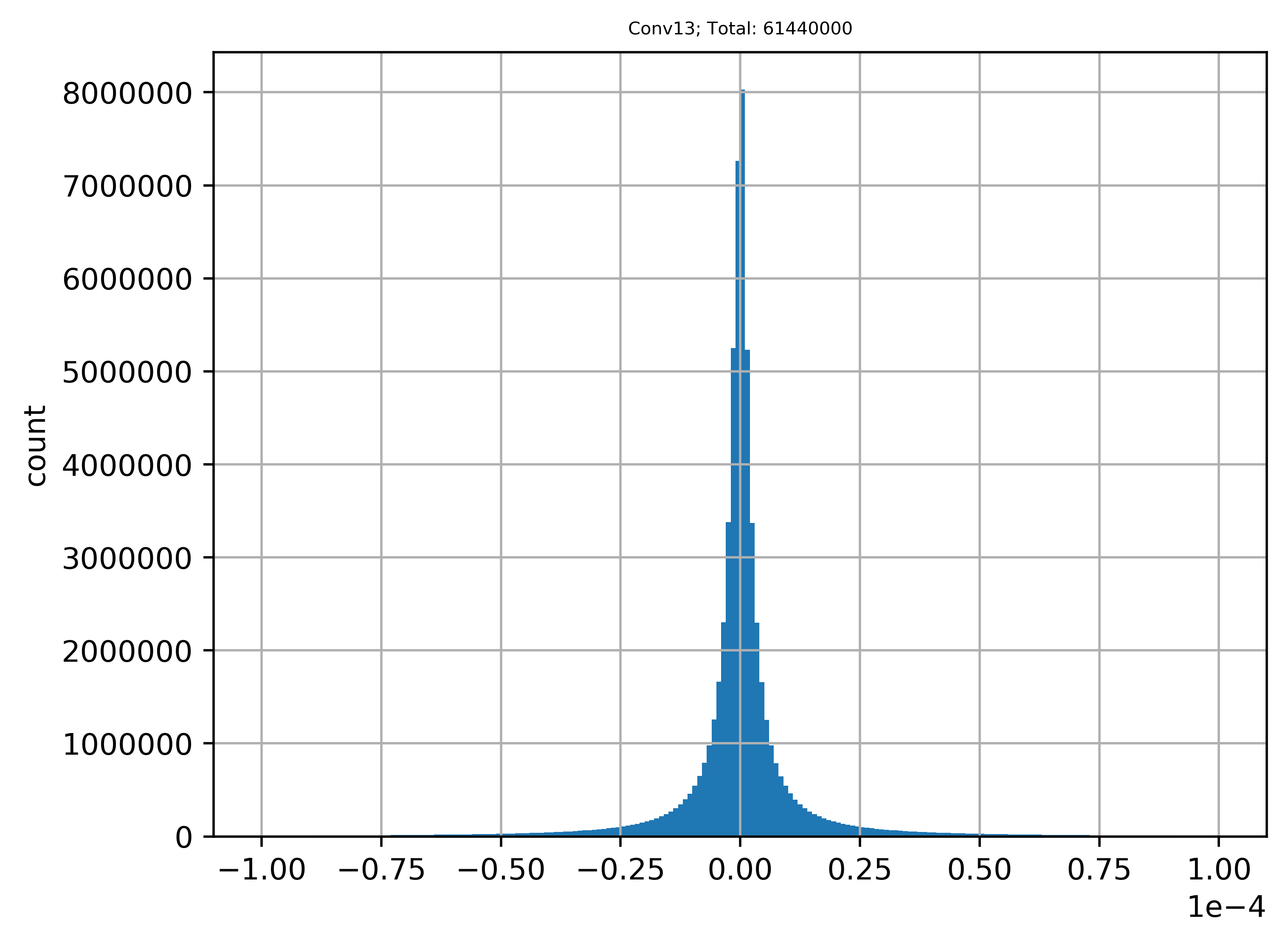} &
			\includegraphics[height=0.1\textwidth]{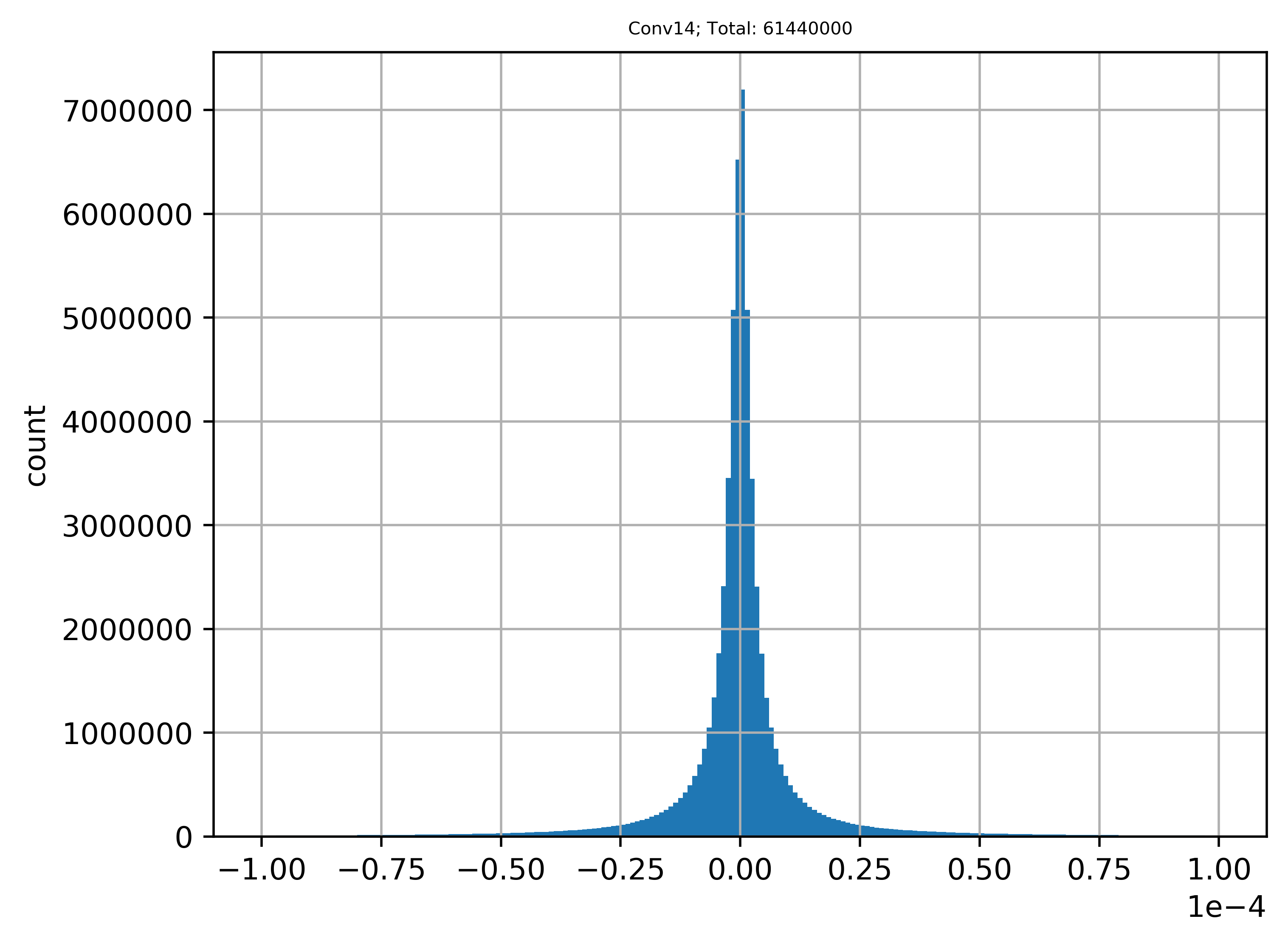} \\
			($a_8$) & ($a_9$) & ($a_{10}$) & ($a_{11}$) & ($a_{12}$) & ($a_{13}$) & ($a_{14}$) \\
			\includegraphics[height=0.1\textwidth]{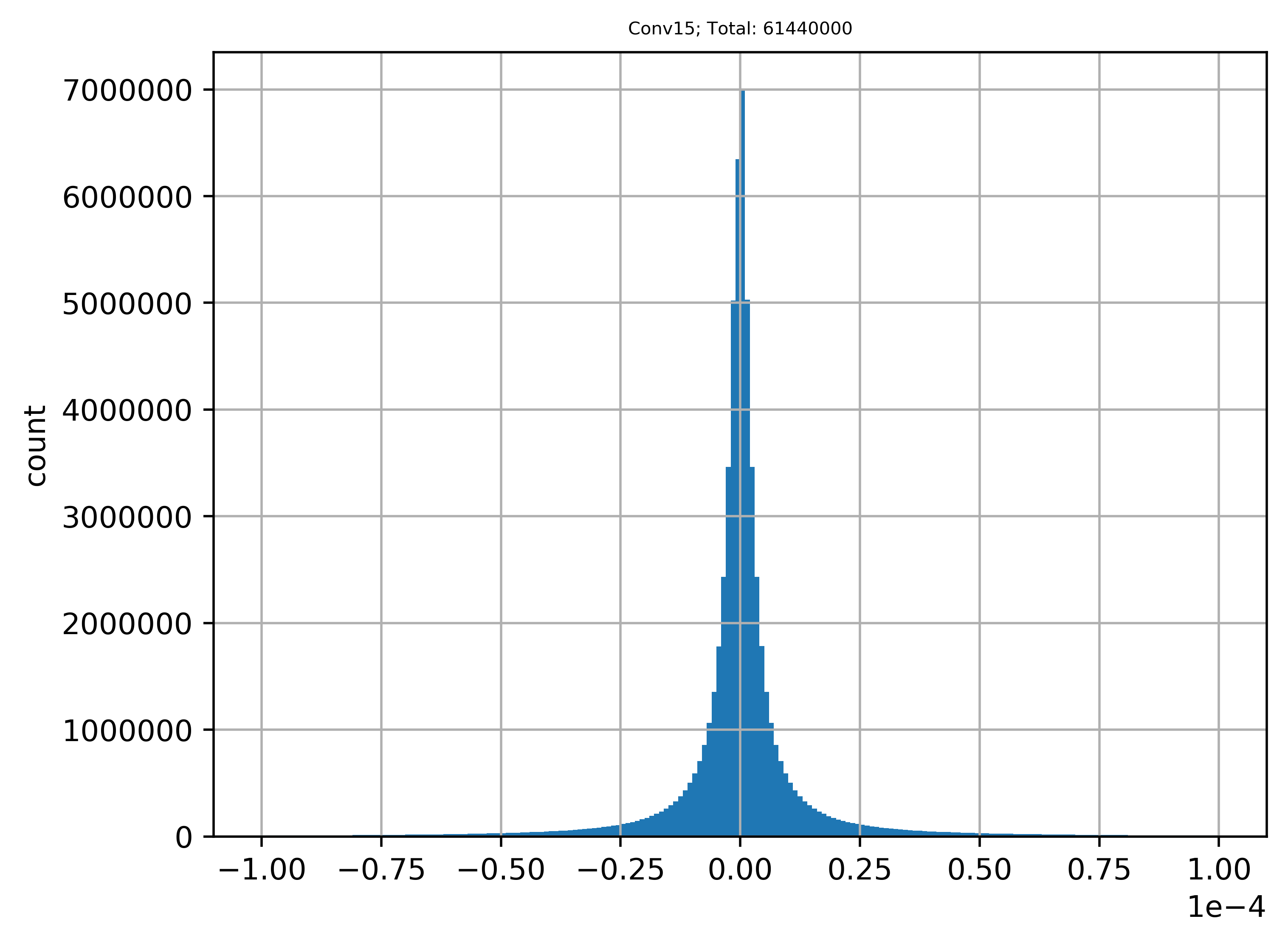} &
			\includegraphics[height=0.1\textwidth]{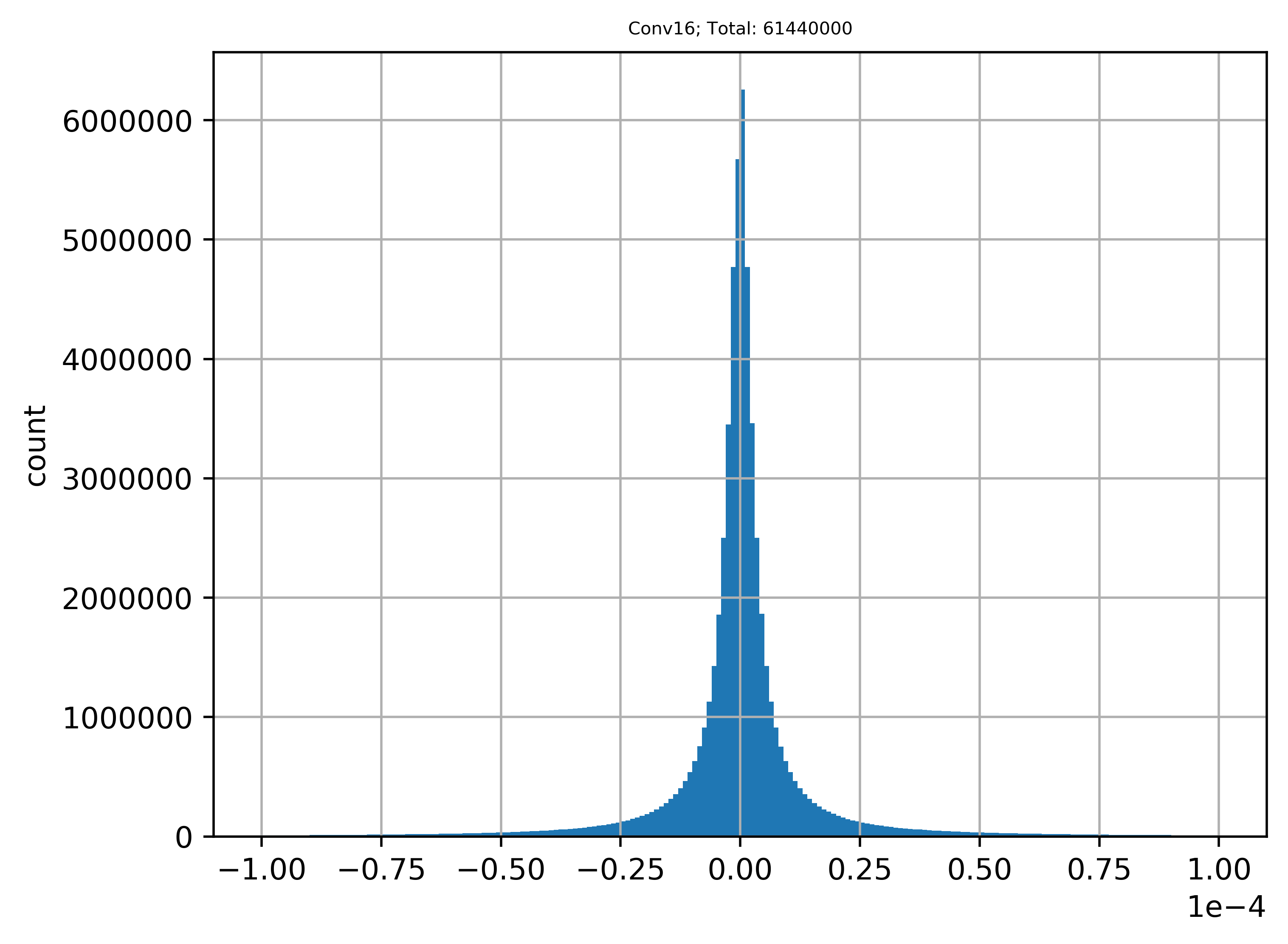} &
			\includegraphics[height=0.1\textwidth]{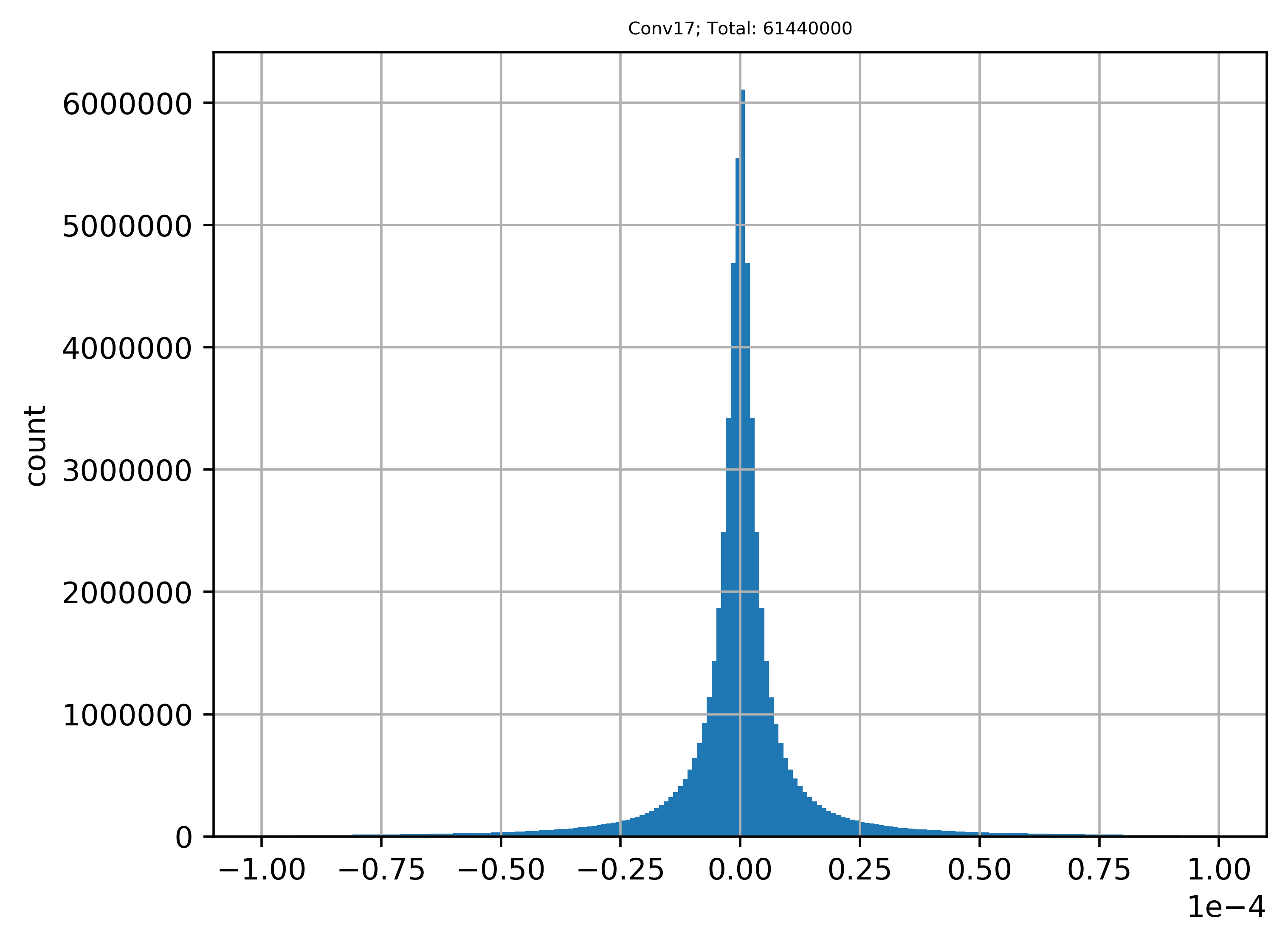} &
			\includegraphics[height=0.1\textwidth]{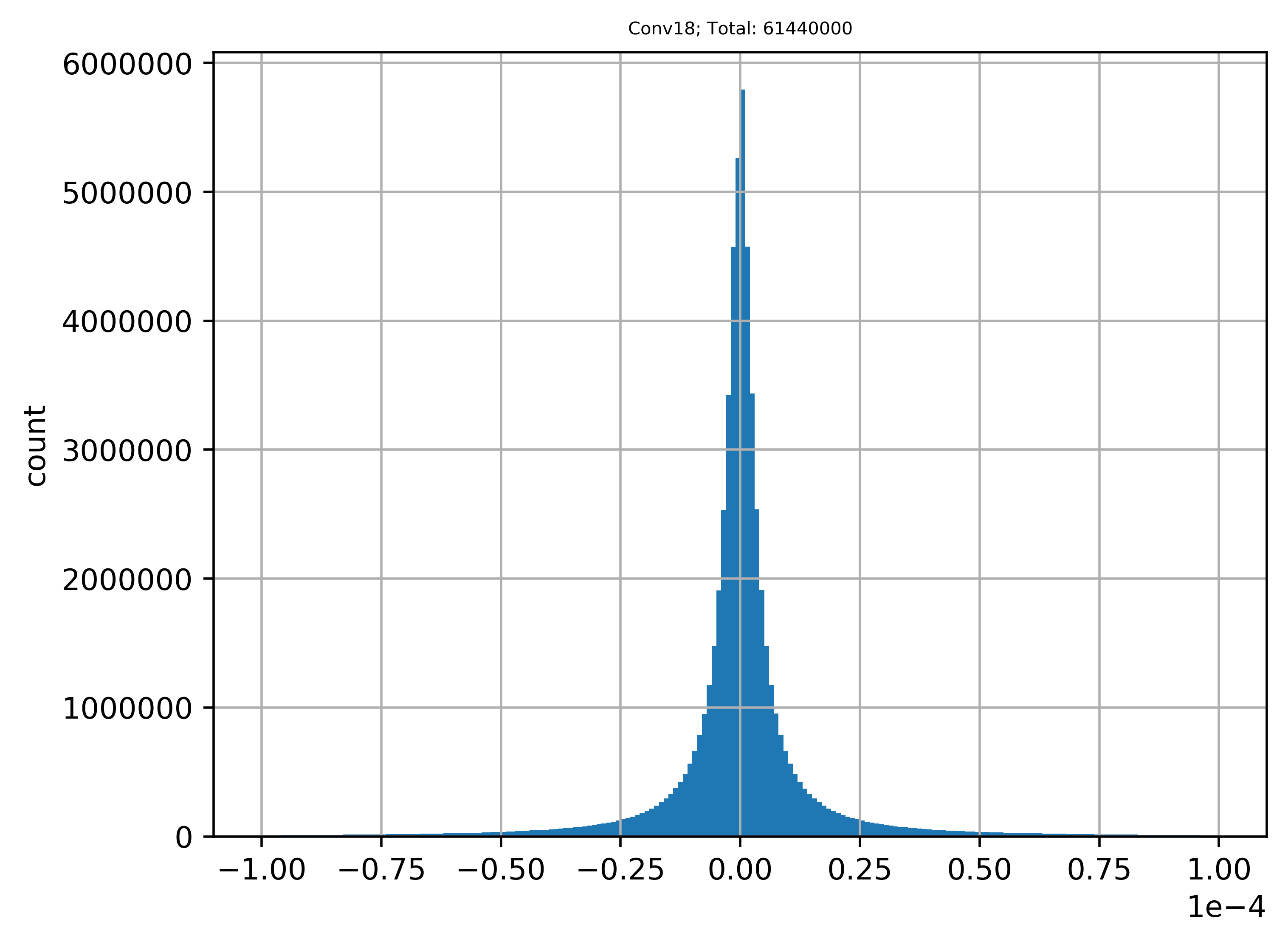} &
			\includegraphics[height=0.1\textwidth]{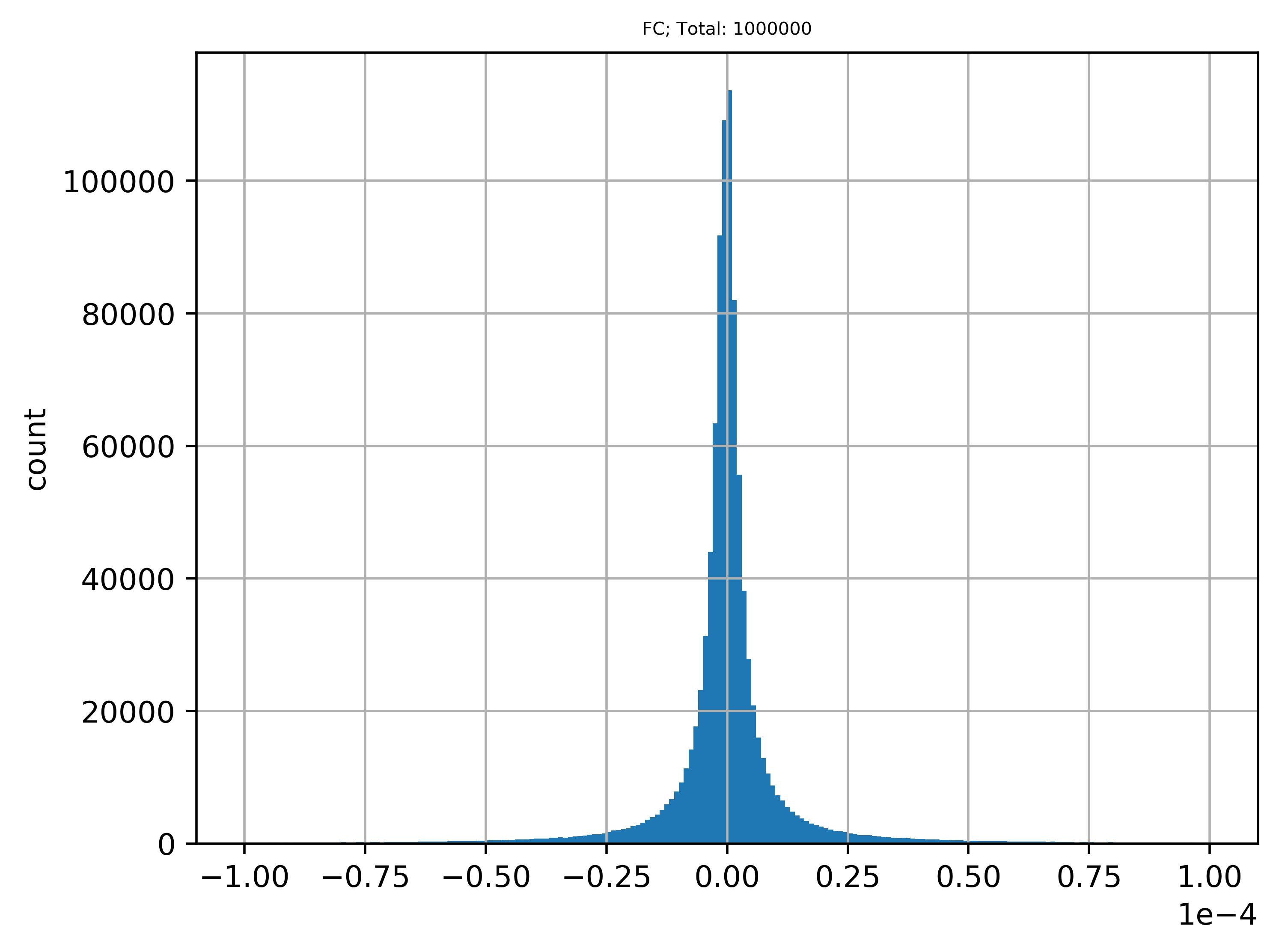} & & \\
			($a_{15}$) & ($a_{16}$) & ($a_{17}$) & ($a_{18}$) & ($a_{19}$) & & \\
		\end{tabular}
		\caption{ResNet20 (CIFAR-100). Histograms of relative errors (Eq.\eqref{eq:relative_err}) between units directly computed by CNN ($\mathbf{c}_{l}$) and their values reconstructed by our method ($\hat{\mathbf{c}}_{l}$) on CIFAR-100 using ResNet20. Details are similar to Fig.\ref{fig:Statistics:cifar10:resnet20} except that the quantity of relative errors in $a_{19}$ is $1m$. Percentages of $\mathbf{\epsilon}_{l} \le 1\%$ for all subplots are listed in Table \ref{table:percentage_of_errs:cifar100}.}
		\label{fig:Statistics:cifar100:resnet20}
	\end{figure}

\begin{figure}[!h]
		\centering
		\begin{tabular}{@{}c@{}c@{}c@{}c@{}c@{}c@{}c@{}}
			\includegraphics[height=0.1\textwidth]{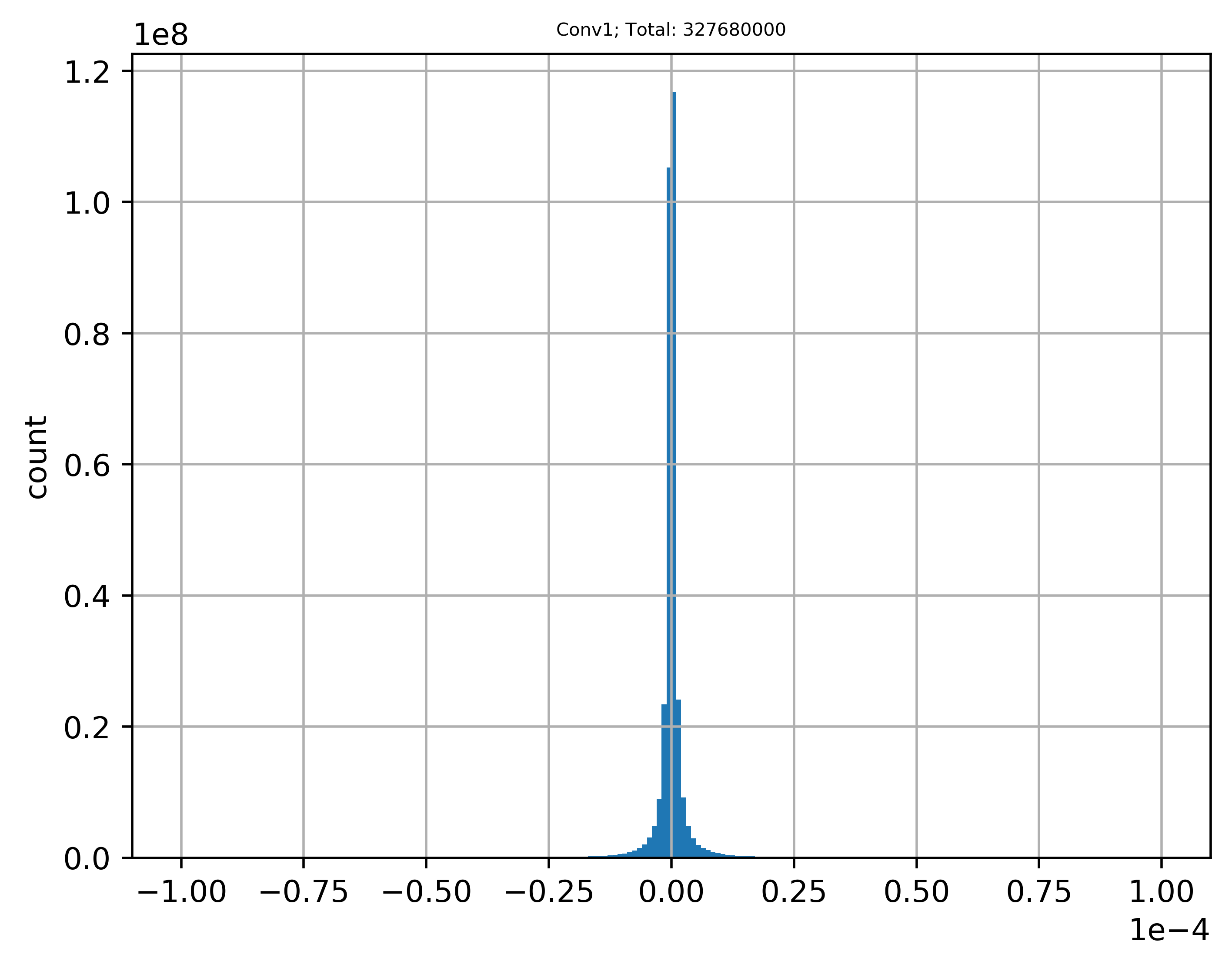} & 
			\includegraphics[height=0.1\textwidth]{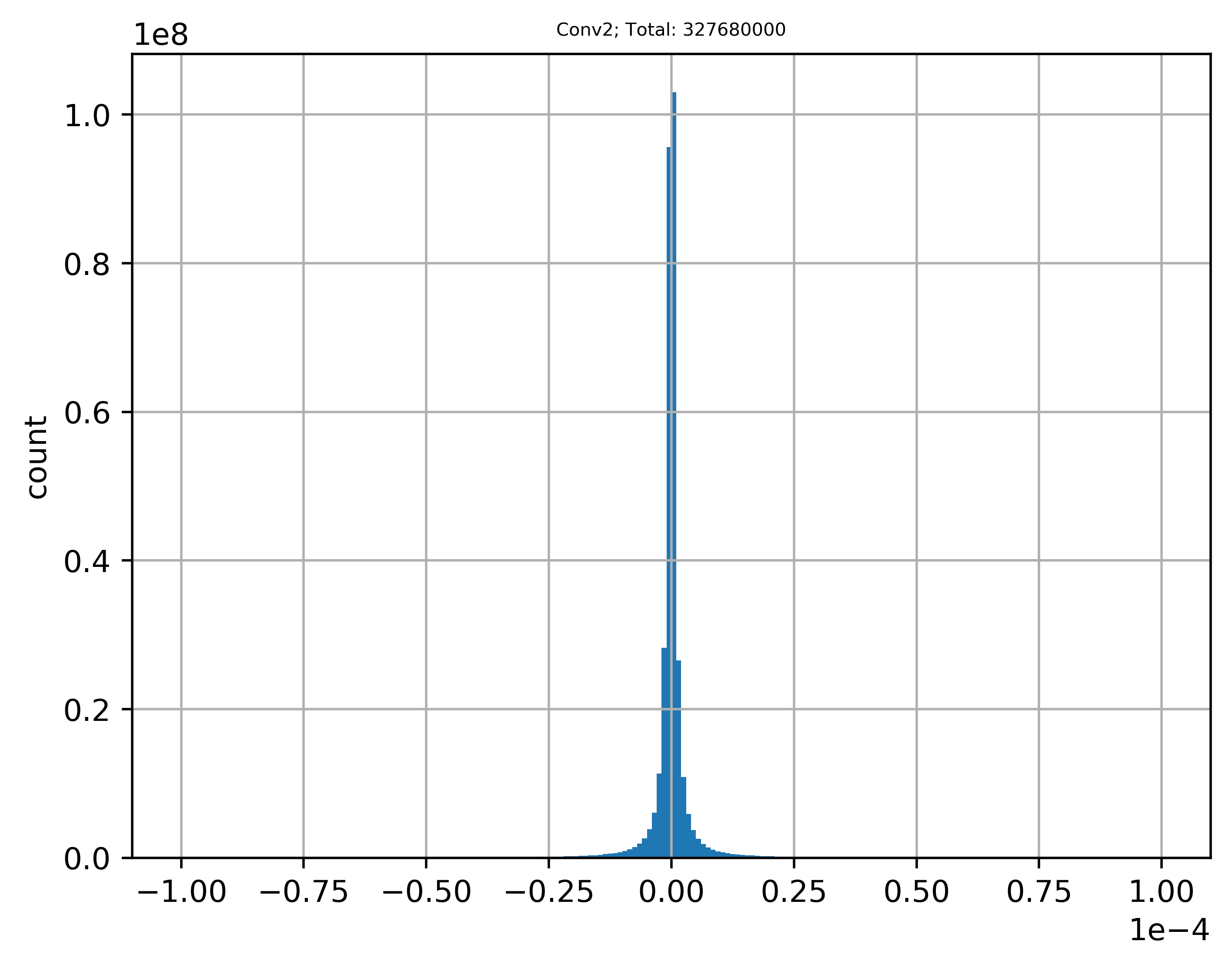} &
			\includegraphics[height=0.1\textwidth]{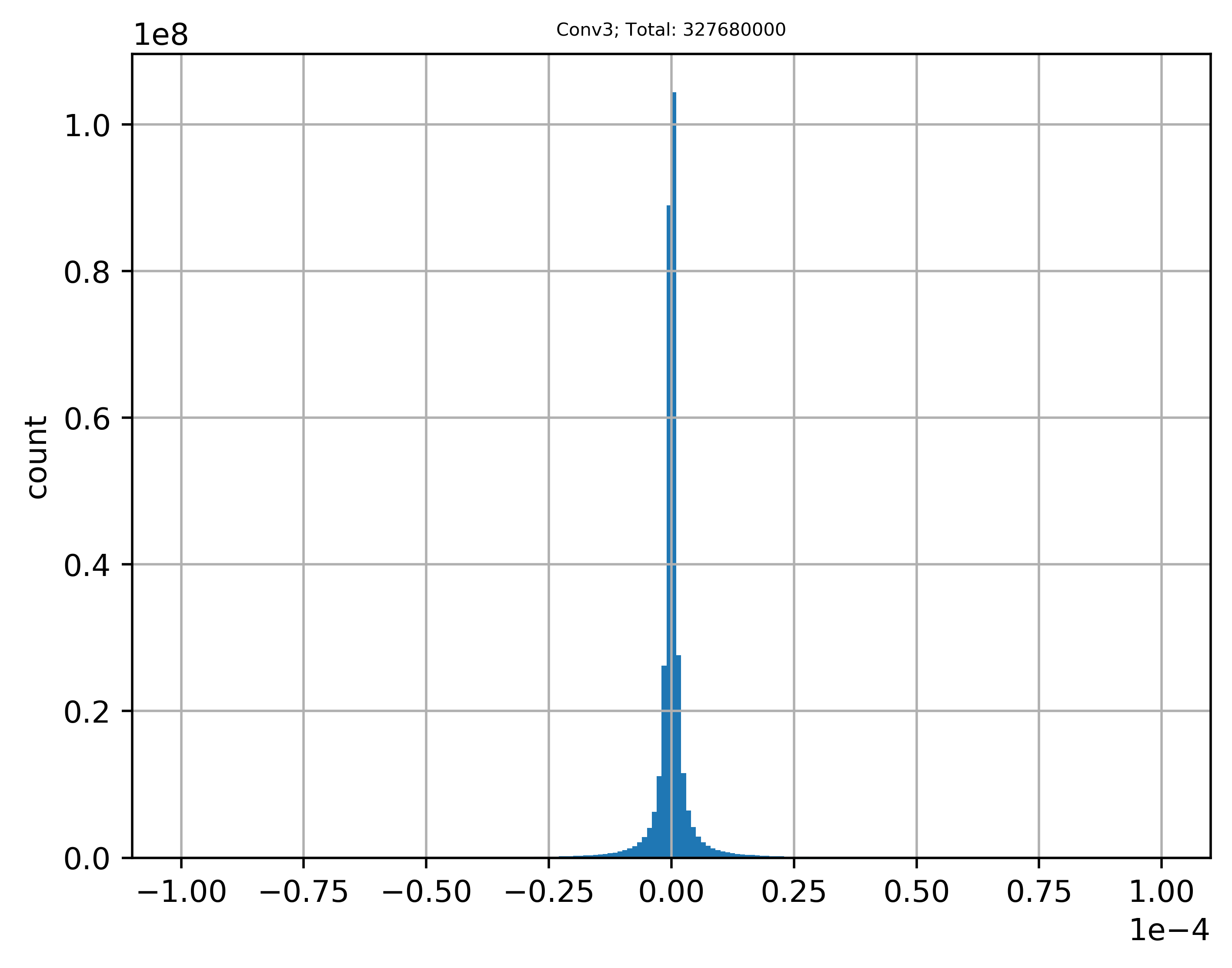} &
			\includegraphics[height=0.1\textwidth]{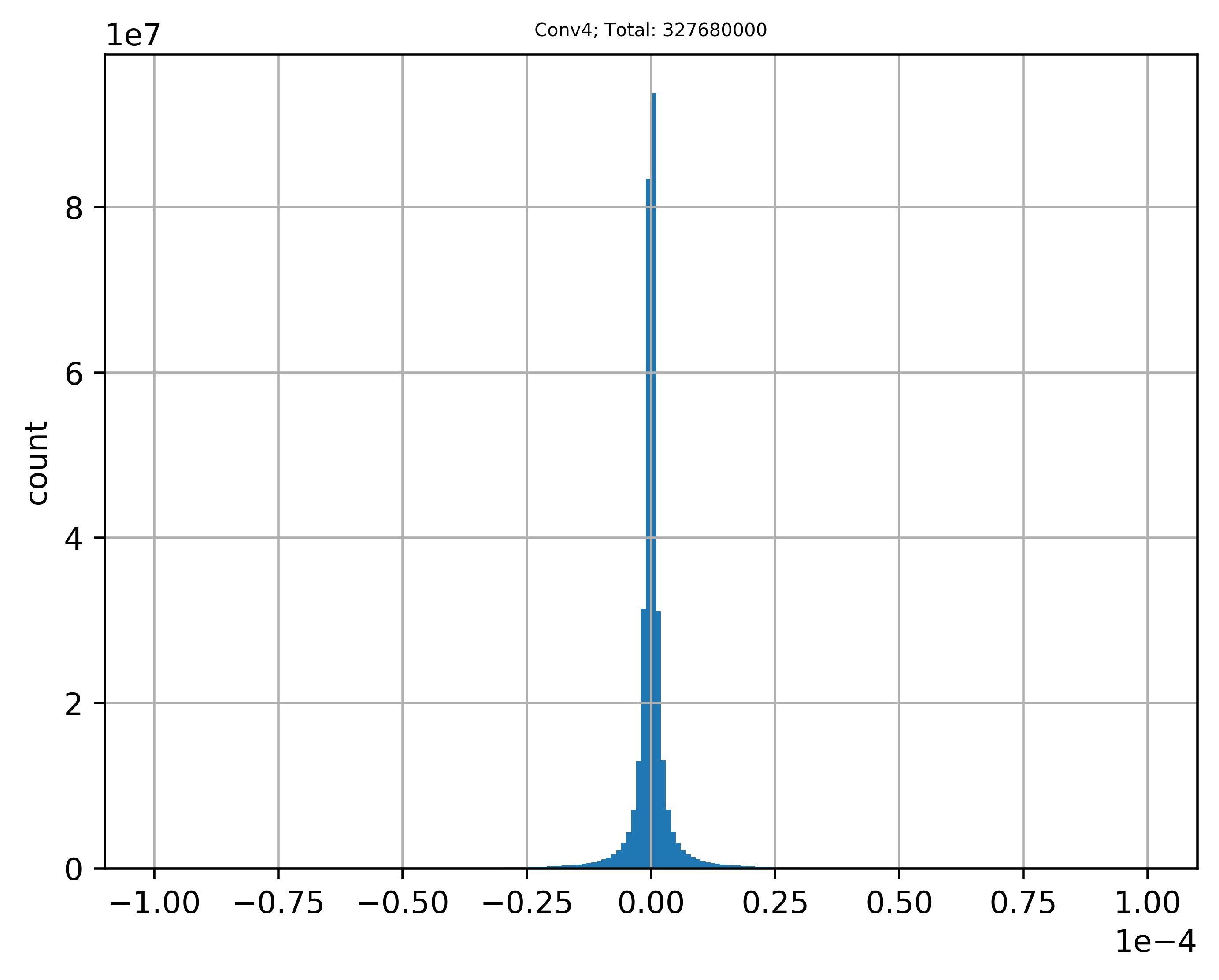} &
			\includegraphics[height=0.1\textwidth]{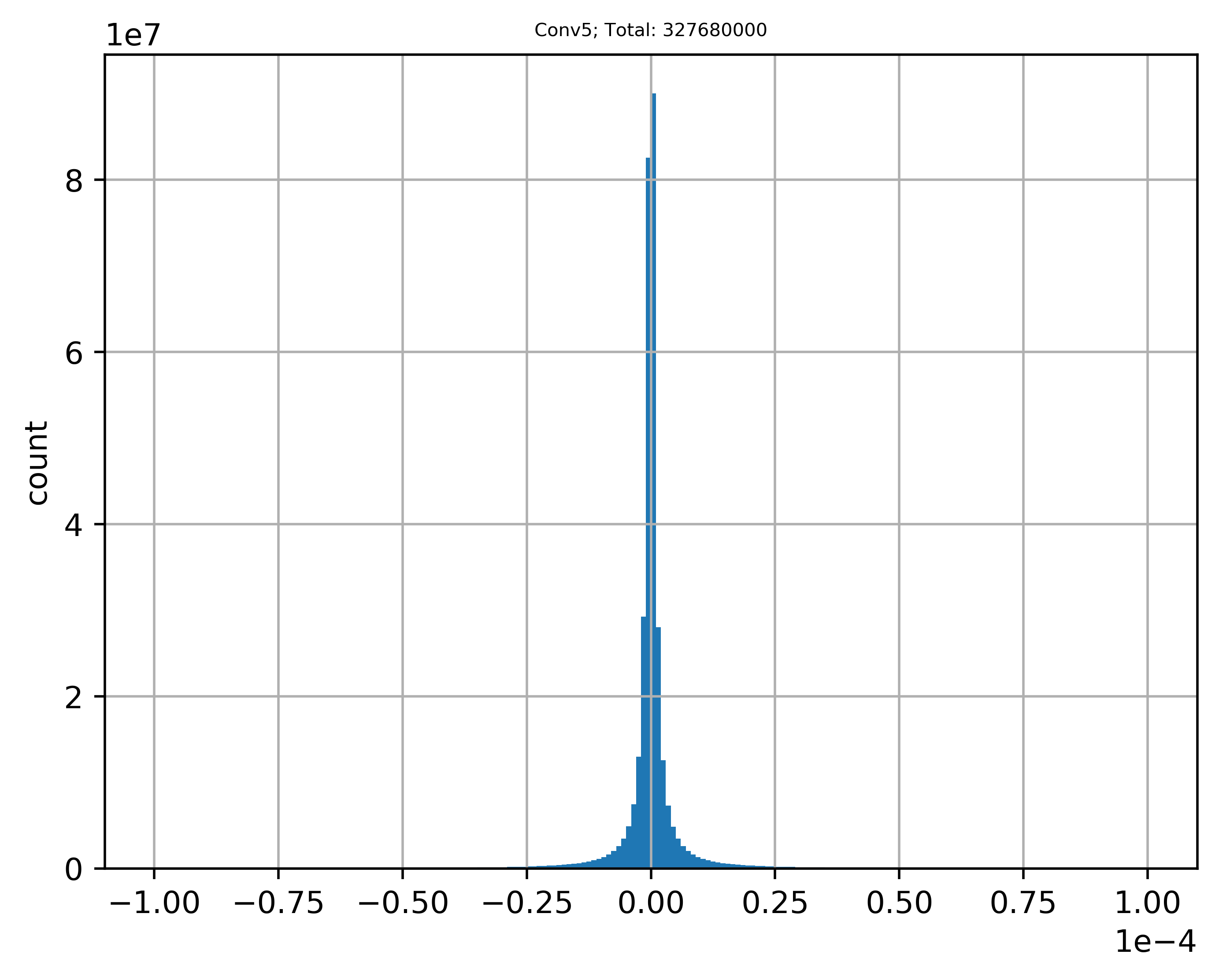} &
			\includegraphics[height=0.1\textwidth]{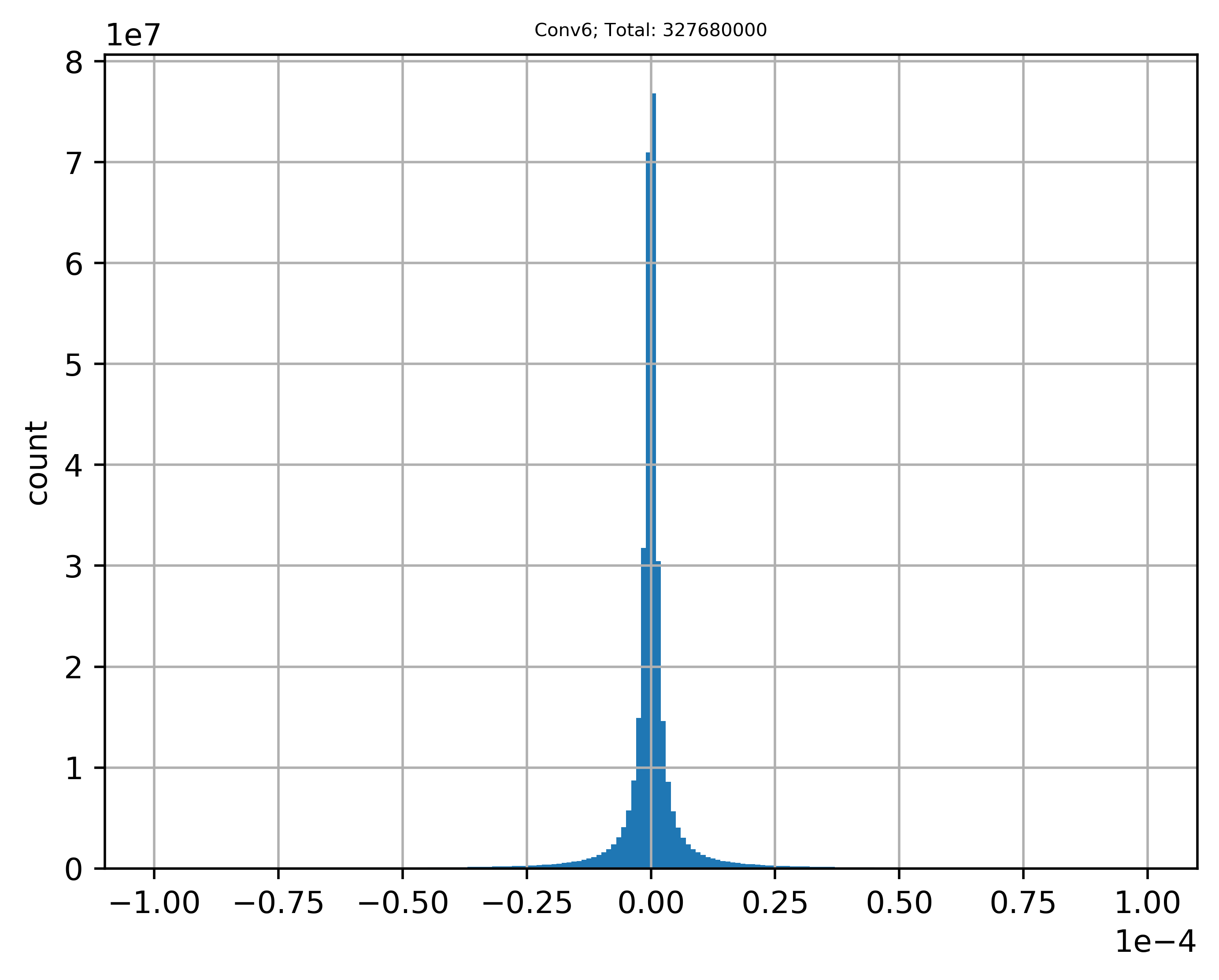} &
			\includegraphics[height=0.1\textwidth]{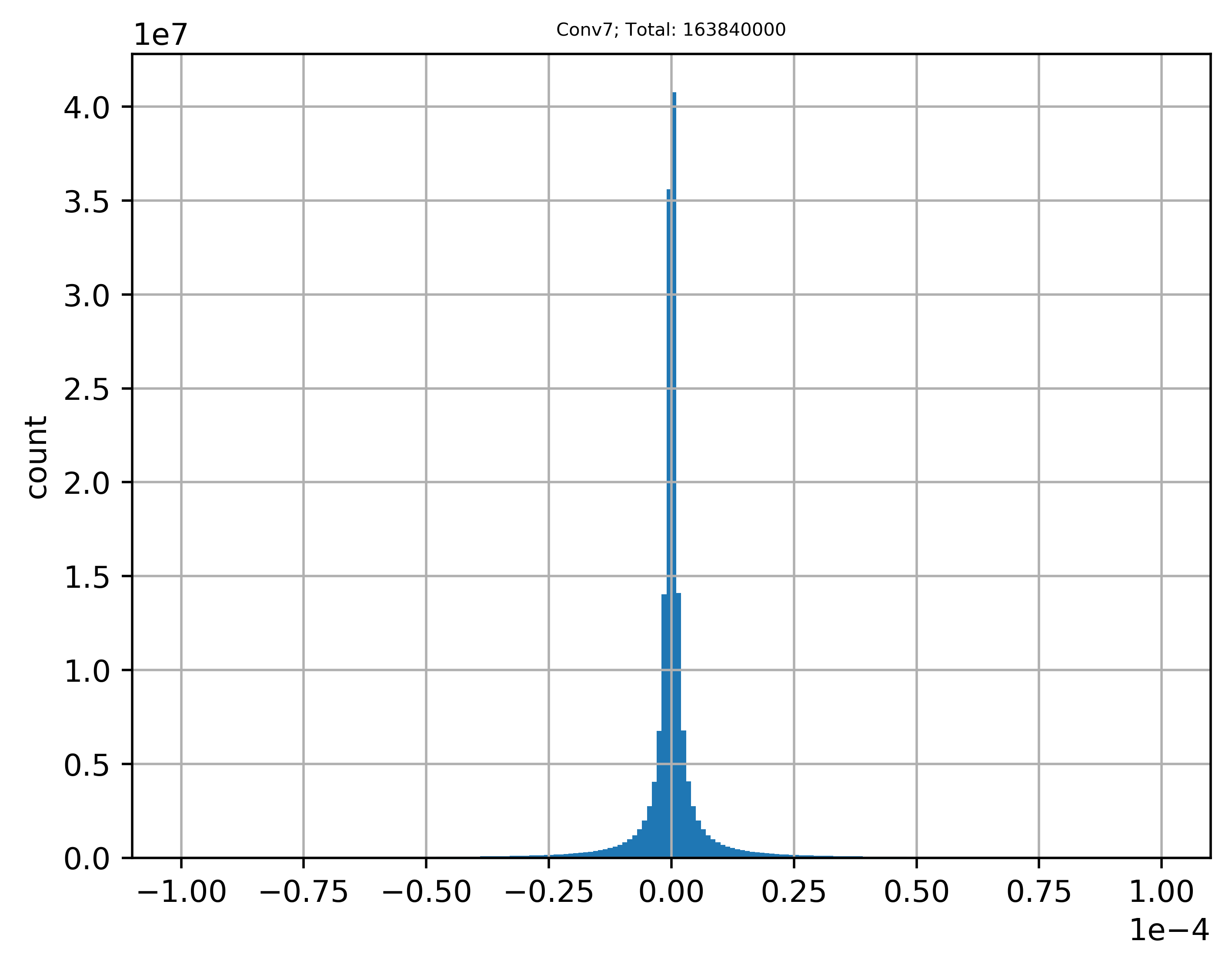} \\
			($a_1$) & ($a_2$) & ($a_3$) & ($a_4$) & ($a_5$) & ($a_6$) & ($a_7$) \\
			\includegraphics[height=0.1\textwidth]{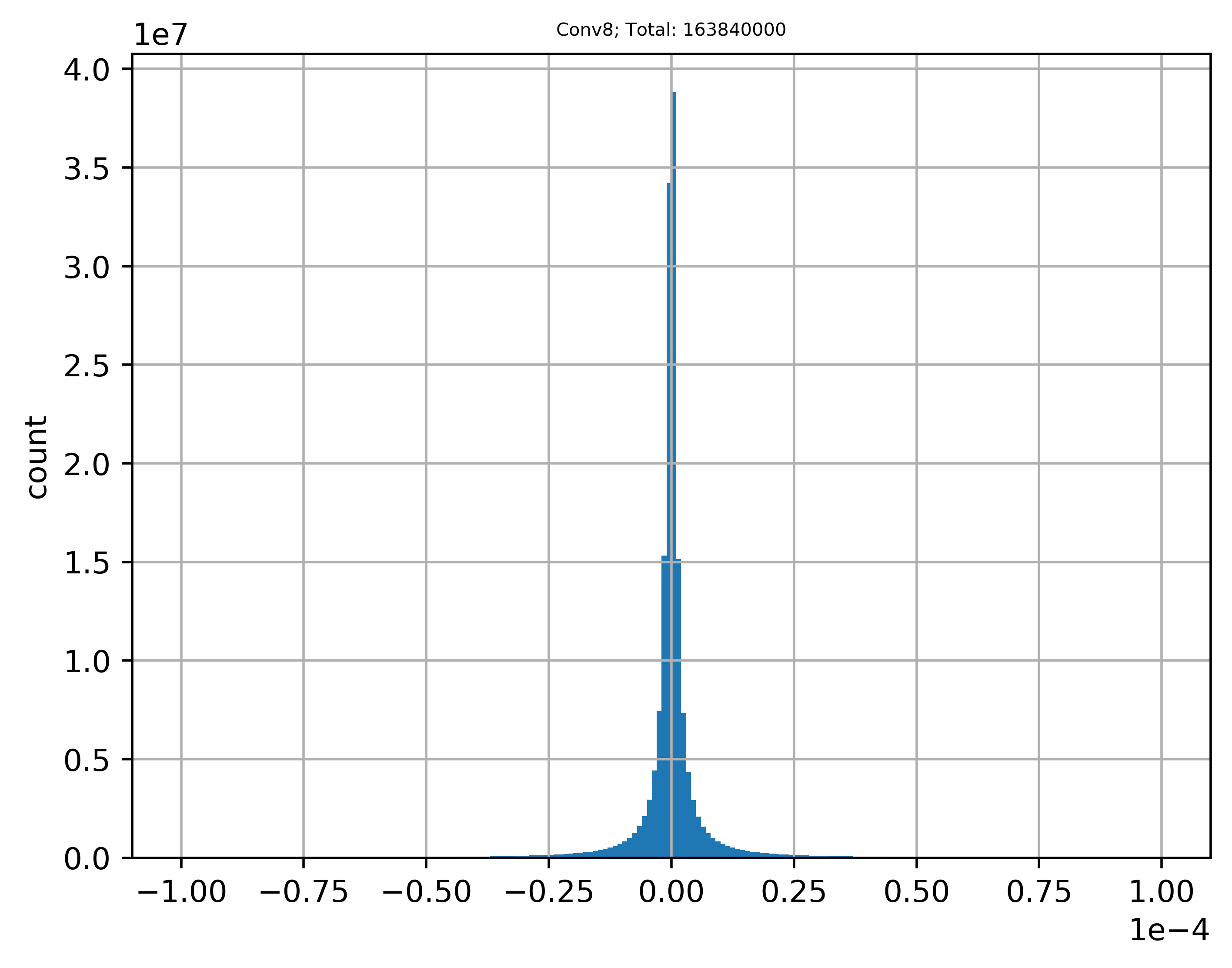} &
			\includegraphics[height=0.1\textwidth]{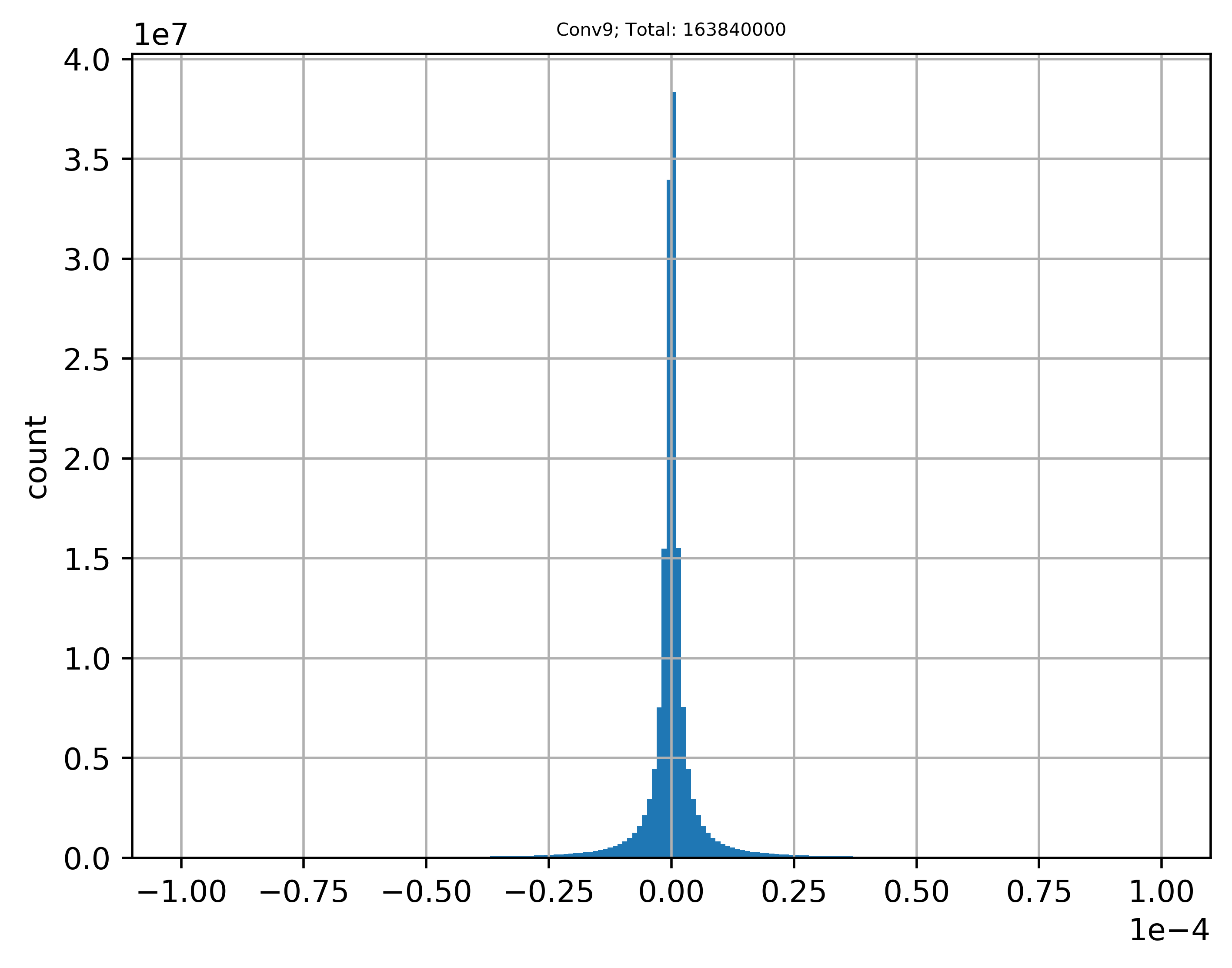} &
			\includegraphics[height=0.1\textwidth]{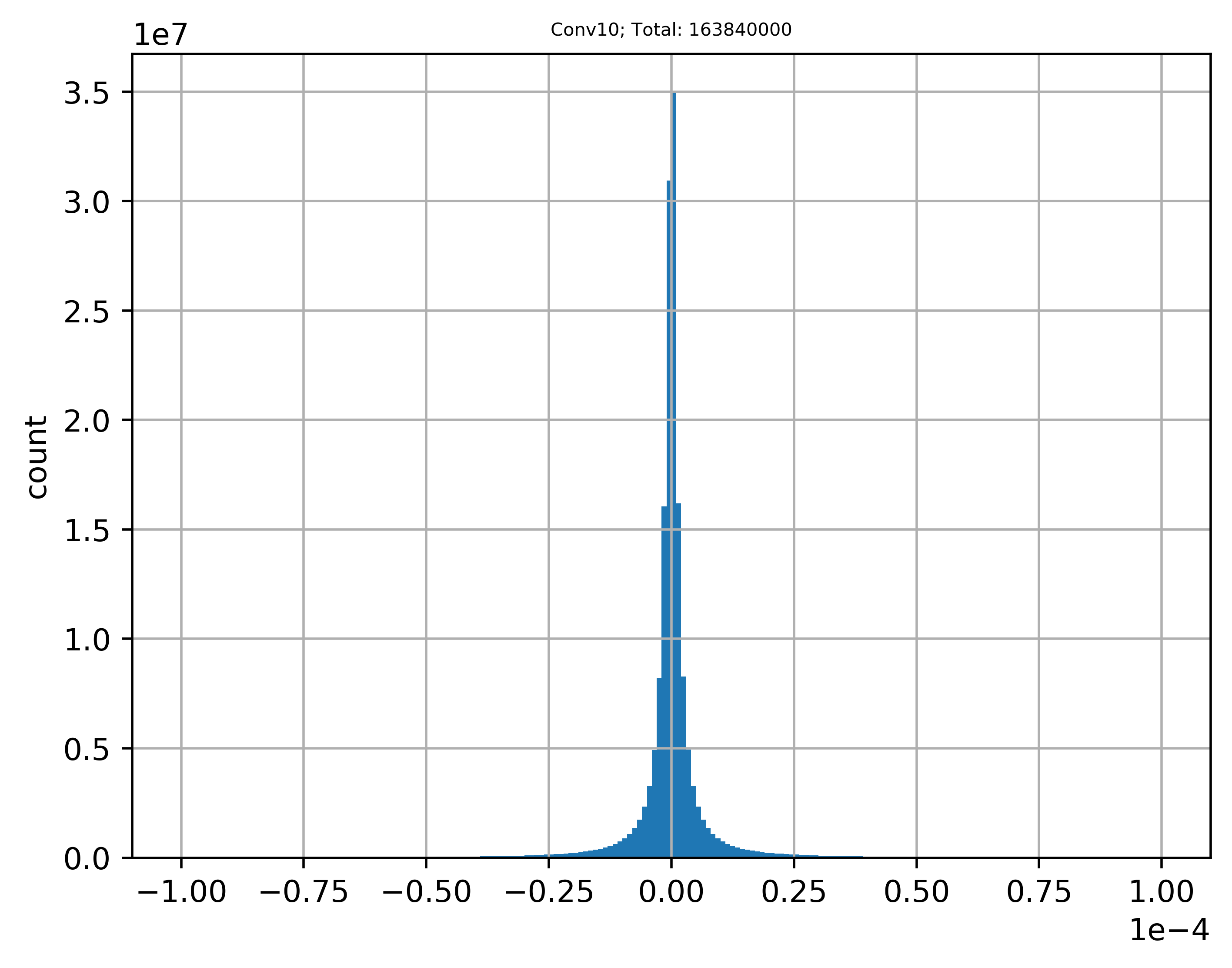} &
			\includegraphics[height=0.1\textwidth]{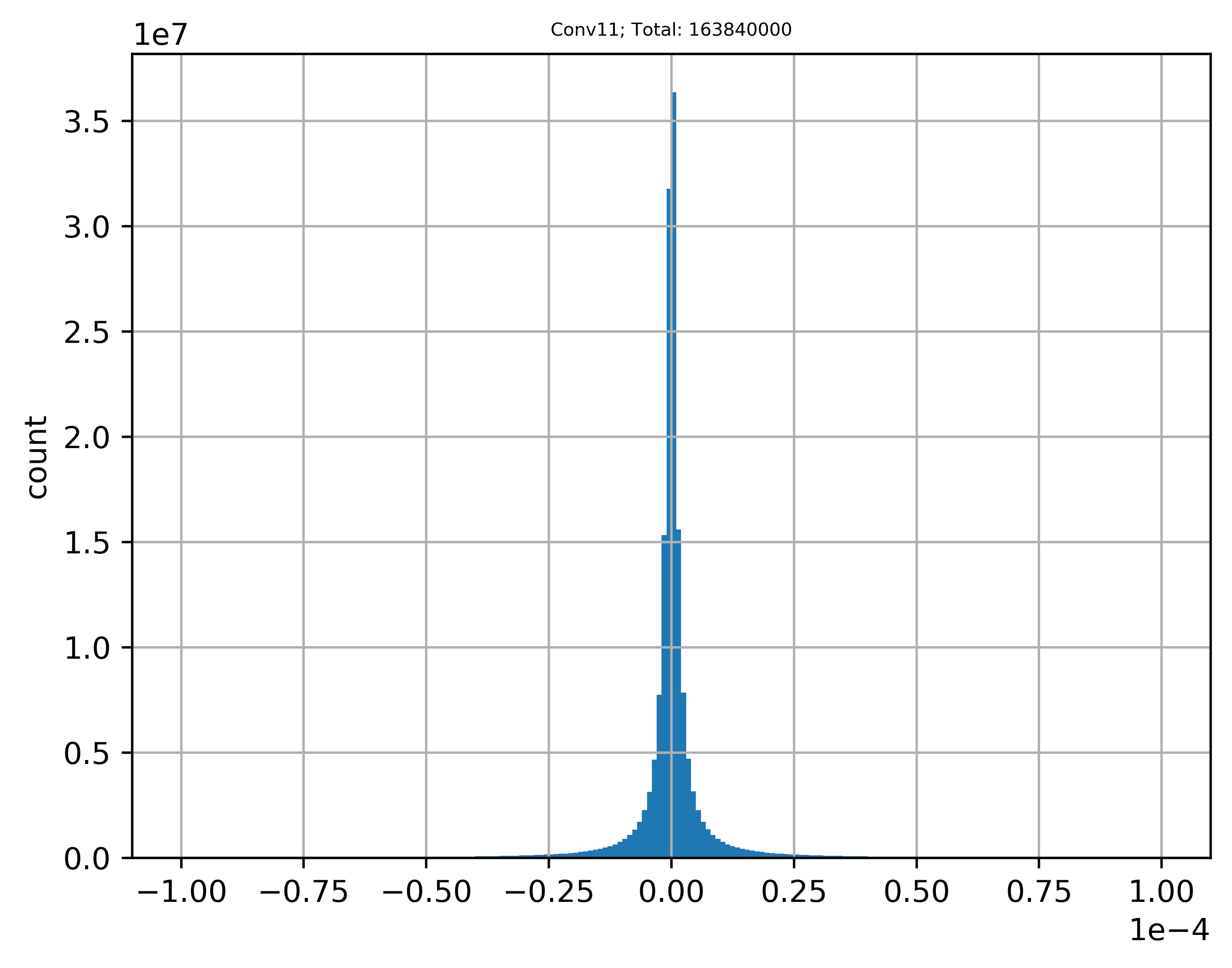} &
			\includegraphics[height=0.1\textwidth]{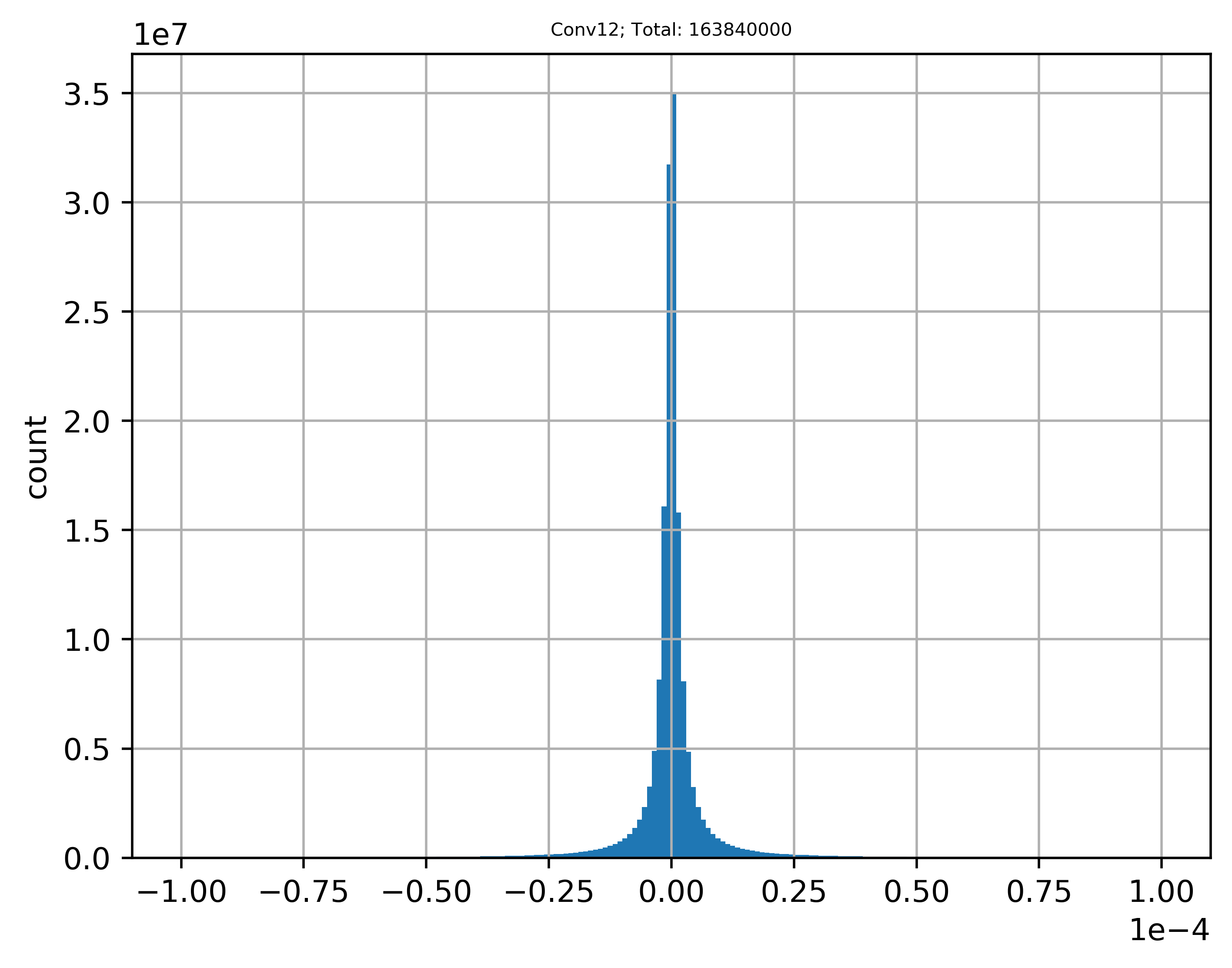} &
			\includegraphics[height=0.1\textwidth]{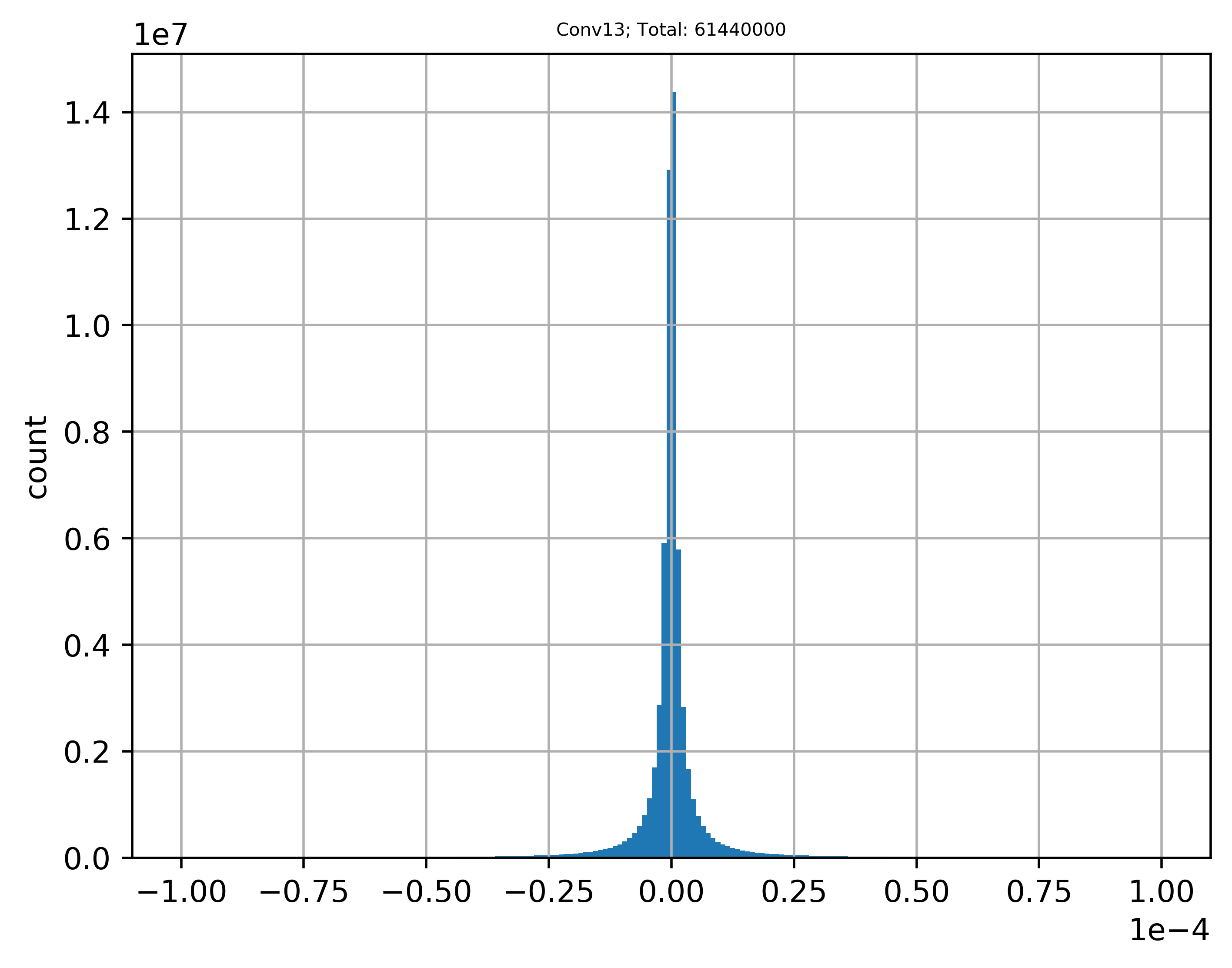} &
			\includegraphics[height=0.1\textwidth]{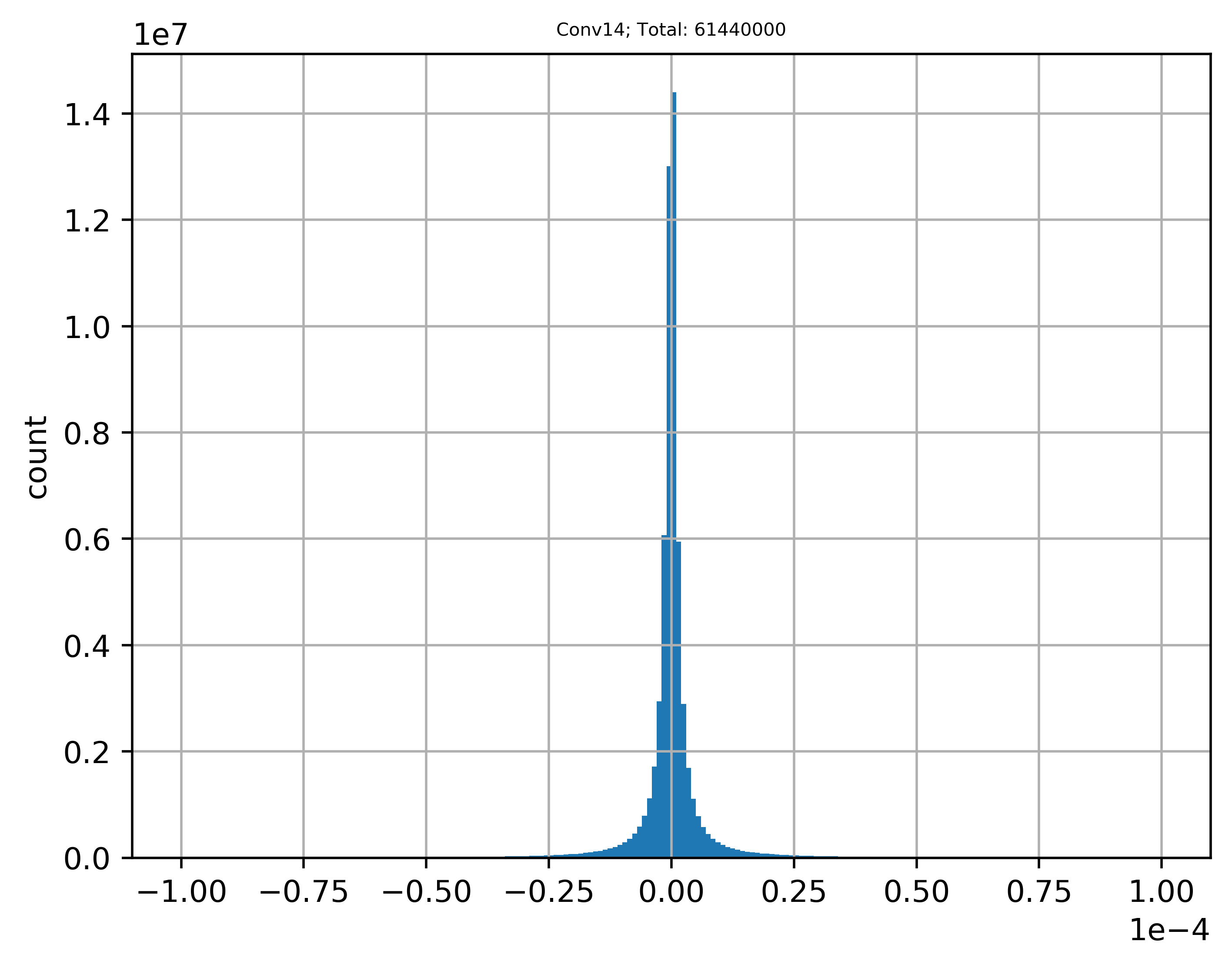} \\
			($a_8$) & ($a_9$) & ($a_{10}$) & ($a_{11}$) & ($a_{12}$) & ($a_{13}$) & ($a_{14}$) \\
			\includegraphics[height=0.1\textwidth]{Statistics/cifar10/resnet20_fixup/Conv15_rel_err_distrib.png} &
			\includegraphics[height=0.1\textwidth]{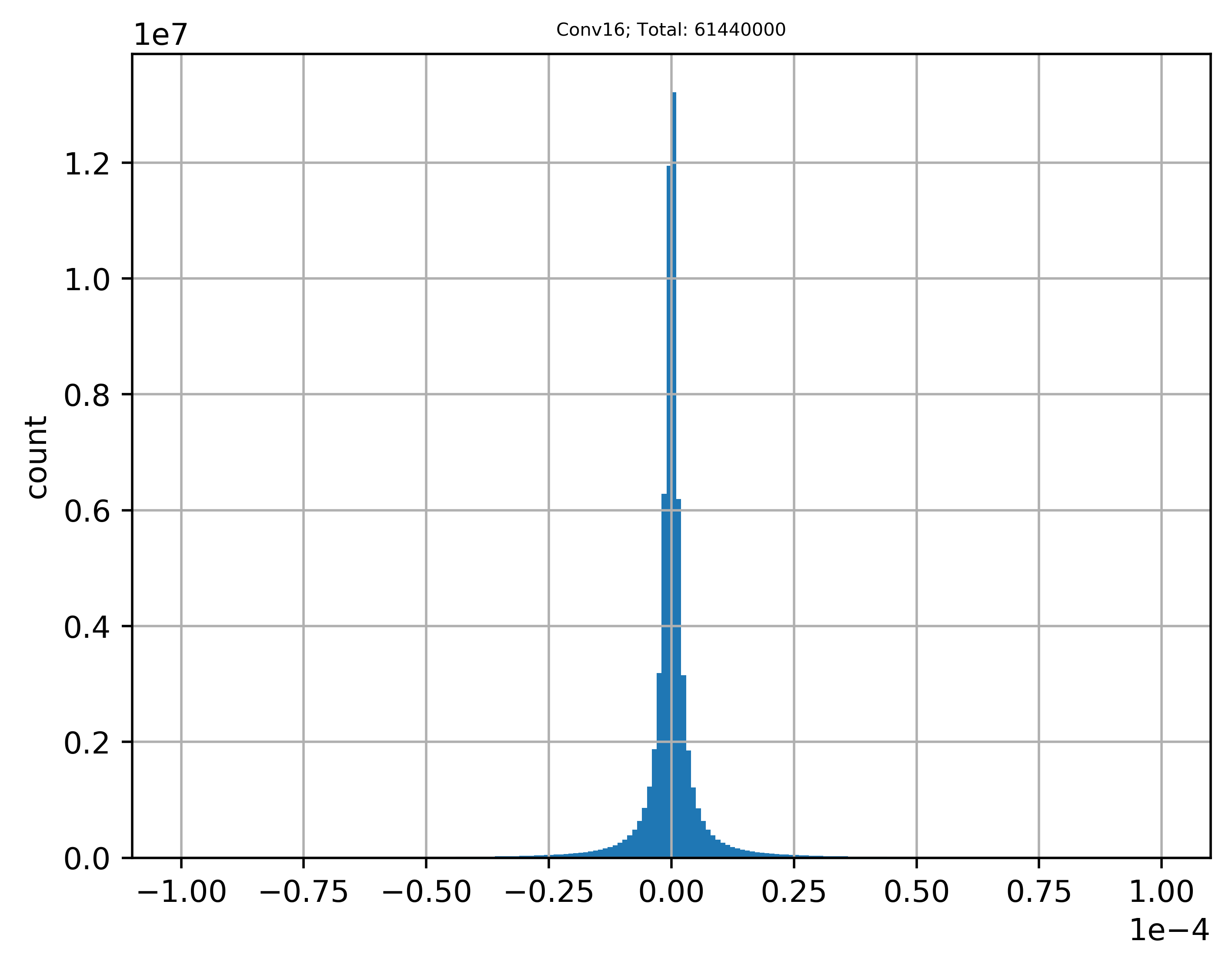} &
			\includegraphics[height=0.1\textwidth]{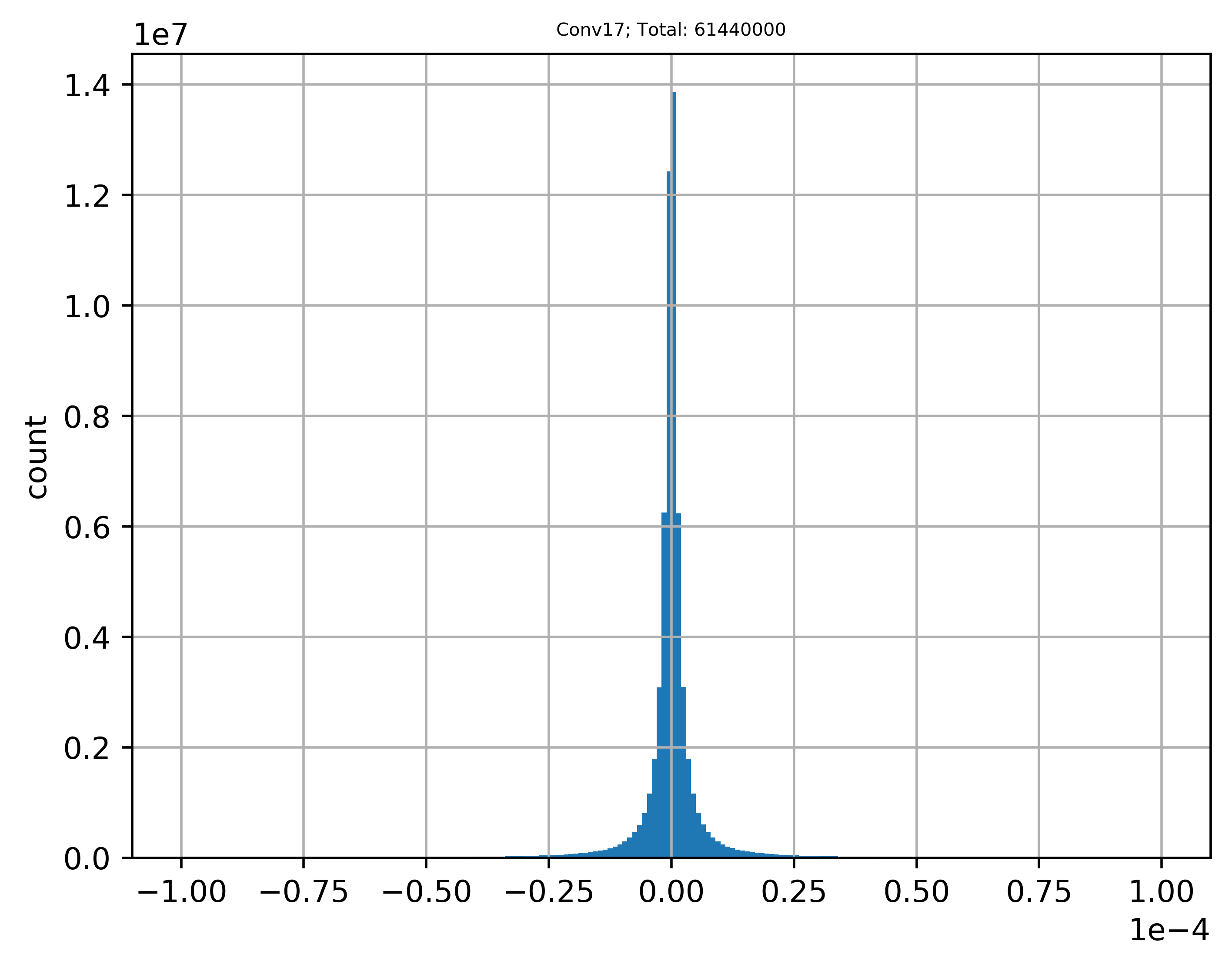} &
			\includegraphics[height=0.1\textwidth]{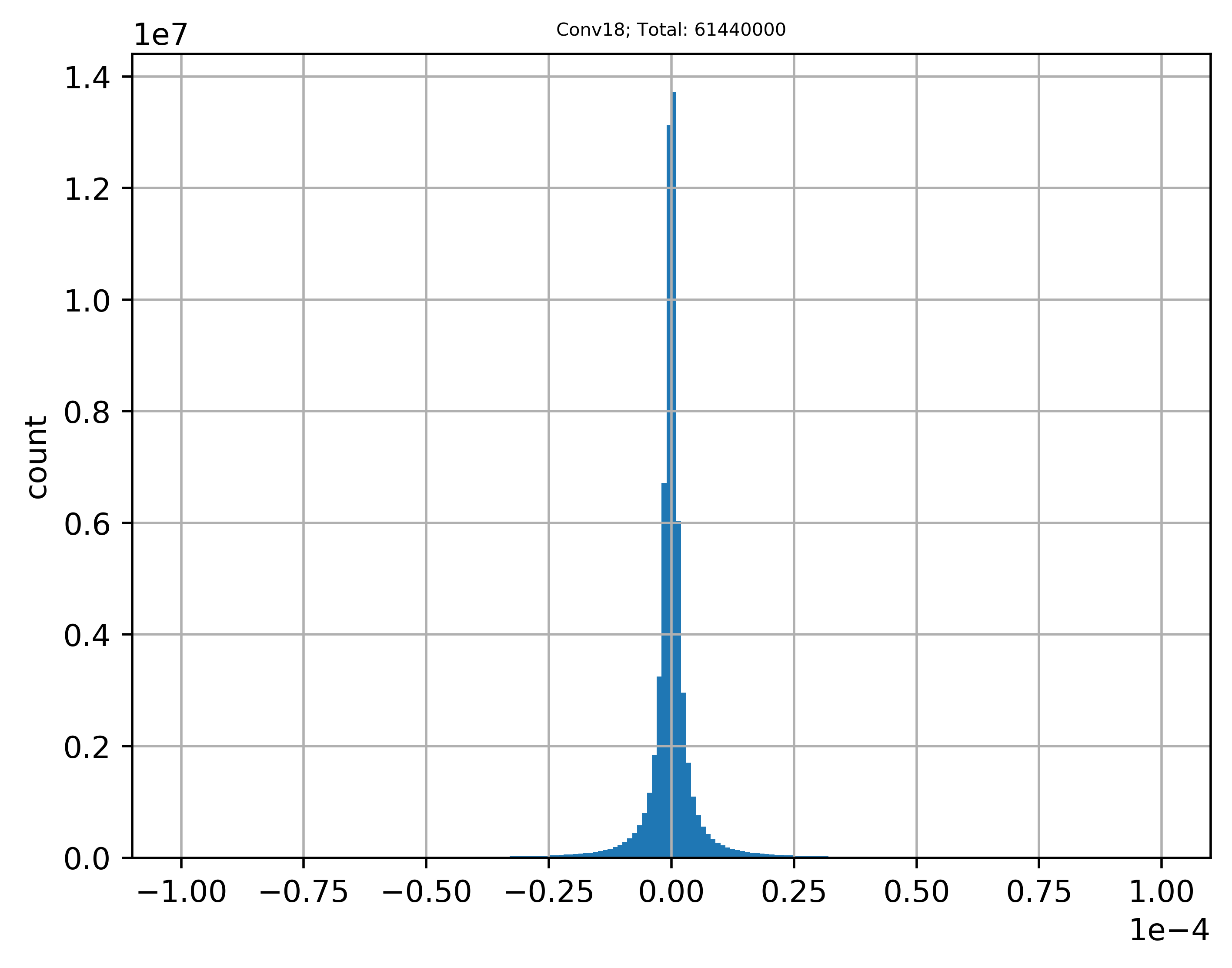} &
			\includegraphics[height=0.1\textwidth]{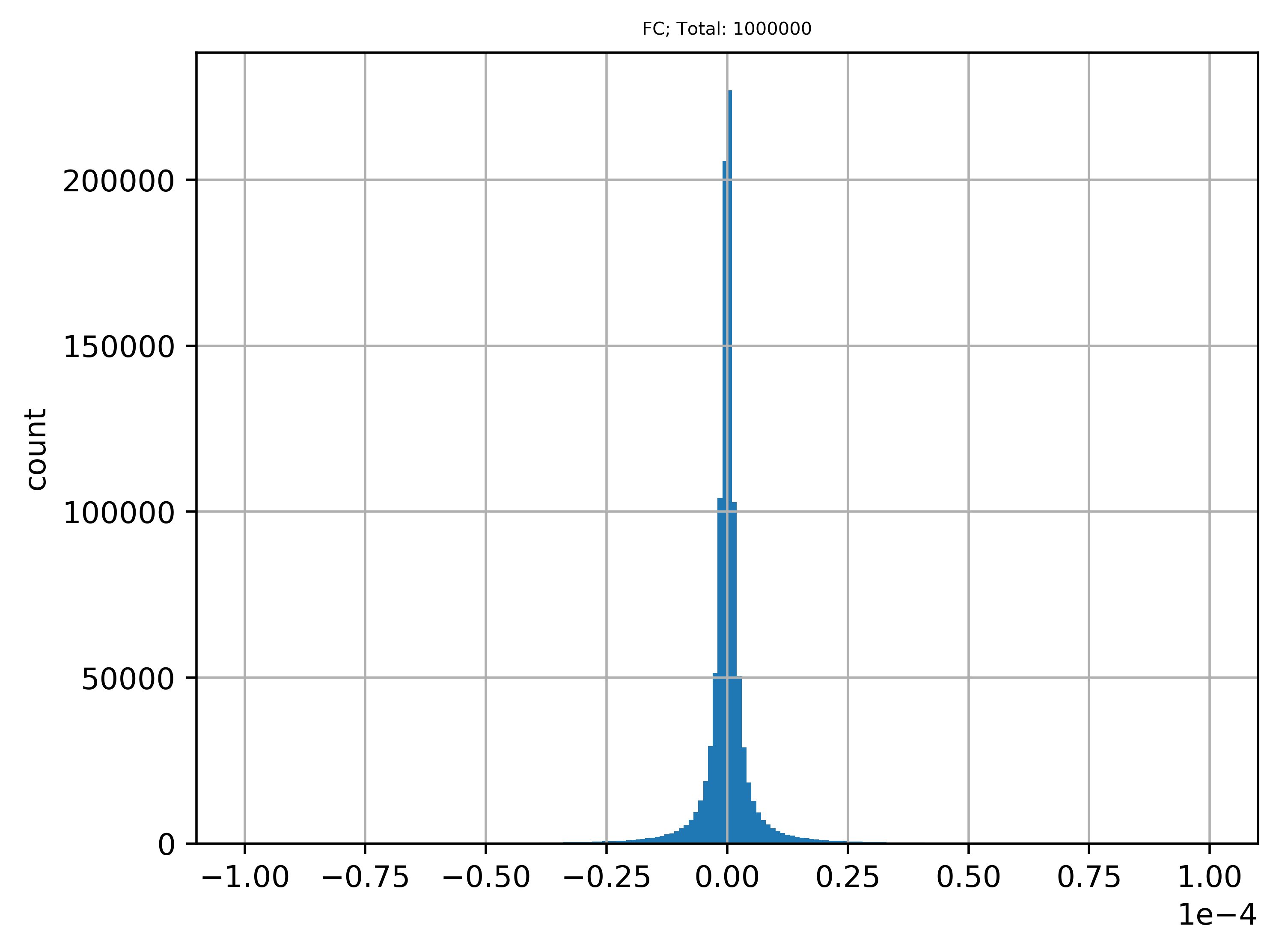} & & \\
			($a_{15}$) & ($a_{16}$) & ($a_{17}$) & ($a_{18}$) & ($a_{19}$) & & \\
		\end{tabular}
		\caption{ResNet20-Fixup (CIFAR-100). Histograms of relative errors (Eq.\eqref{eq:relative_err}) between units directly computed by CNN ($\mathbf{c}_{l}$) and their values reconstructed by our method ($\hat{\mathbf{c}}_{l}$) on CIFAR-100 using ResNet20-Fixup. Details are similar to Fig.\ref{fig:Statistics:cifar10:resnet20} except that the quantity of relative errors in $a_{19}$ is $1m$. Percentages of $\mathbf{\epsilon}_{l} \le 1\%$ for all subplots are listed in Table \ref{table:percentage_of_errs:cifar100}.}
		\label{fig:Statistics:cifar100:resnet20_fixup}
	\end{figure}

We verify our CNN unit activation value reconstruction approach on three models with CIFAR-10 and CIFAR-100 test sets. Relative errors $\mathbf{\epsilon}_{l}$ are collected layer by layer over the $10k$ test samples for each CNN model. All relative errors are shown as histograms in Figures \ref{fig:Statistics:cifar10:vgg7}, \ref{fig:Statistics:cifar10:resnet20}, \ref{fig:Statistics:cifar10:resnet20_fixup}, \ref{fig:Statistics:cifar100:vgg7}, \ref{fig:Statistics:cifar100:resnet20}, and \ref{fig:Statistics:cifar100:resnet20_fixup}. VGG7 has 6 histograms (5 convs + 1 FC), and ResNet20/ResNet20-Fixup each have 19 histograms (18 convs + 1 FC). Percentages of units with reconstruction error $\mathbf{\epsilon}_{l} \le 1 \%$ collected from all models' layers are listed in Tables \ref{table:percentage_of_errs:cifar10} and \ref{table:percentage_of_errs:cifar100}. We also illustrate several examples in Section \ref{Appendix:Pictures} (Appendix). In summary, all three models show that over 99.97\% of the units have relative reconstruction errors $\le 1 \%$. These results experimentally validate that $\mathbf{z}([\mathbf{x}_{n}; \mathbf{x}_{b}]) = \frac{[\mathbf{x}_{n}; \mathbf{x}_{b}]}{8}$ achieves Eq.\eqref{eq:approx}.
	
	Therefore, the soundness of our theory has been experimentally confirmed.
	
	\begin{table}[h]
		\centering
		\begin{tabular}{|c|c|c|c|}
			\hline
			$\mathbf{\epsilon}_{l}$/ \% / Models & VGG7 & ResNet20 & ResNet20-Fixup \\
			\hline
			$\mathbf{\epsilon}_{1}$ & 99.9987 & 99.9999 & 99.9972 \\
			\hline
			$\mathbf{\epsilon}_{2}$ & 99.9978 & 99.9994 & 99.9992 \\
			\hline
			$\mathbf{\epsilon}_{3}$ & 99.9985 & 99.999 & 99.9993 \\
			\hline
			$\mathbf{\epsilon}_{4}$ & 99.9935 & 99.9977 & 99.9989 \\
			\hline
			$\mathbf{\epsilon}_{5}$ & 99.9933 & 99.9974 & 99.9974 \\
			\hline
			$\mathbf{\epsilon}_{6}$ & 99.991 (FC) & 99.9919 & 99.9975 \\
			\hline
			$\mathbf{\epsilon}_{7}$ & N/A & 99.9835 & 99.9914 \\
			\hline
			$\mathbf{\epsilon}_{8}$ & N/A & 99.9839 & 99.9946 \\
			\hline
			$\mathbf{\epsilon}_{9}$ & N/A & 99.986 & 99.9954 \\
			\hline
			$\mathbf{\epsilon}_{10}$ & N/A & 99.9877 & 99.9959 \\
			\hline
			$\mathbf{\epsilon}_{11}$ & N/A & 99.9835 & 99.9953 \\
			\hline
			$\mathbf{\epsilon}_{12}$ & N/A & 99.9871 & 99.9955 \\
			\hline
			$\mathbf{\epsilon}_{13}$ & N/A & 99.9807 & 99.9935 \\
			\hline
			$\mathbf{\epsilon}_{14}$ & N/A & 99.9808 & 99.9935 \\
			\hline
			$\mathbf{\epsilon}_{15}$ & N/A & 99.978 & 99.9943 \\
			\hline
			$\mathbf{\epsilon}_{16}$ & N/A & 99.9781 & 99.9934 \\
			\hline
			$\mathbf{\epsilon}_{17}$ & N/A & 99.9764 & 99.9929 \\
			\hline
			$\mathbf{\epsilon}_{18}$ & N/A & 99.978 & 99.9927 \\
			\hline
			$\mathbf{\epsilon}_{19}$ & N/A & 99.989 (FC) & 99.992 (FC) \\
			\hline
		\end{tabular}
		\caption{Results Summary (CIFAR10). Percentages (\%) of units with reconstruction error $\mathbf{\epsilon}_{l} \le  1 \%$ collected from all layers (involved in verification experiments) of VGG7/ResNet20/ResNet20-Fixup over the CIFAR-10 test set ($10k$ samples). N/A indicates that the layer does not exist in the model (VGG7 has only 6 layers, and its Conv0 is not valid in our analysis scope).}
		\label{table:percentage_of_errs:cifar10}
	\end{table}
	
	\begin{table}[h]
		\centering
		\begin{tabular}{|c|c|c|c|}
			\hline
			$\mathbf{\epsilon}_{l}$/ \% / Models & VGG7 & ResNet20 & ResNet20-Fixup \\
			\hline
			$\mathbf{\epsilon}_{1}$ & 99.9988 & 99.9997 &  99.9997 \\
			\hline
			$\mathbf{\epsilon}_{2}$ & 99.9978 & 99.9995 &  99.9995 \\
			\hline
			$\mathbf{\epsilon}_{3}$ & 99.998 & 99.9988 &  99.9993 \\
			\hline
			$\mathbf{\epsilon}_{4}$ & 99.9928 & 99.9963 &  99.9992 \\
			\hline
			$\mathbf{\epsilon}_{5}$ & 99.993 & 99.9968 &  99.9989 \\
			\hline
			$\mathbf{\epsilon}_{6}$ & 99.9807 (FC) & 99.9833 &  99.9967 \\
			\hline
			$\mathbf{\epsilon}_{7}$ & N/A & 99.9789 &  99.9896 \\
			\hline
			$\mathbf{\epsilon}_{8}$ & N/A & 99.9751 &  99.9951 \\
			\hline
			$\mathbf{\epsilon}_{9}$ & N/A & 99.9878 &  99.9952 \\
			\hline
			$\mathbf{\epsilon}_{10}$ & N/A & 99.9808 &  99.9953 \\
			\hline
			$\mathbf{\epsilon}_{11}$ & N/A & 99.979 &  99.994 \\
			\hline
			$\mathbf{\epsilon}_{12}$ & N/A & 99.9839 &  99.9953 \\
			\hline
			$\mathbf{\epsilon}_{13}$ & N/A & 99.9767 &  99.9921 \\
			\hline
			$\mathbf{\epsilon}_{14}$ & N/A & 99.9746 &  99.9935 \\
			\hline
			$\mathbf{\epsilon}_{15}$ & N/A & 99.9744 &  99.9941 \\
			\hline
			$\mathbf{\epsilon}_{16}$ & N/A & 99.9726 &  99.993 \\
			\hline
			$\mathbf{\epsilon}_{17}$ & N/A & 99.9705 &  99.9935 \\
			\hline
			$\mathbf{\epsilon}_{18}$ & N/A & 99.9705 &  99.9949 \\
			\hline
			$\mathbf{\epsilon}_{19}$ & N/A & 99.9777 (FC) &  99.9896 (FC) \\
			\hline
		\end{tabular}
		\caption{Results Summary (CIFAR100). Percentages (\%) of units with reconstruction error $\mathbf{\epsilon}_{l} \le  1 \%$ collected from all layers of VGG7/ResNet20/ResNet20-Fixup over the CIFAR-100 test set ($10k$ samples). Check Table \ref{table:percentage_of_errs:cifar10} for details.}
		\label{table:percentage_of_errs:cifar100}
	\end{table}
	
	\section{Conclusions}
	\label{Conclusions}
	Given a unit in a convolved feature map or a class prediction from the FC layer inside a CNN, its value can be precisely reconstructed through the sum of two dot-products in the input end, one of which is the input image $\mathbf{x}_{n}$ multiplied by the reconstructed $\mathbf{H}^{Adj, I}(\mathbf{z([\mathbf{x}_{n}: \mathbf{x}_{b}])})$, and the other one is the concatenated layers' bias $\mathbf{x}_{b}$ multiplied by $\mathbf{H}^{Adj, b}(\mathbf{z([\mathbf{x}_{n}: \mathbf{x}_{b}])})$, where the operator tensors $\mathbf{H}^{Adj, I}, \mathbf{H}^{Adj, b}$ can be directly accessed from our AdjointBackMapV2, Algorithm.\ref{algorithm:RM4_to_RM0}. To find this answer, we slightly extend the image space $\mathcal{X}_{I}$ to a bigger one, $\mathcal{X}$, and aim to overcome the bias-free restriction in our earlier work \cite{wan2022adjointbackmap}. Thanks to Adjoint operators and the Riesz Representation, we can project weights from a high-level layer back to the joint space of images and bias to reconstruct an effective hypersurface that replicates a unit of a convolved feature map or the predicted value of the FC layer, through five upgraded reconstruction modes (RMs), as long as the conditions in Section \ref{Model:Algorithm} are satisfied. Both theoretical analysis and experiments on three CNN models verify the soundness of our AdjBackMapV2. We expect our theory might shed light on unveiling CNN's inner workings.
	
	\section{Acknowledgements}
	\label{Acknowledgements}
	Siu Wun Cheung's research was performed at Texas A\&M University. This manuscript was prepared by Lawrence Livermore National Laboratory under Contract DE-AC52-07NA27344 and LLNL-JRNL-848797.

	\bibliography{refs}\label{reference}

	\newpage
	\renewcommand{\theequation}{\thesection.\arabic{equation}}
        \setcounter{equation}{0}
	\renewcommand\thefigure{\thesection.\arabic{figure}}
        \setcounter{figure}{0}
        \renewcommand\thetable{\thesection.\arabic{table}}
        \setcounter{table}{0}
        
	\begin{appendices}
		\section{Notations}
		\label{Appendix:Notations}
		We will use the following notations.
		\begin{enumerate}
		  \item ``Eq'' refers to an equation in the main text;
		  \item ``eq'' refers to an equation in the appendix;
            \item ``Table'' refers to a table in the main text;
		  \item ``table'' refers to a table in the appendix;
		  \item ``Fig'' refers to a figure in the main text;
            \item ``fig'' refers to a figure in the appendix;
		  \item ``Algorithm'' refers to an algorithm in the main text.
		\end{enumerate}
	\end{appendices}

	\begin{appendices}
		\section{Hardware and software for verification experiments}
		\label{Appendix:hw_SW4VerificationExpts}
		Verification experiments (figures \ref{fig:Statistics:cifar10:vgg7}, \ref{fig:Statistics:cifar10:resnet20}, \ref{fig:Statistics:cifar10:resnet20_fixup}, \ref{fig:Statistics:cifar100:vgg7}, \ref{fig:Statistics:cifar100:resnet20}, \ref{fig:Statistics:cifar100:resnet20_fixup}) were conducted on Intel 10920X (VGG7/ResNet20) and 9940X (ResNet20-Fixup) CPUs. Both set up ran TensorFlow 1.15.4 with AVX-2, AVX-512, and FMA3 instruction sets enabled (built from source).
        \end{appendices}

        \begin{appendices}
		\section{Proof of Eq.\eqref{eq:approx}}
		\label{Appendix:Proof_of_eq_approx}
		We show $\mathbf{z}([\mathbf{x}_{n}; \mathbf{x}_{b}]) = k[\mathbf{x}_{n}; \mathbf{x}_{b}] (k \in \mathbb{R}^{+})$ achieves the third equality (``$=$'') in Eq.\eqref{eq:approx} if the CNN is activated with ReLU or Leaky ReLU units.
		
		\paragraph{Notation and concepts}
		Let $\mathbf{N}: \mathcal{X}_{I} \to \mathbb{R}^{K}$ be a CNN (without the last softmax layer) with $L$ convolutional layers, and $\mathbb{R}^{H \times W \times C}$ be an instance of $\mathcal{X}_{I}$ whose $H, W, C$ denote height, width, and number of color channels of images. Suppose $\mathbf{x}_{n} \in \mathbb{R}^{H \times W \times C}$ being an input image. The $l^{th}$ layer has convolutional kernels $\mathbf{w}_{r_{l, 1} \times r_{l, 2} \times c_{l, in} \times c_{l, out}}$ and bias $\mathbf{b}_{l}$ ($0 \le l \le L-1$). We use $\mathbf{c}_{l}(\mathbf{x}_{n})$ to depict the convolved feature maps with bias added in the $l^{th}$ layer, and $\hat{\mathbf{c}}_{l}(\mathbf{x}_{n})$ for the vectorization of $\mathbf{c}_{l}(\mathbf{x}_{n})$, where $\hat{\cdot}$ is a vectorization operator. We use $\mathbf{p}_{l}(\mathbf{x}_{n})$ depicts the activated feature maps in the $l^{th}$ layer (the layer has an activation ${\sigma}_{l}$)), and $\hat{\mathbf{p}}_{l}(\mathbf{x}_{n})$ depicts the vectorization of $\mathbf{p}_{l}(\mathbf{x}_{n})$. Their shapes are $\mathbf{c}_{l}, \mathbf{p}_{l} \in \mathbb{R}^{h_l \times w_l \times c_{l, in}}$ and $\hat{\mathbf{c}}_{l}, \hat{\mathbf{p}}_{l}, \mathbf{b}_{l} \in \mathbb{R}^{m_l}$ where $m_l = h_l w_l c_{l, in}$. Specifically, $(h_{-1}, w_{-1}, c_{-1, in}) = (H, W, C)$, $m_{-1}=HWC$, and $\hat{\mathbf{p}}_{-1}(\mathbf{x}_{n}) = \hat{\mathbf{c}}_{-1}(\mathbf{x}_{n}) = \hat{\mathbf{x}}_{i} \in \mathbb{R}^{m_{-1}}$. Then, their relationships can be described as below,
		\begin{equation} \label{eq:proof_of_eq_approx:separatedforms}
			\begin{split}
				\mathbf{c}_{l}(\mathbf{x}_{n}) = \mathbf{p}_{l-1}(\mathbf{x}_{n}) \circledast \mathbf{w}_{r_{l, 1} \times r_{l, 2} \times c_{l, in} \times c_{l, out}} + \mathbf{b}_{l}\\
				= \hat{\mathbf{c}}_{l}(\mathbf{x}_{n}) 
				= \mathbf{A}_{l}(\hat{\mathbf{p}}_{l-1}(\mathbf{x}_{n}))
				= \mathbf{w}_{l}(\hat{\mathbf{p}}_{l-1}(\mathbf{x}_{n})) + \mathbf{b}_{l}, \\
				\mathbf{p}_{l}(\mathbf{x}_{n}) = {\sigma}_{l}({\mathbf{c}_{l}}(\mathbf{x}_{n})) = {\sigma}_{l}(\hat{\mathbf{c}}_{l}(\mathbf{x}_{n})) = \hat{\mathbf{p}}_{l}(\mathbf{x}_{n}),
			\end{split}
		\end{equation}
		where $\mathbf{w}_{l} \in \mathbb{R}^{m_t \times m_{t-1}}$ is the matrix representation for the layer's convoltion, and $\mathbf{A}_{l}$ is an affine operator that describes the convolution and bias addition. The neural network parameters, $\{(\mathbf{w}_{l}, \mathbf{b}_{l})\}_{l=0}^{l=L-1}$ define the composition of convolutions, average/max poolings (average pooling is equivalent to convolve with $r_{l, 1} \times r_{l, 2}$-sized averaging kernels, and max poolings is equivalent to convolve with $r_{l, 1} \times r_{l, 2}$-sized one-hot kernels), and batch normalizations inside the CNN $\mathbf{N}$. We formalize the unit-wise activation function ${\sigma}_{l}$ as following,
		\begin{equation}\label{eq:proof_of_eq_approx:act}
			{\sigma}_{l}(c) = max(c, 0) + {\gamma}_{l} min(c, 0), \forall c \in \mathbb{R},
		\end{equation}
		where the leakiness ${\gamma}_{l} = 0$ implies ${\sigma}_{l}$ being ReLU, and $0 < {\gamma}_{l} < 1$ implies ${\sigma}_{l}$ being Leaky ReLU. The convolutional layers are terminated at $l = L-1$, and the overall final output $\mathbf{N}(\mathbf{x}_{l}) \in \mathbb{R}^{K}$ is given by,
		\begin{equation} \label{eq:proof_of_eq_approx:finalout}
			\begin{split}
				\mathbf{c}_{L}(\mathbf{x}_{n})
				= \hat{\mathbf{c}}_{L}(\mathbf{x}_{n})
				= \mathbf{A}_{L}(\hat{\mathbf{c}}_{L-1}(\mathbf{x}_{n})) = \mathbf{w}_{L}\hat{\mathbf{c}}_{L-1}(\mathbf{x}_{n}) + \mathbf{b}_{L} \\
				\mathbf{N}(\mathbf{x}_{n})
				= \hat{\mathbf{p}}_{L}(\mathbf{x}_{n})
				= {\sigma}_{L}(\hat{\mathbf{c}}_{L}(\mathbf{x}_{n})).
			\end{split}
		\end{equation}
		Combine the above, we will have,
		\begin{equation} \label{eq:proof_of_eq_approx:composite}
			\mathbf{N}(\mathbf{x}_{n}) = ({\sigma}_{L} \circ \mathbf{A}_{L} \circ {\sigma}_{L-1} \circ \mathbf{A}_{L-1} \circ {...} {\sigma}_{1} \circ \mathbf{A}_{1} \circ {\sigma}_{0} \circ \mathbf{A}_{0})(\hat{\mathbf{p}}_{-1}(\mathbf{x}_{n}))
		\end{equation}
		
		\paragraph{Representation of the equivalent topology in Fig.\ref{fig:Model:cnn_with_bias_and_its_equivalence}}
		We use $M = \sum_{l=0}^{L}m_{l}$ which is equal to $\mathbf{x}_{b}$'s dimensions. For $0 \le l \le L$, we introduce a restriction operator $\mathbf{r}_l: \mathbb{R}^{M} \to \mathbb{R}^{m_l}$, by
		\begin{equation} \label{eq:proof_of_eq_approx:restriction_op}
			\mathbf{r}_l(\mathbf{v}_{b}) 
			= \mathbf{y}_{l}, \forall \mathbf{v}_{b}
			= [\mathbf{y}_{0}; \mathbf{y}_{1}, ..., \mathbf{y}_{L}] \in \mathbb{R}^{M} \mbox{ with } \mathbf{y}_{l} \in \mathbb{R}^{m_l}.
		\end{equation}
		Note that $\mathbf{v}_{b}$ is a variable vector while $\mathbf{x}_{b}$ is fixed. Then, we define a sequence of auxiliary mappings $\{\mathbf{\mathcal{c}}_{l}\}_{l=0}^{l=L-1}$ by the recurrence relation: for $[\hat{\mathbf{x}}_{i}; \mathbf{v}_{b}] \in \mathbb{R}^{m_{-1}} \bigtimes \mathbb{R}^{M}$,
		\begin{equation} \label{eq:proof_of_eq_approx:representation4equivalent_topology}
			\begin{split}
				\mathbf{\mathcal{c}}_{0}([\hat{\mathbf{x}}_{i}; \mathbf{v}_{b}])
				= \mathbf{w}_{0}\hat{\mathbf{x}}_{i} + \mathbf{r}_0(\mathbf{v}_{b}), \\
				\mathbf{\mathcal{c}}_{l}([\hat{\mathbf{x}}_{i}; \mathbf{v}_{b}])
				= \mathbf{w}_{l}({\sigma}_{l-1}(\mathbf{\mathcal{c}}_{l-1}([\hat{\mathbf{x}}_{i}; \mathbf{v}_{b}]))) + \mathbf{r}_l(\mathbf{v}_{b}), 0 \le l \le L-1, \\
				\mathbf{\mathcal{c}}_{L}([\hat{\mathbf{x}}_{i}; \mathbf{v}_{b}])
				= \mathbf{w}_{L}({\sigma}_{L-1}(\mathbf{\mathcal{c}}_{L-1}([\hat{\mathbf{x}}_{i}; \mathbf{v}_{b}]))) + \mathbf{r}_{L}(\mathbf{v}_{b}).
			\end{split}
		\end{equation}
		Combining eq.\eqref{eq:proof_of_eq_approx:separatedforms}, \eqref{eq:proof_of_eq_approx:finalout}, we get,
		\begin{equation} \label{eq:proof_of_eq_approx:verify_equivalent_topology}
			\mathbf{c}_{l}(\mathbf{x}_{n}) = \hat{\mathbf{c}}_{l}(\mathbf{x}_{n})
			= \mathbf{\mathcal{c}}_{l}([\hat{\mathbf{x}}_{i}; \mathbf{x}_{b}]),
		\end{equation}
		where $\mathbf{v}_{b}$ is replaced by $\mathbf{x}_{b}$.
		
		\paragraph{Proof of Eq.\eqref{eq:approx}}
		From eq.\eqref{eq:proof_of_eq_approx:restriction_op}, for any $[\hat{\mathbf{x}}_{i}; \mathbf{v}_{b}] \in \mathbb{R}^{m_{-1}} \bigtimes \mathbb{R}^{M}$, we define a sequence of matrices $\{\mathbf{J}_{l}([\hat{\mathbf{x}}_{i}; \mathbf{v}_{b}])\}_{l=0}^{L}$ using the recurrence relation:
		\begin{equation} \label{eq:proof_of_eq_approx:jacobian}
			\begin{split}
				\mathbf{J}_{0}([\hat{\mathbf{x}}_{i}; \mathbf{v}_{b}])
				= [\mathbf{w}_{0}, \mathbf{R}_0] \in \mathbb{R}^{m_0 \times (m_{-1} + M)}, \\
				\mathbf{J}_{l}([\hat{\mathbf{x}}_{i}; \mathbf{v}_{b}])
				= \mathbf{w}_{l}\mathbf{\Sigma}_{l-1}([\hat{\mathbf{x}}_{i}; \mathbf{v}_{b}])\mathbf{J}_{l-1}([\hat{\mathbf{x}}_{i}; \mathbf{v}_{b}]) + [\mathbf{O}_{l}, \mathbf{R}_{l}], 0 \le l \le L-1\\
				\mathbf{J}_{L}([\hat{\mathbf{x}}_{i}; \mathbf{v}_{b}])
				= \mathbf{w}_{L}\mathbf{\Sigma}_{L-1}([\hat{\mathbf{x}}_{i}; \mathbf{v}_{b}])\mathbf{J}_{L-1}([\hat{\mathbf{x}}_{i}; \mathbf{v}_{b}]) + [\mathbf{O}_{L}, \mathbf{R}_{L}],
			\end{split}
		\end{equation}
		where $[\cdot, \cdot]$ is an operator that concatenates two matrices; $\mathbf{R}_{l} \in \mathbb{R}^{m_t \times M}$ for $0 \le l \le L-1$, and $\mathbf{R}_{L} \in \mathbb{R}^{k \times M}$, is the matrix representations of the restriction operator $\mathbf{r}_{l}$; $\mathbf{O}_{l} \in \mathbb{R}^{m_t \times m_{-1}}$ is the zero matrix; ${\mathbf{\Sigma}}_{l}([\hat{\mathbf{x}}_{i}; \mathbf{v}_{b}]) \in \mathbb{R}^{m_t \times m_t}$ is the Jacobian matrix of ReLU or Leaky ReLU in the $l^{th}$ convolutional layer, which is precisely given by,
		\begin{equation} \label{eq:proof_of_eq_approx:act_jacobian}
			{\mathbf{\Sigma}}_{l}([\hat{\mathbf{x}}_{i}; \mathbf{v}_{b}]) = diag\left(\frac{1 + {\gamma}_{l}}{2} + \frac{1 - {\gamma}_{l}}{2} sgn(\mathbf{\mathcal{c}}_{l}([\hat{\mathbf{x}}_{i}; \mathbf{v}_{b}]))\right).
		\end{equation}
		In fact, $\mathbf{J}_{l}([\hat{\mathbf{x}}_{i}; \mathbf{v}_{b}])$ is the Jacobian matrix of $\mathbf{\mathcal{c}}_{l}$ at $[\hat{\mathbf{x}}_{i}; \mathbf{v}_{b}]$. We point out that eq.\eqref{eq:proof_of_eq_approx:act_jacobian} performs a hierarchal separation of the domain in the sense that, given a sequence of binary vectors $\mathbf{e} = \{e_l\}_{l=0}^{L}$ with $e_l \in \{-1, 1\}^{m_l}$ for $0 \le l \le L$, the Jacobian matrix $\mathbf{J}_{l}$ shares the same value on the subdomain ${\Omega}_l(\mathbf{e})$, where
		\begin{equation} \label{eq:proof_of_eq_approx:subdomain}
			\begin{split}
				{\Omega}_0(\mathbf{e})
				= \mathbb{R}^{m_{-1}} \bigtimes \mathbb{R}^{M}, \\
				{\Omega}_l(\mathbf{e})
				= \{[\hat{\mathbf{x}}_{i}; \mathbf{v}_{b}] \in {\Omega}_{l-1}(\mathbf{e}): sgn(\mathbf{\mathcal{c}}_{l-1}([\hat{\mathbf{x}}_{i}; \mathbf{v}_{b}])) = e_l \}, 0 \le l \le L.
			\end{split}
		\end{equation}
		As a direct consequence, for any $0 \le l \le L$ and $k \in \mathbb{R}^{+}$, we will have,
		\begin{equation} \label{eq:proof_of_eq_approx:main}
			\mathbf{J}_{l}(k[\hat{\mathbf{x}}_{i}; \mathbf{v}_{b}])
			= \mathbf{J}_{l}([\hat{\mathbf{x}}_{i}; \mathbf{v}_{b}]).
		\end{equation}
		Note that $\mathbf{v}_{b} = \mathbf{x}_{b}$ is the case we discussed in the Section \ref{Model}.
		
		With the above definitions, we can rewrite eq.\eqref{eq:proof_of_eq_approx:representation4equivalent_topology} as,
		\begin{equation} \label{eq:proof_of_eq_approx:combine_equivalenttopology_with_jacobian}
			\mathbf{\mathcal{c}}_{l}([\hat{\mathbf{x}}_{i}; \mathbf{v}_{b}])
			= \mathbf{J}_{l}([\hat{\mathbf{x}}_{i}; \mathbf{v}_{b}])[\hat{\mathbf{x}}_{i}; \mathbf{v}_{b}].
		\end{equation}
		Together with the fact that $\mathbf{J}_{l}([\hat{\mathbf{x}}_{i}; \mathbf{v}_{b}])$ is piecewise constant, eq.\eqref{eq:proof_of_eq_approx:combine_equivalenttopology_with_jacobian} reveals that $\mathbf{\mathcal{c}}_{l}$ is a piecewise linear function which passes through the origin in $\mathbb{R}^{m_{-1} + M}$. Furthermore, eq.\eqref{eq:proof_of_eq_approx:separatedforms}, \eqref{eq:proof_of_eq_approx:finalout} can be rewritten as,
		\begin{equation} \label{eq:proof_of_eq_approx:combine_CNN_with_jacobian}
			\mathbf{c}_{l}(\mathbf{x}_{n}) = \mathbf{J}_{l}(k[\hat{\mathbf{p}}_{-1}(\mathbf{x}_{n}); \mathbf{x}_{b}])[\hat{\mathbf{p}}_{-1}(\mathbf{x}_{n}); \mathbf{x}_{b}], 0 \le l \le L, k \in \mathbb{R}^{+},
		\end{equation}
		which proves the third equality in Eq.\eqref{eq:approx}. For any piecewise linear activation function $\sigma(x)$, the following property holds,
		\begin{equation}\label{eq:proof_of_eq_approx:piecewise_act}
			\sigma(x_0) = x_0 \times \eval{\frac{d \sigma(x)}{x}}_{x=x_0}, x_0 \in \mathbb{R}.
		\end{equation}
		This implies that the proof works for any piecewise linear activation $\sigma(x)$ (whose derivative is piecewise constant) as long as we apply $k=1$ and properly replace ${\mathbf{\Sigma}}_l$ in eq.\eqref{eq:proof_of_eq_approx:act_jacobian} according to the $\sigma(x)$ selected.
	\end{appendices}

	\begin{appendices}
		\section{Tables}
        \label{Appendix:Tables}
		\begin{table*}[!h]
			\centering
			\begin{threeparttable}
				\begin{adjustbox}{width=0.8\textwidth}
					\begin{tabular}{|c|c|c|c|}
						\hline
						Original Layer & Equivalent Layer & Parameters & Out-ch Feature Maps \\
						\hline
						Input ($\mathbf{x}$) & Input ($[\mathbf{x}; \mathbf{b}^{'}]$) & N/S & N/S \\
						\hline
						Conv0 ($\mathbf{w}_{0}$) & Conv0 ($\mathbf{w}_{0}^{'}$) & N/S & N/S \\
						\hline
						BN0 & B0 ($\mathbf{b}^{'}[\mathbf{b}_{0}^{'}]$) & N/S & N/S \\
						\hline
						\multicolumn{2}{|c|}{ReLU0} & N/S & N/S \\
						\hline
						Conv1 ($\mathbf{w}_{1}$) & Conv1 ($\mathbf{w}_{1}^{'}$) & $3 \times 3 \times 32 \times 32$ & $32 \times 32 \times 32 (= 32, 768)$ \\
						\hline
						BN1 & B1 ($\mathbf{b}^{'}[\mathbf{b}_{1}^{'}]$) & N/S & N/S \\
						\hline
						\multicolumn{2}{|c|}{ReLU1} & N/S & N/S \\
						\hline
						\multicolumn{2}{|c|}{Avg-pool-2} & N/S & N/S \\
						\hline
						Conv2 ($\mathbf{w}_{2}$) & Conv2 ($\mathbf{w}_{2}^{'}$) & $3 \times 3 \times 32 \times 64$ & $16 \times 16 \times 64 (= 16,384)$ \\
						\hline
						BN2 & B2 ($\mathbf{b}^{'}[\mathbf{b}_{2}^{'}]$) & N/S & N/S \\
						\hline
						\multicolumn{2}{|c|}{ReLU2} & N/S & N/S \\
						\hline
						Conv3 ($\mathbf{w}_{3}$) & Conv3 ($\mathbf{w}_{3}^{'}$) & $3 \times 3 \times 64 \times 64$ & $16 \times 16 \times 64 (= 16,384)$ \\
						\hline
						BN3 & B3 ($\mathbf{b}^{'}[\mathbf{b}_{3}^{'}]$) & N/S & N/S \\
						\hline
						\multicolumn{2}{|c|}{ReLU3} & N/S & N/S \\
						\hline
						\multicolumn{2}{|c|}{Avg-pool-2} & N/S & N/S \\
						\hline
						Conv4 ($\mathbf{w}_{4}$) & Conv4 ($\mathbf{w}_{4}^{'}$) & $3 \times 3 \times 64 \times 96$ & $8 \times 8 \times 96 (= 6,144)$ \\
						\hline
						BN4 & B4 ($\mathbf{b}^{'}[\mathbf{b}_{4}^{'}]$) & N/S & N/S \\
						\hline
						\multicolumn{2}{|c|}{ReLU4} & N/S & N/S \\
						\hline
						Conv5 ($\mathbf{w}_{5}$) & Conv5 ($\mathbf{w}_{5}^{'}$) & $3 \times 3 \times 96 \times 96$ & $8 \times 8 \times 96 (= 6,144)$ \\
						\hline
						BN5 & B5 ($\mathbf{b}^{'}[\mathbf{b}_{5}^{'}]$) & N/S & N/S \\
						\hline
						\multicolumn{2}{|c|}{ReLU5} & N/S & N/S \\
						\hline
						\multicolumn{2}{|c|}{global-pool (g\_p)} & N/S & N/S \\
						\hline
						\multicolumn{2}{|c|}{FC ($\mathbf{w}_{fc}$)} & $96 \times 10$ & $10$ \\
						\hline
						\multicolumn{2}{|c|}{ReLU6} & N/S & N/S \\
						\hline
					\end{tabular}
				\end{adjustbox}
				\centering
				\caption{Parameters of VGG7.}
				\begin{tablenotes}
					\item $\mathbf{w}^{'}_{i}$ and $\mathbf{b}^{'}_{i}$ are computed using Eq.\eqref{eq:BN_reduced};
					\item $\mathbf{b}^{'}$ is a sequentical concatenation of $\mathbf{b}_{0}^{'} \sim \mathbf{b}_{5}^{'}$;
					\item Avg-pool-2: average pooling with window size of $3$ and stride size of $2$; 
					\item N/S: it is not necessary for our method.
				\end{tablenotes}
				\label{table:params:vgg7}
			\end{threeparttable}
		\end{table*}

		\begin{table*}[!h]
			\centering
			\begin{threeparttable}
				\begin{adjustbox}{width=0.87\textwidth}
					\begin{tabular}{|c|c|c|c|c|c|}
						\hline
						Equivalent Layer & $RM_4$ & $RM_3$ & $RM_2$ & $RM_1$ & $RM_0$ \\
						\hline
						Conv0 & N/S & N/S & N/S & N/S & N/S \\
						\hline
						Conv1 & $d_{in} \times 32 \times 32 \times 32 \times 32$ & $d_{in} \times 32 \times 32$ & $d_{in} \times 32 \times 32 \times 32$ & $d_{in} \times 32$ & N/S \\
						\hline
						Conv2 & $d_{in} \times 16 \times 16 \times 32 \times 64$ & $d_{in} \times 32 \times 64$ & $d_{in} \times 16 \times 16 \times 64$ & $d_{in} \times 64$ & N/S \\
						\hline
						Conv3 & $d_{in} \times 16 \times 16 \times 64 \times 64$ & $d_{in} \times 64 \times 64$ & $d_{in} \times 16 \times 16 \times 64$ & $d_{in} \times 64$ & N/S \\
						\hline
						Conv4 & $d_{in} \times 8 \times 8 \times 64 \times 96$ & $d_{in} \times 64 \times 96$ & $d_{in} \times 8 \times 8 \times 96$ & $d_{in} \times 96$ & N/S \\
						\hline
						Conv5 & $d_{in} \times 8 \times 8 \times 96 \times 96$ & $d_{in} \times 96 \times 96$ & $d_{in} \times 8 \times 8 \times 96$ & $d_{in} \times 96$ & N/S \\
						\hline
						FC & N/S & N/S & N/S & N/S & $d_{in} \times 10$ \\
						\hline
					\end{tabular}
				\end{adjustbox}
				\centering
				\caption{Dimensions of $\mathbf{H}^{Adj}(\frac{[\mathbf{x}; \mathbf{x}_{b}]}{8})$ with different RMs on VGG7.}
				\begin{tablenotes}
					\item $d_{in} = (32 \times 32 \times 3 + 384)$ where $384 = (32 + 64 + 96) \times 2$;
					\item N/S: it is not necessary for our method.
				\end{tablenotes}
				\label{table:dims_vgg7}
			\end{threeparttable}
		\end{table*}

		\begin{table*}[!h]
			\centering
			\begin{threeparttable}
				\begin{adjustbox}{width=0.7\textwidth}
					\begin{tabular}{|c|c|c|c|c|}
						\hline
						Block (shortcut) & Original Layer & Equivalent Layer & Parameters & Out-ch Feature Maps \\
						\hline
						& Input ($\mathbf{x}$) & Input ($[\mathbf{x}; \mathbf{b}^{'}]$) & N/S & N/S \\
						\hline
						& Conv0 ($\mathbf{w}_{0}$) & Conv0 ($\mathbf{w}_{0}^{'}$) & N/S & N/S \\
						\hline
						& BN0 & B0 ($\mathbf{b}^{'}[\mathbf{b}_{0}^{'}]$) & N/S & N/S \\
						\hline
						& \multicolumn{2}{|c|}{Leaky\_ReLU0} & N/S & N/S \\
						\hline
						\multirow{2}{*}{\shortstack{Residual 0 \\ (identity)}}
						& Conv1 ($\mathbf{w}_{1}$) & Conv1 ($\mathbf{w}_{1}^{'}$) & $3 \times 3 \times 32 \times 32$ & $32 \times 32 \times 32 (32,768)$ \\
						\cline{2-5}
						& BN1 & B1 ($\mathbf{b}^{'}[\mathbf{b}_{1}^{'}]$) & N/S & N/S \\
						\cline{2-5}
						& \multicolumn{2}{|c|}{Leaky\_ReLU1} & N/A & N/S \\
						\cline{2-5}
						& Conv2 ($\mathbf{w}_{2}$) & Conv2 ($\mathbf{w}_{2}^{'}$) & $3 \times 3 \times 32 \times 32$ & $32 \times 32 \times 32 (32,768)$ \\
						\cline{2-5}
						& BN2 & B2 ($\mathbf{b}^{'}[\mathbf{b}_{2}^{'}]$) & N/S & N/S \\
						\hline
						& \multicolumn{2}{|c|}{Leaky\_ReLU2} & N/S & N/S \\
						\hline
						\multirow{2}{*}{\shortstack{Residual 1 \\ (identity)}}
						& Conv3 ($\mathbf{w}_{3}$) & Conv3 ($\mathbf{w}_{3}^{'}$) & $3 \times 3 \times 32 \times 32$ & $32 \times 32 \times 32 (32,768)$ \\
						\cline{2-5}
						& BN3 & B3 ($\mathbf{b}^{'}[\mathbf{b}_{3}^{'}]$) & N/S & N/S \\
						\cline{2-5}
						& \multicolumn{2}{|c|}{Leaky\_ReLU3} & N/S & N/S \\
						\cline{2-5}
						& Conv4 ($\mathbf{w}_{4}$) & Conv4 ($\mathbf{w}_{4}^{'}$) & $3 \times 3 \times 32 \times 32$ & $32 \times 32 \times 32 (32,768)$ \\
						\cline{2-5}
						& BN4 & B4 ($\mathbf{b}^{'}[\mathbf{b}_{4}^{'}]$) & N/S & N/S \\
						\hline
						& \multicolumn{2}{|c|}{Leaky\_ReLU4} & N/S & N/S \\
						\hline
						\multirow{2}{*}{\shortstack{Residual 2 \\ (identity)}}
						& Conv5 ($\mathbf{w}_{5}$) & Conv5 ($\mathbf{w}_{5}^{'}$) & $3 \times 3 \times 32 \times 32$ & $32 \times 32 \times 32 (32,768)$ \\
						\cline{2-5}
						& BN5 & B5 ($\mathbf{b}^{'}[\mathbf{b}_{5}^{'}]$) & N/S & N/S \\
						\cline{2-5}
						& \multicolumn{2}{|c|}{Leaky\_ReLU5} & N/S & N/S \\
						\cline{2-5}
						& Conv6 ($\mathbf{w}_{6}$) & Conv6 ($\mathbf{w}_{6}^{'}$) & $3 \times 3 \times 32 \times 32$ & $32 \times 32 \times 32 (32,768)$ \\
						\cline{2-5}
						& BN6 & B6 ($\mathbf{b}^{'}[\mathbf{b}_{6}^{'}]$) & N/S & N/S \\
						\hline
						& \multicolumn{2}{|c|}{Leaky\_ReLU6} & N/S & N/S \\
						\hline
						\multirow{2}{*}{\shortstack{Residual 3 \\ (avg-pool+pad)}}
						& Conv7 ($\mathbf{w}_{7}$, $s=2$) & Conv7 ($\mathbf{w}_{7}^{'}$, $s=2$) & $3 \times 3 \times 32 \times 64$ & $16 \times 16 \times 64 (16,384)$ \\
						\cline{2-5}
						& BN7 & B7 ($\mathbf{b}^{'}[\mathbf{b}_{7}^{'}]$) & N/S & N/S \\
						\cline{2-5}
						& \multicolumn{2}{|c|}{Leaky\_ReLU7} & N/S & N/S \\
						\cline{2-5}
						& Conv8 ($\mathbf{w}_{8}$) & Conv8 ($\mathbf{w}_{8}^{'}$) & $3 \times 3 \times 64 \times 64$ & $16 \times 16 \times 64 (16,384)$ \\
						\cline{2-5}
						& BN8 & B8 ($\mathbf{b}^{'}[\mathbf{b}_{8}^{'}]$) & N/S & N/S \\
						\hline
						& \multicolumn{2}{|c|}{Leaky\_ReLU8} & N/S & N/S \\
						\hline
						\multirow{2}{*}{\shortstack{Residual 4 \\ (identity)}}
						& Conv9 ($\mathbf{w}_{9}$) & Conv9 ($\mathbf{w}_{9}^{'}$) & $3 \times 3 \times 64 \times 64$ & $16 \times 16 \times 64 (16,384)$ \\
						\cline{2-5}
						& BN9 & B9 ($\mathbf{b}^{'}[\mathbf{b}_{9}^{'}]$) & N/S & N/S \\
						\cline{2-5}
						& \multicolumn{2}{|c|}{Leaky\_ReLU9} & N/S & N/S \\
						\cline{2-5}
						& Conv10 ($\mathbf{w}_{10}$) & Conv10 ($\mathbf{w}_{10}^{'}$) & $3 \times 3 \times 64 \times 64$ & $16 \times 16 \times 64 (16,384)$ \\
						\cline{2-5}
						& BN10 & B10 ($\mathbf{b}^{'}[\mathbf{b}_{10}^{'}]$) & N/S & N/S \\
						\hline
						& \multicolumn{2}{|c|}{Leaky\_ReLU10} & N/S & N/S \\
						\hline
						\multirow{2}{*}{\shortstack{Residual 5 \\ (identity)}}
						& Conv11 ($\mathbf{w}_{11}$) & Conv11 ($\mathbf{w}_{11}^{'}$) & $3 \times 3 \times 64 \times 64$ & $16 \times 16 \times 64 (16,384)$ \\
						\cline{2-5}
						& BN11 & B11 ($\mathbf{b}^{'}[\mathbf{b}_{11}^{'}]$) & N/S & N/S \\
						\cline{2-5}
						& \multicolumn{2}{|c|}{Leaky\_ReLU11} & N/S & N/S \\
						\cline{2-5}
						& Conv12 ($\mathbf{w}_{12}$) & Conv12 ($\mathbf{w}_{12}^{'}$) & $3 \times 3 \times 64 \times 64$ & $16 \times 16 \times 64 (16,384)$ \\
						\cline{2-5}
						& BN12 & B12 ($\mathbf{b}^{'}[\mathbf{b}_{12}^{'}]$) & N/S & N/S \\
						\hline
						& \multicolumn{2}{|c|}{Leaky\_ReLU12} & N/S & N/S \\
						\hline
						\multirow{2}{*}{\shortstack{Residual 6 \\ (avg-pool+pad)}}
						& Conv13 ($\mathbf{w}_{13}$, $s=2$) & Conv13 ($\mathbf{w}_{13}^{'}$, $s=2$) & $3 \times 3 \times 64 \times 96$ & $8 \times 8 \times 96 (6,144)$ \\
						\cline{2-5}
						& BN13 & B13 ($\mathbf{b}^{'}[\mathbf{b}_{13}^{'}]$) & N/S & N/S \\
						\cline{2-5}
						& \multicolumn{2}{|c|}{Leaky\_ReLU13} & N/A & N/S \\
						\cline{2-5}
						& Conv14 ($\mathbf{w}_{14}$) & Conv14 ($\mathbf{w}_{14}^{'}$) & $3 \times 3 \times 96 \times 96$ & $8 \times 8 \times 96 (6,144)$ \\
						\cline{2-5}
						& BN14 & B14 ($\mathbf{b}^{'}[\mathbf{b}_{14}^{'}]$) & N/S & N/S \\
						\hline
						& \multicolumn{2}{|c|}{Leaky\_ReLU14} & N/S & N/S \\
						\hline
						\multirow{2}{*}{\shortstack{Residual 7 \\ (identity)}}
						& Conv15 ($\mathbf{w}_{15}$) & Conv15 ($\mathbf{w}_{15}^{'}$) & $3 \times 3 \times 96 \times 96$ & $8 \times 8 \times 96 (6,144)$ \\
						\cline{2-5}
						& BN15 & B15 ($\mathbf{b}^{'}[\mathbf{b}_{15}^{'}]$) & N/S & N/S \\
						\cline{2-5}
						& \multicolumn{2}{|c|}{Leaky\_ReLU15} & N/S & N/S \\
						\cline{2-5}
						& Conv16 ($\mathbf{w}_{16}$) & Conv16 ($\mathbf{w}_{16}^{'}$) & $3 \times 3 \times 96 \times 96$ & $8 \times 8 \times 96 (6,144)$ \\
						\cline{2-5}
						& BN16 & B16 ($\mathbf{b}^{'}[\mathbf{b}_{16}^{'}]$) & N/S & N/S \\
						\hline
						& \multicolumn{2}{|c|}{Leaky\_ReLU16} & N/S & N/S \\
						\hline
						\multirow{2}{*}{\shortstack{Residual 8 \\ (identity)}}
						& Conv17 ($\mathbf{w}_{17}$) & Conv17 ($\mathbf{w}_{17}^{'}$) & $3 \times 3 \times 96 \times 96$ & $8 \times 8 \times 96 (6,144)$ \\
						\cline{2-5}
						& BN17 & B17 ($\mathbf{b}^{'}[\mathbf{b}_{17}^{'}]$) & N/S & N/S \\
						\cline{2-5}
						& \multicolumn{2}{|c|}{Leaky\_ReLU17} & N/S & N/S \\
						\cline{2-5}
						& Conv18 ($\mathbf{w}_{18}$) & Conv18 ($\mathbf{w}_{18}^{'}$) & $3 \times 3 \times 96 \times 96$ & $8 \times 8 \times 96 (6,144)$ \\
						\cline{2-5}
						& BN18 & B18 ($\mathbf{b}^{'}[\mathbf{b}_{18}^{'}]$) & N/S & N/S \\
						\hline
						& \multicolumn{2}{|c|}{Leaky\_ReLU18} & N/S & N/S \\
						\hline
						& \multicolumn{2}{|c|}{g\_p} & N/S & N/S \\
						\hline
						& \multicolumn{2}{|c|}{FC ($\mathbf{w}_{fc}$)} & $96 \times 10$ & $10$ \\
						\hline
						& \multicolumn{2}{|c|}{Leaky\_ReLU19} & N/S & N/S \\
						\hline
					\end{tabular}
				\end{adjustbox}
				\centering
				\caption{Parameters of ResNet20 (refer to table \ref{table:params:vgg7}).}
				\begin{tablenotes}
					\item $\mathbf{b}^{'}$ is a sequentical concatenation of $\mathbf{b}_{0}^{'} \sim \mathbf{b}_{18}^{'}$;
					\item avg-pool: average pooling with window size of $1$ and stride size of $2$;
					\item pad: padding zero channels to match the quantity of out channels for summation.
				\end{tablenotes}
				\label{table:params:resnet20}
			\end{threeparttable}
		\end{table*}

		\begin{table*}[!h]
			\centering
			\begin{threeparttable}
				\begin{adjustbox}{width=0.5\height}
					\begin{tabular}{|c|c|c|c|c|}
						\hline
						Block (shortcut) & Original Layer & Equivalent Layer & Parameters & Out-channel Feature Maps \\
						\hline
						& Input ($\mathbf{x}$) & Input ($[\mathbf{x}; \mathbf{b}^{'}]$) & N/S & N/S \\
						\hline
						& Conv0 ($\mathbf{w}_{0}$) & Conv0 ($\mathbf{w}_{0}$) & N/S & N/S \\
						\hline
						& B0 ($b_0$) & B0 ($\mathbf{b}^{'}[b_{0}]$) & N/S & N/S \\
						\hline
						& \multicolumn{2}{|c|}{ReLU0} & N/S & N/S \\
						\hline
						\multirow{2}{*}{\shortstack{Residual 0 \\ (identity)}}
						& B1 ($b_1$) & B1 ($\mathbf{b}^{'}[b_1]$) & N/S & N/S \\
						\cline{2-5}
						& Conv1 ($\mathbf{w}_{1}$) & Conv1 ($\mathbf{w}_{1}$) & $3 \times 3 \times 32 \times 32$ & $32 \times 32 \times 32 (32,768)$ \\
						\cline{2-5}
						& B2 ($b_2$) & B2 ($\mathbf{b}^{'}[b_2]$) & N/S & N/S \\
						\cline{2-5}
						& \multicolumn{2}{|c|}{ReLU1} & N/S & N/S \\
						\cline{2-5}
						& B3 ($b_3$) & B3 ($\mathbf{b}^{'}[b_3]$) & N/S & N/S \\
						\cline{2-5}
						& Conv2 ($\mathbf{w}_{2}$) & \multirow{2}{*}{\shortstack{Conv2 \\ ($\mathbf{w}^{'}_{2} = \mathbf{w}_{2} \times M_2 $)}} & \multirow{2}{*}{\shortstack{$3 \times 3 \times 32 \times 32$}} & \multirow{2}{*}{\shortstack{$32 \times 32 \times 32 (32,768)$}} \\
						\cline{2-2}
						& M2 ($M_2$) & & & \\
						\cline{2-5}
						& B4 ($b_4$) & B4 ($\mathbf{b}^{'}[b_4]$) & N/S & N/S \\
						\hline
						& \multicolumn{2}{|c|}{ReLU2} & N/S & N/S \\
						\hline
						\multirow{2}{*}{\shortstack{Residual 1 \\ (identity)}}
						& B5 ($b_5$) & B5 ($\mathbf{b}^{'}[b_5]$) & N/S & N/S \\
						\cline{2-5}
						& Conv3 ($\mathbf{w}_{3}$) & Conv3 ($\mathbf{w}_{3}$) & $3 \times 3 \times 32 \times 32$ & $32 \times 32 \times 32 (32,768)$ \\
						\cline{2-5}
						& B6 ($b_6$) & B6 ($\mathbf{b}^{'}[b_6]$) & N/S & N/S \\
						\cline{2-5}
						& \multicolumn{2}{|c|}{ReLU3} & N/S & N/S \\
						\cline{2-5}
						& B7 ($b_7$) & B7 ($\mathbf{b}^{'}[b_7]$) & N/S & N/S \\
						\cline{2-5}
						& Conv4 ($\mathbf{w}_{4}$) & \multirow{2}{*}{\shortstack{Conv4 \\ ($\mathbf{w}^{'}_{4} = \mathbf{w}_{4} \times M_4 $)}} & \multirow{2}{*}{\shortstack{$3 \times 3 \times 32 \times 32$}} & \multirow{2}{*}{\shortstack{$32 \times 32 \times 32 (32,768)$}} \\
						\cline{2-2}
						& M4 ($M_4$) & & & \\
						\cline{2-5}
						& B8 ($b_8$) & B8 ($\mathbf{b}^{'}[b_8]$) & N/S & N/S \\
						\hline
						& \multicolumn{2}{|c|}{ReLU4} & N/S & N/S \\
						\hline
						\multirow{2}{*}{\shortstack{Residual 2 \\ (identity)}}
						& B9 ($b_9$) & B9 ($\mathbf{b}^{'}[b_9]$) & N/S & N/S \\
						\cline{2-5}
						& Conv5 ($\mathbf{w}_{5}$) & Conv5 ($\mathbf{w}_{5}$) & $3 \times 3 \times 32 \times 32$ & $32 \times 32 \times 32 (32,768)$ \\
						\cline{2-5}
						& B10 ($b_{10}$) & B10 ($\mathbf{b}^{'}[b_{10}]$) & N/S & N/S \\
						\cline{2-5}
						& \multicolumn{2}{|c|}{ReLU5} & N/S & N/S \\
						\cline{2-5}
						& B11 ($b_{11}$) & B11 ($\mathbf{b}^{'}[b_{11}]$) & N/S & N/S \\
						\cline{2-5}
						& Conv6 ($\mathbf{w}_{6}$) & \multirow{2}{*}{\shortstack{Conv6 \\ ($\mathbf{w}^{'}_{6} = \mathbf{w}_{6} \times M_6 $)}} & \multirow{2}{*}{\shortstack{$3 \times 3 \times 32 \times 32$}} & \multirow{2}{*}{\shortstack{$32 \times 32 \times 32 (32,768)$}} \\
						\cline{2-2}
						& M6 ($M_6$) & & & \\
						\cline{2-5}
						& B12 ($b_{12}$) & B12 ($\mathbf{b}^{'}[b_{12}]$) & N/S & N/S \\
						\hline
						& \multicolumn{2}{|c|}{ReLU6} & N/S & N/S \\
						\hline
						\multirow{2}{*}{\shortstack{Residual 3 \\ (avg-pool+pad)}}
						& B13 ($b_{13}$) & B13 ($\mathbf{b}^{'}[b_{13}]$) & N/S & N/S \\
						\cline{2-5}
						& Conv7 ($\mathbf{w}_{7}, s=2$) & Conv7 ($\mathbf{w}_{7}, s=2$) & $3 \times 3 \times 32 \times 64$ & $16 \times 16 \times 64 (16,384)$ \\
						\cline{2-5}
						& B14 ($b_{14}$) & B14 ($\mathbf{b}^{'}[b_{14}]$) & N/S & N/S \\
						\cline{2-5}
						& \multicolumn{2}{|c|}{ReLU7} & N/S & N/S \\
						\cline{2-5}
						& B15 ($b_{15}$) & B15 ($\mathbf{b}^{'}[b_{15}]$) & N/S & N/S \\
						\cline{2-5}
						& Conv8 ($\mathbf{w}_{8}$) & \multirow{2}{*}{\shortstack{Conv8 \\ ($\mathbf{w}^{'}_{8} = \mathbf{w}_{8} \times M_8 $)}} & \multirow{2}{*}{\shortstack{$3 \times 3 \times 64 \times 64$}} & \multirow{2}{*}{\shortstack{$16 \times 16 \times 64 (16,384)$}} \\
						\cline{2-2}
						& M8 ($M_8$) & & & \\
						\cline{2-5}
						& B16 ($b_{16}$) & B16 ($\mathbf{b}^{'}[b_{16}]$) & N/S & N/S \\
						\hline
						& \multicolumn{2}{|c|}{ReLU8} & N/S & N/S \\
						\hline
						\multirow{2}{*}{\shortstack{Residual 4 \\ (identity)}}
						& B17 ($b_{17}$) & B17 ($\mathbf{b}^{'}[b_{17}]$) & N/S & N/S \\
						\cline{2-5}
						& Conv9 ($\mathbf{w}_{9}$) & Conv9 ($\mathbf{w}_{9}$) & $3 \times 3 \times 64 \times 64$ & $16 \times 16 \times 64 (16,384)$ \\
						\cline{2-5}
						& B18 ($b_{18}$) & B18 ($\mathbf{b}^{'}[b_{18}]$) & N/S & N/S \\
						\cline{2-5}
						& \multicolumn{2}{|c|}{ReLU9} & N/S & N/S \\
						\cline{2-5}
						& B19 ($b_{19}$) & B19 ($\mathbf{b}^{'}[b_{19}]$) & N/S & N/S \\
						\cline{2-5}
						& Conv10 ($\mathbf{w}_{10}$) & \multirow{2}{*}{\shortstack{Conv10 \\ ($\mathbf{w}^{'}_{10} = \mathbf{w}_{10} \times M_{10} $)}} & \multirow{2}{*}{\shortstack{$3 \times 3 \times 64 \times 64$}} & \multirow{2}{*}{\shortstack{$16 \times 16 \times 64 (16,384)$}} \\
						\cline{2-2}
						& M10 ($M_{10}$) & & & \\
						\cline{2-5}
						& B20 ($b_{20}$) & B20 ($\mathbf{b}^{'}[b_{20}]$) & N/S & N/S \\
						\hline
						& \multicolumn{2}{|c|}{ReLU10} & N/S & N/S \\
						\hline
						\multirow{2}{*}{\shortstack{Residual 5 \\ (identity)}}
						& B21 ($b_{21}$) & B21 ($\mathbf{b}^{'}[b_{21}]$) & N/S & N/S \\
						\cline{2-5}
						& Conv11 ($\mathbf{w}_{11}$) & Conv11 ($\mathbf{w}_{11}$) & $3 \times 3 \times 64 \times 64$ & $16 \times 16 \times 64 (16,384)$ \\
						\cline{2-5}
						& B22 ($b_{22}$) & B22 ($\mathbf{b}^{'}[b_{22}]$) & N/S & N/S \\
						\cline{2-5}
						& \multicolumn{2}{|c|}{ReLU11} & N/S & N/S \\
						\cline{2-5}
						& B23 ($b_{23}$) & B23 ($\mathbf{b}^{'}[b_{23}]$) & N/S & N/S \\
						\cline{2-5}
						& Conv12 ($\mathbf{w}_{12}$) & \multirow{2}{*}{\shortstack{Conv12 \\ ($\mathbf{w}^{'}_{12} = \mathbf{w}_{12} \times M_{12} $)}} & \multirow{2}{*}{\shortstack{$3 \times 3 \times 64 \times 64$}} & \multirow{2}{*}{\shortstack{$16 \times 16 \times 64 (16,384)$}} \\
						\cline{2-2}
						& M12 ($M_{12}$) & & & \\
						\cline{2-5}
						& B24 ($b_{24}$) & B24 ($\mathbf{b}^{'}[b_{24}]$) & N/S & N/S \\
						\hline
						& \multicolumn{2}{|c|}{ReLU12} & N/S & N/S \\
						\hline
						\multirow{2}{*}{\shortstack{Residual 6 \\ (avg-pool+pad)}}
						& B25 ($b_{25}$) & B25 ($\mathbf{b}^{'}[b_{25}]$) & N/S & N/S \\
						\cline{2-5}
						& Conv13 ($\mathbf{w}_{13}, s=2$) & Conv13 ($\mathbf{w}_{13}, s=2$) & $3 \times 3 \times 64 \times 96$ & $8 \times 8 \times 96 (6,144)$ \\
						\cline{2-5}
						& B26 ($b_{26}$) & B26 ($\mathbf{b}^{'}[b_{26}]$) & N/S & N/S \\
						\cline{2-5}
						& \multicolumn{2}{|c|}{ReLU13} & N/S & N/S \\
						\cline{2-5}
						& B27 ($b_{27}$) & B27 ($\mathbf{b}^{'}[b_{27}]$) & N/S & N/S \\
						\cline{2-5}
						& Conv14 ($\mathbf{w}_{14}$) & \multirow{2}{*}{\shortstack{Conv14 \\ ($\mathbf{w}^{'}_{14} = \mathbf{w}_{14} \times M_{14} $)}} & \multirow{2}{*}{\shortstack{$3 \times 3 \times 96 \times 96$}} & \multirow{2}{*}{\shortstack{$8 \times 8 \times 96 (6,144)$}} \\
						\cline{2-2}
						& M14 ($M_{14}$) & & & \\
						\cline{2-5}
						& B28 ($b_{28}$) & B28 ($\mathbf{b}^{'}[b_{28}]$) & N/S & N/S \\
						\hline
						& \multicolumn{2}{|c|}{ReLU14} & N/S & N/S \\
						\hline
						\multirow{2}{*}{\shortstack{Residual 7 \\ (identity)}}
						& B29 ($b_{29}$) & B29 ($\mathbf{b}^{'}[b_{29}]$) & N/S & N/S \\
						\cline{2-5}
						& Conv15 ($\mathbf{w}_{15}$) & Conv15 ($\mathbf{w}_{15}$) & $3 \times 3 \times 96 \times 96$ & $8 \times 8 \times 96 (6,144)$ \\
						\cline{2-5}
						& B30 ($b_{30}$) & B30 ($\mathbf{b}^{'}[b_{30}]$) & N/S & N/S \\
						\cline{2-5}
						& \multicolumn{2}{|c|}{ReLU15} & N/S & N/S \\
						\cline{2-5}
						& B31 ($b_{31}$) & B31 ($\mathbf{b}^{'}[b_{31}]$) & N/S & N/S \\
						\cline{2-5}
						& Conv16 ($\mathbf{w}_{16}$) & \multirow{2}{*}{\shortstack{Conv16 \\ ($\mathbf{w}^{'}_{16} = \mathbf{w}_{16} \times M_{16} $)}} & \multirow{2}{*}{\shortstack{$3 \times 3 \times 96 \times 96$}} & \multirow{2}{*}{\shortstack{$8 \times 8 \times 96 (6,144)$}} \\
						\cline{2-2}
						& M16 ($M_{16}$) & & & \\
						\cline{2-5}
						& B32 ($b_{32}$) & B32 ($\mathbf{b}^{'}[b_{32}]$) & N/S & N/S \\
						\hline
						& \multicolumn{2}{|c|}{ReLU16} & N/S & N/S \\
						\hline
						\multirow{2}{*}{\shortstack{Residual 8 \\ (identity)}}
						& B33 ($b_{33}$) & B33 ($\mathbf{b}^{'}[b_{33}]$) & N/S & N/S \\
						\cline{2-5}
						& Conv16 ($\mathbf{w}_{16}$) & Conv16 ($\mathbf{w}_{16}$) & $3 \times 3 \times 96 \times 96$ & $8 \times 8 \times 96 (6,144)$ \\
						\cline{2-5}
						& B34 ($b_{34}$) & B34 ($\mathbf{b}^{'}[b_{34}]$) & N/S & N/S \\
						\cline{2-5}
						& \multicolumn{2}{|c|}{ReLU17} & N/S & N/S \\
						\cline{2-5}
						& B35 ($b_{35}$) & B35 ($\mathbf{b}^{'}[b_{35}]$) & N/S & N/S \\
						\cline{2-5}
						& Conv18 ($\mathbf{w}_{18}$) & \multirow{2}{*}{\shortstack{Conv18 \\ ($\mathbf{w}^{'}_{18} = \mathbf{w}_{18} \times M_{18} $)}} & \multirow{2}{*}{\shortstack{$3 \times 3 \times 96 \times 96$}} & \multirow{2}{*}{\shortstack{$8 \times 8 \times 96 (6,144)$}} \\
						\cline{2-2}
						& M18 ($M_{18}$) & & & \\
						\cline{2-5}
						& B36 ($b_{36}$) & B36 ($\mathbf{b}^{'}[b_{36}]$) & N/S & N/S \\
						\hline
						& \multicolumn{2}{|c|}{ReLU18} & N/S & N/S \\
						\hline
						& \multicolumn{2}{|c|}{g\_p} & N/S & N/S \\
						\hline
						& \multicolumn{2}{|c|}{FC ($\mathbf{w}_{fc}$)} & $96 \times 10$ & $10$ \\
						\hline
						& \multicolumn{2}{|c|}{ReLU19} & N/S & N/S \\
						\hline
					\end{tabular}
				\end{adjustbox}
				\centering
				\caption{Parameters in ResNet20-Fixup (refer to table \ref{table:params:resnet20})}
				\begin{tablenotes}
					\item $M_i$ denotes a multiplier \cite{zhang2019fixup};
					\item $\mathbf{b}^{'}$ is a sequentical concatenation of $b_0 \sim b_{36}$.
				\end{tablenotes}
				\label{table:params:resnet20_fixup}
			\end{threeparttable}
		\end{table*}

		\begin{table*}[!h]
			\centering
			\begin{threeparttable}
				\begin{adjustbox}{width=0.87\textwidth}
					\begin{tabular}{|c|c|c|c|c|c|}
						\hline
						Equivalent Layer & $RM_4$ & $RM_3$ & $RM_2$ & $RM_1$ & $RM_0$ \\
						\hline
						Conv0 & N/S & N/S & N/S & N/S & N/S \\
						\hline
						Conv1 & $d_{in} \times 32 \times 32 \times 32 \times 32$ & $d_{in} \times 32 \times 32$ & $d_{in} \times 32 \times 32 \times 32$ & $d_{in} \times 32$ & N/S \\
						\hline
						Conv2 & $d_{in} \times 32 \times 32 \times 32 \times 32$ & $d_{in} \times 32 \times 32$ & $d_{in} \times 32 \times 32 \times 32$ & $d_{in} \times 32$ & N/S \\
						\hline
						Conv3 & $d_{in} \times 32 \times 32 \times 32 \times 32$ & $d_{in} \times 32 \times 32$ & $d_{in} \times 32 \times 32 \times 32$ & $d_{in} \times 32$ & N/S \\
						\hline
						Conv4 & $d_{in} \times 32 \times 32 \times 32 \times 32$ & $d_{in} \times 32 \times 32$ & $d_{in} \times 32 \times 32 \times 32$ & $d_{in} \times 32$ & N/S \\
						\hline
						Conv5 & $d_{in} \times 32 \times 32 \times 32 \times 32$ & $d_{in} \times 32 \times 32$ & $d_{in} \times 32 \times 32 \times 32$ & $d_{in} \times 32$ & N/S \\
						\hline
						Conv6 & $d_{in} \times 32 \times 32 \times 32 \times 32$ & $d_{in} \times 32 \times 32$ & $d_{in} \times 32 \times 32 \times 32$ & $d_{in} \times 32$ & N/S \\
						\hline
						Conv7 & $d_{in} \times 16 \times 16 \times 32 \times 64$ & $d_{in} \times 32 \times 64$ & $d_{in} \times 16 \times 16 \times 64$ & $d_{in} \times 64$ & N/S \\
						\hline
						Conv8 & $d_{in} \times 16 \times 16 \times 64 \times 64$ & $d_{in} \times 64 \times 64$ & $d_{in} \times 16 \times 16 \times 64$ & $d_{in} \times 64$ & N/S \\
						\hline
						Conv9 & $d_{in} \times 16 \times 16 \times 64 \times 64$ & $d_{in} \times 64 \times 64$ & $d_{in} \times 16 \times 16 \times 64$ & $d_{in} \times 64$ & N/S \\
						\hline
						Conv10 & $d_{in} \times 16 \times 16 \times 64 \times 64$ & $d_{in} \times 64 \times 64$ & $d_{in} \times 16 \times 16 \times 64$ & $d_{in} \times 64$ & N/S \\
						\hline
						Conv11 & $d_{in} \times 16 \times 16 \times 64 \times 64$ & $d_{in} \times 64 \times 64$ & $d_{in} \times 16 \times 16 \times 64$ & $d_{in} \times 64$ & N/S \\
						\hline
						Conv12 & $d_{in} \times 16 \times 16 \times 64 \times 64$ & $d_{in} \times 64 \times 64$ & $d_{in} \times 16 \times 16 \times 64$ & $d_{in} \times 64$ & N/S \\
						\hline
						Conv13 & $d_{in} \times 8 \times 8 \times 64 \times 96$ & $d_{in} \times 64 \times 96$ & $d_{in} \times 8 \times 8 \times 96$ & $d_{in} \times 96$ & N/S \\
						\hline
						Conv14 & $d_{in} \times 8 \times 8 \times 96 \times 96$ & $d_{in} \times 96 \times 96$ & $d_{in} \times 8 \times 8 \times 96$ & $d_{in} \times 96$ & N/S \\
						\hline
						Conv15 & $d_{in} \times 8 \times 8 \times 96 \times 96$ & $d_{in} \times 96 \times 96$ & $d_{in} \times 8 \times 8 \times 96$ & $d_{in} \times 96$ & N/S \\
						\hline
						Conv16 & $d_{in} \times 8 \times 8 \times 96 \times 96$ & $d_{in} \times 96 \times 96$ & $d_{in} \times 8 \times 8 \times 96$ & $d_{in} \times 96$ & N/S \\
						\hline
						Conv17 & $d_{in} \times 8 \times 8 \times 96 \times 96$ & $d_{in} \times 96 \times 96$ & $d_{in} \times 8 \times 8 \times 96$ & $d_{in} \times 96$ & N/S \\
						\hline
						Conv18 & $d_{in} \times 8 \times 8 \times 96 \times 96$ & $d_{in} \times 96 \times 96$ & $d_{in} \times 8 \times 8 \times 96$ & $d_{in} \times 96$ & N/S \\
						\hline
						FC & N/S & N/S & N/S & N/S & $d_{in} \times 10$ \\
						\hline
					\end{tabular}
				\end{adjustbox}
				\centering
				\caption{Dimensions of $\mathbf{H}^{Adj}(\frac{[\mathbf{x}; \mathbf{x}_{b}]}{8})$ with different RMs on ResNet20/ResNet20-Fixup.}
				\begin{tablenotes}
					\item For ResNet20: $d_{in} = (32 \times 32 \times 3 + 1,152)$ where $1,152 = (32 + 64 + 96) \times 6$;
					\item For ResNet20-Fixup: $d_{in} = (32 \times 32 \times 3 + 37)$.
				\end{tablenotes}
				\label{table:dims_ResNet20_Fixup}
			\end{threeparttable}
		\end{table*}
	\end{appendices}

        \begin{appendices}
        \section{Pictures}
        \label{Appendix:Pictures}
\begin{figure}[!h]
		\centering
		\begin{tabular}{@{}c@{}c@{}c@{}c@{}}
		  \includegraphics[width=0.22\textwidth]{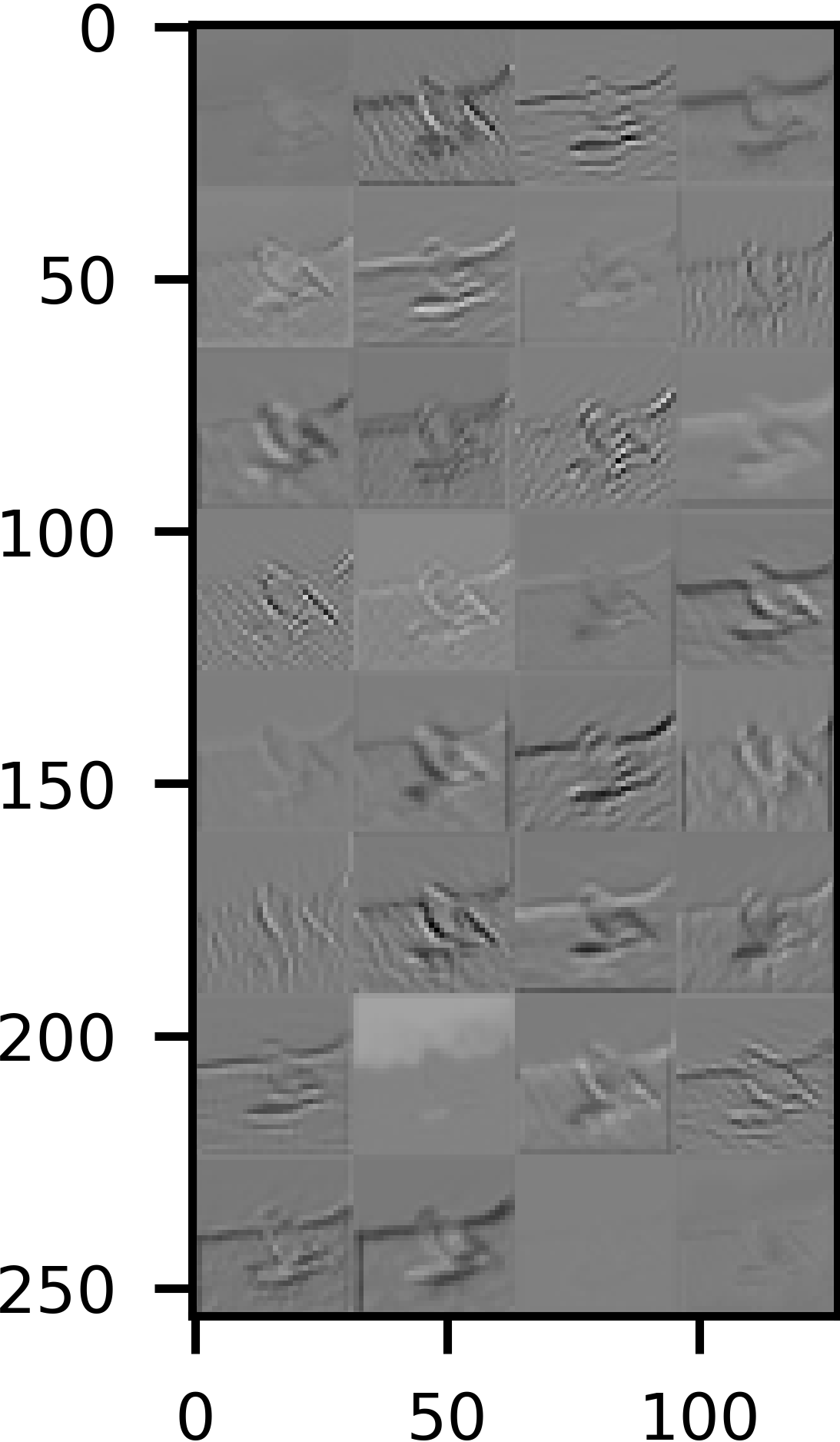} &
            \quad \includegraphics[width=0.22\textwidth]{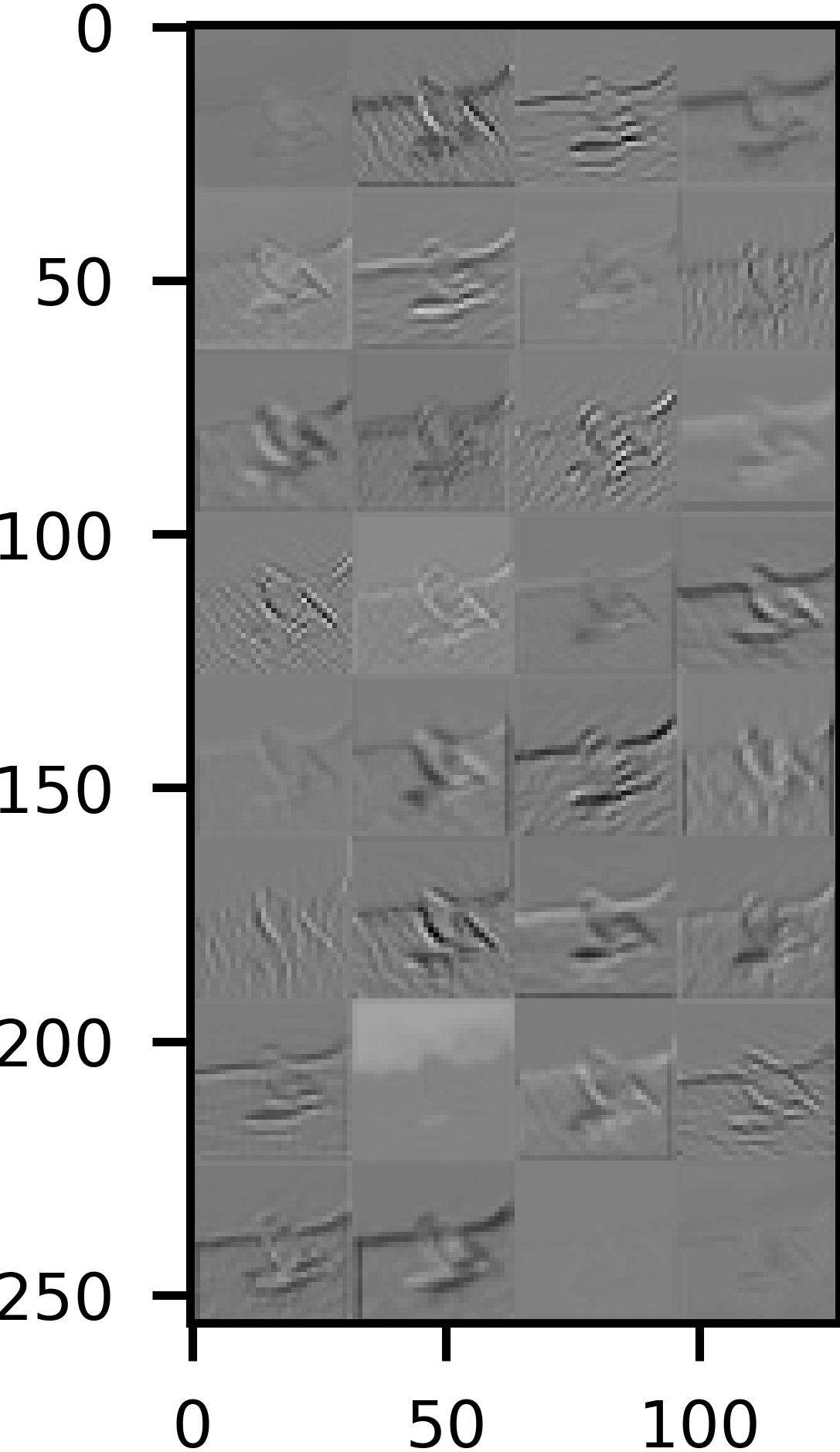} &
		  \quad \includegraphics[width=0.22\textwidth]{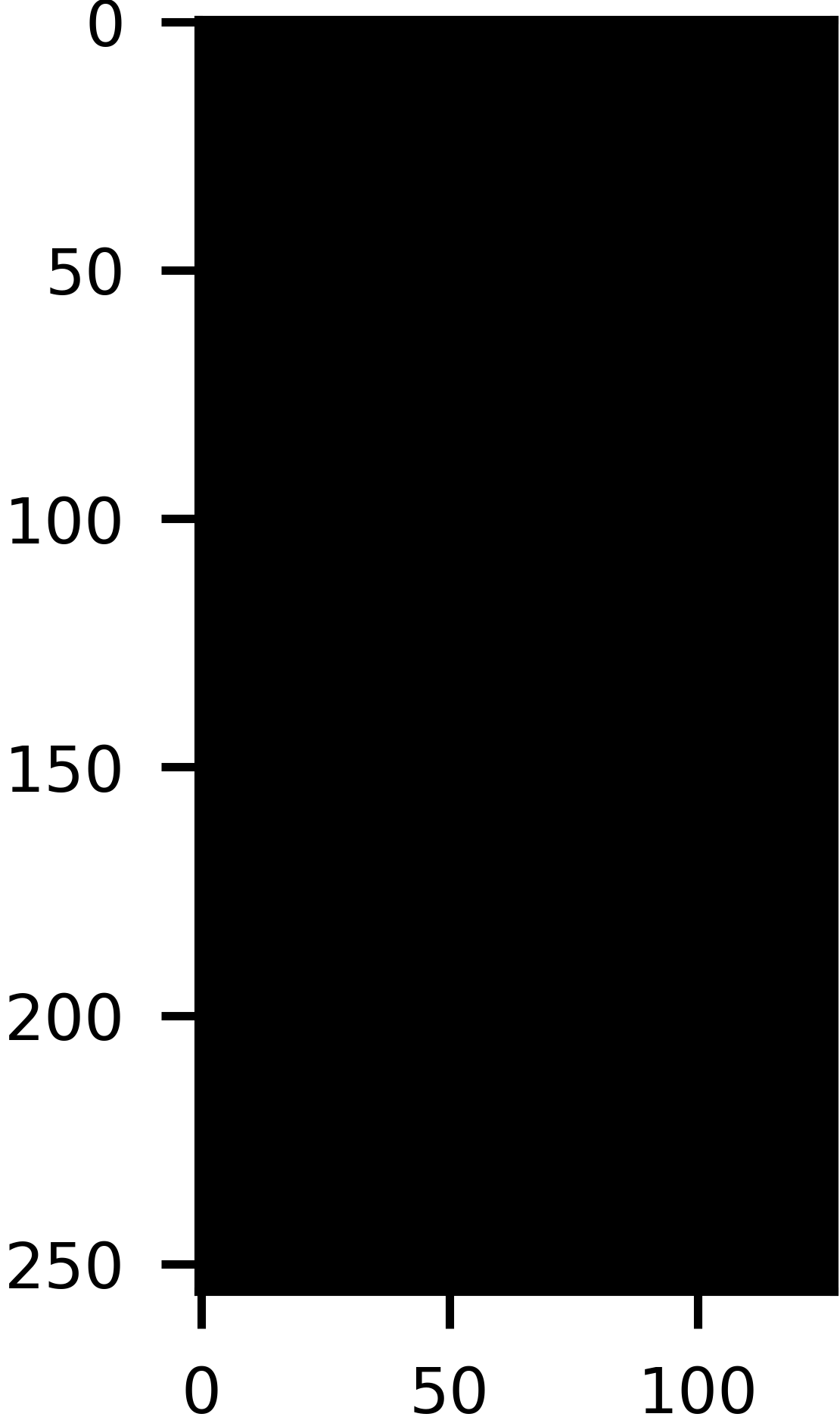} &
            \quad \includegraphics[width=0.22\textwidth]{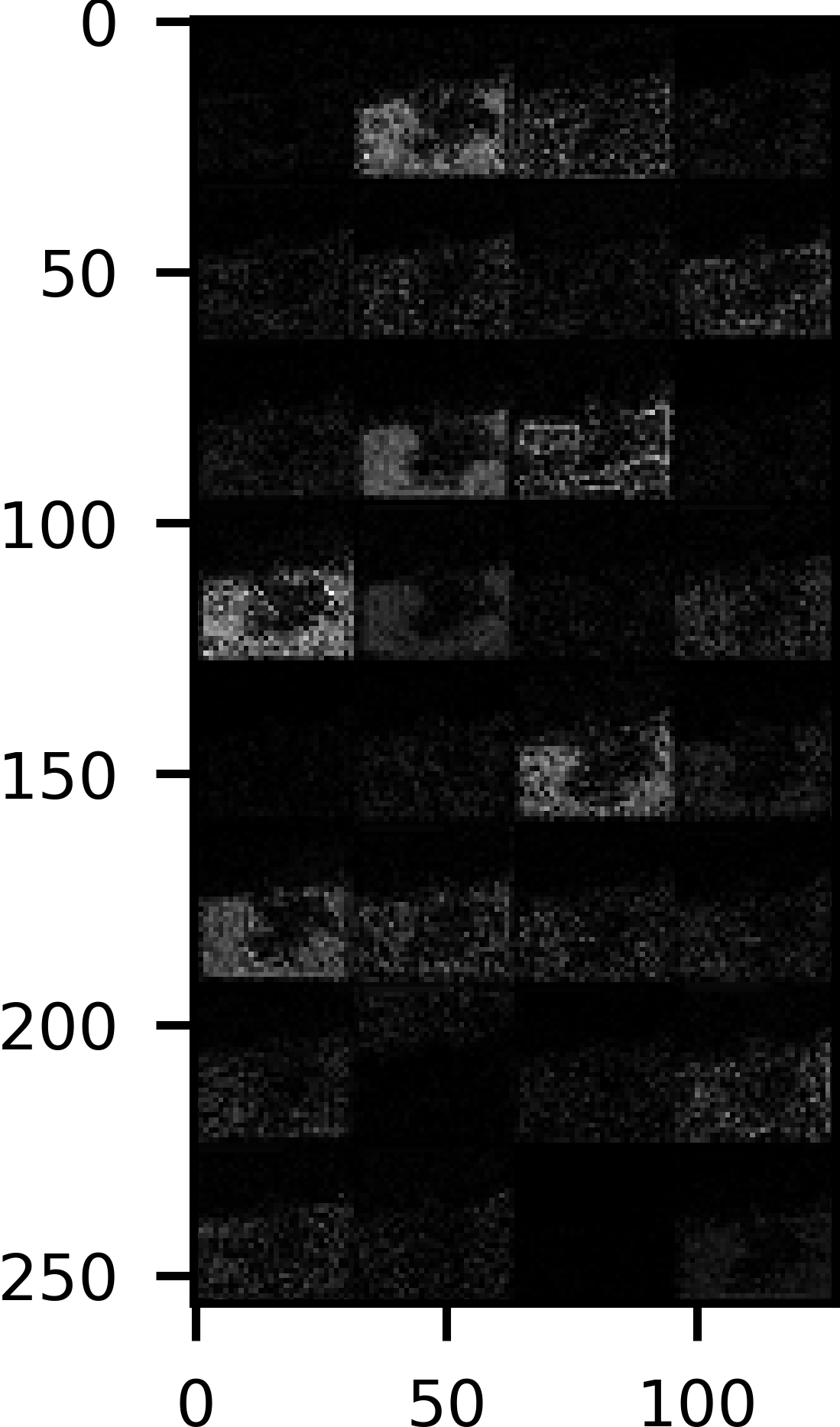} \\
		\end{tabular}
		\caption{VGG7 (CIFAR-10). From left to right: Original VGG7's feature maps (also see the first line of Eq.\eqref{eq:verify_conv}), the reconstructed feature maps (also see the second line of Eq.\eqref{eq:verify_conv}), their difference, and the normalized difference. The first three pictures are under the same scale for visualization; The difference would normally look black (near zero), so we normalized it as the last one  (in the normalized plot, the maximum value is usually around $10^{-7}$). These $32$ feature maps come from the Conv$1$ layer (See table \ref{table:params:vgg7}; Also, please note that the Conv$1$ layer is the second layer, the very first layer is Conv$0$) of VGG7. Each has a resolution of $32 \times 32$, and we group every four feature maps in a row (so each picture has eight rows). We use the CIFAR-10 test set. The input image for VGG7 and the layer for reconstruction are randomly selected. It illustrated that the reconstructed feature maps could recover the original feature maps precisely.}
		\label{fig:featurescomparison:vgg7:cifar10}
	\end{figure}

        We illustrate original CNN's feature maps and their reconstructed ones using Eq.\ref{eq:verify_conv} through several examples (fig.\ref{fig:featurescomparison:vgg7:cifar10} $\sim$ fig.\ref{fig:featurescomparison:resnet20_fixup:cifar100}). These examples verify that the precision of our reconstruction is very high.
    
        \begin{figure}[!h]
		\centering
		\begin{tabular}{@{}c@{}c@{}c@{}c@{}}
		  \includegraphics[width=0.22\textwidth]{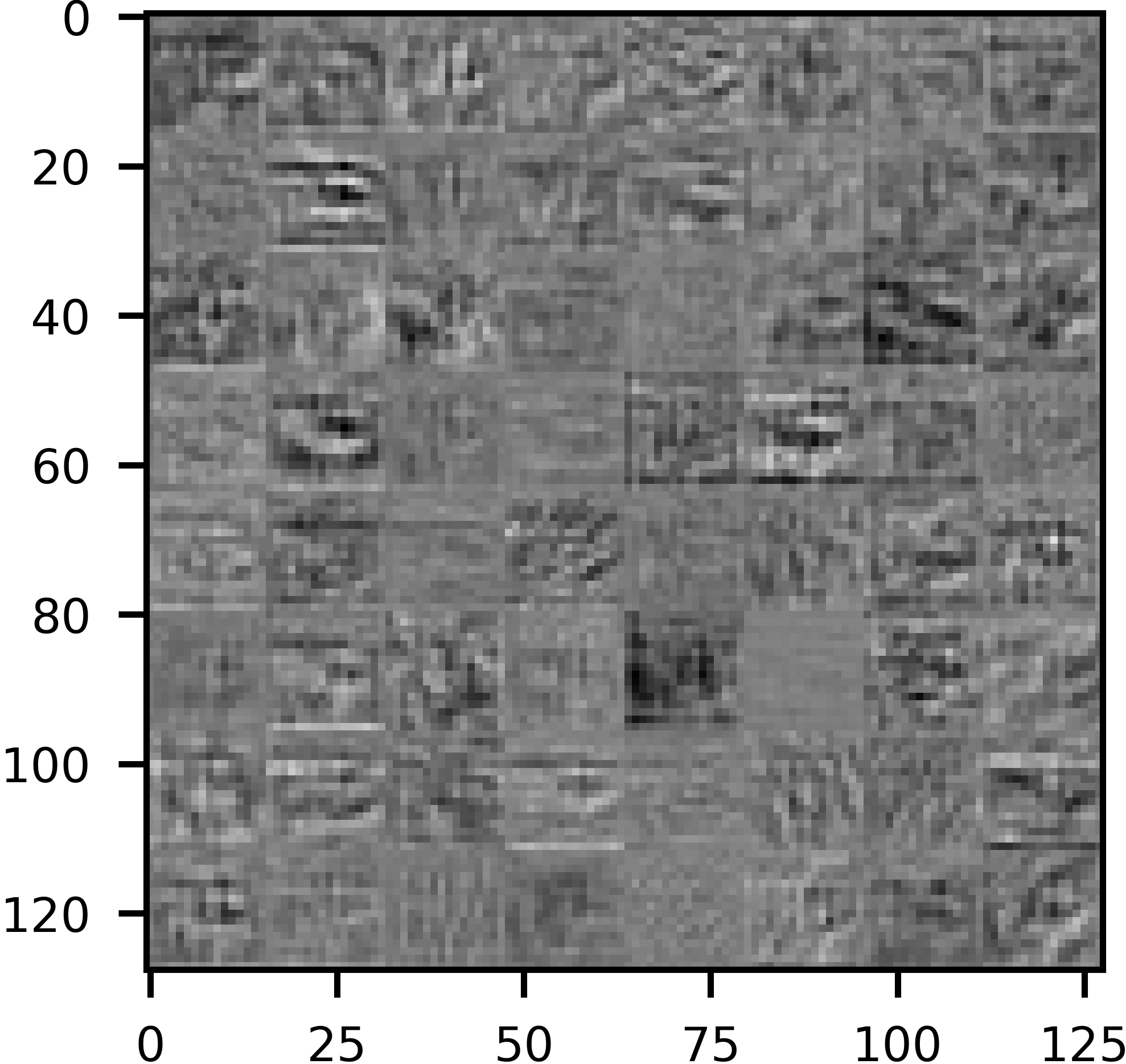} &
            \quad \includegraphics[width=0.22\textwidth]{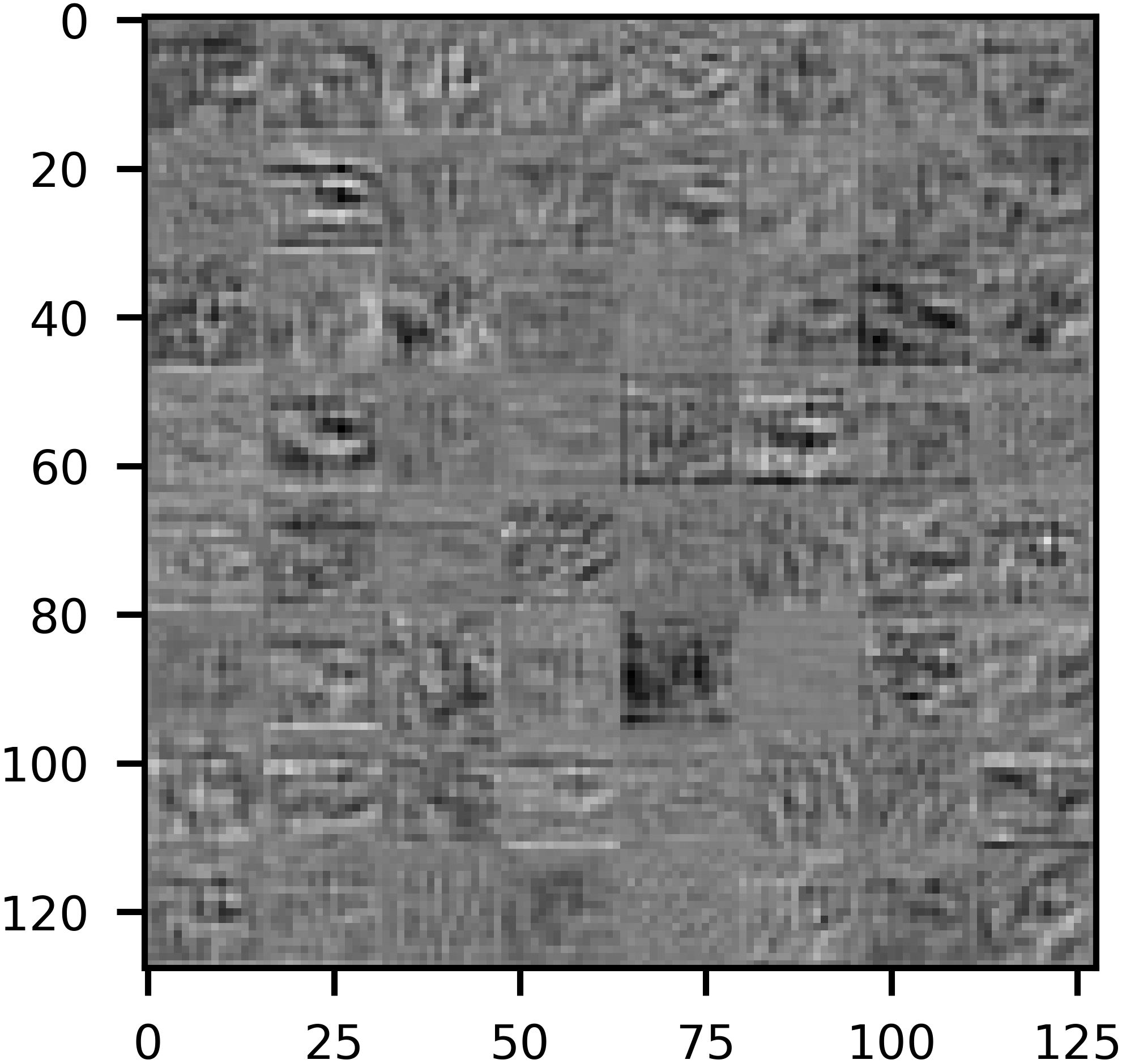} &
		  \quad \includegraphics[width=0.22\textwidth]{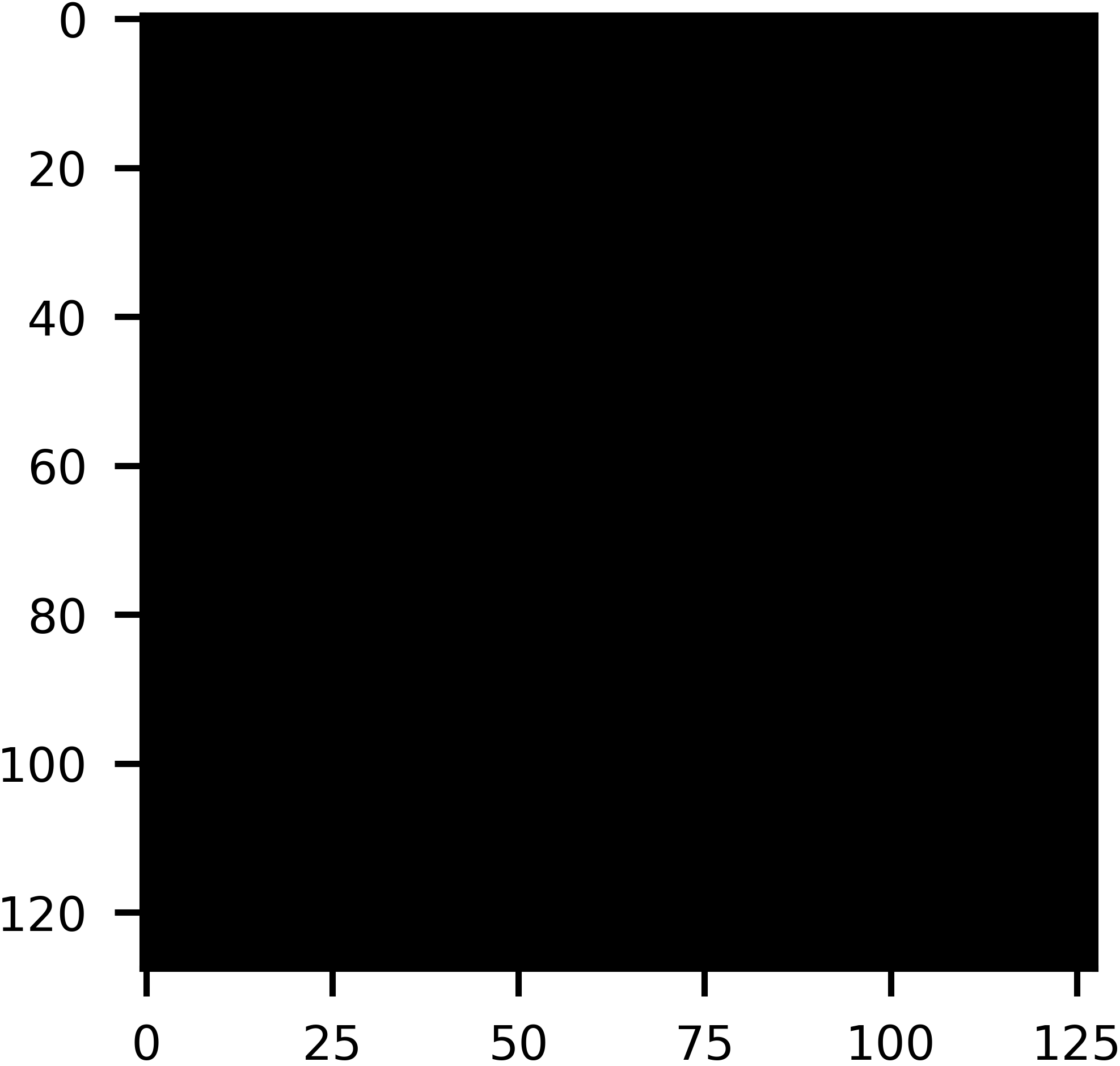} &
            \quad \includegraphics[width=0.22\textwidth]{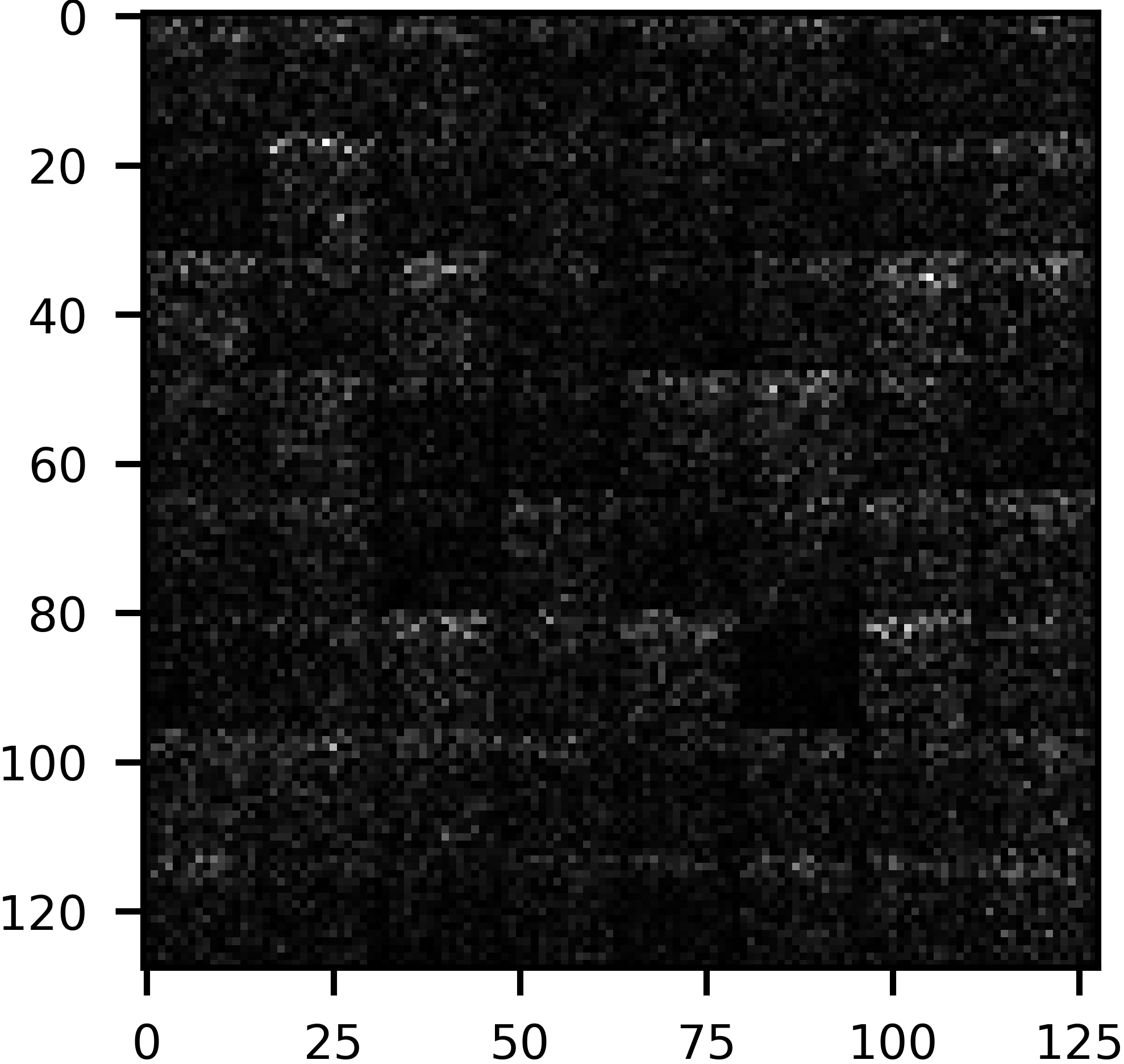}
		\end{tabular}
		\caption{ResNet20 (CIFAR-10). From left to right: Original ResNet20's feature maps, the reconstructed feature maps, their difference, and the normalized difference. These $64$ feature maps come from the Conv$10$ layer (See table \ref{table:params:resnet20}) of ResNet20. Each has a resolution of $16 \times 16$, and we group every eight feature maps in a row (so each picture has eight rows). Check fig.\ref{fig:featurescomparison:vgg7:cifar10} for details.}
		\label{fig:featurescomparison:resnet20:cifar10}
	\end{figure}

        \begin{figure}[!h]
		\centering
		\begin{tabular}{@{}c@{}c@{}c@{}c@{}}
		  \includegraphics[width=0.22\textwidth]{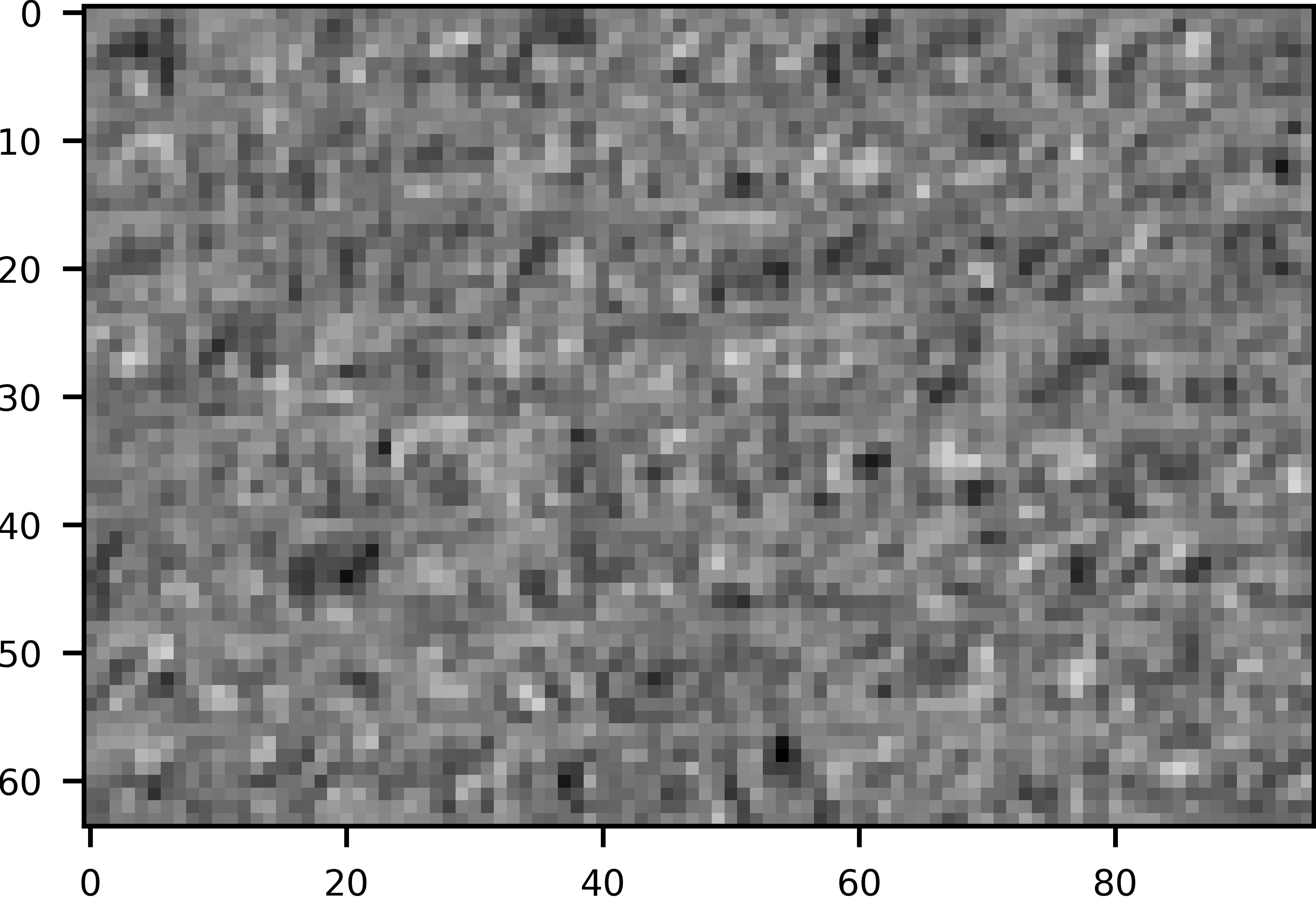} &
            \quad \includegraphics[width=0.22\textwidth]{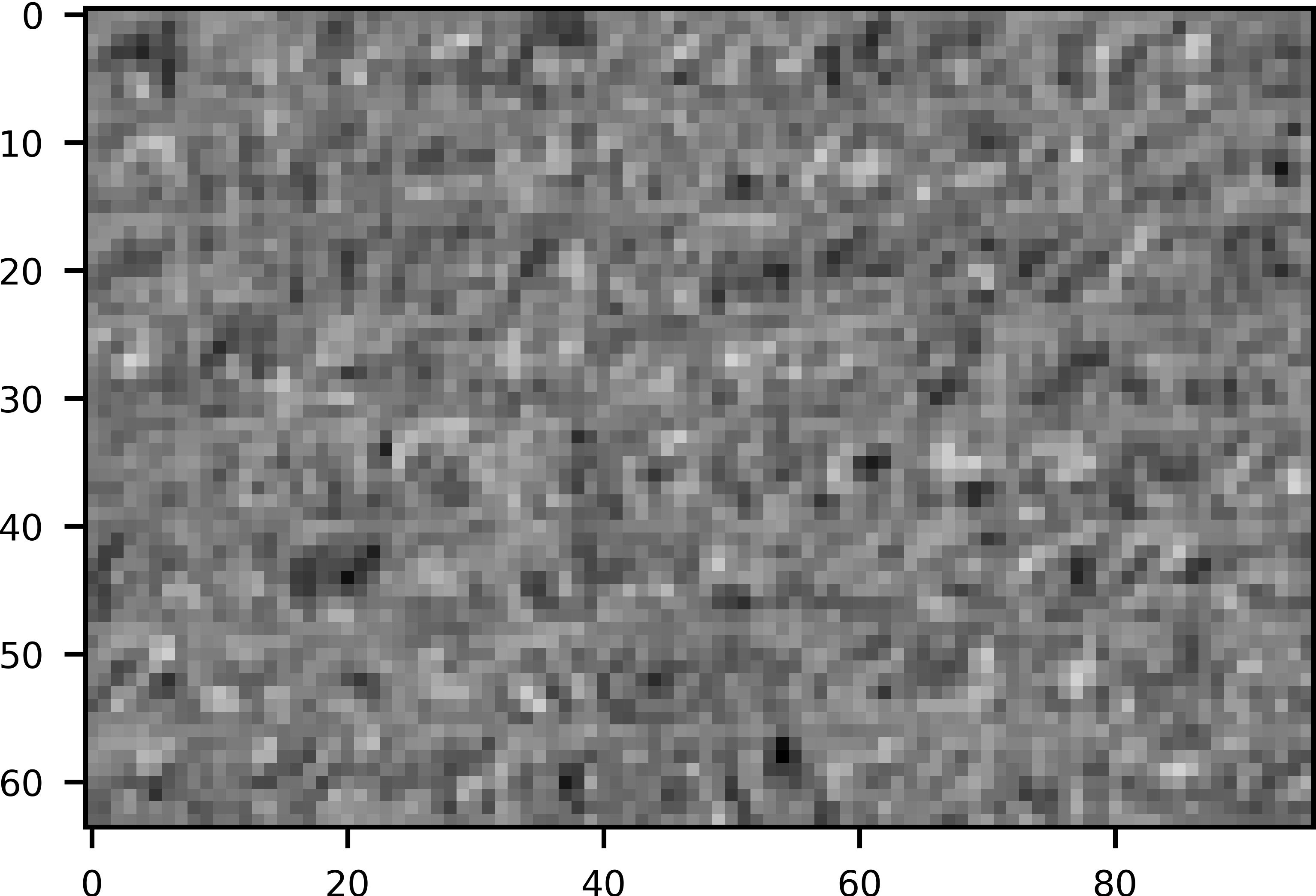} &
		  \quad \includegraphics[width=0.22\textwidth]{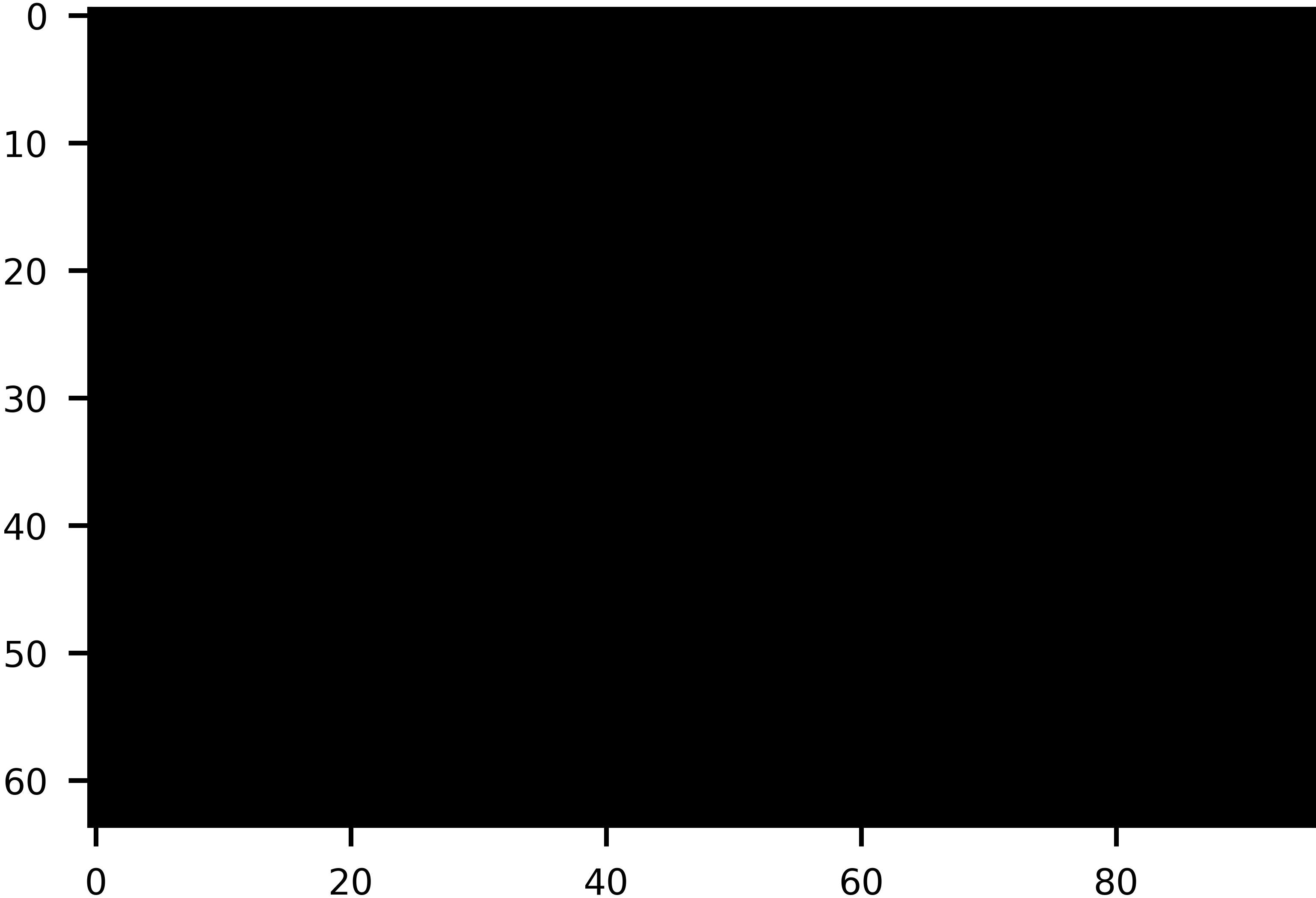} &
            \quad \includegraphics[width=0.22\textwidth]{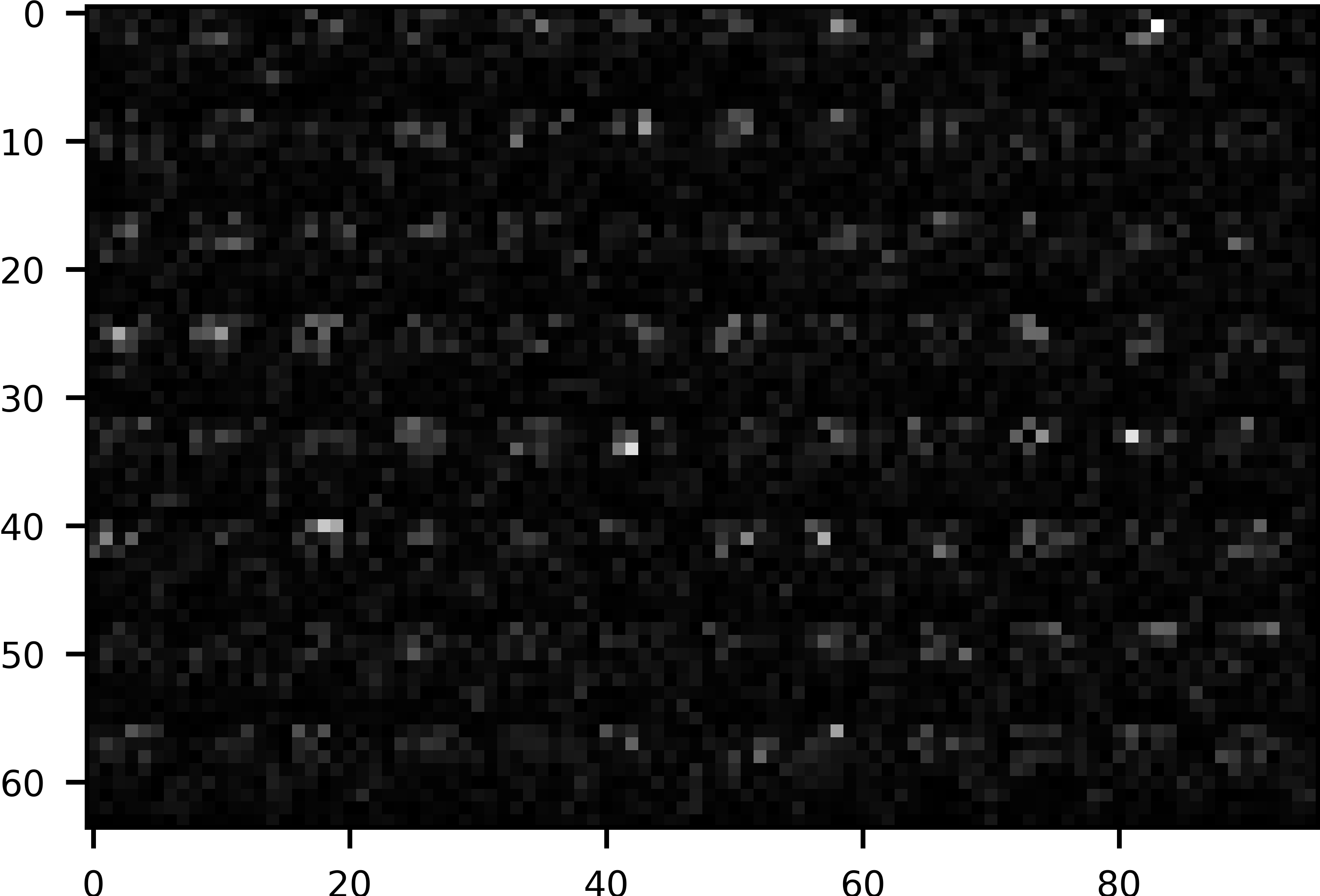} \\
		\end{tabular}
		\caption{ResNet20-Fixup (CIFAR-10). From left to right: Original ResNet20-Fixup's feature maps, the reconstructed feature maps, and their difference, and the normalized difference. These $96$ feature maps come from the Conv$15$ layer (See table \ref{table:params:resnet20_fixup}) of ResNet20-Fixup. Each has a resolution of $8 \times 8$, and we group every twelve feature maps in a row (so each picture has eight rows). Check fig.\ref{fig:featurescomparison:vgg7:cifar10} for details.}
		\label{fig:featurescomparison:resnet20_fixup:cifar10}
	\end{figure}

        \begin{figure}[!h]
		\centering
		\begin{tabular}{@{}c@{}c@{}c@{}c@{}}
		  \includegraphics[width=0.22\textwidth]{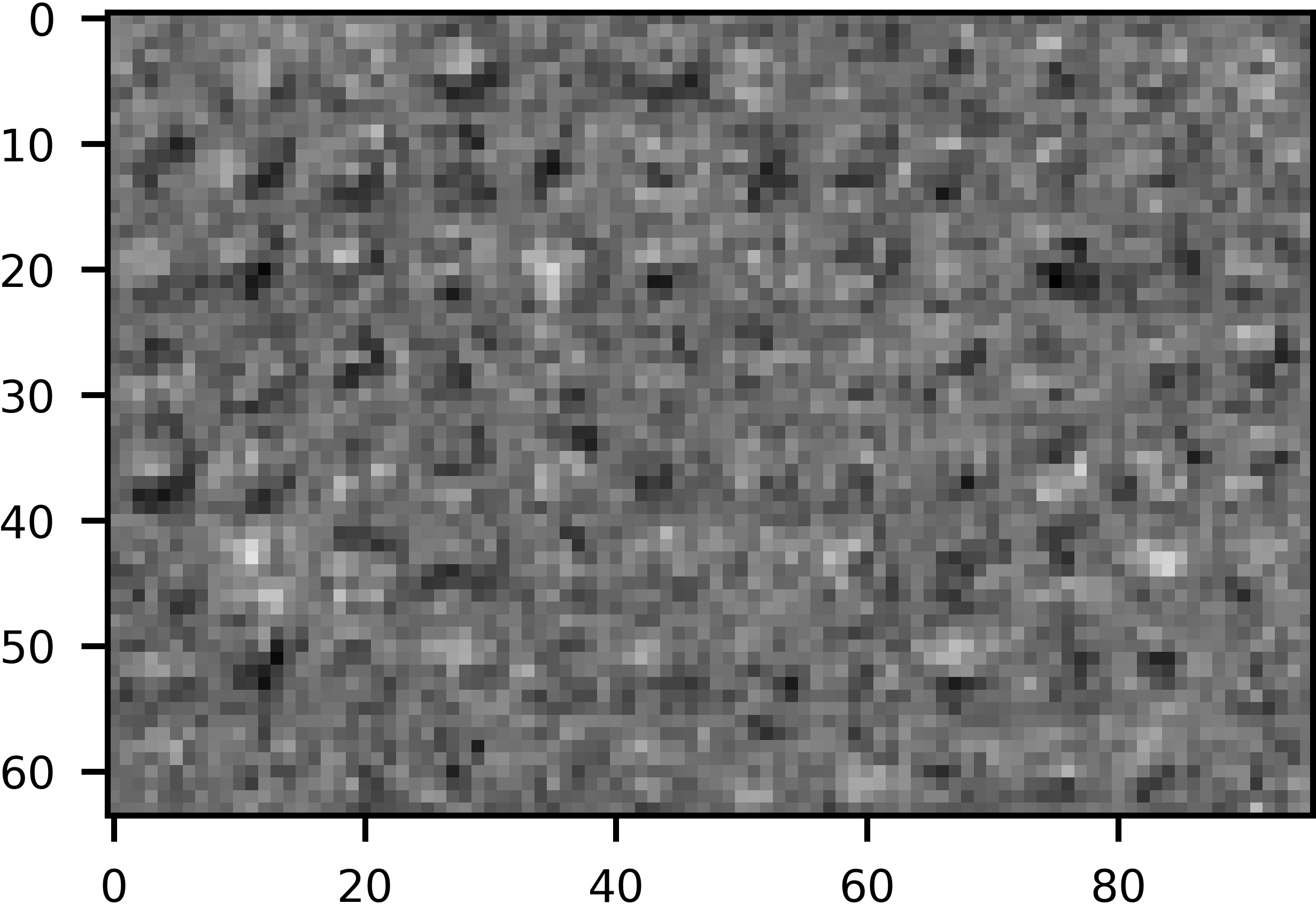} &
            \quad \includegraphics[width=0.22\textwidth]{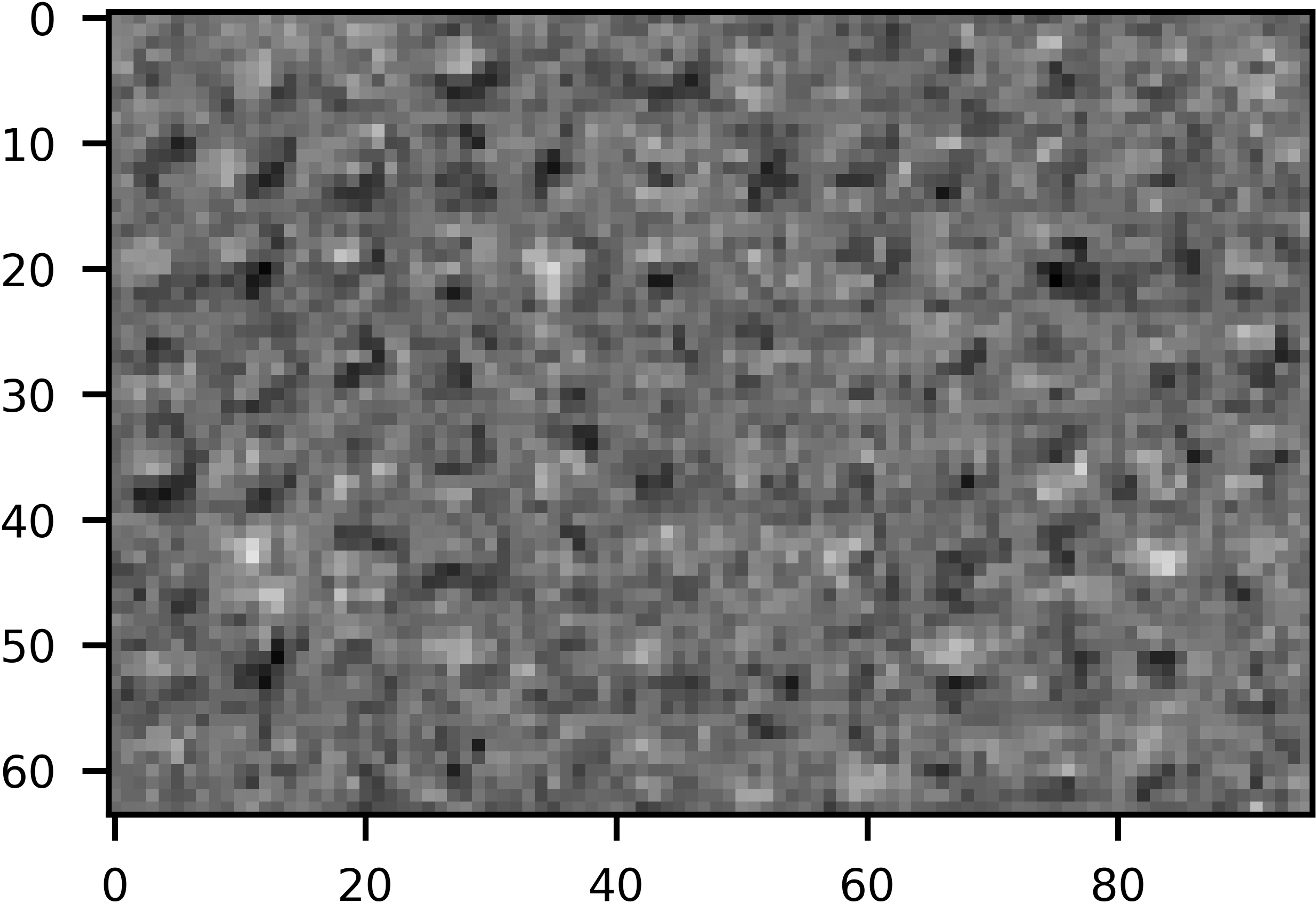} &
		  \quad \includegraphics[width=0.22\textwidth]{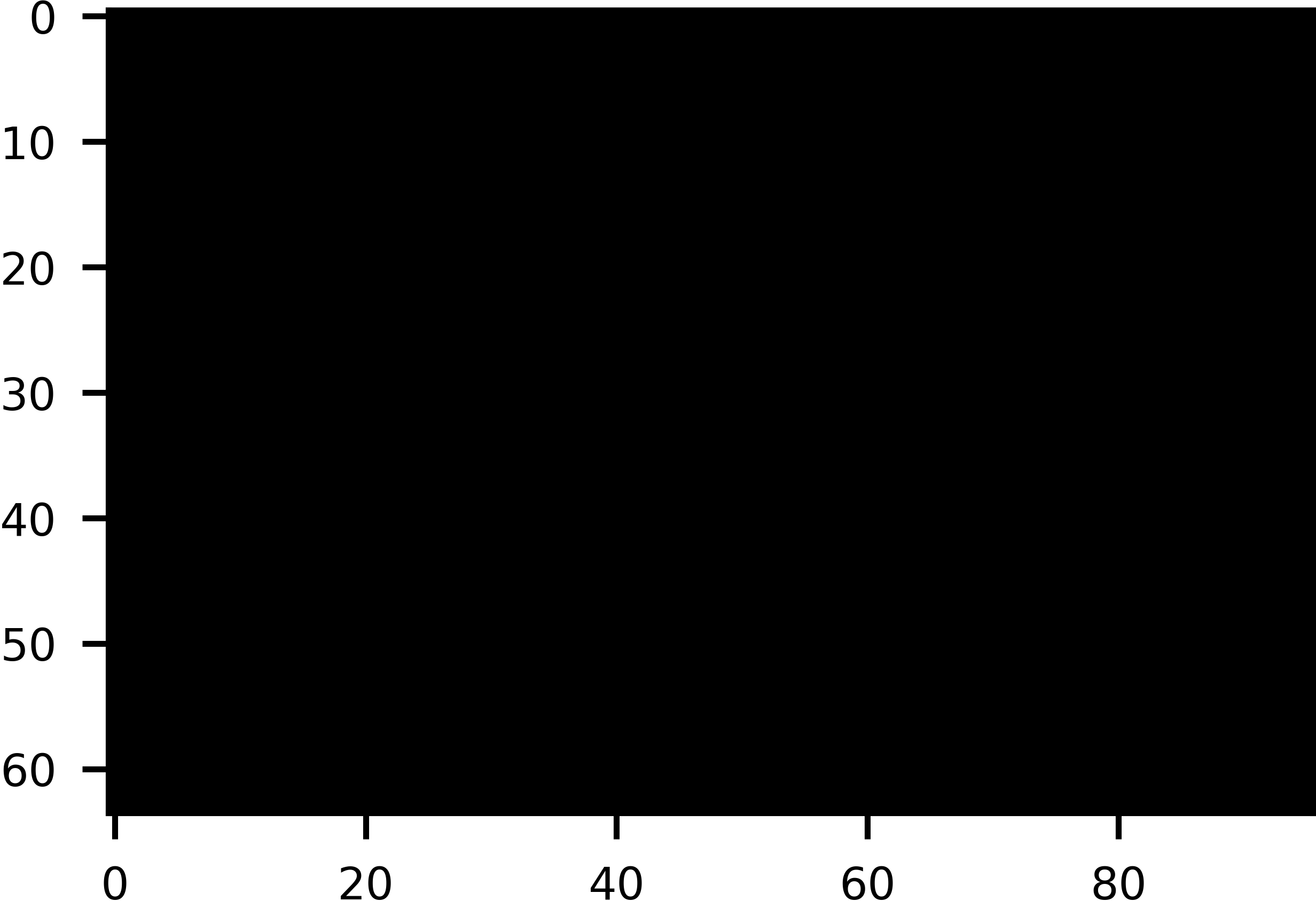} &
            \quad \includegraphics[width=0.22\textwidth]{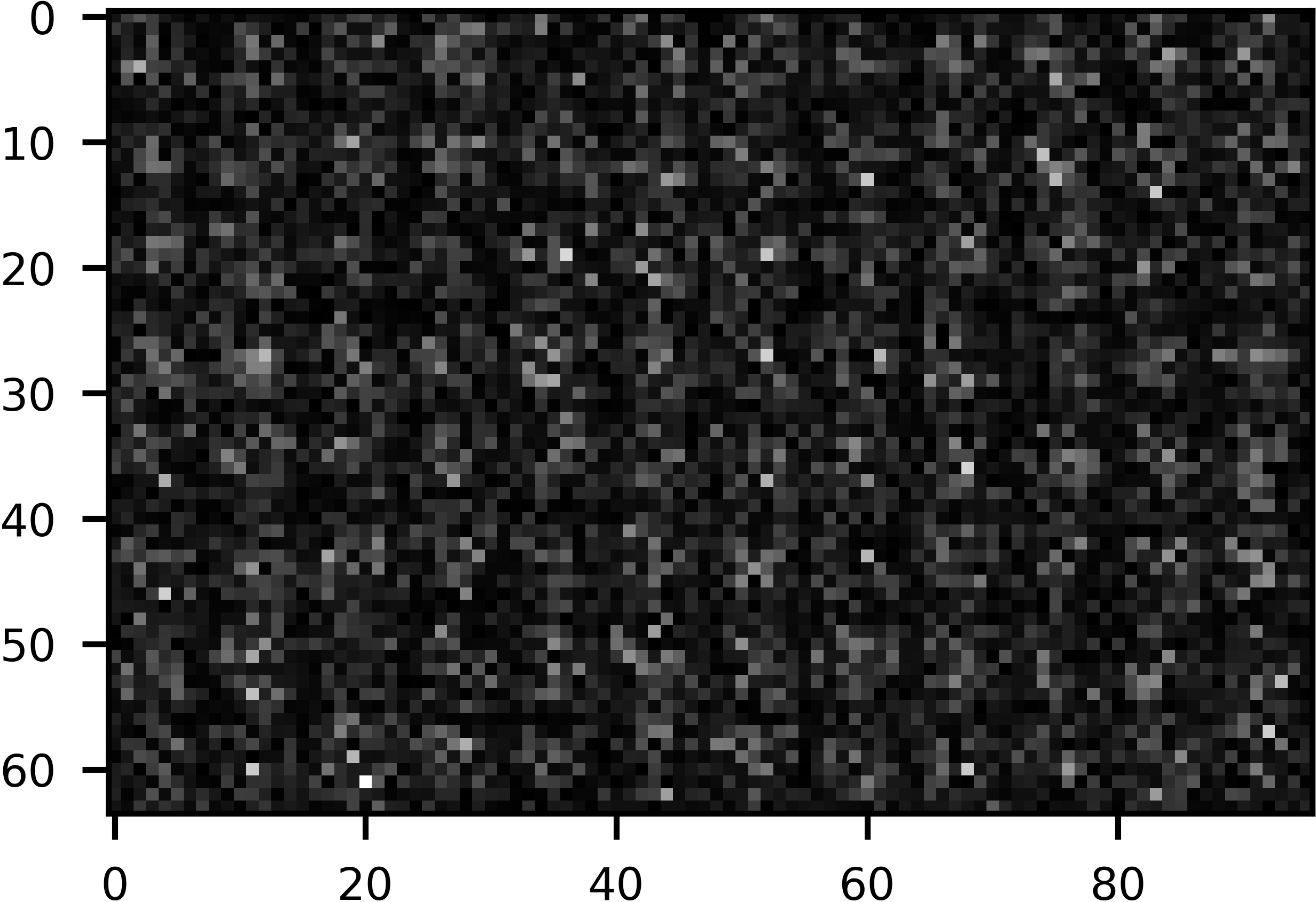} \\
		\end{tabular}
		\caption{VGG7 (CIFAR-100). From left to right, Original VGG7's feature maps (also see the first line of Eq.\eqref{eq:verify_conv}), the reconstructed feature maps (also see the second line of Eq.\eqref{eq:verify_conv}), their difference, and the normalized difference. These $96$ feature maps come from the Conv$5$ layer of VGG7. Each has a resolution of $8 \times 8$, and we group every twelve feature maps in a row (so each picture has eight rows). We use the CIFAR-100 test set. See fig.\ref{fig:featurescomparison:vgg7:cifar10} for other details.}
		\label{fig:featurescomparison:vgg7:cifar100}
	\end{figure}

        \begin{figure}[!h]
		\centering
		\begin{tabular}{@{}c@{}c@{}c@{}c@{}}
		  \includegraphics[width=0.22\textwidth]{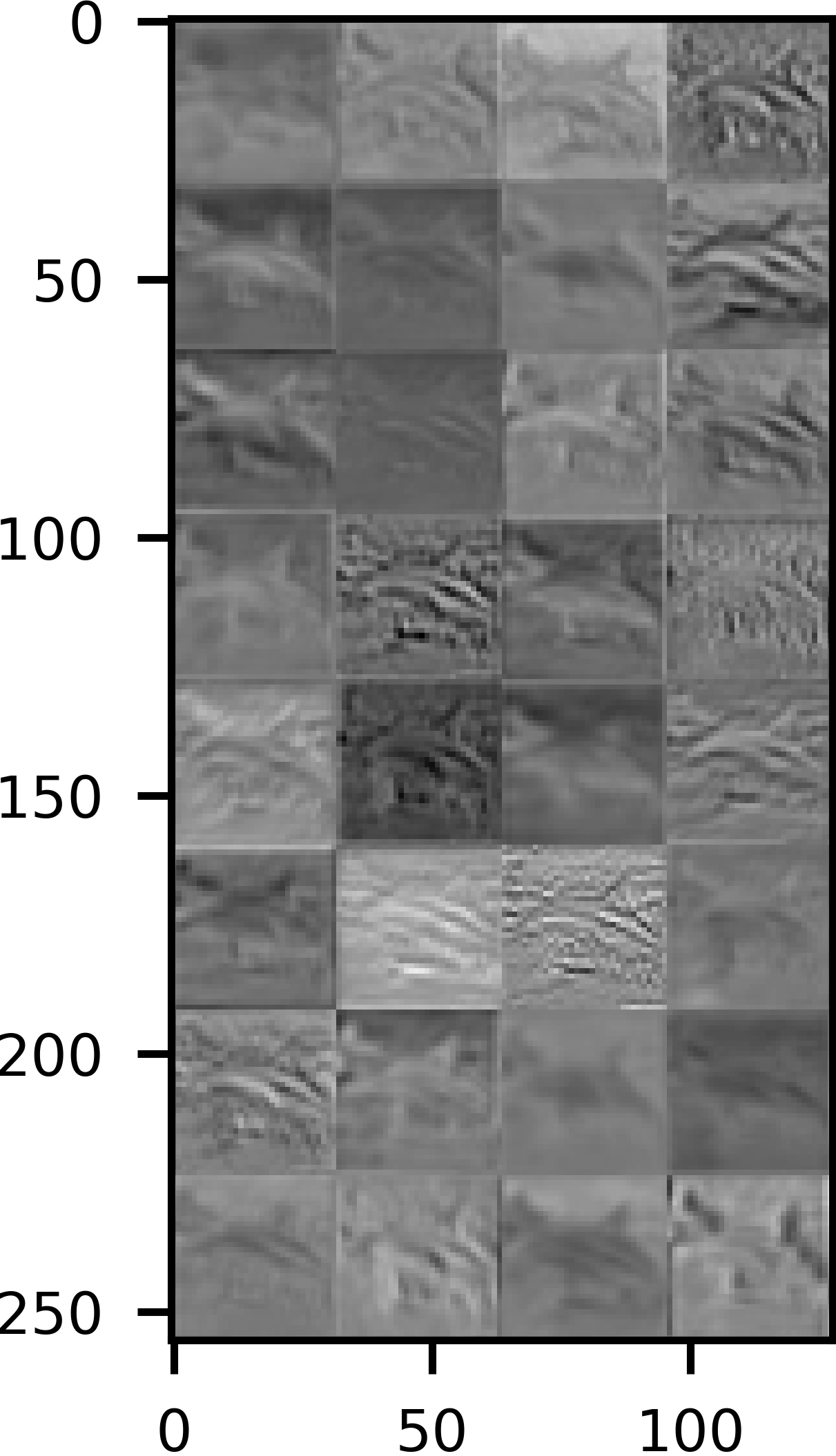} &
            \quad \includegraphics[width=0.22\textwidth]{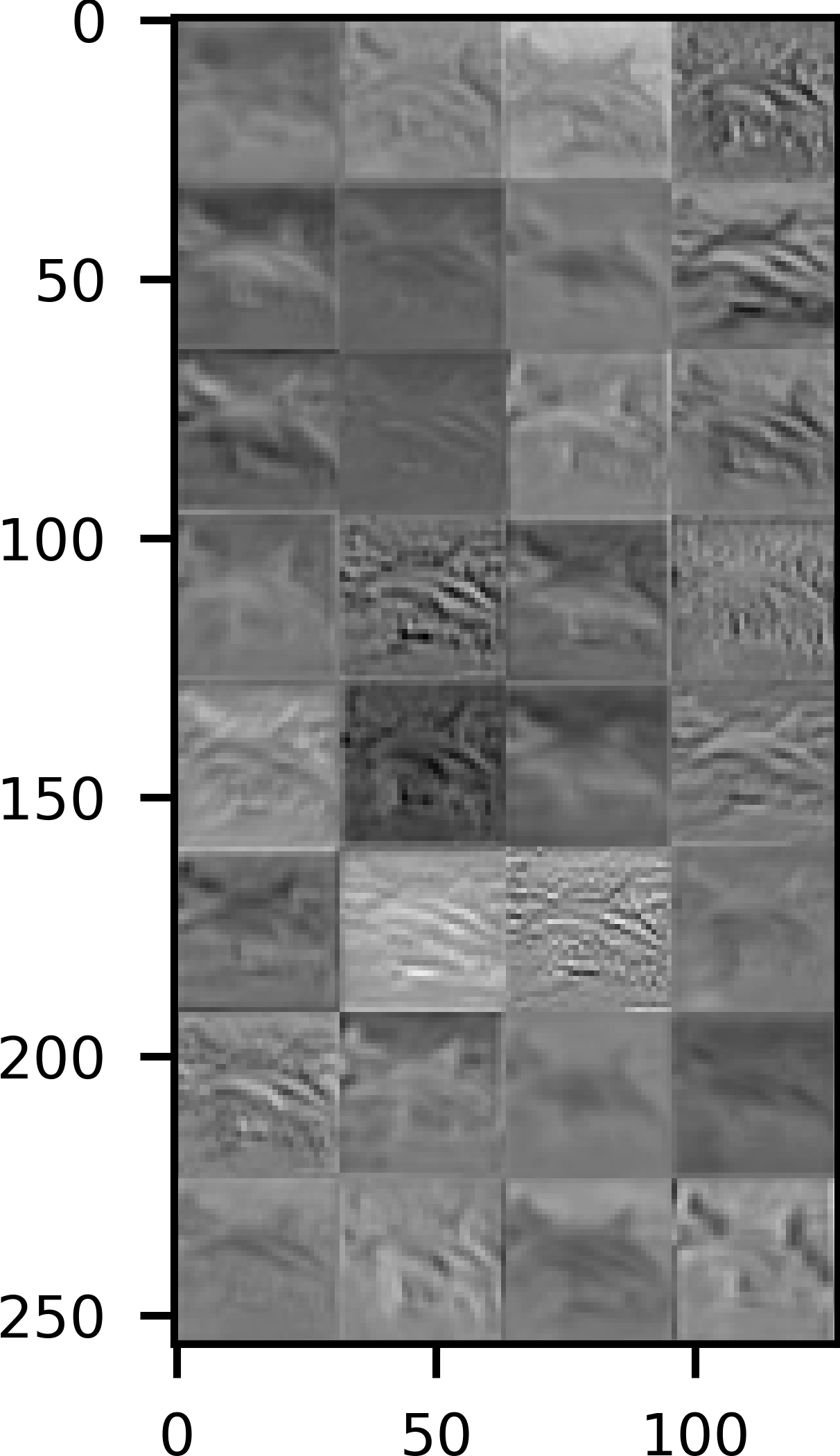} &
		  \quad \includegraphics[width=0.22\textwidth]{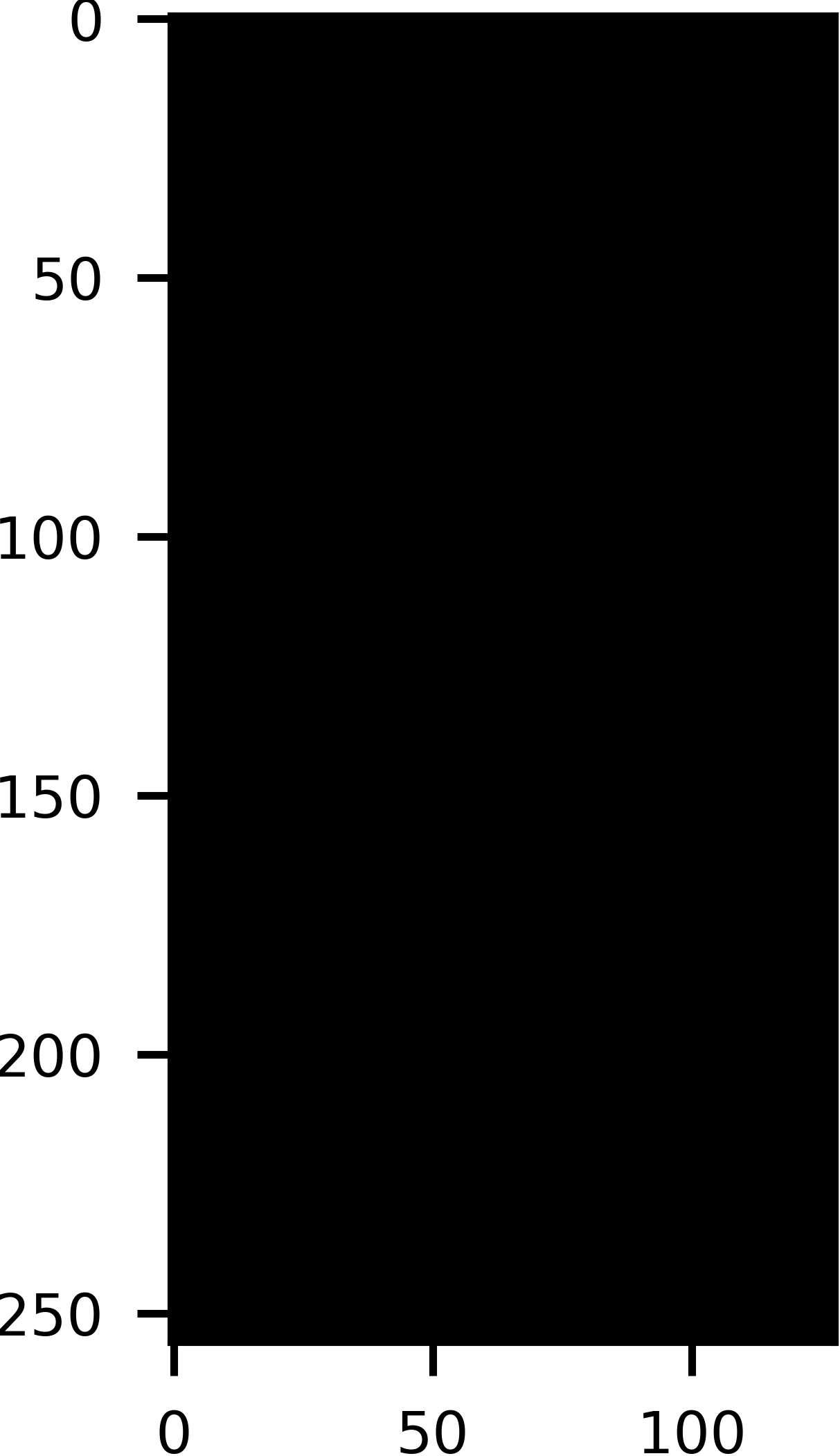} &
            \quad \includegraphics[width=0.22\textwidth]{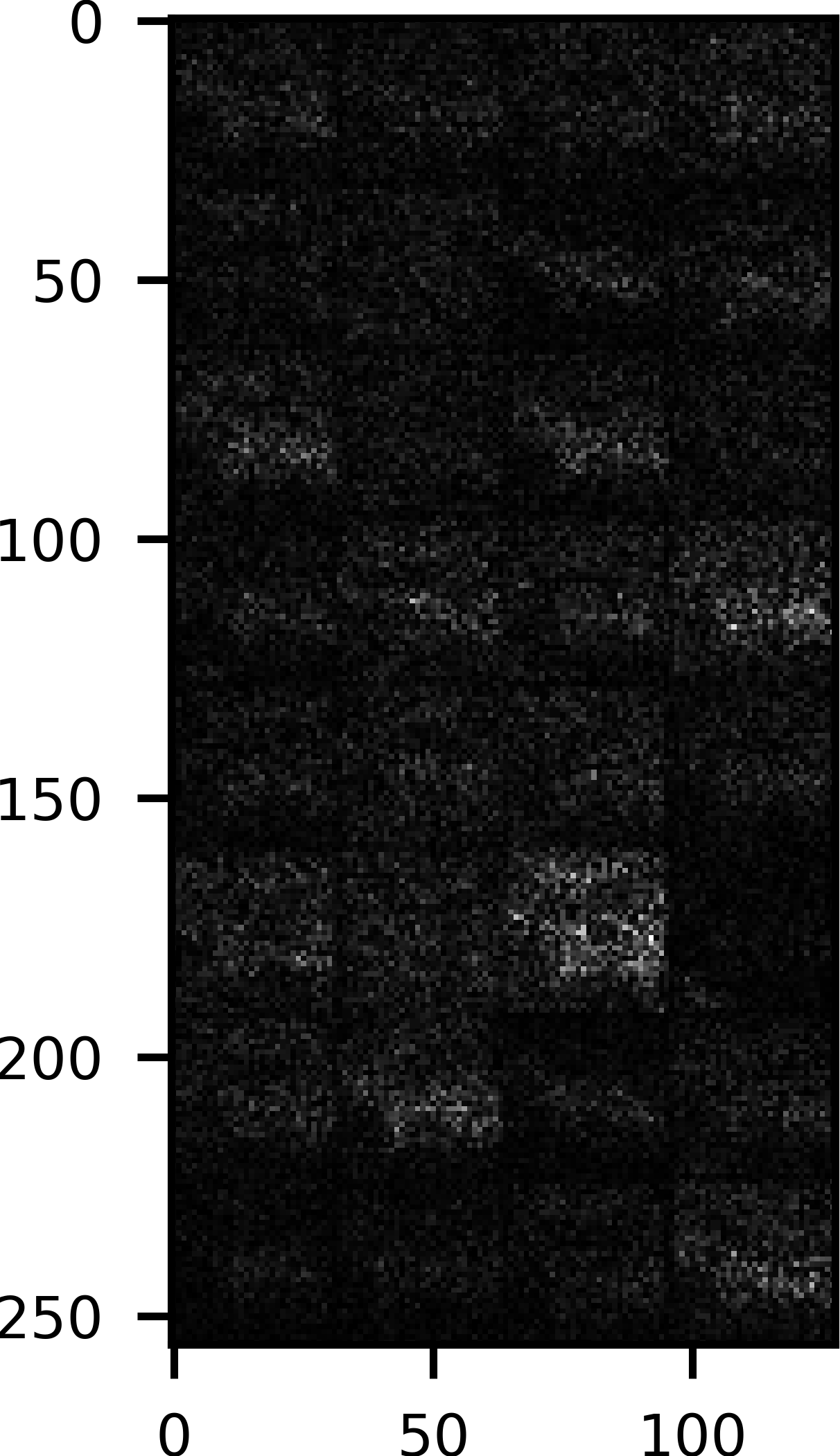}
		\end{tabular}
		\caption{ResNet20 (CIFAR-100). From left to right: Original ResNet20's feature maps, the reconstructed feature maps, their difference, and the normalized difference. These $32$ feature maps come from the Conv$3$ layer (See table \ref{table:params:resnet20}) of ResNet20. Each has a resolution of $32 \times 32$, and we group every four feature maps in a row (so each picture has eight rows). Check fig.\ref{fig:featurescomparison:vgg7:cifar100} for details.}
		\label{fig:featurescomparison:resnet20:cifar100}
	\end{figure}

        \begin{figure}[!h]
		\centering
		\begin{tabular}{@{}c@{}c@{}c@{}c@{}}
		  \includegraphics[width=0.22\textwidth]{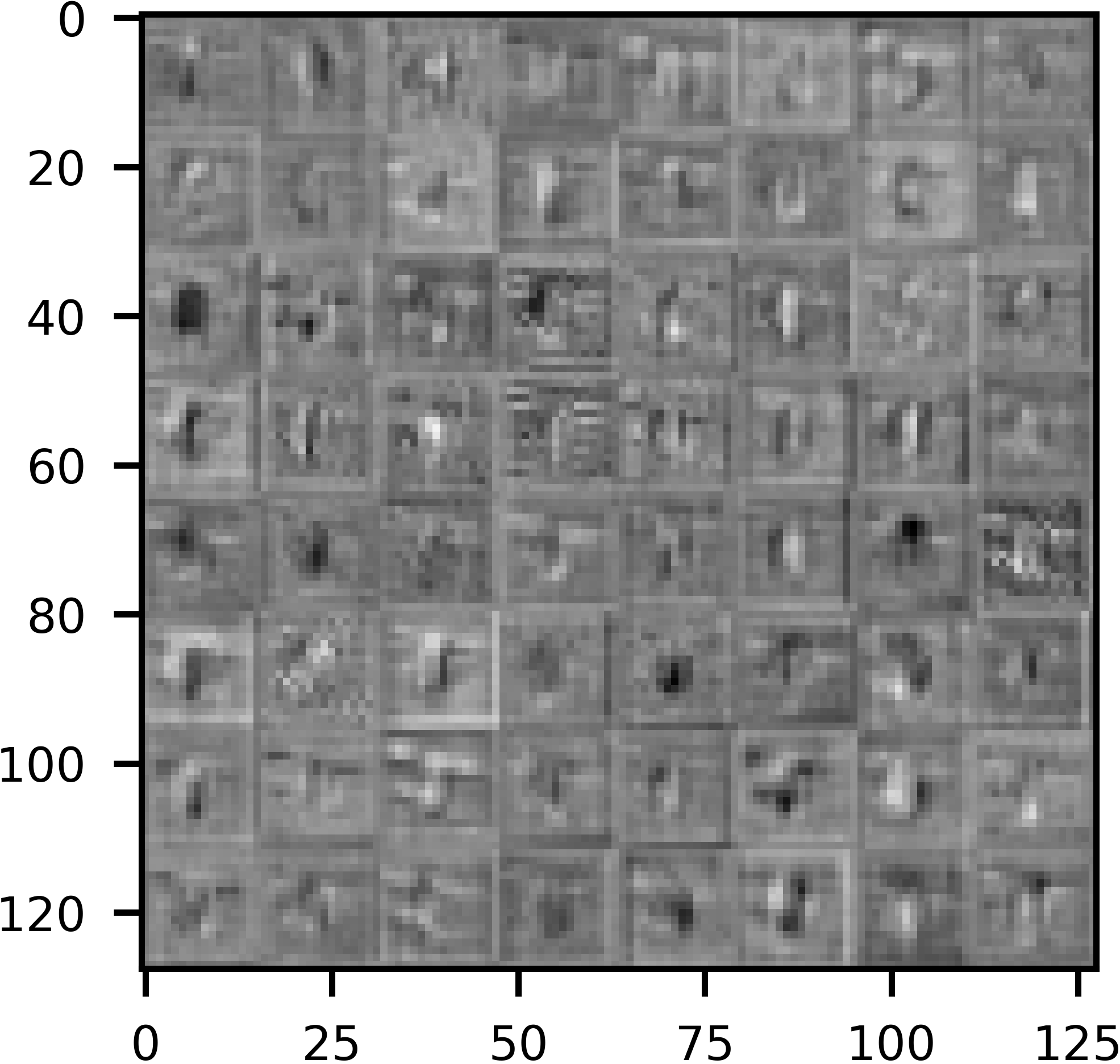} &
            \quad \includegraphics[width=0.22\textwidth]{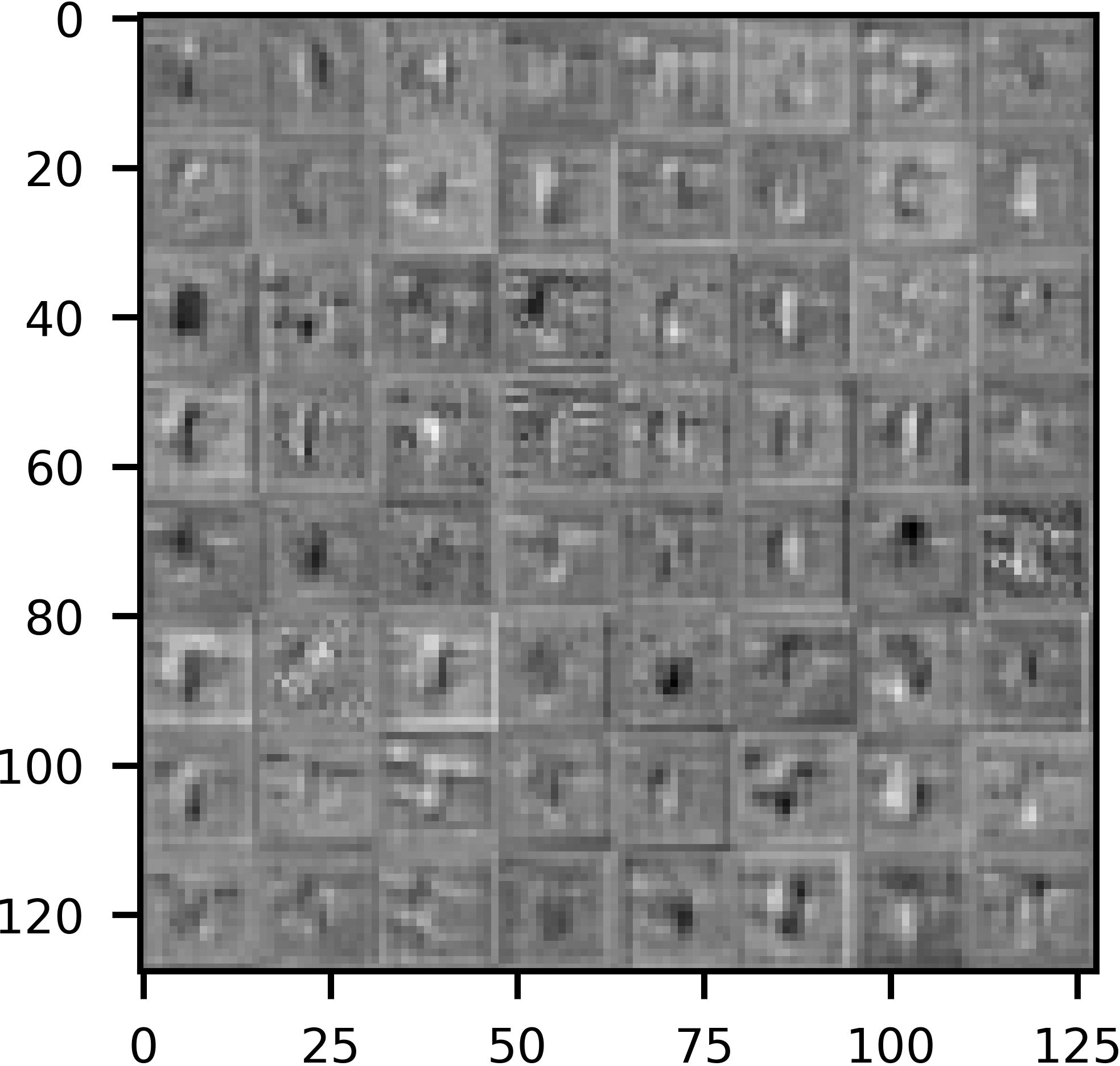} &
		  \quad \includegraphics[width=0.22\textwidth]{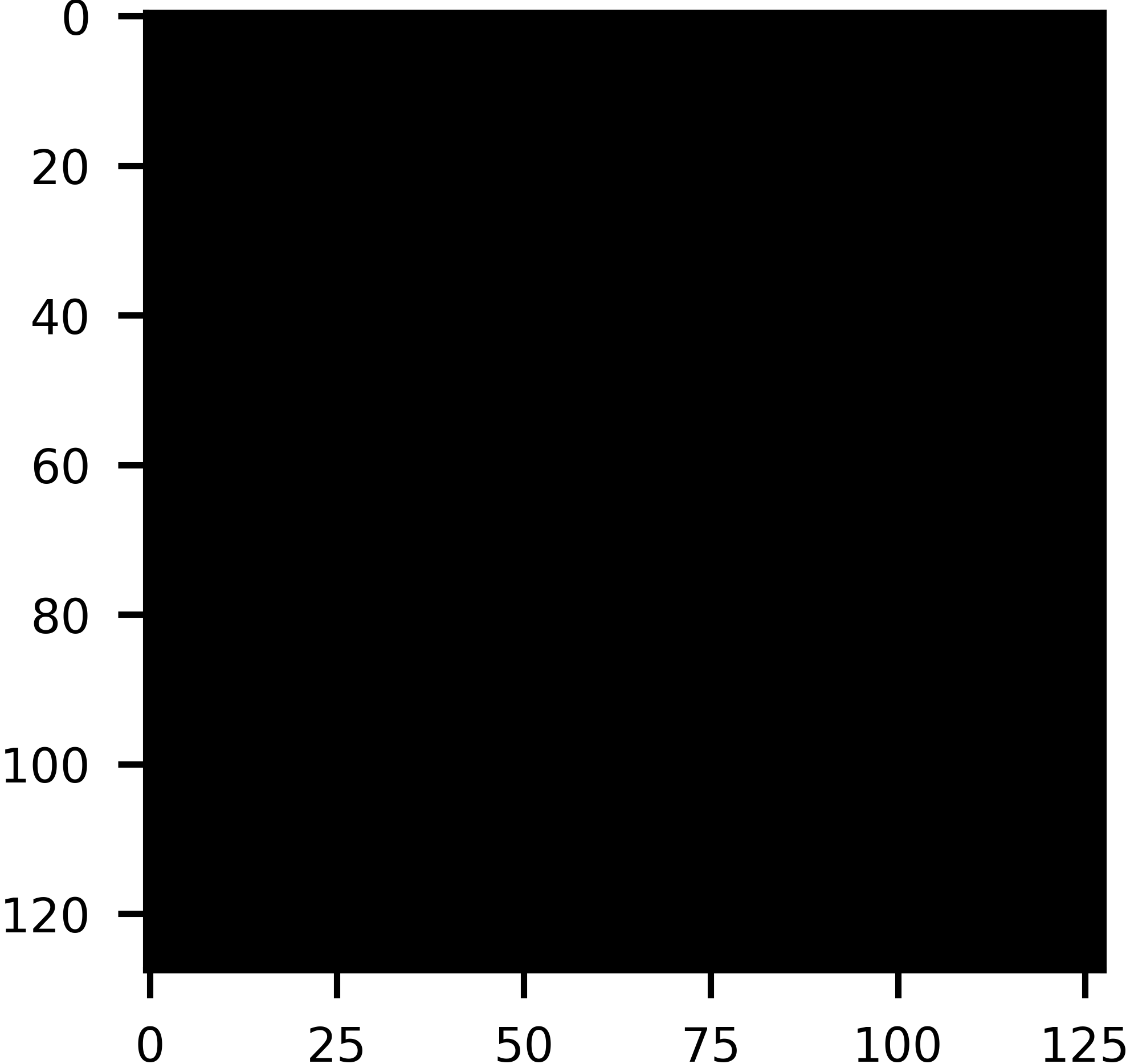} &
            \quad \includegraphics[width=0.22\textwidth]{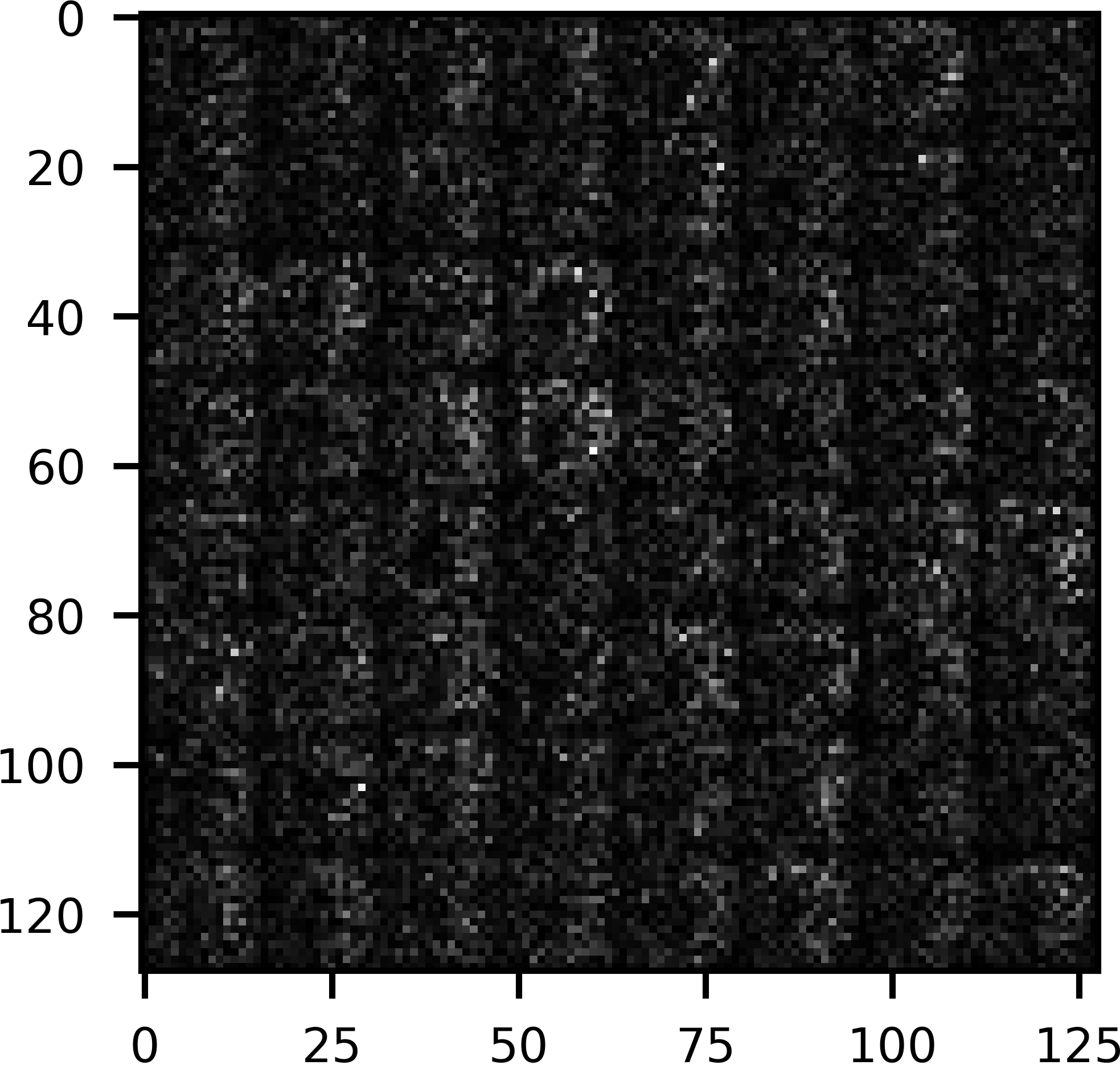} \\
		\end{tabular}
		\caption{ResNet20-Fixup (CIFAR-100). From left to right: Original ResNet20-Fixup's feature maps, the reconstructed feature maps, their difference, and the normalized difference. These $64$ feature maps come from the Conv$12$ layer (See table \ref{table:params:resnet20_fixup}) of ResNet20-Fixup. Each has a resolution of $16 \times 16$, and we group every eight feature maps in a row (so each picture has eight rows). Check fig.\ref{fig:featurescomparison:vgg7:cifar100} for details.}
		\label{fig:featurescomparison:resnet20_fixup:cifar100}
	\end{figure}
        \end{appendices}
\end{document}